\documentclass[sigconf]{acmart}

\AtBeginDocument{%
  \providecommand\BibTeX{{%
    Bib\TeX}}}

\setcopyright{acmlicensed}
\copyrightyear{2018}
\acmYear{2018}
\acmDOI{XXXXXXX.XXXXXXX}
\acmConference[Conference acronym 'XX]{Make sure to enter the correct
  conference title from your rights confirmation email}{June 03--05,
  2018}{Woodstock, NY}
\acmISBN{978-1-4503-XXXX-X/2018/06}

\usepackage{amsmath,amsfonts,bm}

\def\Figref#1{Figure~\ref{#1}}

\def\Secref#1{Section~\ref{#1}}

\def\eqref#1{equation~\ref{#1}}
\def\Eqref#1{Equation~\ref{#1}}

\def\Algref#1{Algorithm~\ref{#1}}

\def\1{\bm{1}}

\def\vh{{\bm{h}}}

\def\vp{{\bm{p}}}

\def\mA{{\bm{A}}}

\def\mH{{\bm{H}}}
\def\mI{{\bm{I}}}

\def\mM{{\bm{M}}}

\def\mP{{\bm{P}}}

\def\mS{{\bm{S}}}

\def\mW{{\bm{W}}}
\def\mX{{\bm{X}}}

\DeclareMathAlphabet{\mathsfit}{\encodingdefault}{\sfdefault}{m}{sl}
\SetMathAlphabet{\mathsfit}{bold}{\encodingdefault}{\sfdefault}{bx}{n}

\usepackage{hyperref}
\usepackage{url}

\usepackage{graphicx}
\usepackage{subcaption} 
\usepackage{booktabs}
\usepackage{makecell}
\usepackage{tabularx} 
\usepackage{wrapfig}
\usepackage{comment}
\usepackage{siunitx}
\usepackage{arydshln}
\usepackage{xspace}
\usepackage{algorithm}
\usepackage{algpseudocode}
\usepackage{amsmath, amssymb}
\usepackage{hyperref}
\usepackage{multirow}
\usepackage{arydshln}
\usepackage{enumitem}
\usepackage{placeins}

\newcommand{\jx}[1]{\textcolor{orange}{[JX: #1]}}
\newcommand{\zz}[1]{\textcolor{red}{[ZZ: #1]}}

\newcommand{\batwo}{BA-2Motifs\xspace}
\newcommand{\bahg}{BA-HouseGrid\xspace}
\newcommand{\spmotif}{SPMotif\xspace}
\newcommand{\bahandg}{BA-HouseAndGrid\xspace}
\newcommand{\bahorg}{BA-HouseOrGrid\xspace}
\newcommand{\alca}{Alkane-Carbonyl\xspace}
\newcommand{\flca}{Fluorid-Carbonyl\xspace}
\newcommand{\ben}{Benzene\xspace}
\newcommand{\bavolume}{BA-Motif-Volume\xspace}
\newcommand{\bahgvolume}{House-Grid-Volume\xspace}
\newcommand{\bacounting}{BA-Motif-Counting\xspace}
\newcommand{\horgvolume}{House-OrGrid-Volume\xspace}
\newcommand{\tri}{Triangles\xspace}
\newcommand{\crip}{Crippen\xspace}

\newcommand{\gnne}{GNNExplainer\xspace}
\newcommand{\pge}{PGExplainer\xspace}
\newcommand{\metagnn}{MetaGNN\xspace}
\newcommand{\match}{MatchExplainer\xspace}
\newcommand{\mixupex}{MixupExplainer\xspace}
\newcommand{\proxyex}{ProxyExplainer\xspace}
\newcommand{\tagex}{TAGExplainer\xspace}
\newcommand{\regex}{RegExplainer\xspace}
\newcommand{\grad}{GRAD\xspace}
\newcommand{\att}{ATT\xspace}
\begin{document}

\title{Beyond Soft Masks: Hard-Perturbation Mixup Explainer for Robust GNN Explainability}
\author{Jialiang Yin}
\authornote{Jialiang Yin and Zheng Zhao contributed equally to this work.}
\email{15940556039@stu.xjtu.edu.cn}
\affiliation{%
  \institution{
  Xi'an Jiaotong University}
  \city{Xi'an}
  \country{China}
}

\author{Zheng Zhao}
\authornotemark[1]
\email{zheng.zhao@stu.xjtu.edu.cn}
\affiliation{%
  \institution{
  Xi'an Jiaotong University}
  \city{Xi'an}
  \country{China}
}

\author{Linsey Pang}
\email{panglinsey@gmail.com}
\affiliation{%
  \institution{PayPal}
  \city{San Jose}
  \country{USA}
}

\author{Bo Dong}
\email{dong.bo@xjtu.edu.cn}
\affiliation{%
  \institution{
  Xi'an Jiaotong University}
  \city{Xi'an}
  \country{China}
}

\author{Bin Shi}
\authornote{Corresponding authors: Bin Shi and Jiaxing Zhang.}
\email{shibin@xjtu.edu.cn}
\affiliation{%
  \institution{
  Xi'an Jiaotong University}
  \city{Xi'an}
  \country{China}
}

\author{Jiaxing Zhang}
\authornotemark[2]
\email{tabzhangjx@gmail.com}
\affiliation{%
  \institution{}
    \city{Bellevue}
     \country{USA}}



\renewcommand{\shortauthors}{Yin et al.}
\newcommand{\ie}{\emph{i.e.}}
\newcommand{\eg}{\emph{e.g.}}
\newcommand{\fix}{\marginpar{FIX}}
\newcommand{\new}{\marginpar{NEW}}
\newcommand{\stitle}[1]{\vspace{1ex}\noindent{\bf #1}}

\begin{abstract}
Graph Neural Networks (GNNs) have demonstrated remarkable performance across a range of applications involving graph-structured data, particularly in high-stakes domains. However, the opaque nature of their decision-making processes limits their trustworthiness and broader adoption. 
Existing post-hoc explanation methods aim to improve explainability by identifying subgraphs that influence GNN predictions and adopt mixup strategies to alleviate the out-of-distribution (OOD) issue caused by using subgraphs for prediction. Yet, these approaches typically rely on soft masks, which are inherently unable to fully eliminate label-irrelevant information, allowing redundant structures to leak into the mixup process and hindering the resolution of the OOD problem, thereby degrading explanation fidelity. In this work, we propose HPME, a \underline{H}ard-\underline{P}erturbation \underline{M}ixup \underline{E}xplanation framework grounded in a generalized Graph Information Bottleneck, which leverages graph pooling to extract discrete explanatory subgraphs and to yield an information-capacity bound to thoroughly compress label-irrelevant components.
Furthermore, we introduce a novel mixup strategy built upon structure-level replacement, generating in-distribution explanations to effectively mitigate the distribution shift.
Extensive experiments on diverse tasks demonstrate that HPME achieves state-of-the-art performance in generating robust and interpretable explanations across both synthetic and real-world datasets. The complete code and datasets can be found in our anonymous repository: 
\href{https://anonymous.4open.science/r/HPME-main-051D}{https://anonymous.4open.science/r/HPME-main-051D}.

\end{abstract}





\maketitle

\section{Introduction}
Graph Neural Networks (GNNs) are well-suited for processing graph-structured data~\citep{scarselli2008graph} and are widely applied to tasks such as community detection~\citep{huang2018overlapping}, traffic flow prediction~\citep{lei2022modeling}, recommendation systems~\citep{fan2020graph,du2022metakg}, and molecular modeling~\citep{gasteiger2021gemnet,liu2023generative}. Despite their success, the ``black-box'' nature of GNN decision-making hinders their adoption in high-stakes domains including healthcare~\citep{choi2020learning}, fraud detection~\citep{dou2020enhancing}, and drug discovery~\citep{qu2025rise}.

\begin{figure}[t]
\centering
\includegraphics[width=0.98\linewidth]{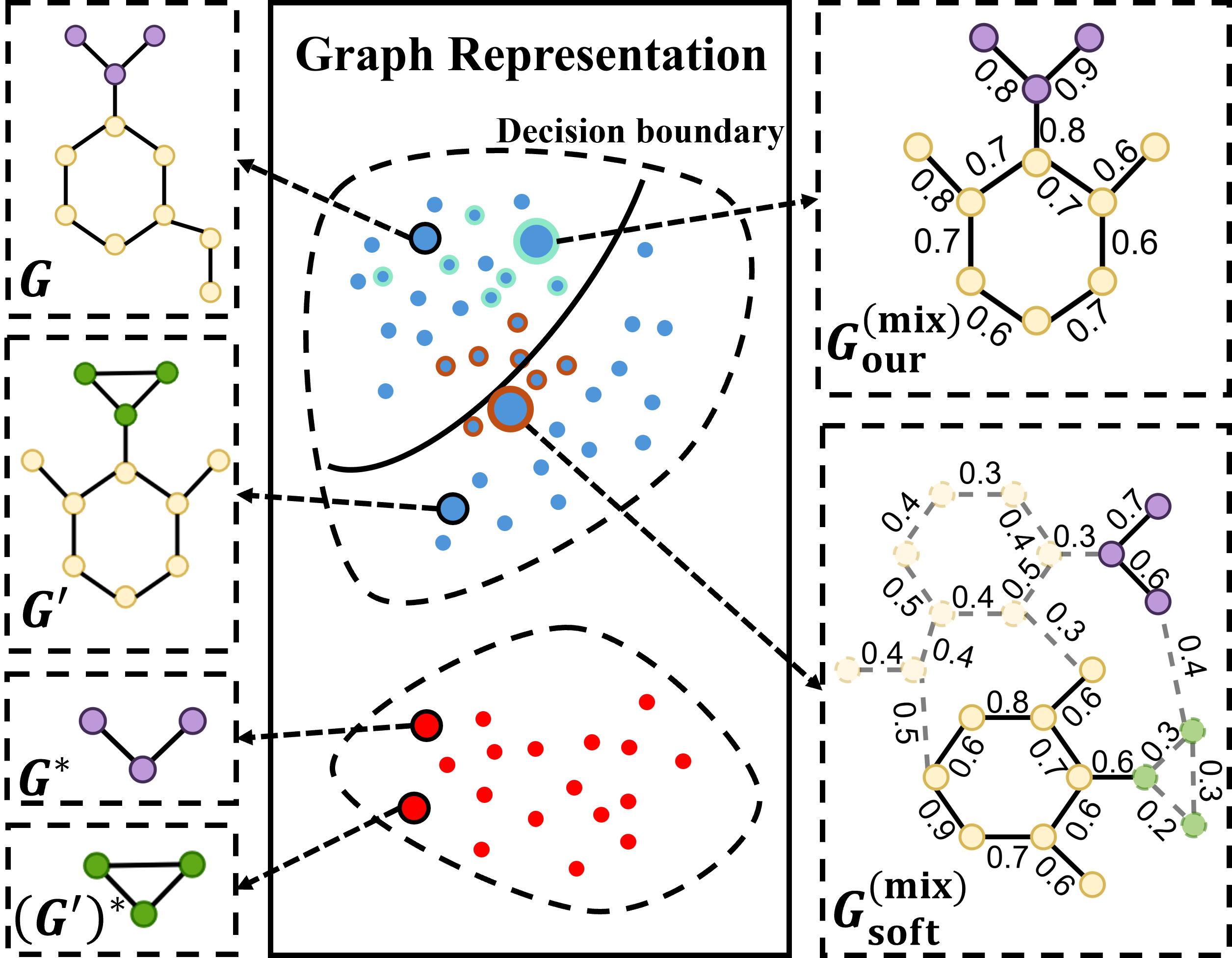}
\caption{Intuitive illustration of the OOD problem and mixup strategies.
The left part shows two original graphs $G$ and $G'$ with their corresponding explanatory subgraphs $G^{*}$ and $(G')^{*}$, while the right part presents mixup graphs generated by our structural mixup and the prior soft-mask-based mixup, denoted as $G^{(\text{mix})}_{\text{our}}$ and $G^{(\text{mix})}_{\text{soft}}$.
The middle part illustrates the distribution in the latent space, where the blue and red regions represent the original graphs and explanatory subgraphs. $G^{*}$ and $(G')^{*}$ deviate from the original graph distribution, whereas both $G^{(\text{mix})}_{\text{our}}$ and $G^{(\text{mix})}_{\text{soft}}$ fall inside it, with the latter drifting away from the decision cluster of $G$ due to redundant information. 
}
\label{OODMixup}
\vspace{-5pt}
\end{figure}

To address this challenge, recent research has focused on improving model explainability through model-level~\citep{yuan2020xgnn} or instance-level~\citep{ying2019gnnexplainer,luo2020parameterized} post-hoc explanatory methods, which aim to uncover the decision logic of GNNs by identifying explanatory subgraphs. Based on the Information Bottleneck (IB) principle~\citep{tishby2000information,tishby2015deep}, a previous work~\citep{wu2020graph} developed the Graph Information Bottleneck (GIB) framework, which formulates the explanation task as follows:
\begin{equation}
\label{eq:gib}
\arg \min_{G^{*}} I(G, G^{*}) - \alpha I(G^{*}, Y),
\end{equation}
where the objective aims to maximize the mutual information $I(G^{*}, Y)$ between the explanation $G^{*}$ and label $Y$, while minimizing $I(G, G^{*})$ to constrain the explanation size from the original graph $G$, with $\alpha$ balancing the two terms. Under the GIB framework, GFlowExplainer~\citep{li2023dag} learns to select informative nodes that preserve predictive relevance while ensuring compactness. MATCHExplainer~\citep{wu2023rethinking} adopts a non-parametric approach, matching shared subgraph patterns to identify concise explanatory subgraphs.

However, minimizing $I(G, G^{*})$ encourages the extraction of $G^{*}$ by removing label-irrelevant information from $G$, which may result in a distribution shift and lead to the out-of-distribution (OOD) problem~\citep{wang2021towards,yuan2021explainability}. To alleviate the distribution shift, 
MixupExplainer~\citep{zhang2023mixupexplainer} blends the explanatory subgraph with a label-irrelevant subgraph to generate augmented in-distribution graph instances. ProxyExplainer~\citep{chen2024generating} leverages graph autoencoders to reconstruct the explanatory and label-irrelevant subgraphs, and fuses them to generate proxy graphs.
Despite mitigating the OOD issue, existing methods still exhibit the following limitations, as illustrated in Figure~\ref{OODMixup}.
First, soft-mask-based explanatory methods struggle to drive edge weights toward a truly binary form, causing the extracted explanatory subgraphs to retain residual label-irrelevant information, thereby preventing the mixup graph \(G^{(\text{mix})}_{\text{soft}}\) from forming a clean composition of purely explanatory and label-irrelevant subgraphs.
Second, existing mixup strategies introduce uninterpretable edges when connecting explanatory and label-irrelevant subgraphs. For example, MixupExplainer relies on random sampling to establish such connections, while ProxyExplainer generates them via learned graph decoders. These synthetic and redundant edges lead to distributional shifts in the mixup graphs and ultimately compromise the fidelity of the explanations.

To address these challenges, we propose the HPME, a Hard-Perturbation Mixup Explanation framework for robust GNN explainability. HPME is built upon a generalized Graph Information Bottleneck with hard perturbations, an extension of GIB that explicitly integrates the graph pooling process into the optimization of $I(G, G^*)$, naturally imposing information compression to extract discrete explanatory subgraphs, and enabling thorough suppression of redundant structures inherent in soft-mask-based methods. 
To further address the distribution shift induced by estimating $I(G^*, Y)$, we design a novel structural mixup strategy built upon graph pooling, which fuses explanatory and label-irrelevant subgraphs via structural replacement, generating in-distribution, structurally faithful mixup graphs that preserve natural connectivity and improve explanation fidelity. The key contributions are as follows:


\noindent~$\bullet$ We extend the GIB framework by incorporating hard perturbations, leading to a generalized Graph Information Bottleneck that enables principled structural compression of explanations. This formulation leverages graph pooling to impose an explicit information-capacity bound, naturally extracting discrete explanatory subgraphs while suppressing label-irrelevant information.

\noindent~$\bullet$ We design a novel mixup strategy that structurally replaces explanatory subgraphs rather than soft mixup. This approach constructs mixup graphs that are structurally faithful to the original graphs, effectively mitigating the distribution shift problem while avoiding the generation of synthetic and uninterpretable edges.


\noindent~$\bullet$ Comprehensive experiments on synthetic and real-world datasets demonstrate that HPME consistently outperforms existing methods across diverse tasks, achieving up to a 30.1\% improvement in AUC, while maintaining explanation consistency and generalization.


\section{Related Work}
\stitle{GNN Explainability and Out-Of-Distribution. }
Current GNN explanatory methods~\citep{shin2024page,wang2022gnninterpreter,yu2024mage,zhang2022gstarx,wu2023rethinking,li2023dag, zhang2025confexplainer} aim to enhance model decision transparency through model-level or instance-level methods~\citep{yuan2022explainability}. Model-level methods~\citep{yuan2020xgnn} enhance GNN explainability by generating input-independent explanations, while instance-level methods~\citep{yuan2021explainability} identify input features most relevant to the model prediction. In this work, we focus on post-hoc instance-level methods. However, such methods face growing OOD challenges~\citep{zheng2023towards,fang2023evaluating}.
While some works~\citep{fang2024regularization,faber2020contrastive} have already made attempts, but they require that the model be retrained.
Separately, traditional mixup-based data augmentation methods~\citep{zhang2023mixupexplainer,chen2024generating,zhang2024regexplainer} also alleviate the OOD problem, while still struggle to ensure the quality of mixup graphs for approximating the original distribution.

\stitle{Graph Pooling. }
Graph pooling methods typically capture global information through hierarchical compression to obtain generalizable and expressive representations. Some methods learn a score function from node features and select the top $r$ nodes to form a new subgraph~\citep{gao2019graph}. For example, SAGPooling~\citep{lee2019self} refines node scoring by integrating structural context via GNNs. GSAPool~\citep{zhang2021hierarchical} enhances pooling process by dynamically fusing node features and topology information. DMIPool~\citep{zhao2023graph} further captures multi-level dependencies from dual-view representations to improve robustness. MLAP~\citep{itoh2022multi} preserves cross-layer structural information through layer-wise pooling, enhancing node distinguishability. Our method introduces pooling strategies to apply hard perturbations to graph structure, extracting more distinguishable explanatory and label-irrelevant subgraphs, ultimately enabling the mixup graphs to more faithfully approximate the original distribution.
\section{Preliminary}
\subsection{Notations and Problem Formulation}
We give a graph as $G=(\mathcal{V}, \mathcal{E}, \mX, \mA)$\footnote{To simplify notation, we denote graph as $G = (\mX, \mA)$ in the remainder of this paper.} from a graph dataset $\mathcal{G}$, where $\mathcal{V} = \{v_1, v_2, \ldots, v_n\}$ denotes the node set with $n$ denoting the number of nodes, and $\mathcal{E}\in\mathcal{V} \times \mathcal{V}$ denotes the edge set. The graph feature matrix is denoted by $\mX \in \mathbb{R}^{n \times d}$, where $d$ is the feature dimension and the $i$-th row $x_i \in \mathbb{R}^{i \times d}$ corresponds to node $v_i$. The adjacency matrix $\mA \in \{0, 1\}^{n \times n}$ of $G$ determines the edge set $\mathcal{E}$ such that $A_{ij} = 1$ if there is an edge$(i, j)\in\mathcal{E}$ and $A_{ij} = 0$ otherwise. In this paper, we focus on the graph-level classification and regression tasks, where node-level tasks can be converted to graph-level problems~\citep{ying2019gnnexplainer,luo2020parameterized}. Notably, each graph $G_{i}$ has a label $Y_{i}\in\mathcal{Y}$, where $i\in\{1,\ldots,\mathcal{N}\}$, $\mathcal{Y}$ represents the label set and $\mathcal{N}$ is the number of graphs in the dataset, with a pretrained GNN model $f$ to make prediction. 
We use the node embeddings matrix $\mH$ as the input for our explainer network $P_\theta$ and hard perturbation operator $Q_\phi$, which is extracted prior to the readout operation of the GNN $f(\cdot)$. In addition, we define $f_{\text{enc}}(\mX, \mA)$ as the component of $f$ that generates the graph representation $\vh_{G}$.

\begin{figure*}[t]
    \centering
    \includegraphics[width=\linewidth]{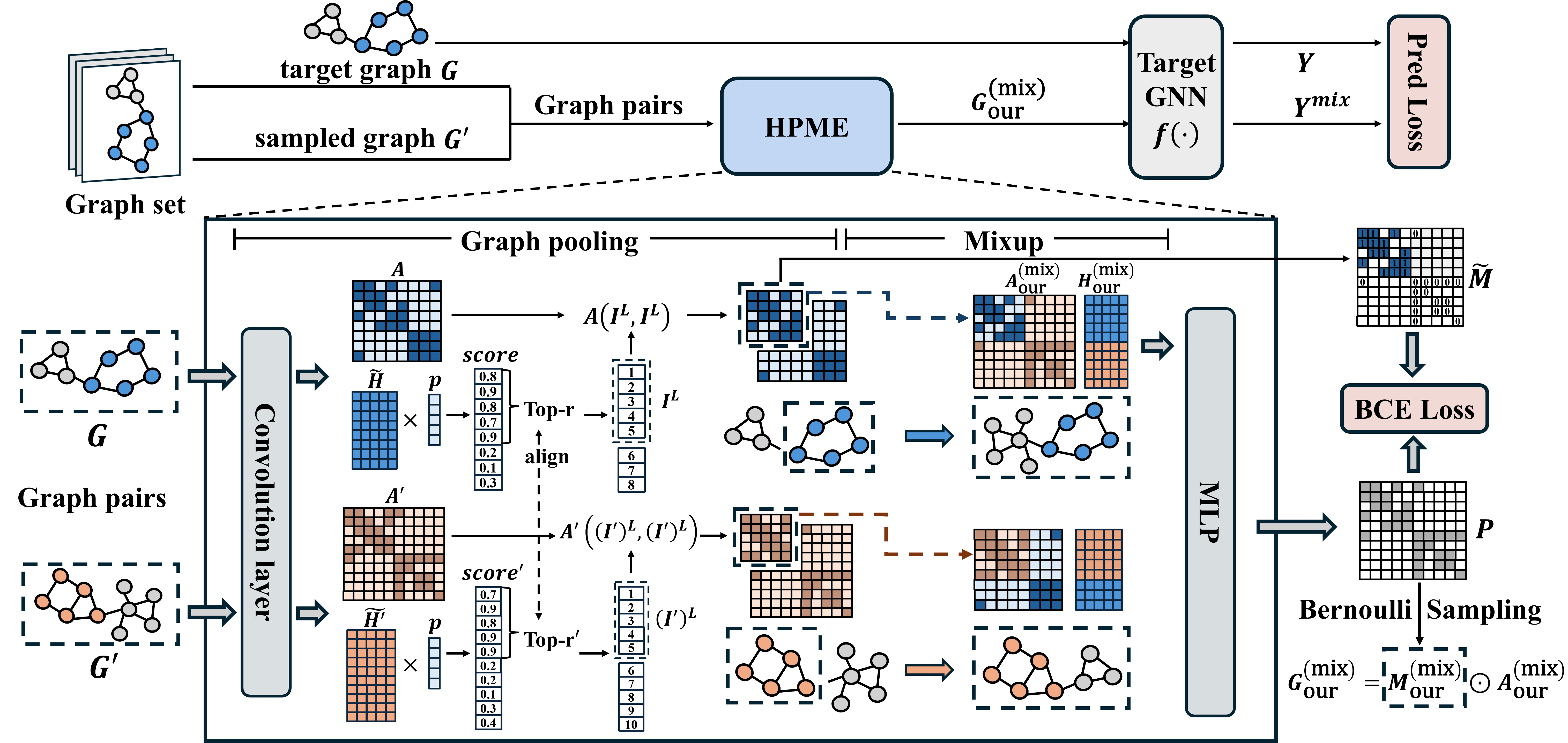}
    \caption{Overview of the HPME framework. HPME takes a target graph $G$ and a randomly sampled neighbor graph $G'$ as inputs. Graph pooling selects the top-$r$ nodes based on similarity scores to crop the adjacency matrix and obtain the pooled subgraph. Structural mixup then exchanges the pooled structures between $G$ and $G'$ to construct the mixup graph. An MLP is applied to the mixup graph to predict the edge existence probability $\mP$. Edge weights $\mW$ are sampled from the predicted probability distribution $\mP$, where $\mW$ is applied to the explanatory regions and $1-\mW$ to the label-irrelevant regions to construct the edge mask $\mM^{(\text{mix})}_{\text{our}}$. The final mixup graph is obtained as $G^{(\text{mix})}_{\text{our}} = (\mH^{(\text{mix})}_{\text{our}}, \mM^{(\text{mix})}_{\text{our}} \odot \mA^{(\text{mix})}_{\text{our}})$. The prediction loss evaluates the predictive fidelity of $G^{(\text{mix})}_{\text{our}}$ with respect to the target model $f$, while the BCE loss aligns the predicted edge probability distribution $\mP$ with the pooled subgraph distribution $\tilde{\mM}$.}
    \label{fig:wide}
\end{figure*}

\textbf{Problem 1 }(Post-hoc Instance-level GNN Explanatory Method). Given a pretrained GNN model $f$ and an arbitrary graph $G \in \mathcal{G}$, the objective of \emph{post-hoc instance-level GNN explanatory method} is to find a subgraph $G^{*}\in G$ that explain the prediction of $f$ on $G$.


\section{Methods}

In this section, we first present a Generalized Graph Information Bottleneck, a new theoretical formulation that extends the GIB framework with hard structural perturbations to effectively compress task-irrelevant information while preserving predictive fidelity. 
Based on this formulation, we instantiate the framework via a graph-pooling-based parameterization network to extract discrete explanatory subgraphs, and propose a structural mixup strategy to generate in-distribution mixup graphs to alleviate the distribution shift induced by subgraph prediction. 
Finally, we describe the overall training objective, which jointly optimizes prediction fidelity and structural alignment to train the explainer. 
An overview of the proposed framework is illustrated in Figure~\ref{fig:wide}.


\subsection{Generalized Graph Information Bottleneck with Hard Perturbations}
According to previous works~\citep{wu2020graph, ying2019gnnexplainer}, the mutual information term $I(G^*, Y)$ in ~\Eqref{eq:gib} can be decomposed as $I(G^*, Y) = H(Y) - H(Y | G^*)$. Since the marginal entropy $H(Y)$ is determined solely by the dataset and remains constant during optimization, maximizing the mutual information is equivalent to minimizing the conditional entropy $H(Y | G^*)$. Consequently, by substituting $I(G^*, Y)$ in ~\Eqref{eq:gib}, the optimization objective is defined as:
\begin{equation}
    \label{eq:gib_derived}
    \arg \min_{G^*} I(G, G^*) + \alpha H(Y | G^*).
\end{equation}
In the following parts, we introduce a hard perturbation parameterization network via graph pooling to optimize $I(G, G^*)$ and a structural mixup strategy to faithfully estimate $H(Y | G^*)$.

\subsubsection{Optimization of $I(G, G^*)$ via Hard Perturbations}
Minimizing the mutual information term $I(G, G^*)$ in the GIB objective corresponds to the structural compression process that removes label-irrelevant information from the explanation. Existing post-hoc explanation methods typically adopt a parameterized explainer $P_\theta(G^*_\theta|G)$ to extract soft mask explanations, and enforce information compression through additional regularization terms, such as size constraints \citep{ying2019gnnexplainer, luo2020parameterized} or Kullback--Leibler (KL) divergence objectives \citep{miao2022interpretable,chen2024generating}. However, due to the continuous and probabilistic nature of soft masks, these approaches often struggle to completely eliminate label-irrelevant structures during optimization, leading to residual redundancy in the generated explanations\citep{funke2021hard,ma2025c2explainer}.

To overcome the above limitation, we introduce a hard perturbation operator $Q_\phi(G^*_{\text{pool}} | G)$ implemented by graph pooling. Given an input graph $G$, $Q_\phi$ performs hard node selection to produce a sparsified pooled subgraph $G^*_{\text{pool}}$, which provides a discrete structural reference for compressing explanations. Let $\Omega_N$ denote the space of graphs with $N$ nodes, and let $\Omega_k$ denote the subspace of pooled subgraphs obtained by retaining $k$ selected nodes via pooling, where $k \ll N$. Since pooling maps graphs from $\Omega_N$ to $\Omega_k$, it imposes an inherent information bottleneck on the resulting subgraph. We formalize this property as follows.

\noindent
\textbf{Property 1.}
\textit{$I(G, G^*_{\text{pool}})$ is upper-bounded by the entropy of the reduced subspace.}

The mutual information between the original graph $G$ and the pooled subgraph $G^*_{\text{pool}}$ satisfies the following bound:
\begin{equation}
\label{eq:pool_bound}
I(G, G^*_{\text{pool}}) \le H(G^*_{\text{pool}}) \le \log |\Omega_k|.
\end{equation}

This bound indicates that $G^*_{\text{pool}}$ inherently satisfies the compression requirement in GIB by restricting the information capacity to $\log|\Omega_k|$.
To transfer the bottleneck effect of $Q_\phi$ to the learnable explainer $P_\theta$, we introduce the following property, which establishes that minimizing the divergence between $P_\theta$ and $Q_\phi$ provides a theoretically grounded way to compress $I(G, G^*_\theta)$.

\noindent
\textbf{Property 2.}
\textit{$I(G, G^*_\theta)$ is upper-bounded via distributional alignment between $P_\theta$ and $Q_\phi$.}

For the explanatory subgraph distribution $P_\theta(G^*_\theta | G)$ and the pooled subgraph distribution $Q_\phi(G^*_{\text{pool}} | G)$, the mutual information between $G$ and $G^*_\theta$ is bounded as follows:
\begin{equation}
\label{eq:distill_bound}
I(G, G^*_\theta) \le I(G, G^*_{\text{pool}}) 
+ C \cdot \mathbb{E}_G \left[ \sqrt{KL(Q_\phi(\cdot|G) \,||\, P_\theta(\cdot|G))} \right],
\end{equation}
where $C$ is a constant that depends on the graph size. The detailed proof is provided in Appendix~\ref{sec:proof_prop2}.
Since $I(G, G^*_{\text{pool}})$ is already constrained by the upper bound $\log|\Omega_k|$, the information capacity of $G^*_{\text{pool}}$ is effectively compressed. Therefore, reducing $I(G, G^*_\theta)$ primarily relies on minimizing the divergence term, which correspondingly drives $I(G, G^*_\theta)$ toward the same compressed regime as $I(G, G^*_{\text{pool}})$.
To instantiate the KL divergence, we adopt the Erd\H{o}s--R\'enyi assumption \citep{erdos1960evolution} and model the existence of each edge as an independent Bernoulli variable. Under this formulation, minimizing the KL divergence between $Q_\phi$ and $P_\theta$ is equivalent to minimizing the Binary Cross-Entropy (BCE) loss \citep{chen2023d4explainer}. Accordingly, the divergence term is instantiated as:
\begin{equation}
\label{eq:bce}
    \mathcal{L}_{\text{BCE}} = - \sum_{(u,v) \in \mathcal{E}} \left( q_{uv} \log p_{uv} + (1-q_{uv}) \log (1-p_{uv}) \right),
\end{equation}
where $q_{uv} \in \{0,1\}$ denotes discrete edge variable sampled from $Q_\phi$, and $p_{uv} \in (0,1)$ represents the edge existence probability parameterized by the explainer $P_\theta$.
Consequently, by substituting $I(G, G^*)$ in ~\Eqref{eq:gib_derived} with its derived upper bound and replacing $H(Y | G^*)$ with $H(Y | G^*_{\text{pool}})$, which is formally justified in Appendix~\ref{sec:proof_replacement}, the objective in ~\Eqref{eq:gib_derived} is reformulated as:
\begin{equation}
    \min_{\theta, \phi} I(G, G^*_{\text{pool}}) + C \cdot \mathbb{E}_G \left[ \sqrt{KL(Q_\phi(\cdot|G) \,||\, P_\theta(\cdot|G))} \right] + \alpha H(Y | G^*_{\text{pool}}).
    \label{eq:gib_tractable}
\end{equation}

\subsubsection{Estimation of $H(Y | G^*)$ via Structural Mixup}

To optimize the second term $H(Y | G^*)$ in ~\Eqref{eq:gib_derived}, existing methods typically adopt Cross-Entropy (CE) for classification~\cite{ying2019gnnexplainer, luo2020parameterized} or Mean Squared Error (MSE) for regression~\cite{zhang2024regexplainer}. This approximation is generally implemented by applying the pre-trained model $f$ to the explanatory subgraph $G^*$, resulting in the prediction $Y^* = f(G^*)$. However, this strategy suffers from an OOD issue~\cite{zhang2023mixupexplainer,chen2023d4explainer}, since the model is trained on full graphs while the sparse subgraphs $G^*$ reside in a different distribution space. Consequently, predictions based on $f(G^*)$ can be unreliable and biased.


To address the distributional gap, we follow MixupExplainer~\cite{zhang2023mixupexplainer} and introduce a subgraph $G^\Delta$ sampled from the label-independent distribution $\mathbb{P}_{\Delta}$, where $G^\Delta$ is constructed by removing the pooled subgraph $G^*_{\text{pool}}$ from the original graph $G$. Incorporating variable $G^\Delta$, the optimization objective is reformulated as:
\begin{equation}
    \min_{\theta, \phi} I(G, G^*_{\text{pool}}) + C \cdot \mathbb{E}_G \left[ \sqrt{KL(Q_\phi(\cdot|G) \,||\, P_\theta(\cdot|G))} \right] + \alpha H(Y | G^{(\text{mix})}_{\text{our}}),
    \label{eq:gib_last}
\end{equation}
\begin{equation*}
    \text{where } G^{(\text{mix})}_{\text{our}} = G^*_{\text{pool}} \cup G^\Delta, \quad G^\Delta \sim \mathbb{P}_{\Delta},
\end{equation*}
\begin{equation*}
    \text{s.t. } I(G^\Delta; Y | G^*_{\text{pool}}) = 0.
\end{equation*}

Here, $G^{(\text{mix})}_{\text{our}}$ denotes a mixup graph obtained by structurally splicing the hard-perturbation subgraph $G^*_{\text{pool}}$ with the residual subgraph $G^\Delta$, which remains within the original data distribution while preserving the task-relevant information contained in $G^*_{\text{pool}}$.
We further establish the equivalence between ~\Eqref{eq:gib_last} and ~\Eqref{eq:gib_tractable}. By the chain rule of conditional entropy, we have $H(Y | G^{(\text{mix})}_{\text{our}}) = H(Y | G^*_{\text{pool}}, G^\Delta) = H(Y | G^*_{\text{pool}}) + I(G^\Delta; Y | G^*_{\text{pool}})$. Since the constraint $I(G^\Delta; Y | G^*_{\text{pool}}) = 0$ holds, it follows that $H(Y | G^{(\text{mix})}_{\text{our}}) = H(Y | G^*_{\text{pool}})$. This equivalence indicates that minimizing the conditional entropy on the in-distribution graph $G^{(\text{mix})}_{\text{our}}$ preserves the fidelity of the explanatory subgraph while alleviating the distribution shift issue.

\begin{figure}[t]
    \centering
    \includegraphics[width=\columnwidth]{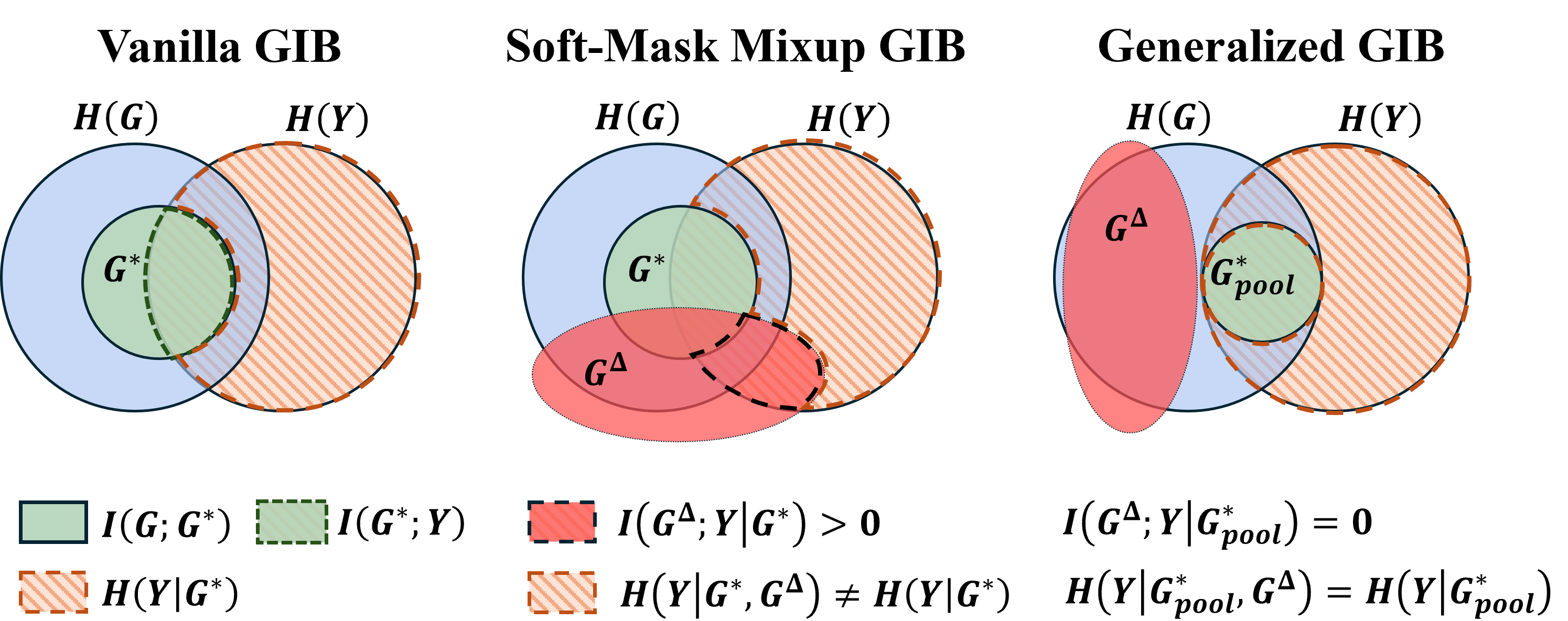}
    \caption{Comparison among vanilla GIB, soft-mask-based mixup GIB, and the proposed Generalized GIB with hard perturbations. Soft-mask-based mixup inevitably produces a residual subgraph $G^\Delta$ that may contain label-relevant structures, resulting in $I(G^\Delta; Y | G^*)>0$ and breaking the equivalence with the vanilla GIB objective. In contrast, our hard-perturbation structural mixup removes $G^*_{\text{pool}}$ through explicit structural pruning, avoiding label-relevant information in $G^\Delta$ and enabling more faithful explanation extraction.}
    \label{fig:gib}
\end{figure}

Unlike soft-mask-based mixup strategies, our hard-perturbation structural mixup constructs $G^\Delta$ by explicitly removing $G^*_{\text{pool}}$ from $G$, ensuring that $G^\Delta$ contains no label-relevant information from the explanation. 
This mechanism effectively enforces $I(G^\Delta; Y | G^*_{\text{pool}})=0$, providing a more reliable equivalence between $H(Y | G^{(\text{mix})}_{\text{our}})$ and $H(Y | G^*_{\text{pool}})$, as illustrated in Figure~\ref{fig:gib}.
Accordingly, we define $\mathcal{L}_{\text{pred}}$ as the prediction loss computed on the mixup graph $G^{(\text{mix})}_{\text{our}}$ using the model output $f(G^{(\text{mix})}_{\text{our}})$.
\subsection{Structural Mixup with Graph Pooling}
Building upon the generalized GIB objective with hard perturbations, HPME instantiates the framework by using a graph pooling strategy to enforce structural compression on the original graph and a structural mixup strategy to mitigate distribution shifts, ultimately generating soft mask for the explanatory subgraph. Each stage is detailed as follows.

\subsubsection{Extracting the Pooled Subgraph $G^*_{\text{pool}}$ }

The graph pooling process can be implemented as multiple pooling and Graph Convolution layers~\citep{kipf2016semi}, as shown below:
\begin{equation}
\label{eq:graph_pooling}
\tilde{\mH}^{l} = \text{Conv}^{l}(\mH^{l-1}, \mA^{l-1}), 
\quad
\mA^{l}, \mH^{l} = \text{pooling}(\tilde{\mH}^{l}, \mA^{l-1}, r_{l}),
\end{equation}
where $l$ denotes the $l$-th layer, $\text{Conv}^{l}$ is a single Graph Convolution layer that updates node embeddings, the pooling layer preserves the Top-$r_{l}$ most essential nodes with $r_{l}$ denoting the ratio of nodes preserved, and $\mH^{l}$ and $\mA^{l}$ represent the node embeddings and adjacency matrix. More specifically, in the $l$-th pooling layer, $\text{score} = \tilde{\mH}^{l} \cdot \vp_{l}/\ \|\vp_{l}\|$ measures the directional similarity between node embeddings $\tilde{\mH}^{l}$ and a learnable projection vector $p_{l}$. The $\lfloor r_l \cdot n_{l-1}\rfloor$ nodes with the highest scores are preserved, where $n_{l{-}1}$ denotes the number of nodes in the subgraph at the $(l-1)$-th layer. The preserved node indices are denoted as $I^{l}$. After $L$ pooling layers, the graph pooling process progressively crops the original graph and yields the final preserved node indices $I^{L}=\{i_{1},...,i_{m}\}\subseteq \{1,...,n\}$, with $m$ denoting the number of preserved node indices. Ultimately, the pooled subgraph is represented as $G^*_{\text{pool}} = (\mH^{*}, \mA^{*})$, where $\mH^{*} = \mH(I^{L}, :)$ and $\mA^{*} = \mA(I^{L},I^{L})$. 
\subsubsection{Structural Mixup for Constructing $G^{(\text{mix})}_{\text{our}}$}
To generate mixup graph that remains close to the original graph distribution, we replace the pooled subgraph from a randomly sampled graph $G'$ with that from the input graph $G$.
Let $\mA'$ and $\mH'$ denote the adjacency and feature matrices of $G'$, and let $(\mA')^* \in \{0,1\}^{m \times m}$ and $\mA^* \in \{0,1\}^{m \times m}$ denote the adjacency matrices of the pooled subgraphs extracted from $G'$ and $G$, respectively, where $m$ is the number of preserved nodes. The $\mS' \in \{0,1\}^{n' \times m}$ is a binary selection matrix that maps the preserved node indices $(I')^L$ of $G'$ to the full node space, constructed as $\mS' = \mI_{n'}[:, (I')^L]$.
We define the mixup graph $G^{(\text{mix})}_{\text{our}} = (\mH^{(\text{mix})}_{\text{our}}, \mA^{(\text{mix})}_{\text{our}})$ as:
\begin{equation}
\label{eq:str_mixup}
\begin{aligned}
\mA^{(\text{mix})}_{\text{our}} &= \mA' - \mS'(\mA')^*\mS'^\top + \mS'\mA^*\mS'^\top, \\
\mH^{(\text{mix})}_{\text{our}} &= \mH' - \mS'(\mH')^* + \mS'\mH^*.
\end{aligned}
\end{equation}

Intuitively, we replace the pooled subgraph of $G'$ with that of $G$ at the same node positions, enabling structure-level replacement while preserving semantic locality. The number of preserved node indices $m$ is aligned across graphs to ensure compatibility of subgraph replacement. 
The mixup graph $G^{(\text{mix})}_{\text{our}}$ is used in~\Eqref{eq:gib_last} to replace $G^*$ for predictive fidelity estimation, thereby mitigating the distribution shift introduced during prediction process.
\subsubsection{Generating the Soft Explanation Mask}

In practice, following the existing works~\cite{luo2020parameterized,luo2024towards}, we apply the binary concrete distributions to approximate the Bernoulli distributions~\cite{maddison2016concrete,jang2016categorical}. Specifically, an MLP is employed to take concatenated node embeddings \( [\vh_i; \vh_j] \) from \( \mH^{\text{(mix)}}_{\text{our}} \) and output the parameter \( w_{ij} \) for each edge\( (i,j) \). And then, the $p_{ij}\in(0,1)$ of Bernoulli distribution in ~\Eqref{eq:bce} can be obtained by $p_{ij} = \frac{\exp(w_{ij})}{1+\exp(w_{ij})}$ and $\hat{w}_{ij} = \sigma((\log \epsilon - \log(1 - \epsilon) + w_{ij})/\tau)$, where $\epsilon \sim \text{Uniform}(0, 1)$, \( \tau \) is the temperature, and $\sigma(\cdot)$ is the Sigmoid function.
The final mixup graph is defined as \( G^{\text{(mix)}}_{\text{our}} = (\mH^{\text{(mix)}}_{\text{our}}, \mM^{\text{(mix)}}_{\text{our}} \odot \mA^{\text{(mix)}}_{\text{our}}) \), where \(\mM_{\text{our}}^{(\text{mix})}\) can be expressed as \( \mM_{\text{our}}^{(\text{mix})} = (1 {-} \mW) \odot (\mA' {-} \mS'(\mA')^* \mS'^\top) {+} \mW \odot (\mS' \mA^* \mS'^\top) \), where the weight matrix \( \mW \) consists of entries \( \hat{w}_{ij} \).
\subsection{Overall Loss Function}

\subsubsection{Binary Cross-Entropy Loss}

To ensure that the generated edge weights are faithful to the explanatory and label-irrelevant subgraphs obtained from the graph pooling process, the BCE loss is defined as follows:
\begin{equation}
\label{eq:bce_loss}
\mathcal{L}_{\text{BCE}} = - \sum_{(i,j) \in \mathcal{E}^{(\text{mix})}} \left[ \tilde{M}_{ij} \log p_{ij} + (1 - \tilde{M}_{ij}) \log (1 - p_{ij}) \right],
\end{equation} 
where \( \tilde{\mM}= \mS' \mA^* \mS'^\top \) denotes the discrete distribution derived from the pooled subgraph, $p_{ij}$ represents the edge existence probability, and $\mathcal{E}^{(\text{mix})}$ denotes the edge set of the mixup graph.

\begin{table*}[t]
\centering
\caption{AUC-ROC edge-level explanation accuracy on seven graph classification and six graph regression datasets. Higher scores indicate better performance. SingleMotif datasets contain a single explanatory structure, whereas MultipleMotif datasets include multiple structures.}
\label{classification_regression}
\renewcommand{\arraystretch}{1.0} 
\begin{tabular*}{\linewidth}{@{\extracolsep{\fill}}lccccccc}
\toprule
& \multicolumn{3}{c}{SingleMotif (Classification)} & \multicolumn{4}{c}{MultipleMotif (Classification)} \\
\cmidrule(lr){2-4} \cmidrule(lr){5-8}
Method & \makecell{BA-\\2motifs} & \makecell{BA-\\HouseGrid} & SPMotif 
& \makecell{BA-\\HouseAndGrid} & \makecell{Alkane-\\Carbonyl} & \makecell{Fluorid-\\Carbonyl} & Benzene \\
\midrule
Grad           & $0.752\pm0.009$ & $0.609\pm0.005$ & $0.644\pm0.007$ & $0.419\pm0.010$ & $0.526\pm0.009$ & $0.687\pm0.006$ & $0.506\pm0.005$ \\
MetaGNN        & $0.665\pm0.197$ & $0.840\pm0.096$ & $0.631\pm0.102$ & $0.806\pm0.069$ & $0.762\pm0.060$ & $0.667\pm0.041$ & $0.658\pm0.175$ \\
GNNExplainer   & $0.512\pm0.004$ & $0.503\pm0.005$ & $0.510\pm0.003$ & $0.507\pm0.005$ & $0.512\pm0.012$ & $0.520\pm0.006$ & $0.497\pm0.004$ \\
PGExplainer    & $0.677\pm0.069$ & $0.611\pm0.221$ & $0.607\pm0.023$ & $0.731\pm0.199$ & $0.619\pm0.373$ & $0.635\pm0.079$ & $0.750\pm0.182$ \\
TAGExplainer   & $0.676\pm0.148$ & $0.848\pm0.039$ & $0.531\pm0.026$ & $0.647\pm0.072$ & $0.842\pm0.059$ & $0.752\pm0.046$ & $0.760\pm0.067$ \\
MatchExplainer & $0.802\pm0.001$ & $0.757\pm0.001$ & $0.499\pm0.000$ & $0.773\pm0.001$ & $0.603\pm0.002$ & $0.779\pm0.000$ & $0.512\pm0.001$ \\
MixupExplainer & $0.878\pm0.107$ & $0.811\pm0.078$ & $0.631\pm0.082$ & $0.804\pm0.190$ & $0.791\pm0.148$ & $0.686\pm0.049$ & $0.796\pm0.148$ \\
ProxyExplainer & $0.896\pm0.029$ & $0.745\pm0.327$ & $0.607\pm0.079$ & $0.704\pm0.191$ & $0.859\pm0.097$ & $0.729\pm0.129$ & $0.809\pm0.129$ \\
Ours           & $\bm{0.971\pm0.021}$ & $\bm{0.965\pm0.026}$ & $\bm{0.748\pm0.123}$ & $\bm{0.979\pm0.004}$ & $\bm{0.944\pm0.020}$ & $\bm{0.807\pm0.011}$ & $\bm{0.861\pm0.004}$ \\
\bottomrule
\end{tabular*}

\vspace{5pt} 

\begin{tabular*}{\linewidth}{@{\extracolsep{\fill}}lcccccc}
\toprule
& \multicolumn{2}{c}{SingleMotif (Regression)} & \multicolumn{4}{c}{MultipleMotif (Regression)} \\
\cmidrule(lr){2-3} \cmidrule(lr){4-7}
Method 
& \makecell{BA-Motif\\-Volume}  
& \makecell{House-Grid\\-Volume}  
& \makecell{BA-Motif\\-Counting}  
& Triangles  
& Crippen  
& \makecell{House-OrGrid\\-Volume} \\
\midrule
Grad           & $0.448\pm0.000$ & $0.544\pm0.000$ & $0.498\pm0.000$ & $0.587\pm0.000$ & $0.540\pm0.000$ & $0.477\pm0.000$ \\
ATT            & $0.512\pm0.002$ & $0.499\pm0.002$ & $0.512\pm0.003$ & $0.512\pm0.003$ & $0.501\pm0.003$ & $0.521\pm0.003$ \\
TAGExplainer   & $0.548\pm0.262$ & $0.856\pm0.105$ & $0.763\pm0.354$ & $0.610\pm0.163$ & $0.486\pm0.004$ & $0.698\pm0.167$ \\
MixupExplainer & $0.741\pm0.205$ & $0.854\pm0.082$ & $0.613\pm0.370$ & $0.561\pm0.139$ & $0.530\pm0.016$ & $0.721\pm0.128$ \\
RegExplainer   & $0.766\pm0.139$ & $0.858\pm0.062$ & $\bm{0.946\pm0.095}$ & $0.560\pm0.132$ & $0.497\pm0.004$ & $0.741\pm0.107$ \\
Ours           & $\bm{0.997\pm0.001}$ & $\bm{0.966\pm0.003}$ & $\bm{0.946\pm0.004}$ & $\bm{0.648\pm0.020}$ & $\bm{0.565\pm0.038}$ & $\bm{0.941\pm0.012}$ \\
\bottomrule
\end{tabular*}
\end{table*}

\subsubsection{Overall Loss Function}

The final overall loss is formulated as follows:
\begin{equation}
\label{eq:finall_loss}
  \mathcal{L} = \mathcal{L}_{\text{pred}} + \beta \mathcal{L}_{\text{BCE}}(\tilde{\mM}, \mP),
\end{equation}
where $\beta$ is a hyperparameter balancing the contribution of two parts, $\tilde{\mM}$ is a discrete distribution matrix derived from the adjacency matrix of the pooled subgraph, and $\mP$ is the predicted edge existence probability matrix consisting of entries $p_{ij}$. $\mathcal{L}_{\text{pred}}$ denotes the prediction loss, which employs $\mathcal{L}_{\text{CE}}(f(G), f(G^{(\text{mix})}_{\text{our}}))$ for classification tasks and $\mathcal{L}_{\text{MSE}}(f(G), f(G^{(\text{mix})}_{\text{our}}))$ for regression tasks, respectively. The training algorithm and computational complexity analysis are provided in Appendix~\ref{sec:alg_detail}.

\section{Experiments}
To evaluate the effectiveness of HPME, we conduct experiments on synthetic and real-world datasets covering both classification and regression tasks. 
First, the basic experimental setup is presented in \Secref{sec:exp:setup}. 
Then, we evaluate the performance of HPME against baselines in identifying explanatory subgraphs in \Secref{sec:exp:performance}. 
Furthermore, the effectiveness of HPME in addressing the distribution shift problem is demonstrated in \Secref{sec:exp:ood} and ablation studies in \Secref{sec:exp:ablation}. 
Lastly, we present case studies in \Secref{sec:exp:case} to examine whether HPME accurately extracts explanatory subgraphs and generates natural in-distribution mixup graphs compared with baseline methods. Additional experiments, including hyperparameter sensitivity analysis, multi-task explanation quality evaluation, backbone models analysis, evaluation with alternative explanation metrics, and case studies can be found in Appendix~\ref{sec:app:moreex}.

\begin{table*}[t]
\centering
\caption{Cosine similarity and euclidean distance are computed between the graph representations of the original graph and those of the ground-truth explanatory subgraph (GT), ProxyExplainer (ProxyE), RegExplainer (RegE), and our method (HPME). Higher values for cosine similarity and lower values for Euclidean distance denote greater distributional similarity. Bold values indicate better performance.}
\label{tab:exp_metrics}

\setlength{\tabcolsep}{2pt}
\renewcommand{\arraystretch}{1.05}
\normalsize

\resizebox{1.0\linewidth}{!}{
\begin{tabular}{
l 
ccc p{2pt}
ccc p{2pt}
ccc p{2pt}
ccc p{2pt}
ccc p{2pt}
ccc
}

\toprule[0.8pt]

& \multicolumn{3}{c}{\textbf{BA-HouseGrid}} &&
  \multicolumn{3}{c}{\textbf{Benzene}} &&
  \multicolumn{3}{c}{\textbf{Fluoride-Carbonyl}} &&
  \multicolumn{3}{c}{\textbf{Crippen}} &&
  \multicolumn{3}{c}{\textbf{BA-Motif-Volume}} &&
  \multicolumn{3}{c}{\textbf{Triangles}} \\

\cmidrule(lr){2-4}
\cmidrule(lr){6-8}
\cmidrule(lr){10-12}
\cmidrule(lr){14-16}
\cmidrule(lr){18-20}
\cmidrule(lr){22-24}

\textbf{Metric} 
& GT & ProxyE & HPME &&
  GT & ProxyE & HPME &&
  GT & ProxyE & HPME &&
  GT & RegE & HPME &&
  GT & RegE & HPME &&
  GT & RegE & HPME \\

\midrule[0.5pt]

\textbf{Cosine} $\uparrow$ 
& 0.688 & 0.802 & \textbf{0.868} &&
  0.821 & 0.565 & \textbf{0.941} &&
  0.918 & 0.723 & \textbf{0.937} &&
  0.869 & 0.847 & \textbf{0.936} &&
  0.696 & 0.876 & \textbf{0.953} &&
  0.897 & 0.953 & \textbf{0.985} \\

\textbf{Euclidean} $\downarrow$
& 1.044 & 0.706 & \textbf{0.661} &&
  0.953 & 1.471 & \textbf{0.578} &&
  0.684 & 1.272 & \textbf{0.604} &&
  0.653 & 0.794 & \textbf{0.431} &&
  1.138 & 0.775 & \textbf{0.458} &&
  0.389 & 0.247 & \textbf{0.143} \\

\bottomrule[0.8pt]

\end{tabular}}
\end{table*}

\begin{figure*}[ht]
  \centering
  \begin{subfigure}[t]{0.15\textwidth}
    \centering
    \includegraphics[width=\linewidth]{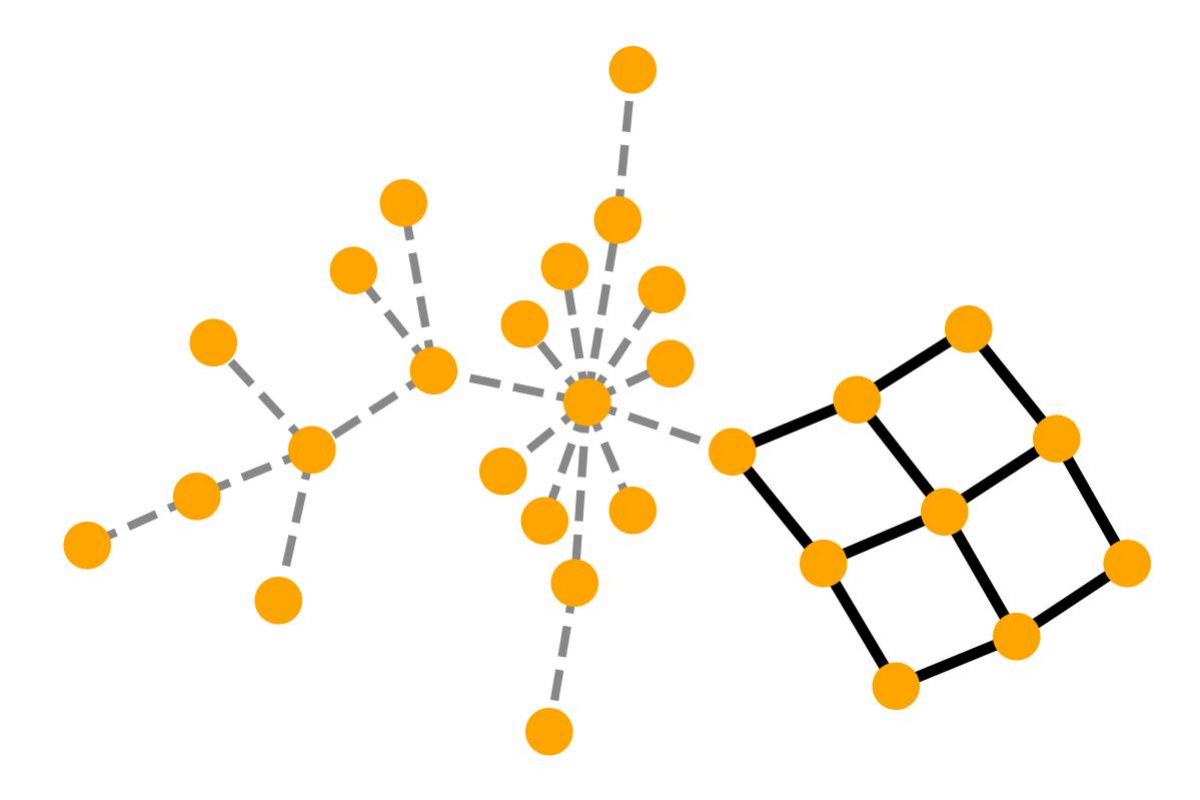}
    \caption{Ground Truth}
  \end{subfigure}
  \hfill
  \begin{subfigure}[t]{0.15\textwidth}
    \centering
    \includegraphics[width=\linewidth]{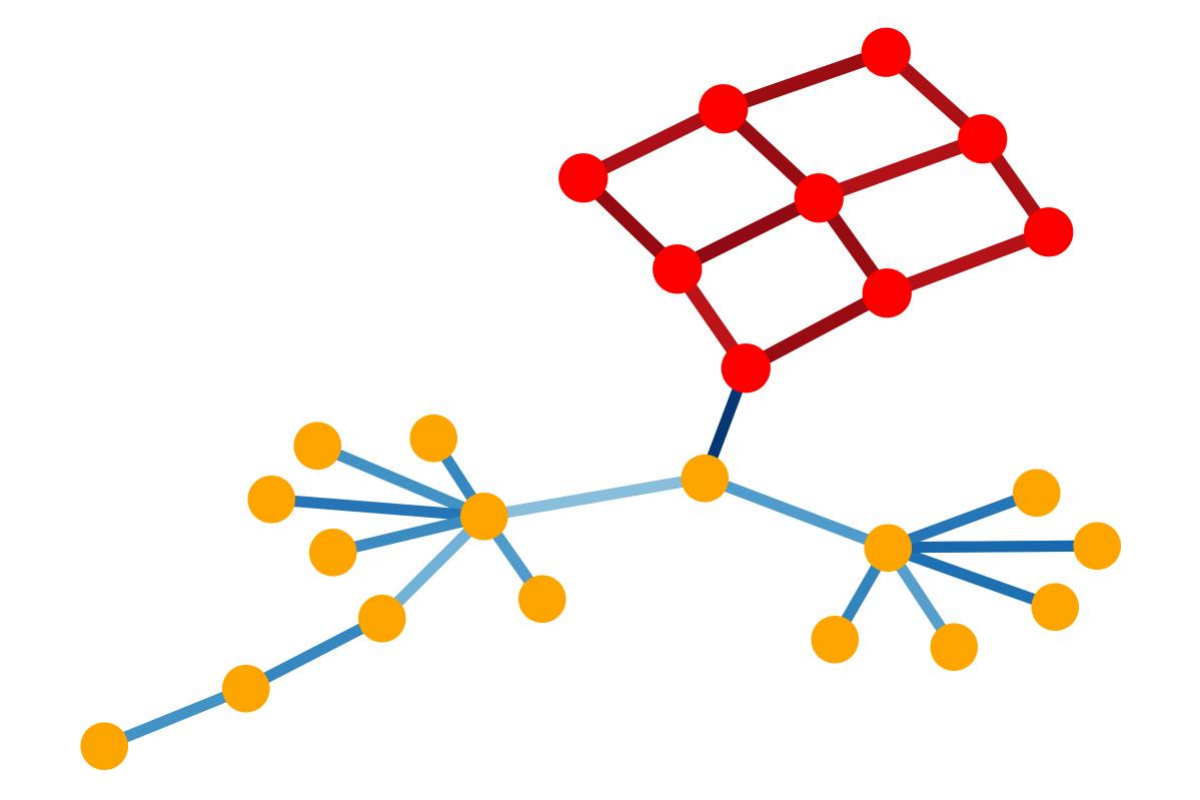}
    \caption{HPME}
  \end{subfigure}
  \hfill
  \begin{subfigure}[t]{0.15\textwidth}
    \centering
    \includegraphics[width=\linewidth]{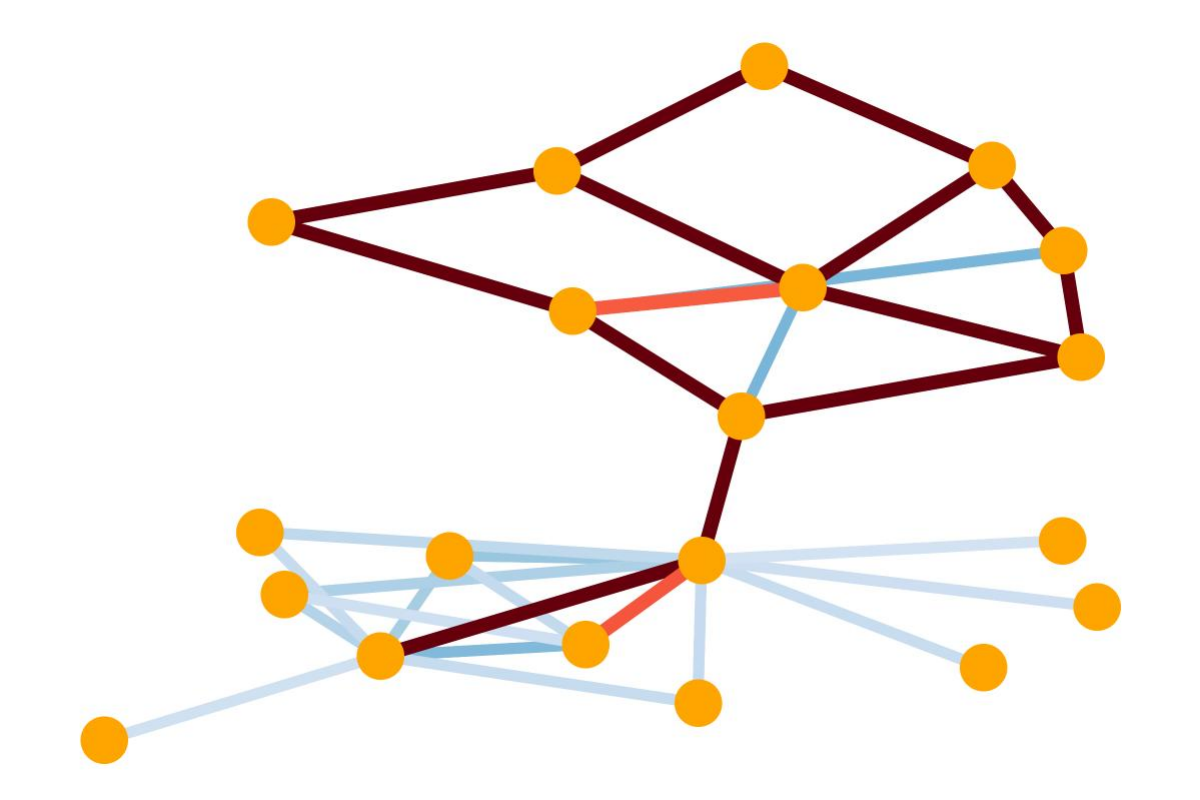}
    \caption{ProxyExplainer}
  \end{subfigure}
  \hfill
  \begin{subfigure}[t]{0.15\textwidth}
    \centering
    \includegraphics[width=\linewidth]{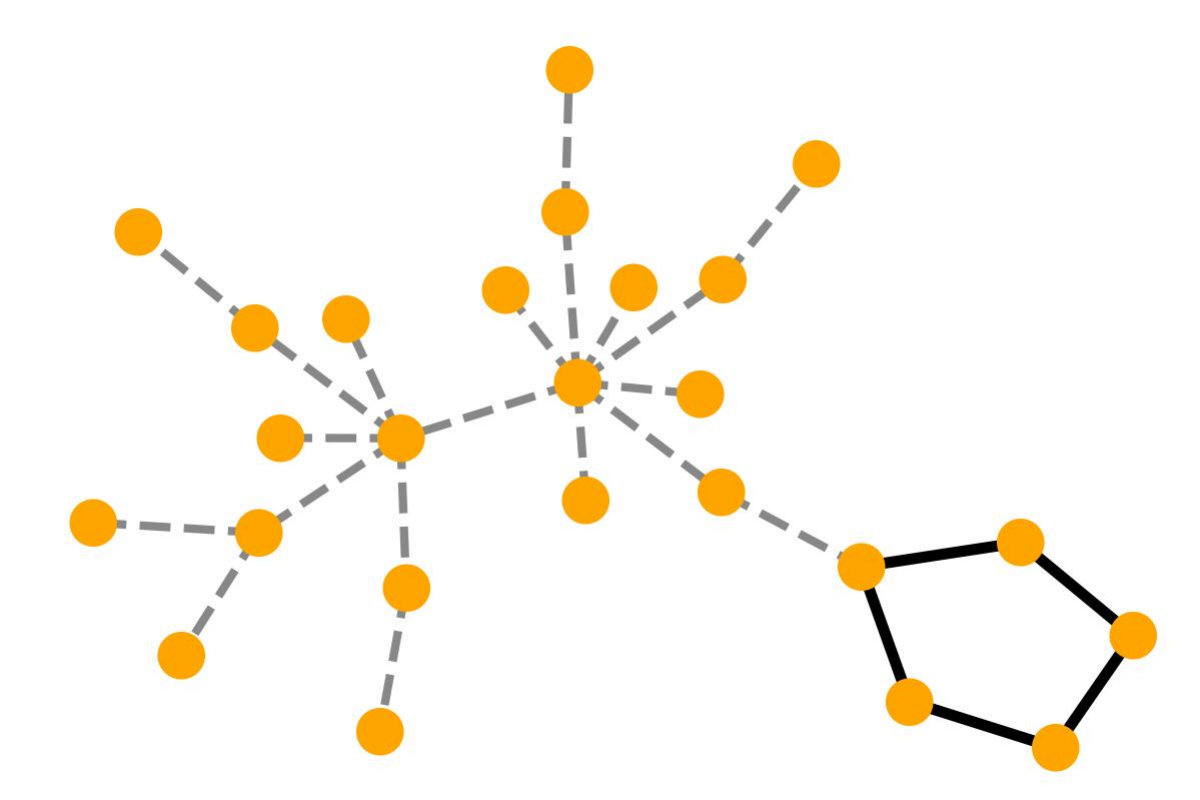}
    \caption{Ground Truth}
  \end{subfigure}
  \hfill
  \begin{subfigure}[t]{0.15\textwidth}
    \centering
    \includegraphics[width=\linewidth]{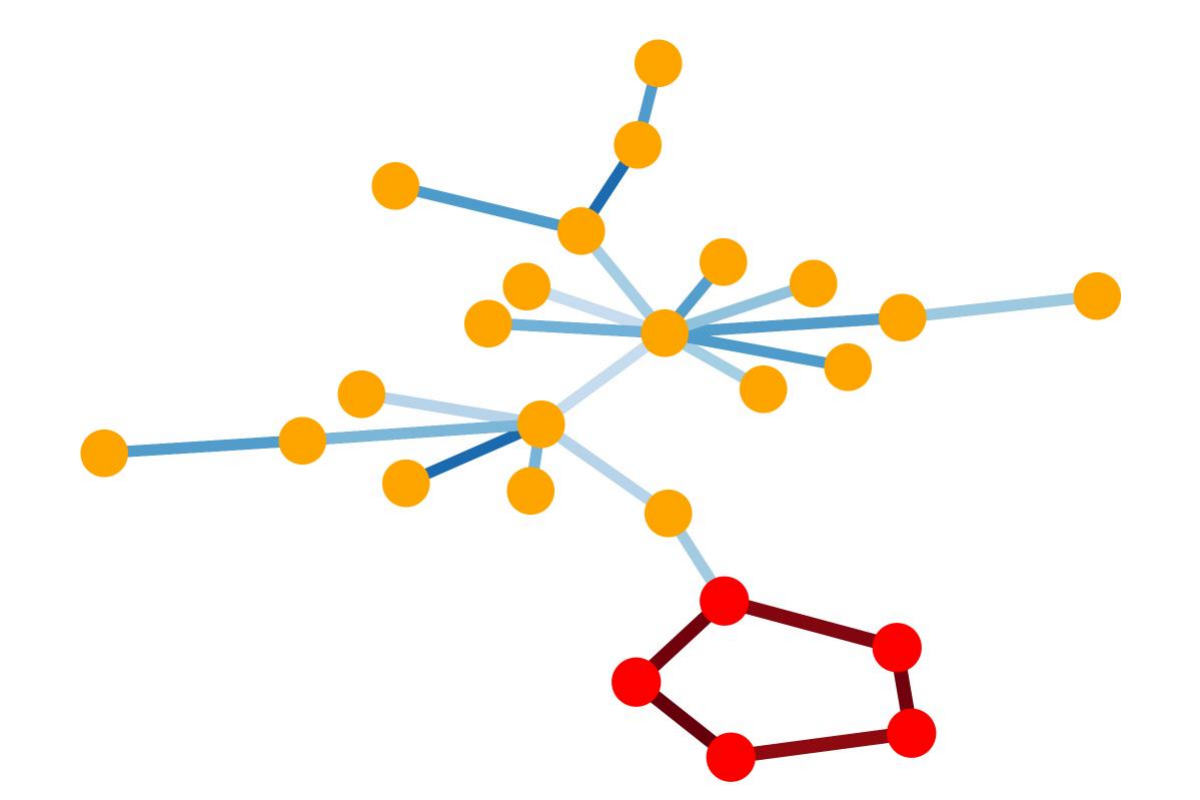}
    \caption{HPME}
  \end{subfigure}
  \hfill
  \begin{subfigure}[t]{0.15\textwidth}
    \centering
    \includegraphics[width=\linewidth]{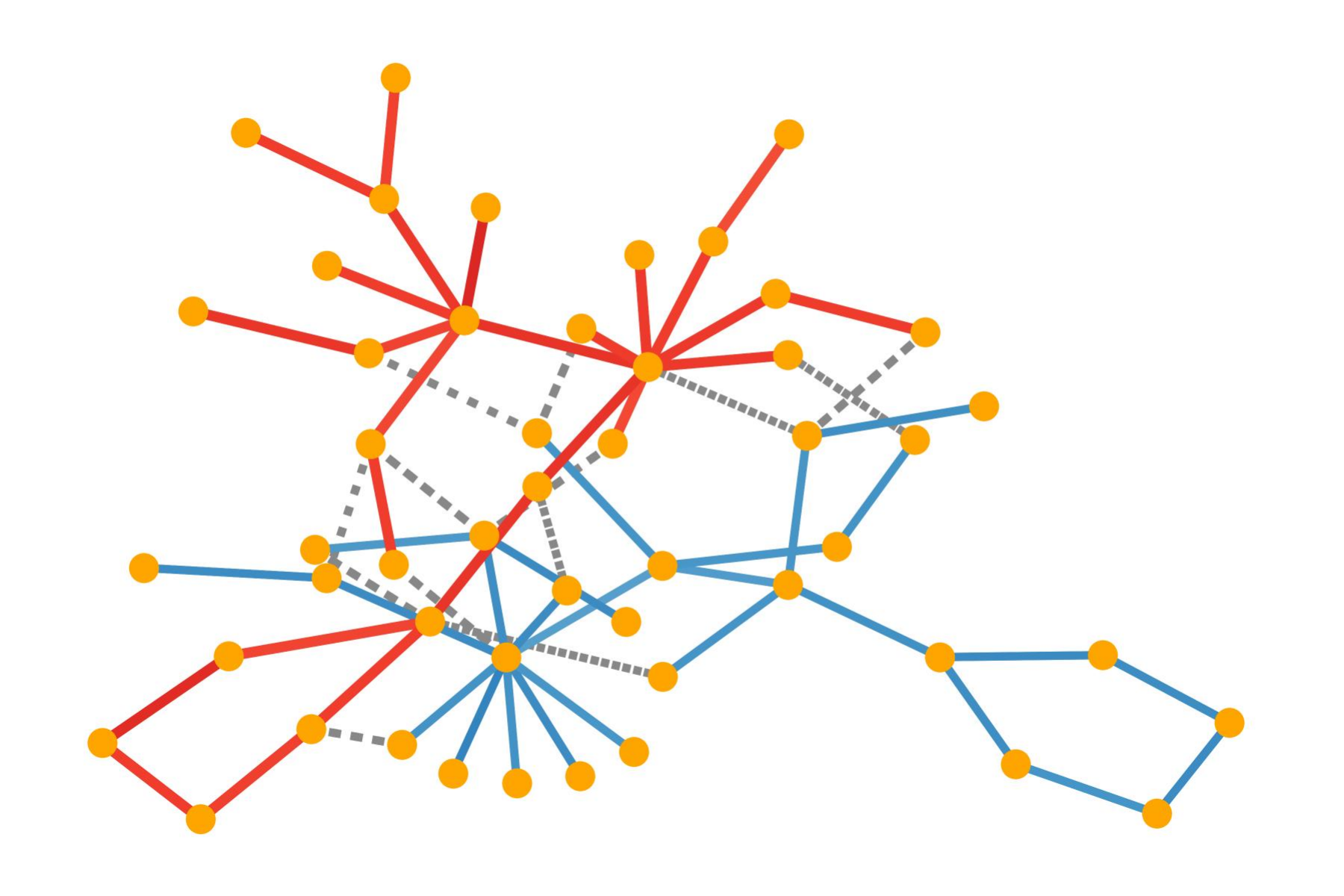}
    \caption{RegExplainer}
  \end{subfigure}

  \caption{Visualization of the ground truth and mixup graphs generated by different methods. Subfigures~(a)--(c) show results on the BA-HouseGrid classification dataset, while Subfigures~(d)--(f) correspond to the BA-Motif-Volume regression dataset. Edge weights are indicated by color intensity.}
  \label{fig:mixup_combined}
\end{figure*}

\subsection{Experiment Settings}
\label{sec:exp:setup}
We evaluate HPME on thirteen datasets with ground-truth explanations, spanning graph classification and regression tasks across both synthetic and real-world molecular data. The synthetic datasets include BA-2Motifs~\citep{luo2020parameterized}, BA-HouseGrid~\citep{amara2023ginx}, BA-HouseAndGrid~\citep{bui2024explaining}, SPMotif~\citep{wu2022discovering}, BA-Motif-Volume, BA-Motif-Counting~\citep{zhang2024regexplainer}, House-Grid-Volume, House-OrGrid-Volume, and 
Triangles~\citep{chen2020can}, while the real-world molecular datasets include Alkane-Carbonyl, Fluoride-Carbonyl, Benzene~\citep{sanchez2020evaluating}, and Crippen~\citep{delaney2004esol}. 
To assess effectiveness, we benchmark HPME against eight representative post-hoc methods, including GNNExplainer~\citep{ying2019gnnexplainer}, PGExplainer~\citep{luo2020parameterized}, TAGExplainer~\citep{xie2022task}, MetaGNN~\citep{spinelli2022meta}, MatchExplainer~\citep{wu2023rethinking}, MixupExplainer~\citep{zhang2023mixupexplainer}, ProxyExplainer~\citep{chen2024generating}, and RegExplainer~\citep{zhang2024regexplainer}, as well as the gradient-based GRAD~\citep{ying2019gnnexplainer} and the attention-based ATT~\citep{velivckovic2017graph}.

Following prior works~\citep{ying2019gnnexplainer,zhang2023mixupexplainer}, each dataset is divided into training, validation, and test sets in a ratio of $0.8$, $0.1$, and $0.1$. A three-layer GCN is adopted as the backbone model, with an additional linear layer for regression tasks. Analyses with other backbone models are reported in Appendix~\ref{sec:backbone_analysis}. Regarding hyperparameters, we apply grid search to determine the loss weight $\beta$ and set the top-$r$ ratios according to the ground-truth explanations. Evaluation is conducted using the AUC-ROC score for ground-truth subgraph identification, while distributional shifts are assessed by comparing graph representations using cosine similarity and euclidean distance. Further experimental details are provided in Appendix~\ref{sec:full_setup}.

\subsection{Quantitative Evaluation}
\label{sec:exp:performance}
In this section, we compare HPME with baselines using the AUC-ROC score.  For MixupExplainer, which was originally designed for graph classification tasks, we replace the CE loss with MSE loss when applying it to regression tasks. All experiments are repeated $10$ times with different random seeds. Table~\ref{classification_regression} reports the average scores and standard deviations for classification and regression tasks, respectively. Overall, our method consistently outperforms all baselines, providing the most accurate explanations across datasets. 
On graph classification datasets, HPME achieves an average improvement of $0.10$ ($11.39\%$) over the best baseline, while on regression datasets it improves by $0.10$ ($13.43\%$) on average, with the highest gain reaching $0.231$ ($30.16\%$). These results highlight the consistent effectiveness of HPME across different task settings.

Compared with representative methods, HPME consistently captures the invariant explanatory factors via the hard perturbation strategy. For instance, although MixupExplainer and RegExplainer mitigate the distribution shift of explanatory subgraphs via the mixup strategy and perform well on a part of the datasets, their reliance on a soft mixup approach introduces redundant edges, failing to maintain stable performance across all datasets.
ProxyExplainer, which adopts a generative approach, often introduces irrelevant edges on complex datasets, resulting in poor performance on MultipleMotif datasets such as BA-HouseAndGrid, as further confirmed by the visualizations in Appendix~\ref{sec:more_explanation}. Furthermore, while TAGExplainer aims for task-agnostic explanation, its reliance on gradient-based techniques for the downstream explainer, coupled with OOD issues in its self-supervised embedding learning, compromises the quality of the extracted explanatory subgraphs on all datasets. In contrast, HPME achieves stable and superior performance across all benchmarks, demonstrating robustness and adaptability. Appendix~\ref{sec:multi-task} provides additional validation on node classification and link prediction tasks.

\begin{figure}[t] 
  \centering
  \includegraphics[width=0.95\linewidth]{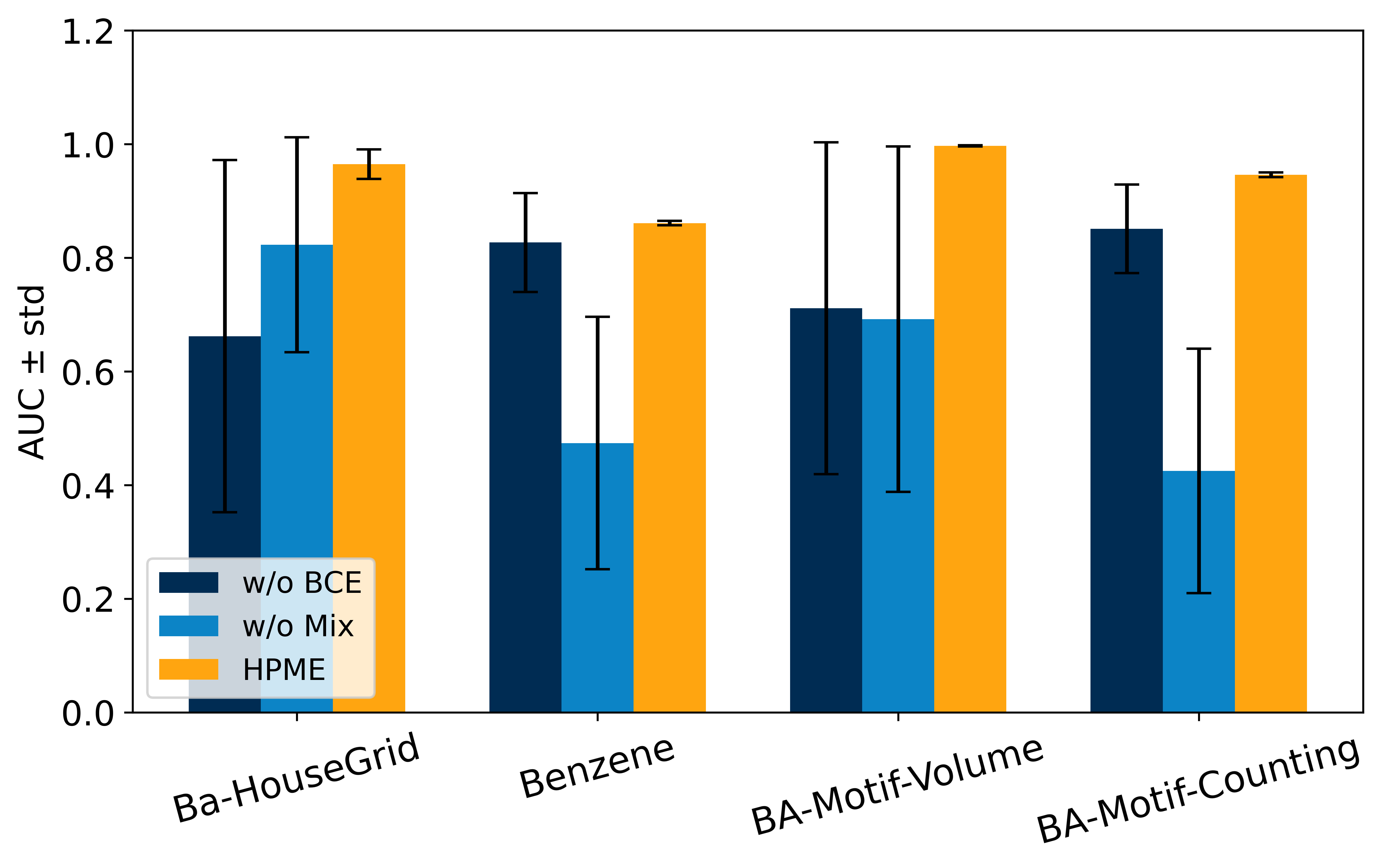}
  \caption{Ablation study of HPME. We evaluate the AUC-ROC performance of the original HPME and its variants. Standard deviations are indicated by black error bars.}
  \label{fig:ablation}
  \vspace{-10pt} 
\end{figure}

\subsection{Alleviating Distribution Shifts}
\label{sec:exp:ood}
In this section, we quantitatively analyze the distance between the original graph $G$ and the ground truth explanatory subgraph $G^*$ to demonstrate the presence of distribution shift in both classification and regression tasks, and to show that our proposed method effectively mitigates this issue.

\begin{figure*}[t]
    \centering
    \begin{subfigure}[b]{0.19\textwidth}
        \includegraphics[width=\linewidth]{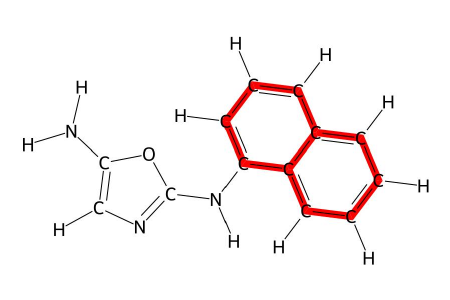}
        \caption{Ground Truth}
    \end{subfigure}
    \begin{subfigure}[b]{0.19\textwidth}
        \includegraphics[width=\linewidth]{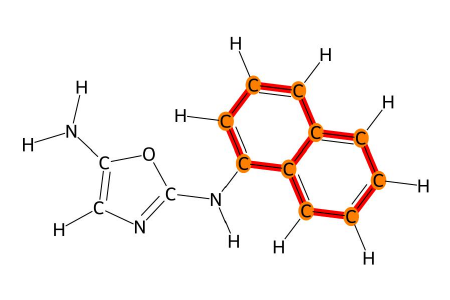}
        \caption{HPME}
    \end{subfigure}
    \begin{subfigure}[b]{0.19\textwidth}
        \includegraphics[width=\linewidth]{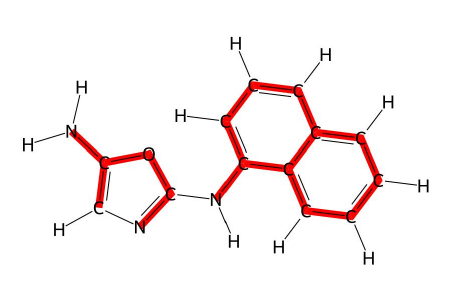}
        \caption{ProxyExplainer}
    \end{subfigure}
    \begin{subfigure}[b]{0.19\textwidth}
        \includegraphics[width=\linewidth]{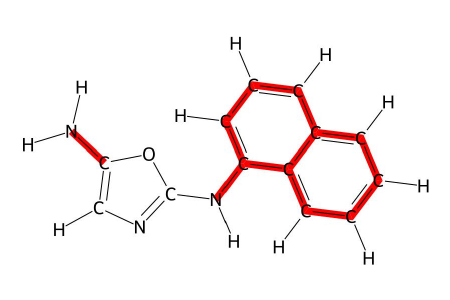}
        \caption{MixupExplainer}
    \end{subfigure}
    \begin{subfigure}[b]{0.19\textwidth}
        \includegraphics[width=\linewidth]{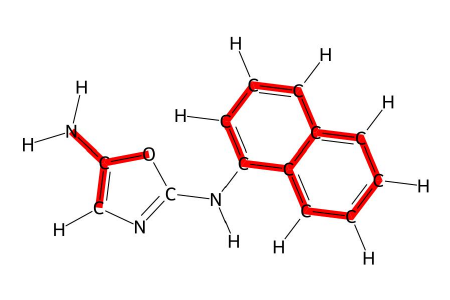}
        \caption{MetaGNN}
    \end{subfigure}

    \vspace{5pt} 

    \begin{subfigure}[b]{0.19\textwidth}
        \includegraphics[width=\linewidth]{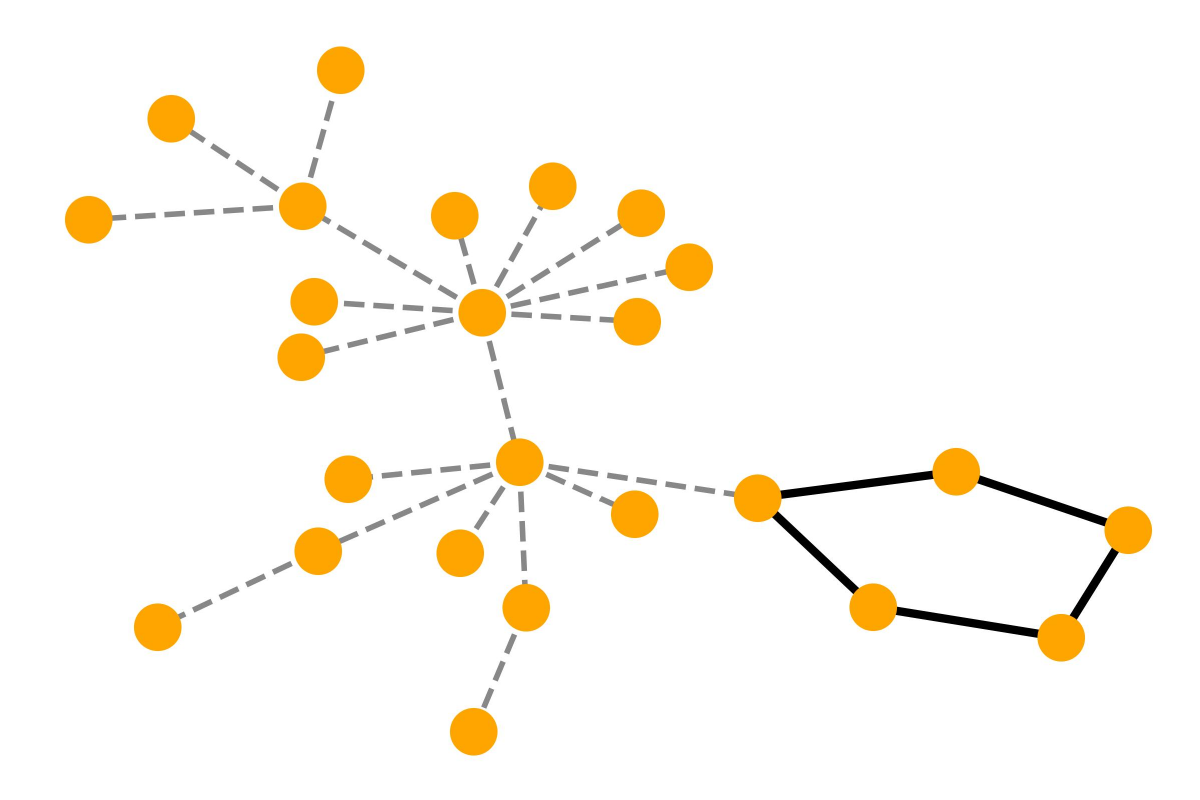}
        \caption{Ground Truth}
    \end{subfigure}
    \begin{subfigure}[b]{0.19\textwidth}
        \includegraphics[width=\linewidth]{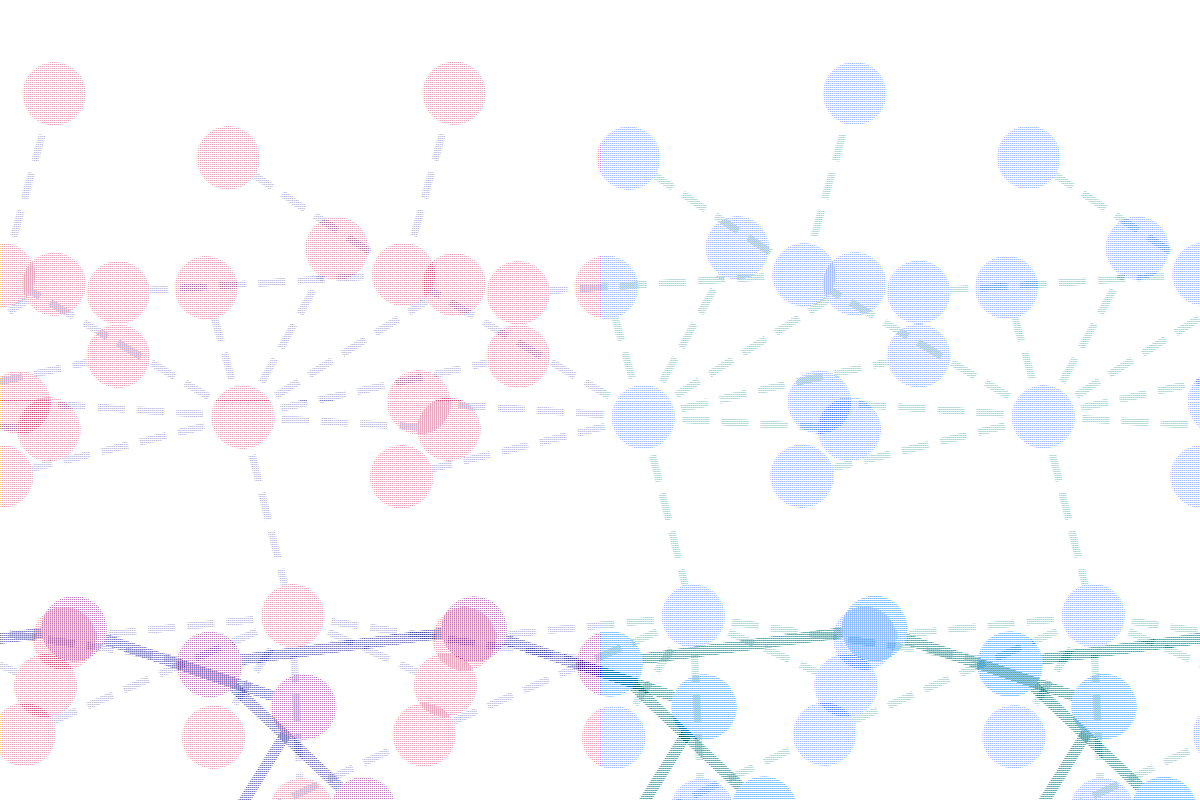}
        \caption{HPME}
    \end{subfigure}
    \begin{subfigure}[b]{0.19\textwidth}
        \includegraphics[width=\linewidth]{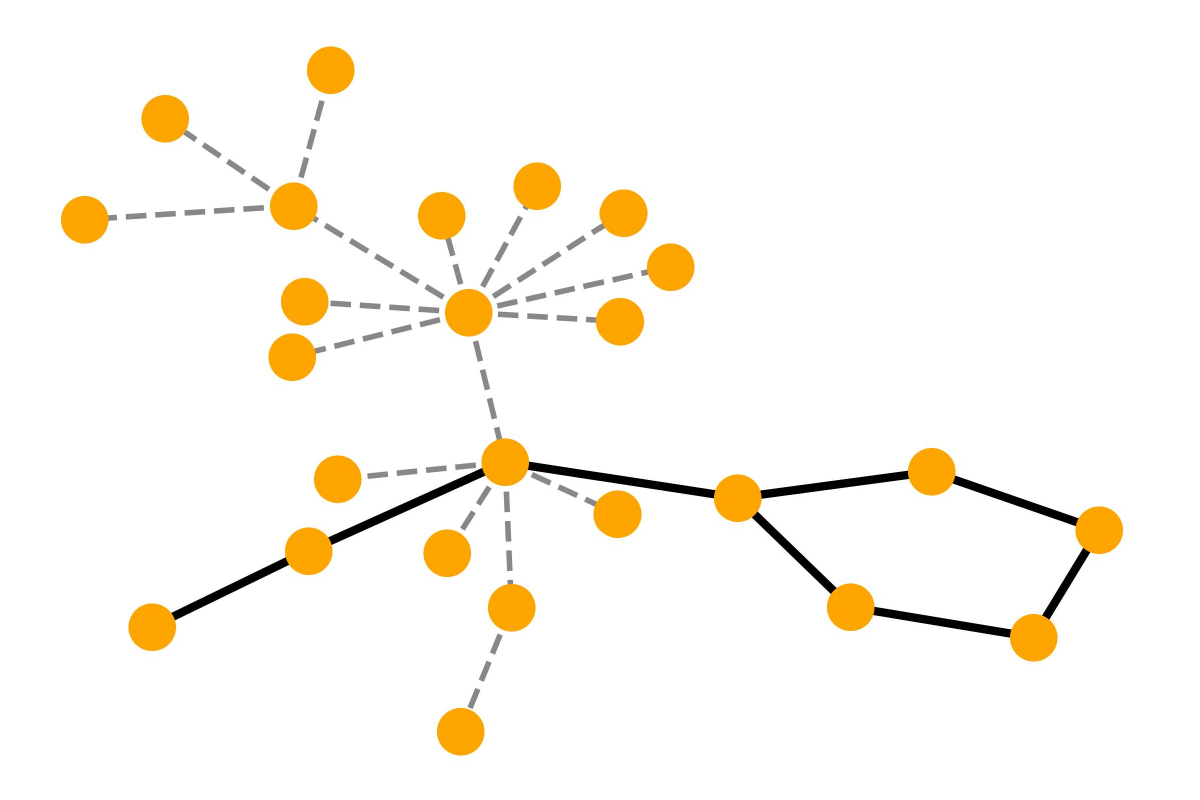}
        \caption{RegExplainer}
    \end{subfigure}
    \begin{subfigure}[b]{0.19\textwidth}
        \includegraphics[width=\linewidth]{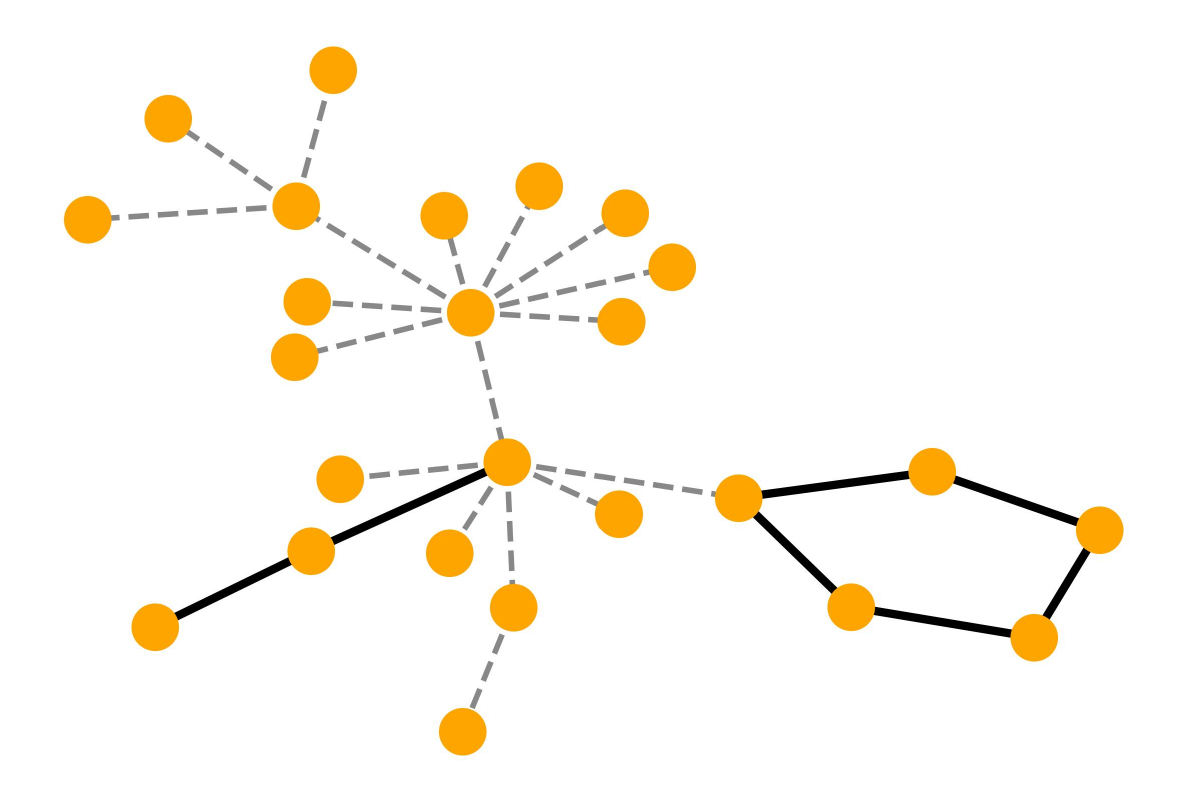}
        \caption{MixupExplainer}
    \end{subfigure}
    \begin{subfigure}[b]{0.19\textwidth}
        \includegraphics[width=\linewidth]{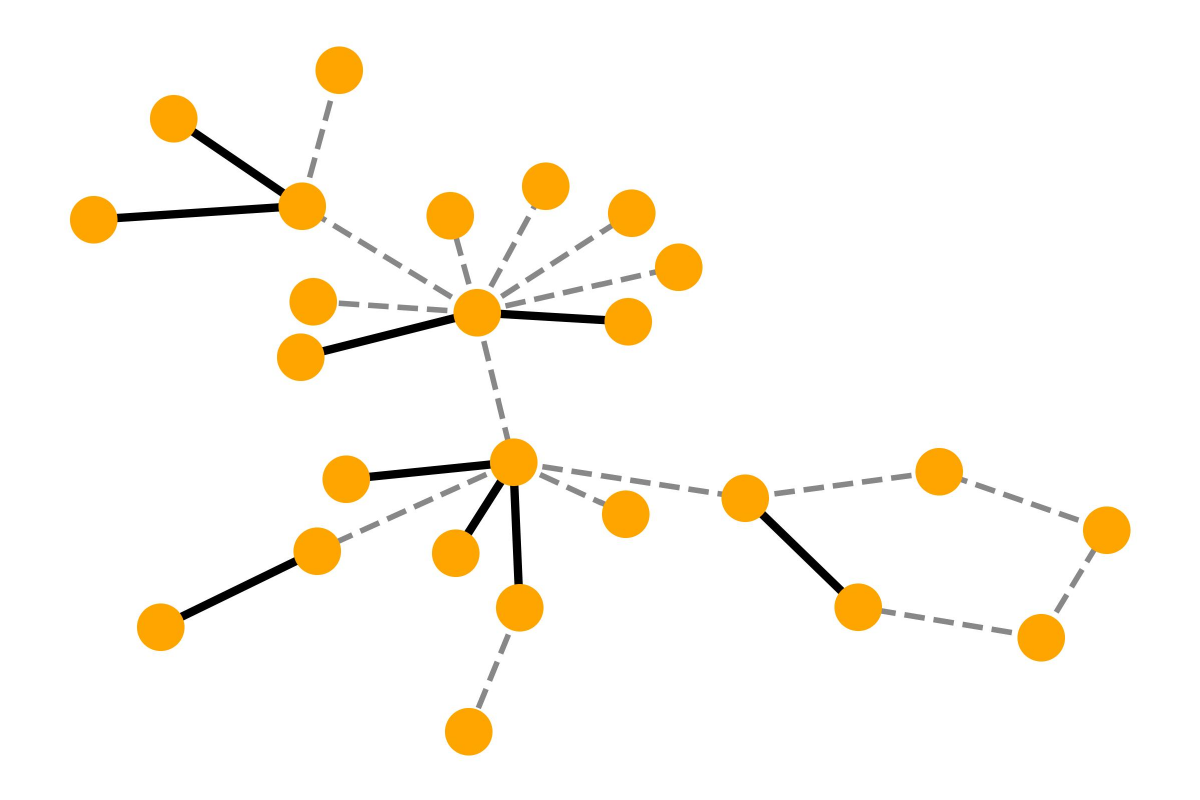}
        \caption{ATT}
    \end{subfigure}

    \caption{Visualization of explanation results obtained by different methods. 
Subfigures~(a)--(e) show results on the Benzene classification dataset, while Subfigures~(f)--(j) correspond to the BA-Motif-Volume regression dataset.}

\label{fig:casestudy:combined}
\end{figure*}

Specifically, we use the output of the penultimate layer of the target model as the graph representation vector and distribution shift is evaluated by computing the average cosine similarity and Euclidean distance between original graph representations and their ground-truth or mixup counterparts.
As shown in Table~\ref{tab:exp_metrics}, ``GT" denotes the average cosine similarity and euclidean distance between the original graph and the ground-truth explanatory subgraph. Meanwhile, ``ProxyE", ``RegE" and ``HPME" denote the same metrics between the original graph and the mixup graphs generated by ProxyExplainer, RegExplainer and our method, respectively.
The results reveal lower similarity and higher euclidean distance between the representations of original graph and ground-truth explanatory subgraph, demonstrating the existence of distributional shift in both classification and regression tasks. 
Moreover, since existing soft-mask-based methods inherently introduce redundant and irrelevant edges, they fail to adequately approximate the original graph distribution in more complex real-world datasets such as Fluorid-Carbonyl and Crippen. In contrast, the mixup graphs generated by our method exhibit higher cosine similarity and lower Euclidean distance to the original graph representations, indicating that the proposed mixup strategy effectively mitigates the distributional shift problem and enables the explainer to more precisely identify explanatory subgraphs, thereby enhancing overall explanation performance. Additional distribution-shift evaluation results on more classification and regression datasets are reported in Appendix~\ref{sec:app:ood}.

\subsection{Ablation Study}
\label{sec:exp:ablation}
In this section, we conduct ablation studies to evaluate the contribution of each component in the HPME framework. Specifically, we design the following variants:  
1) w/o Mix: Removes the structral mixup step after extracting the explanation, directly feeding the explanatory subgraph into the objective function. 
2) w/o BCE: Removes only the information compression term from the training loss. All variants are configured with the same hyperparameters as the original HPME, including the learning rate, number of training epochs, and loss weight $\beta$.

We conduct experiments on representative datasets covering both classification and regression tasks, with BA-HouseGrid and Benzene used for classification, and BA-motif-volume and BA-Motif-Counting used for regression. The results are presented in \Figref{fig:ablation}. Experimental results show that HPME consistently outperforms all its variants across all datasets, indicating that each component contributes positively to the overall performance, especially the proposed structural mixup strategy based on the hard perturbation.

\subsection{Case Studies}
\label{sec:exp:case}

To verify whether our method can effectively capture explanatory subgraphs, we conduct a case study on the real-world classification dataset Benzene and the synthetic regression dataset BA-Motif-Volume. The visualization results are presented in \Figref{fig:casestudy:combined}. Explanatory subgraphs are highlighted with bold red edges for Benzene and bold black for BA-Motif-Volume. 
For HPME, the nodes identified through graph pooling are additionally marked with orange overlays in Benzene and red overlays in BA-Motif-Volume. The extracted explanatory subgraph is obtained by ranking all edges according to their weight scores and retaining the top-$K$ edges, where $K$ is set to the number of edges in the ground-truth explanation. The visualization results demonstrate that consistently extracts the most accurate sparsified explanatory subgraphs, effectively eliminates redundant structural information, and enhances the extraction quality of explanatory subgraphs across both classification and regression tasks. Moreover, the nodes identified through graph pooling perfectly align with the ground-truth nodes, demonstrating that the pooling process is effectively optimized to filter out task-irrelevant structures. Additional results can be found in Appendix~\ref{sec:more_explanation}.

To further validate the effectiveness of the proposed structural mixup strategy, we conduct another case study comparing the quality of the mixup graphs generated by HPME and baseline methods during process, as shown in \Figref{fig:mixup_combined}. Specifically, ProxyExplainer is evaluated on BA-HouseGrid for the classification task, and RegExplainer on BA-Motif-Volume for the regression task, with additional examples provided in Appendix~\ref{sec:more_mixup}. The visualizations are obtained from intermediate results of the final training epoch under the best-performing hyperparameters.

In the mixup graphs, edges in the explanatory subgraph are highlighted in red and those in the label-irrelevant subgraph in blue, with color intensity indicating the weight values in $(0,1)$, while random connections generated by RegExplainer are shown as gray dashed lines. Our structural mixup strategy naturally integrates explanatory and label-irrelevant subgraphs, producing concise mixup graphs that faithfully preserve the original structural distribution. In contrast, ProxyExplainer employs a generative approach that may introduce redundant edges and distort the explanatory structure, while RegExplainer performs a soft mixup strategy via matrix concatenation, resulting in randomly sampled uninterpretable edges and inevitably failing to completely eliminate the inherent redundant information, thereby producing mixup graphs that fail to remain faithful to the original graph distribution. Overall, the visualization results across both classification and regression tasks confirm the superiority of our strategy in constructing in-distribution mixup graphs, which more effectively mitigates the distribution shift problem, enabling the HPME to generate more robust and precise explanations.

\section{Conclusion}
In this work, we systematically investigate the OOD problem in post-hoc instance-level GNN explanation frameworks grounded in the GIB principle, identifying that existing soft-mask-based approaches inherently permit label-irrelevant information to persist within the explanatory subgraphs and leak into the mixup process, an issue which has been largely overlooked in prior studies. Consequently, the generated mixup graphs fail to faithfully approximate the original graph distribution, hindering the resolution of the OOD problem. To address these limitations, we propose HPME, a framework built upon a principled generalized GIB with hard perturbations to extract discrete explanatory subgraphs and effectively eliminate residual label-irrelevant information, which establishes a theoretical foundation for future research on GNN explainability. Moreover, we introduce a novel structural mixup strategy built upon graph pooling that generates naturally connected in-distribution mixup graphs, innovatively alleviating the OOD problem. Extensive experiments on both synthetic and real-world benchmarks demonstrate that HPME consistently outperforms existing baselines in extracting faithful explanations across diverse tasks, and visualization analysis further shows that the proposed structural mixup strategy yields a cleaner composition of explanatory and label-irrelevant subgraphs compared with soft-mask-based methods. 
While HPME demonstrates the superior performance on homogeneous graphs, extending the generalized GIB with hard perturbations to diverse graph structures presents a promising avenue for future research, particularly for capturing complex structural dependencies in real-world systems.

\FloatBarrier
\bibliographystyle{ACM-Reference-Format}
\bibliography{sample-base}

\appendix








\newpage
\section{Notations}
In Table~\ref{tab:symbols}, we summarized the important notations we used and their descriptions in this paper.
\begin{table}[htb]
\centering
\small
\renewcommand{\arraystretch}{1.0}
\setlength{\tabcolsep}{6pt}

\caption{Important symbols and notations.}
\label{tab:symbols}

\begin{tabular}{cl}
\toprule
Symbols & Descriptions \\
\midrule
$G$ & Graph instance \\
$\mX$ & Node feature matrix \\
$\mA$ & Adjacency matrix \\
$Y$ & Label for $G$ \\
$\mathcal{G}$ & Graph dataset \\
$\mathcal{V}, \mathcal{E}$ & Node set and edge set \\
$\mathcal{Y}$ & Label set \\
$v_i$ & The $i$-th node \\
$x_i$ & The feature of node $v_i$\\
$n$ & Number of nodes in $G$ \\
edge$(i,j)$ & Edge from node $i$ to node $j$ \\
$i, j$ & Node indices \\  
$d$ & Feature dimension \\
$G^*, Y^*$ & Optimal explanatory subgraph and its label \\ 
$\mH^*, \mA^*$ & Node embeddings and adjacency matrix of $G^*$ \\
$I(\cdot)$ & Mutual information \\
$\alpha$ & Hyperparameter for MI term in GIB \\
$\mathcal{N}$ & Number of graphs in $\mathcal{G}$ \\
$f(\cdot)$ & To-be-explained GNN model \\
$\mH$ & Node embeddings matrix \\
$P_\theta(\cdot)$ & Explainer network \\
$Q_\phi(\cdot)$ & Hard perturbation operator \\
$G^*_\theta$ & Soft mask explanation \\
$G^*_{\text{pool}}$ & pooled subgraph \\
$\Omega_k$ & Subspace of pooled subgraphs \\
$f_{\text{enc}}(\cdot)$ & Encoder of $f$ generating representations \\
$\vh_G$ & Graph representation of $G$ \\
$r_l$ & Node preservation ratio at layer $l$ \\
$I^L$ & Final preserved node indices after $L$ layers \\
$\text{score}$ & Node scores \\
$p_{l}$ & Learnable projection vector at layer $l$ \\
$N$ & Number of sampled neighbors \\
$G'$ & Sampled neighbor \\
$\mA^{\text{(mix)}}_{\text{our}}$ & Adj. matrix via structural mixup \\
$\mH^{\text{(mix)}}_{\text{our}}$ & Embeddings via structural mixup \\
$G^{\text{(mix)}}_{\text{our}}$ & Graph via structural mixup \\
$S'$ & Binary node preserved matrix from $G'$\\
$\mW$ & Edge weights matrix \\
$\mM^{\text{(mix)}}_{\text{our}}$ & Edge weights mask for mixup graph \\
$\mathcal{L}_{\text{BCE}}$ & Binary cross-entropy loss \\
$\beta$ & Hyperparameter balancing $\mathcal{L}_{\text{pred}}$ and $\mathcal{L}_{\text{BCE}}$ \\
\bottomrule
\end{tabular}
\end{table}

\section{Algorithms}
\label{sec:alg_detail}
\begin{algorithm}[htb]
\caption{Training Explainer}
\label{alg_training}
\begin{algorithmic}[1]
\Statex \hspace{-1em}\textbf{Input:} A graph dataset \( \mathcal{G} \), pretrained GNN model \( f \), explainer network \( P_{\theta}(\cdot) \).
\Statex \hspace{-1em}\textbf{Output:} Trained explainer network \( P_{\theta}(\cdot) \).
\State Initialize explainer network \( P_{\theta}(\cdot) \).
\For{$e \in$ epochs}
    \For{$G \in \mathcal{G}$}
        \State Randomly sample graph \(  G' \) from graph dataset \( \mathcal{G} \)
        \State \(G^{\text{(mix)}}_{\text{our}} \gets \text{Structural Mixup}(G, G') \)
        \State Generate \(\mM^{\text{(mix)}}_{\text{our}}\) with \(G^{\text{(mix)}}_{\text{our}}\) 
        \State Compute \( \mathcal{L}_{\text{pred}}(f(G), f(G)^{\text{mix}}) \)
        \State Compute \( \mathcal{L}_{\text{BCE}} \) with \Eqref{eq:bce_loss}
        \State Compute overall loss \( \mathcal{L} \) with \Eqref{eq:finall_loss}
    \EndFor
    \State Update \( P_{\theta}(\cdot) \) with back propagation
\EndFor
\State \textbf{Return} Trained explainer network \( P_{\theta}(\cdot) \)
\end{algorithmic}
\end{algorithm}

\begin{algorithm}[htb]
\caption{Structural Mixup}
\label{alg_strmixup}
\begin{algorithmic}[1]
\Statex \hspace{-1em}\textbf{Input:} To-be-explained graph \( G \), \(G'\) sampled from graph dataset \( \mathcal{G} \).
\Statex \hspace{-1em}\textbf{Output:} Graph \( G^{\text{(mix)}}_{\text{our}} \).
\State Obtain \(I^L\) via \(L\)-layer graph pooling following \Eqref{eq:graph_pooling}
\State Obtain \((I')^{L}\) via \(L\)-layer graph pooling following \Eqref{eq:graph_pooling}
\State Generate pooled graphs \(G^*_{pool}\) and \((G')^*_{pool}\) based on \(I^{L}\) and \((I')^{L}\)
\State Structural Mixup adjacency \(\mA^{\text{(mix)}}_{\text{our}}\) and node embeddings \(\mH^{\text{(mix)}}_{\text{our}}\) matrix with \Eqref{eq:str_mixup}

\State \textbf{Return} \( G^{\text{(mix)}}_{\text{our}}=(\mH^{\text{(mix)}}_{\text{our}}, \mA^{\text{(mix)}}_{\text{our}})\)
\end{algorithmic}
\end{algorithm}

To address the distribution shift issue in the GIB objective, a graph-pooling-based structural mixup strategy is employed to generate in-distribution, naturally connected mixup graphs \( G^{\text{(mix)}}_{\text{our}} \) that preserve the information of explanatory subgraphs \( G^* \). Accordingly, a prediction loss is adopted as the optimization objective within GIB to approximate the mutual information between \( G^* \) and the label \( Y \). For completeness, we present the pseudocode of our method below, which encompasses both explainer training and the structural mixup strategy.

In \Algref{alg_training}, at each epoch, we first randomly sample a graph \(G'\) from the graph dataset \(\mathcal{G}\) for each graph \( G \). Then, the structural mixup strategy combines the explanatory subgraph \( G^* \) with the label-irrelevant parts of \(G'\), producing the mixup graphs \( G^{\text{(mix)}}_{\text{our}} \). For each mixup graph, an MLP is employed to predict edge weights, approximating the structural information obtained from the graph pooling process to generate the final soft masks \( \mM^{\text{(mix)}}_{\text{our}} \). The explainer is optimized by minimizing the prediction loss and BCE loss \Eqref{eq:bce_loss}, while the overall loss function \Eqref{eq:finall_loss} sums the two objectives with a trade-off parameter. Finally, the explainer parameters \( P_{\theta}(\cdot) \) are updated through backpropagation with the overall loss.

Specifically, 
\Algref{alg_strmixup} shows the structural mixup process. Given a to-be-explained graph \( G \) and a sampled graph \( G' \), we perform \( L \)-layer graph pooling to obtain the preserved node indices \( I^L \) and \((I')^L\). Then, the explanatory subgraph \( G^* \) is represented as \( G^* = (\mH^*, \mA^*) \), where \( \mH^* = \mH(I^L, :) \) and \( \mA^* = \mA(I^L, I^L) \). Similarly, the explanatory subgraph of \( G' \) is obtained in the same way based on \((I')^L\). The adjacency matrix \( \mA^{\text{(mix)}}_{\text{our}} \) and node embeddings \( \mH^{\text{(mix)}}_{\text{our}} \) of the mixup graph are obtained by replacing \((\mA')^*\) with \( \mA^* \), as defined in \Eqref{eq:str_mixup}. Finally, the mixup graph \( G^{\text{(mix)}}_{\text{our}} = (\mH^{\text{(mix)}}_{\text{our}}, 
\mA^{\text{(mix)}}_{\text{our}}) \) is returned.

\stitle{Computational Complexity Analysis. }Given graph \( G \), the time complexity of \( L \)-layer graph pooling is \( O(L \cdot (|\mathcal{E}| d + n d)) \), where \( d \) is the feature dimension, \( \mathcal{E} \) denotes the edge set, and \( n \) is the number of nodes in \( G \). Then, structural replacement replaces the explanatory subgraph via \( m \) node indices, with a complexity of \( O(m^2) \), where \( m \) is the number of final preserved nodes. Edge weights for the mixup graph are generated by an MLP and the time complexity is \( O(|\mathcal{E}^{(\text{mix})}| \cdot d) \), where \(\mathcal{E}^{(\text{mix})}\) denotes the edge set of the mixup graph. Since \( m \ll n \), \( |\mathcal{E}| \ll n^2 \), and the sizes of \( |\mathcal{E}| \) and \( |\mathcal{E}^{(\text{mix})}| \) are comparable, the overall time complexity of our method is dominated by \( O(L \cdot |\mathcal{E}| d) \).

\section{Full Experimental Setup}
\label{sec:full_setup}
Detailed experimental settings are provided in this section, including implementation details, datasets, and baseline methods. All experiments are conducted on a Linux workstation running Ubuntu 22.10 (kernel 5.19.0-46-generic) equipped with 8 NVIDIA GeForce RTX 4090 GPUs (24 GB each). The system uses NVIDIA driver version 535.86.05 and CUDA 12.2. All code is implemented in Python 3.9.21, using PyTorch 2.0.1+cu118, PyTorch Geometric 2.6.1, torch-scatter 2.1.2+pt20cu118, and torch-sparse 0.6.18+pt20cu118.

\subsection{Implementation Details}
Following the configuration in previous work~\citep{ying2019gnnexplainer,zhang2023mixupexplainer}, we divide each dataset into training, validation, and test sets with a ratio of $0.8$, $0.1$, and $0.1$, respectively. We choose GCN as the backbone model for graph-level and node-level tasks, given its stable performance across diverse datasets, with SEAL~\cite{zhang2018link} adopted for link prediction task following established practices. 
The detailed comparison of the prediction performance of GCN and other GNN architectures are provided in Appendix~\ref{sec:backbone_analysis}. A three-layer GCN is used as the target model, and for graph regression tasks, an additional linear layer is appended. Additionally, a two-layer MLP serves as the explanation network. For the compared baselines, we strictly adhered to the original implementations without any architectural modifications. For link prediction, we use SEAL~\cite{zhang2018link} as the base model to be explained. We train the GCN model to a reasonable performance, as shown in the GCN column of Tables~\ref{tab:classification_results} and~\ref{tab:regression_results}. For explanation, we follow the sample selection strategy used in GNNExplainer~\citep{ying2019gnnexplainer} and randomly select $200$ graph instances per dataset to reduce computational cost. All explanation methods are optimized using the Adam optimizer with a weight decay of $5 \times 10^{-4}$~\citep{kingma2014adam}. Regarding hyperparameters, we apply grid search to determine the loss weights $\beta$, and set the top-$r$ ratios according to the ground truth explanations. For baseline methods with overlapping hyperparameters, we adopt a unified setting; otherwise, we retain their default configurations. We ensure consistent learning rates and training epochs across all explanation methods. We evaluate the performance of our proposed method against baseline approaches on ground-truth explanation tasks using the AUC-ROC score and robust fidelity. To quantitatively assess the effectiveness of our method in addressing distributional shifts, we measure the distances between graph representations using cosine similarity and Euclidean distance.
\subsection{Datasets}
\label{sec:dataset_setup}
We evaluate our method on a range of graph datasets, including both synthetic and real-world benchmarks.

\stitle{BA-Shapes~\citep{ying2019gnnexplainer}. }This dataset is based on a BA graph with 80 "house" motifs, where node labels define a 4-class classification task and motif edges provide explanation ground truth.

\stitle{BA-Community~\citep{ying2019gnnexplainer}. }This is an extension of BA-Shapes where two different motifs are attached to the base graph, and nodes across two motifs are assigned different labels, resulting in an 8-class classification task.

\stitle{Tree-Grid~\citep{ying2019gnnexplainer}. }This node classification dataset is based on a single 8-layer balanced binary tree with 80 grid motifs. The task is to distinguish motif nodes from tree nodes, with the motif edges serving as the ground-truth explanations.

\stitle{\batwo~\citep{luo2020parameterized}. }The \batwo dataset comprises 1,000 synthetic graphs, each generated from a Barabási–Albert graph and extended with either a house or a five-cycle motif. The label of each graph is determined by its attached motif, forming a binary classification task in which the motif serves as the ground-truth explanation.

\stitle{\bahg~\citep{bui2024explaining}. }The \bahg dataset employs the house motif and the \(3\times3\)  grid motif. These motifs are chosen to minimize structural overlap and ensure that models learn the full motif structure instead of relying on local substructures for prediction.

\stitle{\spmotif~\citep{wu2022discovering}. } In the \spmotif dataset, each graph consists of a base structure (Tree, Ladder, or Wheel) and a motif (Cycle, House, or Crane). A parameter $b$ controls the degree of the spurious correlation between the base structure and the motif, with $b=\frac{1}{3}$ indicating no spurious correlation. In our experiments, $b=0.7$. The label and ground-truth explanation for each graph are determined solely by the motif it contains. 

\stitle{\bahandg~\citep{bui2024explaining}. }The \bahandg dataset contains Barabási–Albert graph graphs extended with the house motif, the grid motif, or both. Graphs are labeled 1 if they contain both motifs, otherwise, they are labeled 0.

\stitle{\alca~\citep{sanchez2020evaluating}. }The \alca dataset comprises 4,326 molecular graphs partitioned into two classes based on the functional groups. A molecule is labeled positive when it contains both alkane and carbonyl groups, which also serve as the ground-truth explanation.

\stitle{\flca~\citep{sanchez2020evaluating}. }The \flca dataset comprises 8,671 molecular graphs partitioned into two classes based on functional groups. A molecule is labeled positive if it contains both fluoride atoms and a carbonyl group, which also serve as the ground-truth explanation.

\stitle{\ben~\citep{sanchez2020evaluating}. }The \ben dataset comprises 12,000 molecular graphs from the ZINC15~\citep{sterling2015zinc} database, partitioned into two classes based on the presence of benzene rings. A molecule is labeled positive if it contains at least one benzene ring. Each benzene ring in the molecule serves as a distinct ground-truth explanation. 

\stitle{\bavolume~\citep{zhang2024regexplainer}. }In \bavolume dataset, each graph is constructed from a Barabási–Albert graph with an attached five-cycle motif. Node features are assigned random float values in the range [0.00, 100.00], and the regression label for each graph is defined as the sum of node feature values over the motif. 

\stitle{\bahgvolume. }The \bahgvolume dataset is derived from \bahg~\citep{bui2024explaining} by replacing all node features with random floats sampled from [0.00, 100.00]. The regression label for each graph is defined as the sum of node features within its motif. 

\stitle{\bacounting~\citep{zhang2024regexplainer}. }The \bacounting dataset consists of graphs created by attaching a randomly sampled number of five-cycle motifs (with the number varying from $\{0,\ldots,10\}$ in our experiment) to a Barabási–Albert random graph. The number of motifs in each graph serving as its regression label.

\stitle{\tri~\citep{chen2020can}. }The \tri dataset is constructed following the prior work~\citep{chen2020can}, consisting of 5,000 Erdős–Rényi random graphs denoted as \(ER(m, p)\), where \(m = 30\) is the number of nodes in each graph and \(p = 0.1\) is the edge existence probability. The regression label for each graph is the number of triangles it contains. In our experiments, we use a subset of 1,000 graphs randomly sampled from this dataset.

\stitle{\crip~\citep{delaney2004esol}. }The \crip dataset contains 1,127 molecules with corresponding aqueous solubility measurements from the Delaney solubility~\citep{delaney2004esol} dataset and assigns node weights using the Crippen model~\citep{wildman1999prediction}. Following prior work~\citep{sanchez2020evaluating}, we adopt this dataset and generate edge weights as the average of incident node weights.

\stitle{\horgvolume. }The \horgvolume dataset is derived from \bahorg~\citep{bui2024explaining} by replacing all node features with random floats sampled from [0.00, 100.00]. The regression label for each graph is defined as the sum of node features within its motif.
\subsection{Baselines}
\label{sec:baseline_setup}
To assess effectiveness, we incorporate various post-hoc methods, including \gnne, \pge, \metagnn, \match, \mixupex, \proxyex, and \regex, as well as the gradient-based \grad and the attention-based \att.

\stitle{\grad~\citep{ying2019gnnexplainer}. }\grad is a gradient-based method that generates weights to edges and nodes by computing the gradients of the loss function of GNN with respect to the adjacency matrix and node features.

\stitle{\att~\citep{velivckovic2017graph}. }ATT is a graph attention network that learns attention weights for edges in the input graph, with these weights are adopted as a proxy measure of edge importance. 

\stitle{\gnne~\citep{ying2019gnnexplainer}. }\gnne learns soft masks over edges and node features for each instance by minimizing the mutual information between the original graph and the prediction results. The explanatory subgraphs are obtained via element-wise multiplication of the learned soft masks with the original graph.

\stitle{\pge~\citep{luo2020parameterized}. }\pge extends \gnne by parameterizing the explanation generation process with a trainable explainer, and generates a soft mask to produce the explanatory subgraph. 

\stitle{\metagnn~\citep{spinelli2022meta}. }\metagnn trains GNNs to be inherently interpretable by incorporating a meta-explainer, which generates post-hoc explanations during training. 

\stitle{\tagex~\citep{xie2022task}. }\tagex is a task-agnostic explainer trained in a self-supervised manner. It explains GNN predictions by separating the explainer into embedding and downstream components, enabling the explanation of GNN embedding models with unseen downstream tasks and allowing efficient explanation of multitask models.

\stitle{\match~\citep{wu2023rethinking}. }\match explains GNN predictions by matching shared subgraph patterns using graph edit distance (GED) as the similarity metric, and this non-parametric subgraph matching approach inherently avoids optimization bias.

\stitle{\mixupex~\citep{zhang2023mixupexplainer}. } \mixupex mitigates the distribution shift present in previous methods by mixing explanatory subgraphs with the label-irrelevant parts of other randomly sampled graphs in a non-parametric manner.

\stitle{\proxyex~\citep{chen2024generating}. }\proxyex performs autoencoders to reconstruct the explanatory and label-irrelevant parts, then combines them to generate proxy graphs in a parametric manner, approximating the original distribution to mitigate the OOD issue.

\stitle{\regex~\citep{zhang2024regexplainer}. }\regex addresses explainability in graph regression tasks by combining mixup with contrastive learning to mitigate distribution shift and tackle the challenge of continuously ordered labels. 

\begin{table*}[t]
\centering
\caption{Robust Fidelity scores for explanations on three link prediction tasks.}
\label{tab:fidelity_linkpre}
\small
\setlength{\tabcolsep}{5pt}
\renewcommand{\arraystretch}{1.1}

\begin{tabular}{lccccccc}
\toprule
\multirow{2}{*}{Method} & \multicolumn{2}{c}{BA-Shapes} & \multicolumn{2}{c}{BA-Community} & \multicolumn{2}{c}{Tree-Grid} \\
\cmidrule(lr){2-3} \cmidrule(lr){4-5} \cmidrule(lr){6-7}
& $Fid_{\alpha_1,+}$ $\uparrow$ & $Fid_{\alpha_2,-}$ $\downarrow$ & $Fid_{\alpha_1,+}$ $\uparrow$ & $Fid_{\alpha_2,-}$ $\downarrow$ & $Fid_{\alpha_1,+}$ $\uparrow$ & $Fid_{\alpha_2,-}$ $\downarrow$ \\
\midrule
PGExplainer$_{\text{link}}$ & 0.201 $\pm$ 0.197 & 0.285 $\pm$ 0.177 & 0.132 $\pm$ 0.135 & 0.229 $\pm$ 0.101 & 0.112 $\pm$ 0.048 & 0.178 $\pm$ 0.053 \\
HPME$_{\text{link}}$ & \textbf{0.225 $\pm$ 0.139} & \textbf{0.044 $\pm$ 0.077} & \textbf{0.221 $\pm$ 0.101} & \textbf{0.016 $\pm$ 0.038} & \textbf{0.166 $\pm$ 0.057} & \textbf{0.021 $\pm$ 0.028} \\
\bottomrule
\end{tabular}
\end{table*}

\section{Proof of Property 2}
\label{sec:proof_prop2}

For the explanatory subgraph distribution $P_\theta(G^*_\theta | G)$ and the pooled subgraph distribution $Q_\phi(G^*_{\text{pool}} | G)$, let $I_P := I(G;G^*_\theta)$ and $I_Q := I(G;G^*_{\text{pool}})$.
By definition, $I_P = H_P(G^*_\theta) - H_P(G^*_\theta | G)$ and $I_Q = H_Q(G^*_{\text{pool}}) - H_Q(G^*_{\text{pool}} | G)$.
Since $Q_\phi$ is induced by a deterministic pooling operator, $G^*_{\text{pool}}$ is uniquely determined by $G$ and thus $H_Q(G^*_{\text{pool}} | G)=0$, which implies $I_Q = H_Q(G^*_{\text{pool}})$.
Consequently,
\begin{equation}
\begin{aligned}
I_P - I_Q 
&= \big(H_P(G^*_\theta) - H_Q(G^*_{\text{pool}})\big) - H_P(G^*_\theta | G) \\
&\le \big|H_P(G^*_\theta) - H_Q(G^*_{\text{pool}})\big|.
\end{aligned}
\label{eq:ip_iq_reduce}
\end{equation}

Next we bound the marginal entropy gap by a distributional distance.
Let $\Omega$ be the finite discrete space of candidate explanatory subgraphs considered by the explainer under a fixed graph size or budget, and denote $|\Omega|=d$.
Let $\delta := \mathrm{TV}\!\left(P_\theta(G^*_\theta),\,Q_\phi(G^*_{\text{pool}})\right)$ be the total variation distance between the induced marginal distributions over $\Omega$.
By the Audenaert--Fannes inequality:
\begin{equation}
\big|H_P(G^*_\theta) - H_Q(G^*_{\text{pool}})\big|
\le \delta \log(d-1) + h(\delta),
\label{eq:af}
\end{equation}
where $h(\cdot)$ is the binary entropy function.
It remains to control $\delta$ using the conditional KL term in the optimization objective.
For each fixed $G$, define the conditional total variation 
$\delta(G):=\mathrm{TV}\!\left(P_\theta(\cdot| G),\,Q_\phi(\cdot| G)\right)$.
By convexity of total variation under marginalization, $\delta \le \mathbb{E}_G[\delta(G)]$.
Applying Pinsker's inequality pointwise yields:
\begin{equation}
\delta(G) \le \sqrt{\tfrac{1}{2} KL\!\left(Q_\phi(\cdot| G)\,||\,P_\theta(\cdot| G)\right)}.
\label{eq:pinsker_point}
\end{equation}
Taking expectation over $G$ and combining the above gives:
\begin{equation}
\delta \le \mathbb{E}_G[\delta(G)]
\le 
\mathbb{E}_G\!\left[\sqrt{\tfrac{1}{2} KL\!\left(Q_\phi(\cdot| G)\,||\,P_\theta(\cdot| G)\right)}\right].
\label{eq:tv_to_kl}
\end{equation}

Substituting \Eqref{eq:tv_to_kl} into \Eqref{eq:af}, and then into \Eqref{eq:ip_iq_reduce}, we obtain:
\begin{equation}
\begin{aligned}
I(G;G^*_\theta)
&\le I(G;G^*_{\text{pool}}) \\
&\quad + \log(d-1)\cdot \mathbb{E}_G\!\left[\sqrt{\tfrac{1}{2} KL\!\left(Q_\phi(\cdot| G)\,||\,P_\theta(\cdot| G)\right)}\right]
+ h(\delta).
\end{aligned}
\label{eq:mi_bound_with_h}
\end{equation}

When the alignment term $KL(Q_\phi || P_\theta)$ is small, $\delta$ is small by \Eqref{eq:tv_to_kl}. 
Since $h(\delta)$ is uniformly bounded by $\log 2$, whereas $\log(d-1)$ scales with the size of the subgraph space, the term $h(\delta)$ is negligible in comparison and can be absorbed into the constant.
Letting $C := \frac{\log(d-1)}{\sqrt{2}}$, we arrive at the simplified bound:
\begin{equation}
I(G;G^*_\theta) \le I(G;G^*_{\text{pool}})
+ C \cdot \mathbb{E}_G\!\left[\sqrt{KL\!\left(Q_\phi(\cdot| G)\,||\,P_\theta(\cdot| G)\right)}\right].
\end{equation} 
\hfill $\square$

\section{Justification of Replacing $H(Y | G^*)$ with $H(Y | G^*_{\text{pool}})$}
\label{sec:proof_replacement}

In ~\Eqref{eq:gib_tractable}, we evaluate the conditional entropy term using the pooled hard subgraph $G^*_{\text{pool}}$ rather than the soft explanatory subgraph sampled from $P_\theta(\cdot|G)$. 
Since $H(Y | G^*)$ is optimized via a surrogate prediction loss $\ell(Y,f(G^*))$, it suffices to bound the gap between losses on $G^*_\theta$ and $G^*_{\text{pool}}$.

Let $g^{pool}=G^*_{\text{pool}}(G)$ be the deterministic output of hard pooling and define $h(g)=\ell(Y,f(g))$ with $0\le h(g)\le L$. 
Because $Q_\phi(\cdot|G)$ is a delta distribution at $g^{pool}$, we have $\mathbb{E}_{g\sim Q_\phi(\cdot|G)}[h(g)]=h(g^{pool})$, and thus:
\begin{equation}
\left|\mathbb{E}_{g\sim P_\theta(\cdot|G)}[h(g)]-\mathbb{E}_{g\sim Q_\phi(\cdot|G)}[h(g)]\right|
\le 2L\cdot \mathrm{TV}\big(P_\theta(\cdot|G),Q_\phi(\cdot|G)\big).
\end{equation}
By Pinsker's inequality and the symmetry of $\mathrm{TV}(\cdot,\cdot)$:
\begin{equation}
\mathrm{TV}\big(P_\theta(\cdot|G),Q_\phi(\cdot|G)\big)
\le \sqrt{\tfrac{1}{2}\, KL\big(Q_\phi(\cdot|G) \,||\, P_\theta(\cdot|G)\big)}.
\end{equation}
Combining the above yields:
\begin{equation}
\begin{aligned}
\left|\mathbb{E}_{g\sim P_\theta(\cdot|G)}[\ell(Y,f(g))]-\ell(Y,f(g^{pool}))\right| \\
\le L\sqrt{2\, KL\big(Q_\phi(\cdot|G) \,||\, P_\theta(\cdot|G)\big)}.
\end{aligned}
\label{eq:replacement_bound}
\end{equation}

Therefore, minimizing $KL(Q_\phi(\cdot|G) || P_\theta(\cdot|G))$ drives the prediction losses on $G^*_\theta$ and $G^*_{\text{pool}}$ to be close for every input graph $G$. 
Since the conditional entropy $H(Y | G^*)$ is optimized through this prediction loss in both classification and regression settings, the above bound implies that optimizing $H(Y | G^*_{\text{pool}})$ provides an accurate surrogate for optimizing $H(Y | G^*)$. 
This justifies replacing $H(Y | G^*)$ with $H(Y | G^*_{\text{pool}})$ in ~\Eqref{eq:gib_tractable}.

\section{Extensive Experiments}
\label{sec:app:moreex}

\subsection{Multi-task Explanation Quality Evaluation}
\label{sec:multi-task}

HPME presents a general GIB framework that incorporates hard structural perturbations to eliminate the redundant edges inherent in soft masks, thereby achieving superior performance across diverse tasks, including node classification and link prediction. 
For node classification task, we extract a 3-hop ego subgraph centered on the target node as the target graph to be explained. For link prediction task, we follow SEAL~\cite{zhang2018link} to construct a local subgraph around the target edge. HPME then applies the same perturbation-based procedure as in graph-level tasks to identify critical subgraphs within these target graphs as explanations. Further experimental details for both tasks are provided in Appendix~\ref{sec:full_setup}.

\begin{table}[h]
\centering
\caption{AUC-ROC edge-level explanation accuracy on three node classification datasets. Higher scores indicate better performance.}
\label{tab:node_class}
\small 
\renewcommand{\arraystretch}{1.1} 
\begin{tabular*}{\linewidth}{@{\extracolsep{\fill}}lccc}
\toprule
Method & BA-Shapes & BA-Community & Tree-Grid \\
\midrule
PGExplainer & 0.902 $\pm$ 0.027 & 0.653 $\pm$ 0.111 & 0.205 $\pm$ 0.113 \\
MixupExplainer & 0.801 $\pm$ 0.145 & 0.654 $\pm$ 0.113 & 0.179 $\pm$ 0.054 \\
HPME & \textbf{0.991 $\pm$ 0.001} & \textbf{0.906 $\pm$ 0.154} & \textbf{0.678 $\pm$ 0.039} \\
\bottomrule
\end{tabular*}
\end{table}

For the node classification task, we conduct experiments on the BA-Shapes, BA-Community, and Tree-Grid datasets, comparing our method HPME against representative baselines, including PGExplainer~\cite{luo2020parameterized} and MixupExplainer~\cite{zhang2023mixupexplainer}. As shown in Table~\ref{tab:node_class}, HPME consistently outperforms these baselines across all datasets. The most significant improvement is observed on the Tree-Grid dataset, where HPME achieves a performance gain of $0.473$. As ground-truth explanations are unavailable for these datasets in link prediction task, we adopt the robust fidelity metrics $Fid_{\alpha_1,+}$ and $Fid_{\alpha_2,-}$~\cite{zheng2023towards} for evaluation. The results in Table~\ref{tab:fidelity_linkpre} show that HPME$_{\text{link}}$ consistently outperforms PGExplainer$_{\text{link}}$ across all datasets under both metrics. Overall, the experimental results across node classification, link prediction, graph classification, and graph regression demonstrate that, compared to baseline methods, HPME consistently generates in-distribution mixup graphs and produces faithful explanations.
\begin{figure*}[t]
    \centering
    \begin{subfigure}[b]{0.24\textwidth}
        \includegraphics[width=\linewidth]{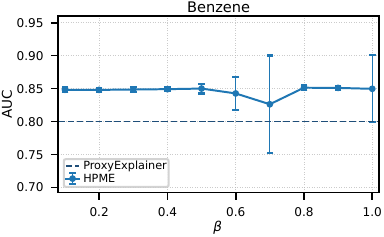}
    \end{subfigure}\hfill
    \begin{subfigure}[b]{0.24\textwidth}
        \includegraphics[width=\linewidth]{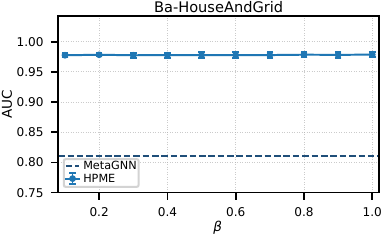}
    \end{subfigure}\hfill
    \begin{subfigure}[b]{0.24\textwidth}
        \includegraphics[width=\linewidth]{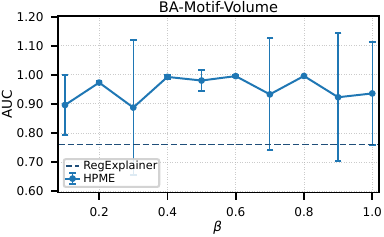}
    \end{subfigure}\hfill
    \begin{subfigure}[b]{0.24\textwidth}
        \includegraphics[width=\linewidth]{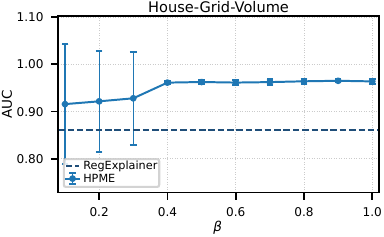}
    \end{subfigure}
    \caption{Sensitivity analysis of hyperparameter $\beta$ for BCE loss.}
    \label{fig:sensitivity}
\end{figure*}
\begin{figure*}[t]
    \centering
    \begin{subfigure}[b]{0.24\textwidth}
        \includegraphics[width=\linewidth]{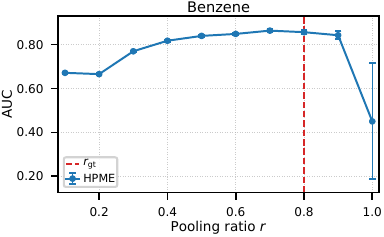}
    \end{subfigure}\hfill
    \begin{subfigure}[b]{0.24\textwidth}
        \includegraphics[width=\linewidth]{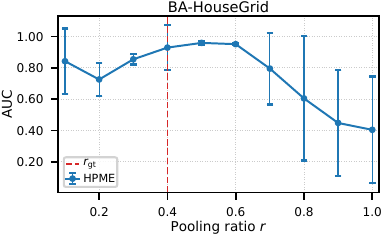}
    \end{subfigure}\hfill
    \begin{subfigure}[b]{0.24\textwidth}
        \includegraphics[width=\linewidth]{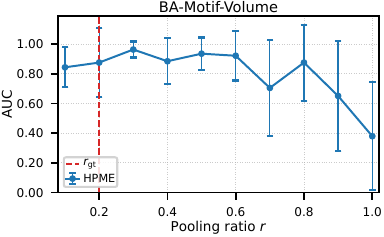}
    \end{subfigure}\hfill
    \begin{subfigure}[b]{0.24\textwidth}
        \includegraphics[width=\linewidth]{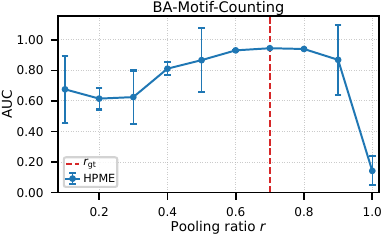}
    \end{subfigure}
    \caption{Sensitivity analysis of pooling ratio $r$.}
    \label{fig:pooling_size_sensitivity}
\end{figure*}

\begin{table*}[t]
\centering
\renewcommand{\arraystretch}{1.1}

\caption{Comparison of GCN, GIN, and GAT on classification datasets using ACC and Macro-F1 metrics.}
\label{tab:classification_results}

\begin{tabular}{lcccccc}
\toprule
\textbf{Dataset} & \multicolumn{2}{c}{\textbf{GCN}} & \multicolumn{2}{c}{\textbf{GIN}} & \multicolumn{2}{c}{\textbf{GAT}} \\
\cmidrule(lr){2-3} \cmidrule(lr){4-5} \cmidrule(lr){6-7}
& ACC & Macro\_F1 & ACC & Macro\_F1 & ACC & Macro\_F1 \\
\midrule
BA-2motifs         & 99.00 & 98.98 & \textbf{100.00} & \textbf{100.00} & 44.00 & 30.55 \\
BA-HouseGrid       & 99.90 & 99.89 & \textbf{100.00} & \textbf{100.00} & 49.10 & 32.93 \\
BA-HouseAndGrid    & 98.40 & 98.39 & \textbf{99.90}  & \textbf{99.89}  & 48.90 & 32.84 \\
SPMotif            & \textbf{96.16} & \textbf{96.15} & 95.88  & 95.89  & 34.05 & 18.98 \\
Alkane-Carbonyl    & \textbf{95.57} & \textbf{95.13} & 92.03  & 91.24  & 94.69 & 94.31 \\
Fluoride-Carbonyl  & 94.35 & 89.99 & \textbf{97.23}  & \textbf{95.03}  & 83.29 & 45.44 \\
Benzene            & 90.58 & 90.53 & \textbf{92.66}  & \textbf{92.65}  & 79.08 & 79.06 \\
\bottomrule
\end{tabular}

\end{table*}

\subsection{Hyper-parameter Sensitivity Study}
We further investigate the sensitivity of our proposed method to the hyperparameter $\beta$, which controls the weight of the BCE loss. Specifically, we conduct experiments on four representative datasets, including the Ba-HouseAndGrid and Benzene classification datasets, as well as the BA-Motif-Volume and House-Grid-Volume regression datasets. We vary $\beta$ from 0.1 to 1.0 in increments of 0.1 while keeping the contrastive learning weight fixed at 1.0. The experiments are conducted under ten random seeds, where the solid lines represent the average AUC of our method and the dashed lines correspond to the second-best performing method. The experimental results are presented in \Figref{fig:sensitivity}. 
The results indicate that HPME maintains stable performance across both classification and regression tasks. For some datasets, the AUC slightly decreases and exhibits larger variance when the weight of the BCE loss is small. This phenomenon can be attributed to the role of the BCE loss in constraining the edge weights to remain faithful to the explanation structure extracted by graph pooling, while simultaneously reducing the weights of non-explanatory structures. Without this constraint, the edge weights may not be effectively optimized, leading to a decline in performance.

We study the sensitivity of HPME to the pooling ratio $r$, which controls the size of the selected subgraph during graph pooling. Specifically, we vary $r$ from $0.1$ to $1.0$ on two classification datasets, Benzene and BA-HouseGrid, and two regression datasets, BA-Motif-Volume and BA-Motif-Counting. For each dataset, $r_{\mathrm{gt}}$ denotes the reference pooling ratio derived from the ground-truth explanation size. As shown in \Figref{fig:pooling_size_sensitivity}, HPME maintains strong performance across a broad range of suitable pooling ratios around $r_{\mathrm{gt}}$. When $r$ is overly large, the selected subgraph may contain redundant structures, while an overly small $r$ may remove critical explanatory substructures. These results demonstrate that HPME is robust to pooling-size choices within reasonable ranges and can effectively resist redundant structures.

\begin{table*}[t]
\centering
\renewcommand{\arraystretch}{1.1}

\caption{Comparison of GCN, GIN, and GAT on regression datasets using RMSE and MAPE metrics. MAPE values are marked with dashes for BA-Motif-Counting and Triangles due to the presence of label-zero samples that make MAPE undefined.}
\label{tab:regression_results}

\begin{tabular}{lcccccc}
\toprule
\textbf{Dataset} & \multicolumn{2}{c}{\textbf{GCN}} & \multicolumn{2}{c}{\textbf{GIN}} & \multicolumn{2}{c}{\textbf{GAT}} \\
\cmidrule(lr){2-3} \cmidrule(lr){4-5} \cmidrule(lr){6-7}
& RMSE & MAPE & RMSE & MAPE & RMSE & MAPE \\
\midrule
BA-Motif-Volume       & \textbf{234.20} & \textbf{7.47}  & 398.96 & 14.94 & 485.31 & 17.45 \\
BA-Motif-Counting     & \textbf{0.39}   & \textemdash & 1.94   & \textemdash & 2.84   & \textemdash \\
Triangles             & \textbf{2.00}   & \textemdash & 2.46   & \textemdash & 2.55   & \textemdash \\
House-Grid-Volume     & \textbf{11.32}  & \textbf{2.45}  & 94.54  & 29.56 & 49.85  & 13.88 \\
House-OrGrid-Volume   & \textbf{34.15}  & \textbf{6.69}  & 152.14 & 28.17 & 53.64  & 10.27 \\
Crippen               & \textbf{0.92}   & \textbf{30.71} & 1.91   & 123.17 & 1.16   & 74.68 \\
\bottomrule
\end{tabular}

\end{table*}

\begin{table*}[ht]
\centering
\caption{Explanation Quality on Classification (Alkane-Carbonyl, BA-HouseGrid) and Regression (House-OrGrid-Volume, BA-Motif-Volume) datasets via SimOAR and Robust Fidelity.}
\label{tab:fidelity_based_metrics}
\renewcommand{\arraystretch}{1.1}
\begin{tabular}{@{}lcccc@{}}
\toprule
\multirow{2}{*}{Method} & Alkane-Carbonyl & BA-HouseGrid & \multicolumn{2}{c}{BA-HouseGrid} \\
\cmidrule(lr){2-3} \cmidrule(lr){4-5}
& SimOAR $\uparrow$ & SimOAR $\uparrow$ & $Fid_{\alpha_1,+}$ $\uparrow$ & $Fid_{\alpha_2,-}$ $\downarrow$ \\
\midrule
GNNExplainer   & 0.780 $\pm$ 0.006 & 0.649 $\pm$ 0.006 & 0.021 $\pm$ 0.049 & 0.241 $\pm$ 0.370 \\
PGExplainer    & 0.802 $\pm$ 0.050 & 0.667 $\pm$ 0.080 & 0.012 $\pm$ 0.036 & 0.319 $\pm$ 0.433 \\
MixupExplainer & 0.839 $\pm$ 0.035 & 0.767 $\pm$ 0.092 & 0.026 $\pm$ 0.045 & 0.281 $\pm$ 0.408 \\
ProxyExplainer & 0.829 $\pm$ 0.011 & 0.720 $\pm$ 0.082 & 0.016 $\pm$ 0.041 & 0.245 $\pm$ 0.392 \\
HPME           & \textbf{0.857 $\pm$ 0.050} & \textbf{0.834 $\pm$ 0.067} & \textbf{0.038 $\pm$ 0.076} & \textbf{0.072 $\pm$ 0.244} \\
\midrule
\multirow{2}{*}{Method} & House-OrGrid-Volume & BA-Motif-Volume & \multicolumn{2}{c}{BA-Motif-Volume} \\
\cmidrule(lr){2-3} \cmidrule(lr){4-5}
& SimOAR $\uparrow$ & SimOAR $\uparrow$ & $Fid_{\alpha_1,+}$ $\uparrow$ & $Fid_{\alpha_2,-}$ $\downarrow$ \\
\midrule
MixupExplainer & 0.810 $\pm$ 0.075 & 0.215 $\pm$ 0.058 & 0.158 $\pm$ 0.315 & 1.298 $\pm$ 1.442 \\
RegExplainer   & 0.752 $\pm$ 0.086 & 0.204 $\pm$ 0.057 & 0.184 $\pm$ 0.332 & 1.162 $\pm$ 1.439 \\
HPME           & \textbf{0.840 $\pm$ 0.094} & \textbf{0.247 $\pm$ 0.056} & \textbf{0.413 $\pm$ 0.396} & \textbf{0.061 $\pm$ 0.117} \\
\bottomrule
\end{tabular}
\end{table*}

\subsection{Backbone Models Analysis}
\label{sec:backbone_analysis}
We conduct a performance analysis of common backbone models employed by explainability methods on both classification and regression tasks. For the classification task, model performance is evaluated using ACC and Macro\_F1 metrics on four synthetic datasets, including BA-2Motifs, BA-HouseGrid, BA-HouseAndGrid, and SPMotif, as well as three real-world datasets including Alkane-Carbonyl, Fluoride-Carbonyl, and Benzene. For the regression task, RMSE and MAPE are adopted as evaluation metrics on five synthetic datasets including BA-Motif-Volume, BA-Motif-Counting, Triangles, House-Grid-Volume, and House-OrGrid-Volume, along with one real-world dataset, Crippen.

Experimental results demonstrate that GCN and GIN achieve strong performance in classification tasks, while GCN consistently attains the best performance across regression benchmarks. Therefore, to ensure that the explanation quality is not confounded by the performance of the backbone model, we choose GCN as the backbone model for explainer evaluation.

\subsection{Explanation Quality Evaluation via Robust Fidelity Metrics}
Additionally, we follow existing works~\cite{yuan2021explainability,bui2024explaining} to evaluate the quality of the identified explanations via fidelity-based metrics. Owing to the reliability issue caused by the OOD problem in standard fidelity metrics~\cite{amara2023ginx}, we utilize the robust fidelity measures SimOAR~\cite{fang2023evaluating} with default perturbation ratio $0.1$ and ($Fid_{\alpha_1,+}$, $Fid_{\alpha_2,-}$)~\cite{zheng2023towards}with default parameters $\alpha_1=0.1$, $\alpha_2=0.9$. Table~\ref{tab:fidelity_based_metrics} demonstrates that the quality of the explanatory subgraphs obtained by HPME outperform all baselines under both metrics.

\begin{table*}[h]
\centering
\caption{Additional distribution-shift evaluation on classification datasets.}
\label{tab:app_ood_classification}
\small
\setlength{\tabcolsep}{4pt}
\renewcommand{\arraystretch}{1.1}
\begin{tabular}{lcccccccc}
\toprule
\multirow{2}{*}{Method}
& \multicolumn{2}{c}{BA-HouseGrid}
& \multicolumn{2}{c}{BA-HouseAndGrid}
& \multicolumn{2}{c}{Benzene}
& \multicolumn{2}{c}{Fluoride-Carbonyl} \\
\cmidrule(lr){2-3} \cmidrule(lr){4-5} \cmidrule(lr){6-7} \cmidrule(lr){8-9}
& Cos. $\uparrow$ & Euc. $\downarrow$
& Cos. $\uparrow$ & Euc. $\downarrow$
& Cos. $\uparrow$ & Euc. $\downarrow$
& Cos. $\uparrow$ & Euc. $\downarrow$ \\
\midrule
GT
& 0.688 $\pm$ 0.102 & 1.044 $\pm$ 0.168
& 0.613 $\pm$ 0.013 & 1.436 $\pm$ 0.040
& 0.821 $\pm$ 0.059 & 0.953 $\pm$ 0.126
& 0.918 $\pm$ 0.040 & 0.684 $\pm$ 0.150 \\
PGExplainer
& 0.687 $\pm$ 0.108 & 1.095 $\pm$ 0.164
& 0.867 $\pm$ 0.064 & 0.899 $\pm$ 0.213
& 0.885 $\pm$ 0.047 & 0.777 $\pm$ 0.136
& 0.898 $\pm$ 0.033 & 0.778 $\pm$ 0.131 \\
MixupExplainer
& 0.781 $\pm$ 0.071 & 0.973 $\pm$ 0.148
& 0.966 $\pm$ 0.019 & 0.477 $\pm$ 0.119
& 0.907 $\pm$ 0.038 & 0.771 $\pm$ 0.147
& 0.895 $\pm$ 0.042 & 0.783 $\pm$ 0.141 \\
ProxyExplainer
& 0.802 $\pm$ 0.034 & 0.706 $\pm$ 0.062
& 0.970 $\pm$ 0.006 & 0.442 $\pm$ 0.040
& 0.565 $\pm$ 0.039 & 1.471 $\pm$ 0.130
& 0.723 $\pm$ 0.039 & 1.272 $\pm$ 0.128 \\
HPME
& \textbf{0.868 $\pm$ 0.069} & \textbf{0.661 $\pm$ 0.176}
& \textbf{0.971 $\pm$ 0.021} & \textbf{0.425 $\pm$ 0.113}
& \textbf{0.941 $\pm$ 0.031} & \textbf{0.578 $\pm$ 0.184}
& \textbf{0.937 $\pm$ 0.035} & \textbf{0.604 $\pm$ 0.144} \\
\bottomrule
\end{tabular}
\end{table*}

\begin{table*}[h]
\centering
\caption{Additional distribution-shift evaluation on regression datasets.}
\label{tab:app_ood_regression}
\small
\setlength{\tabcolsep}{4pt}
\renewcommand{\arraystretch}{1.1}
\begin{tabular}{lcccccccc}
\toprule
\multirow{2}{*}{Method}
& \multicolumn{2}{c}{BA-Motif-Volume}
& \multicolumn{2}{c}{BA-Motif-Counting}
& \multicolumn{2}{c}{Crippen}
& \multicolumn{2}{c}{Triangles} \\
\cmidrule(lr){2-3} \cmidrule(lr){4-5} \cmidrule(lr){6-7} \cmidrule(lr){8-9}
& Cos. $\uparrow$ & Euc. $\downarrow$
& Cos. $\uparrow$ & Euc. $\downarrow$
& Cos. $\uparrow$ & Euc. $\downarrow$
& Cos. $\uparrow$ & Euc. $\downarrow$ \\
\midrule
GT
& 0.696 $\pm$ 0.107 & 1.138 $\pm$ 0.140
& 0.663 $\pm$ 0.077 & 1.196 $\pm$ 0.120
& 0.869 $\pm$ 0.080 & 0.953 $\pm$ 0.154
& 0.897 $\pm$ 0.038 & 0.684 $\pm$ 0.075 \\
PGExplainer
& 0.799 $\pm$ 0.092 & 0.945 $\pm$ 0.169
& 0.949 $\pm$ 0.021 & 0.489 $\pm$ 0.093
& 0.873 $\pm$ 0.077 & 0.677 $\pm$ 0.167
& 0.951 $\pm$ 0.025 & 0.248 $\pm$ 0.059 \\
MixupExplainer
& 0.950 $\pm$ 0.021 & 0.488 $\pm$ 0.098
& 0.844 $\pm$ 0.102 & 0.794 $\pm$ 0.207
& 0.841 $\pm$ 0.099 & 0.807 $\pm$ 0.204
& 0.947 $\pm$ 0.023 & 0.256 $\pm$ 0.045 \\
RegExplainer
& 0.876 $\pm$ 0.063 & 0.775 $\pm$ 0.178
& 0.950 $\pm$ 0.021 & 0.485 $\pm$ 0.093
& 0.847 $\pm$ 0.094 & 0.794 $\pm$ 0.188
& 0.953 $\pm$ 0.017 & 0.247 $\pm$ 0.044 \\
HPME
& \textbf{0.953 $\pm$ 0.027} & \textbf{0.458 $\pm$ 0.169}
& \textbf{0.969 $\pm$ 0.015} & \textbf{0.401 $\pm$ 0.082}
& \textbf{0.936 $\pm$ 0.057} & \textbf{0.431 $\pm$ 0.168}
& \textbf{0.985 $\pm$ 0.014} & \textbf{0.143 $\pm$ 0.051} \\
\bottomrule
\end{tabular}
\end{table*}

\subsection{Additional Results on Distribution Shift}
\label{sec:app:ood}

To further evaluate whether the generated mixup graphs alleviate the distribution shift between the original graph and the explanatory subgraph, we provide complete results on both classification and regression datasets. Following Section~\ref{sec:exp:ood}, we compute the cosine similarity and Euclidean distance between the graph representations of the original graphs and those of the ground-truth explanatory subgraphs or generated mixup graphs. Higher cosine similarity and lower Euclidean distance indicate that the generated graphs are closer to the original graph distribution.

The results in Tables~\ref{tab:app_ood_classification} and~\ref{tab:app_ood_regression} further confirm that directly using the ground-truth explanatory subgraphs leads to clear distribution shifts from the original graphs, as reflected by lower cosine similarity and higher Euclidean distance. Compared with existing explainers, HPME consistently produces mixup graphs with higher cosine similarity and lower Euclidean distance across both classification and regression datasets. These results demonstrate that the proposed hard-perturbation mixup strategy better preserves the distributional characteristics of the original graphs while retaining the explanatory structures.

\subsection{Extensive Case Study} 
\label{sec:more_casestudy}
\subsubsection{Visualization of Additional Explanation Results} 
\label{sec:more_explanation}
\begin{figure*}[h]
    \centering
    \vspace{0.8em}
    \begin{subfigure}[b]{0.19\textwidth}
        \includegraphics[width=\linewidth]{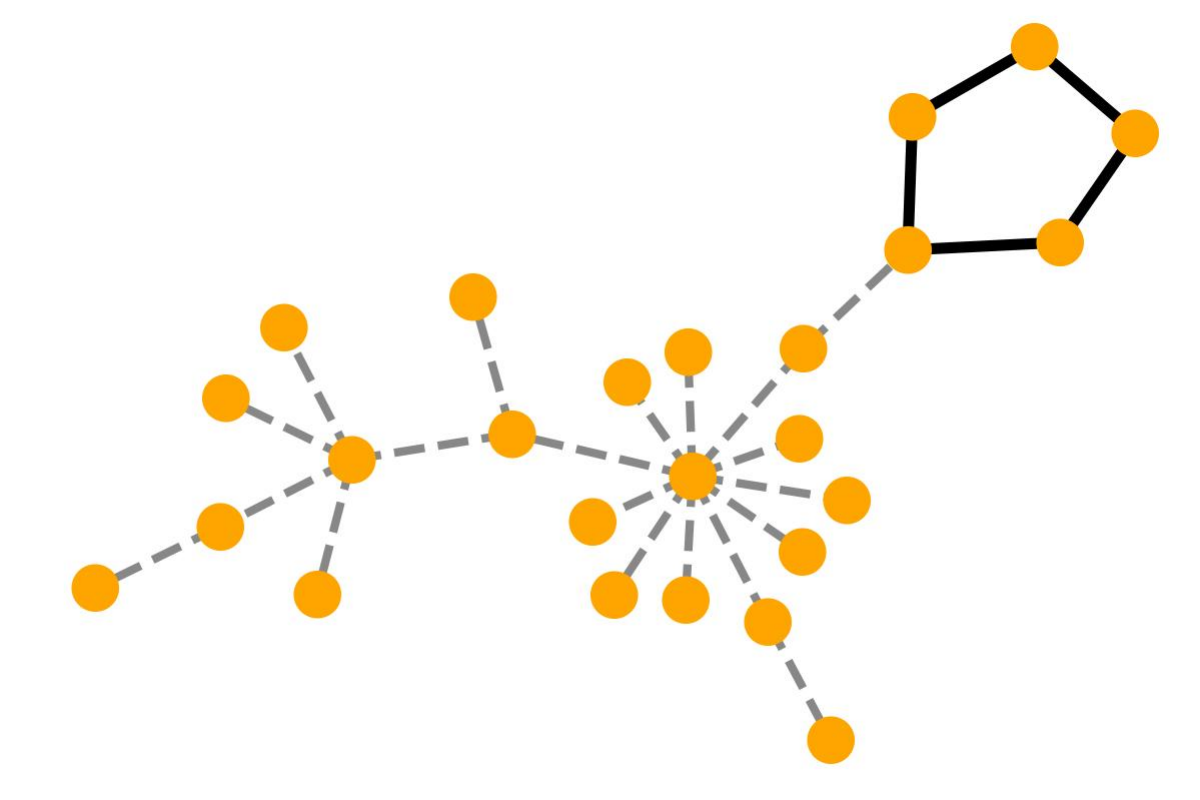}
    \end{subfigure}
    \begin{subfigure}[b]{0.19\textwidth}
        \includegraphics[width=\linewidth]{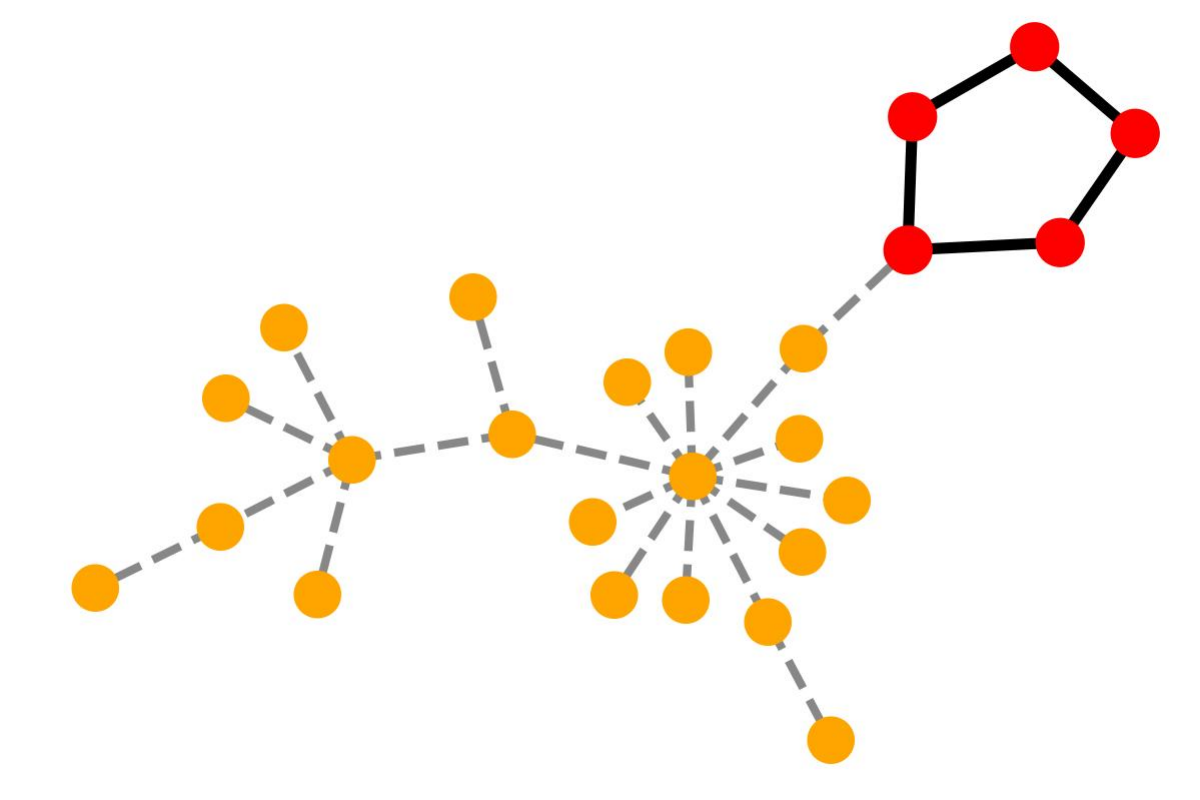}
    \end{subfigure}
    \begin{subfigure}[b]{0.19\textwidth}
        \includegraphics[width=\linewidth]{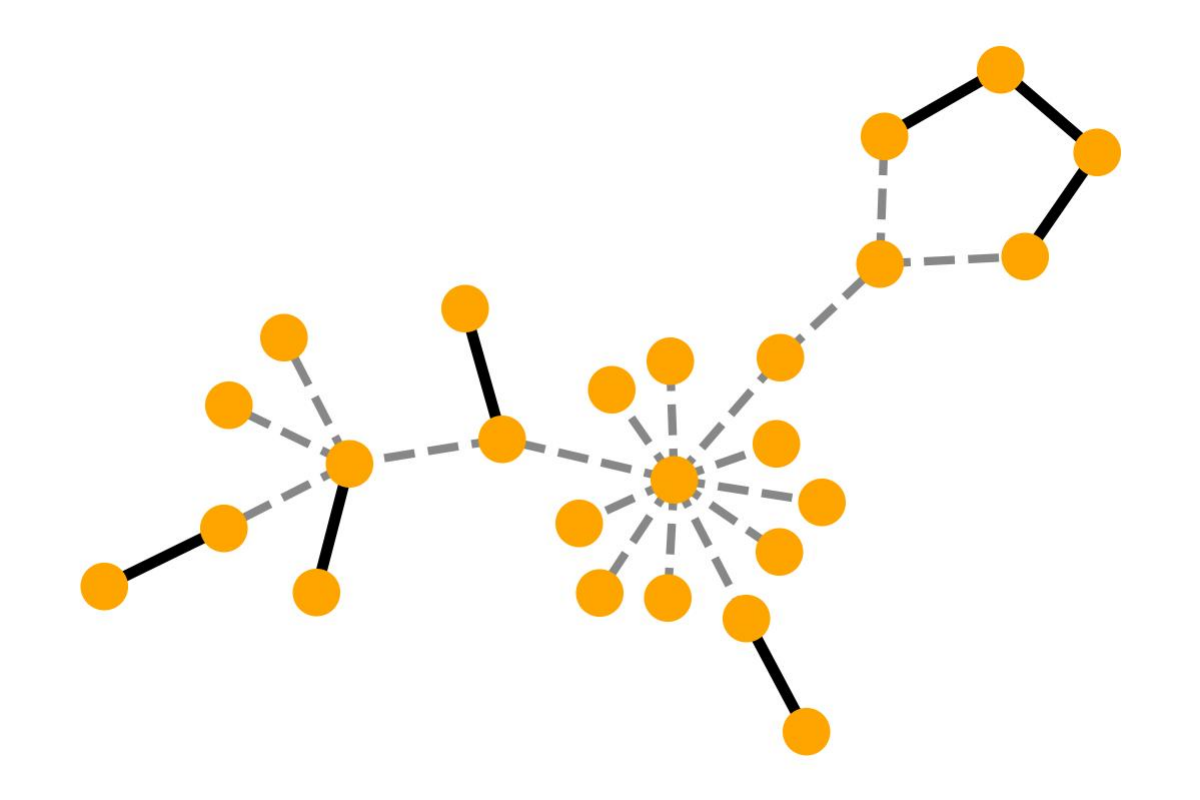}
    \end{subfigure}
    \begin{subfigure}[b]{0.19\textwidth}
        \includegraphics[width=\linewidth]{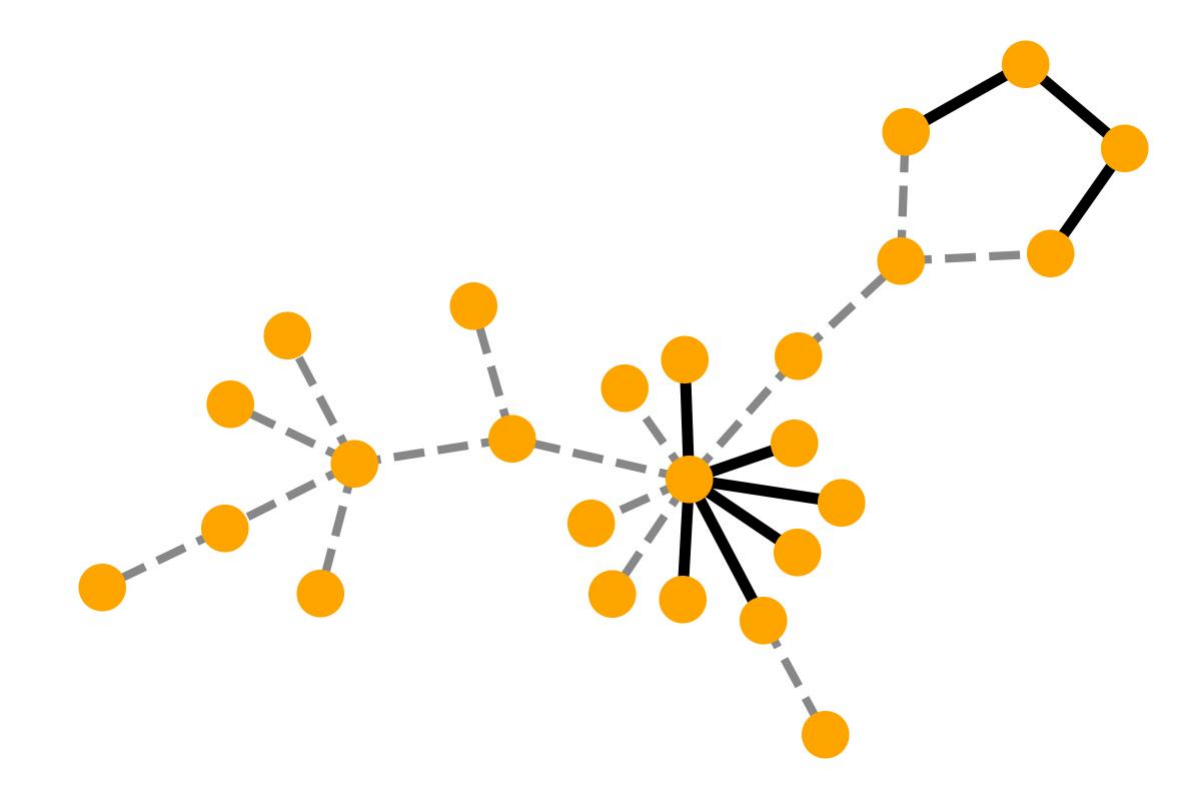}
    \end{subfigure}
    \begin{subfigure}[b]{0.19\textwidth}
        \includegraphics[width=\linewidth]{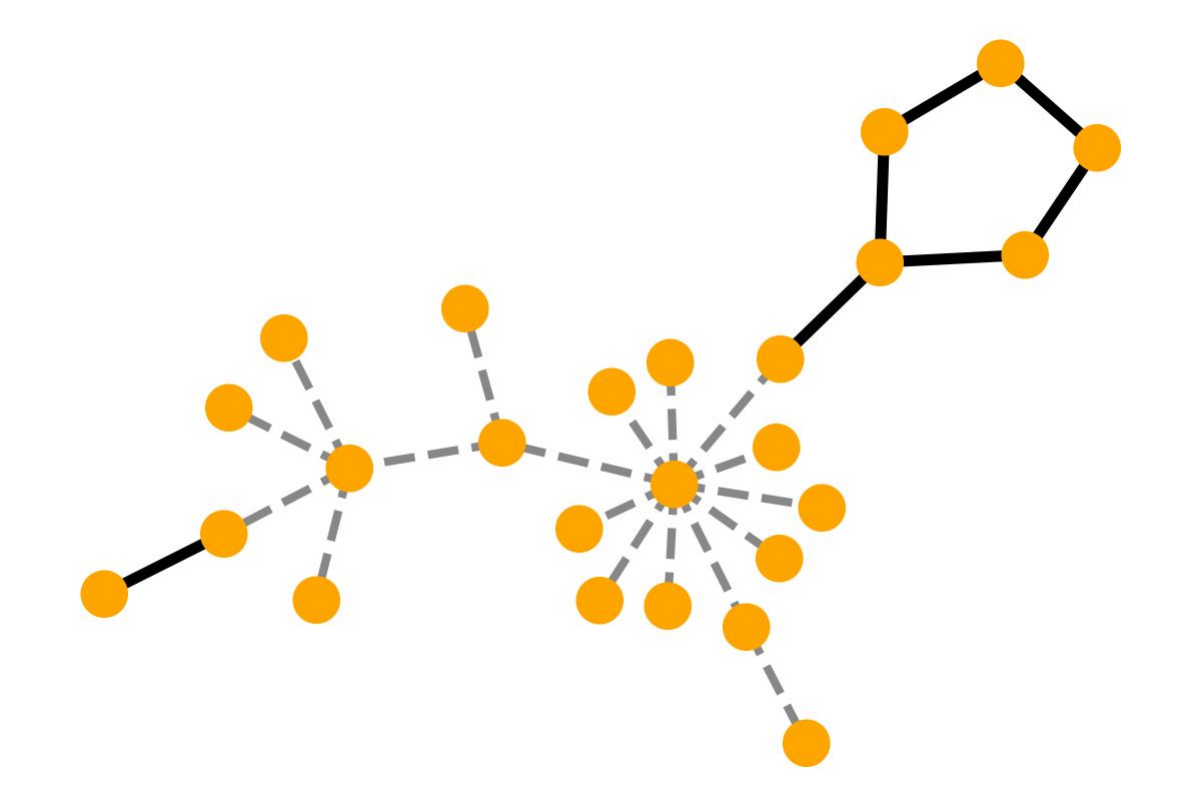}
    \end{subfigure}
    \par\vspace{0.8em}
    \begin{subfigure}[b]{0.19\textwidth}
        \includegraphics[width=\linewidth]{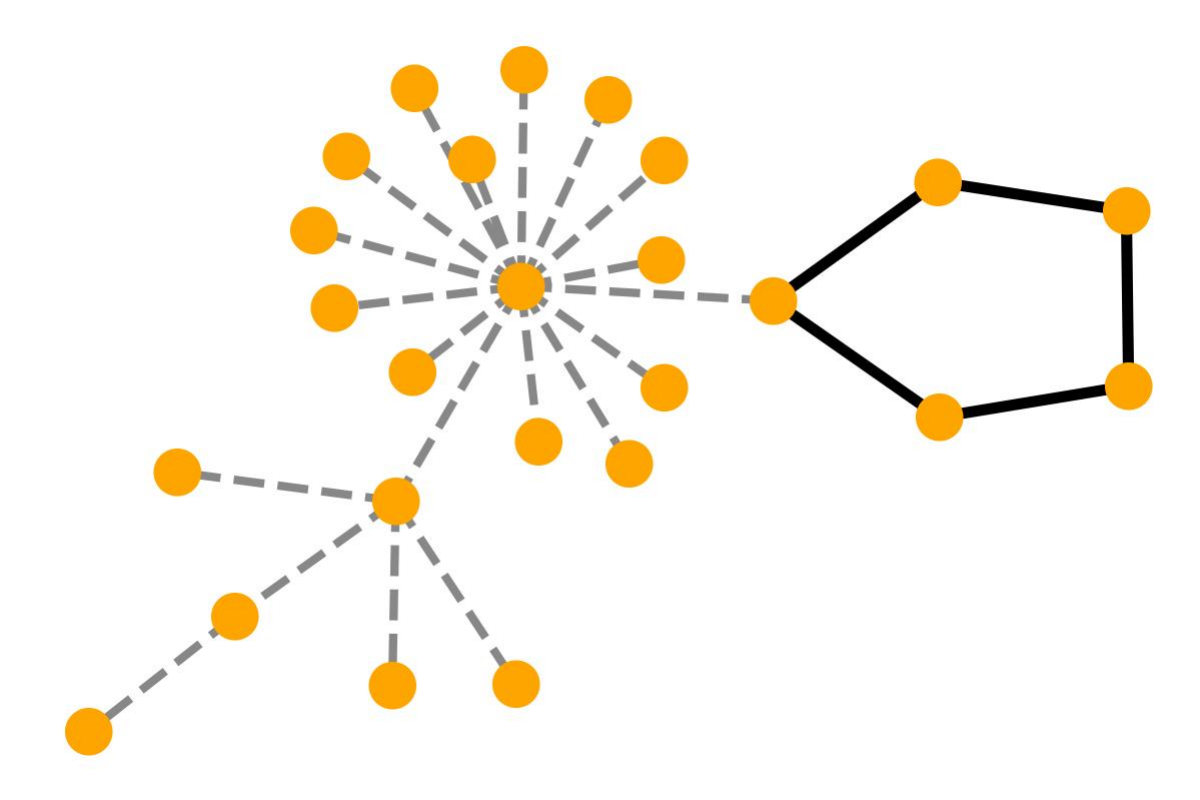}
    \end{subfigure}
    \begin{subfigure}[b]{0.19\textwidth}
        \includegraphics[width=\linewidth]{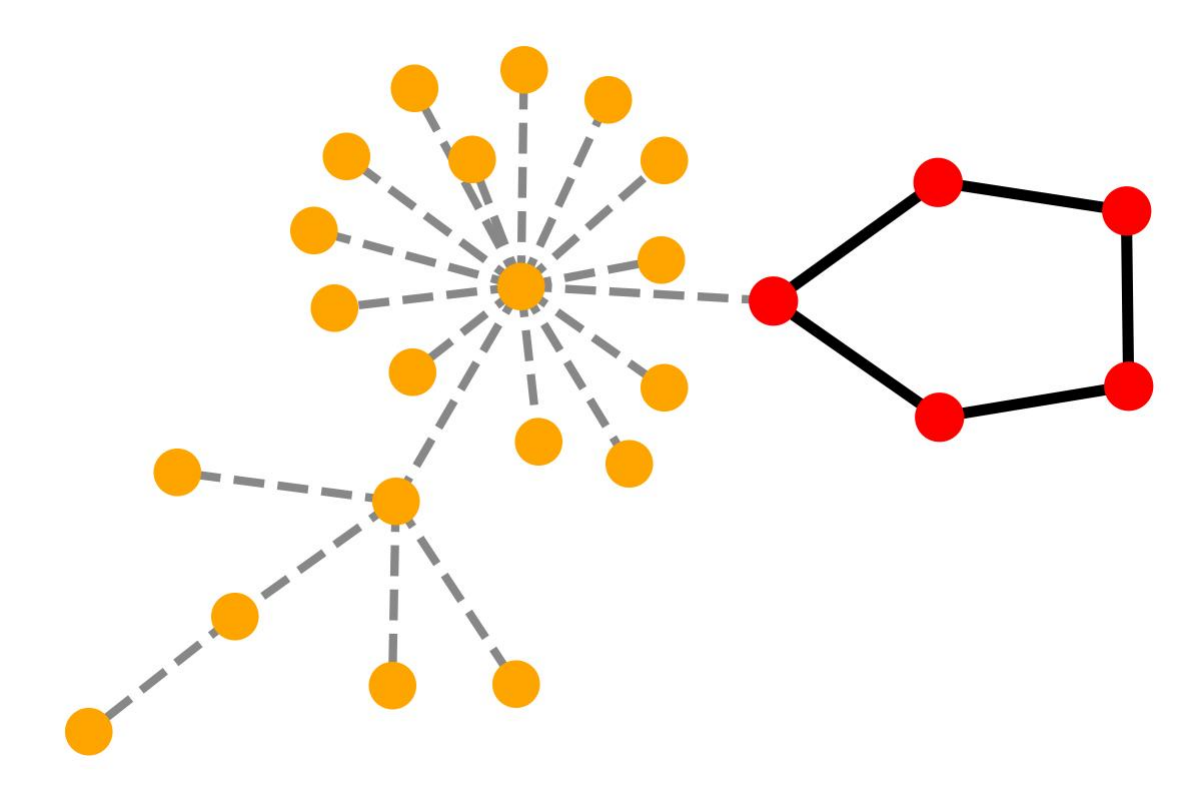}
    \end{subfigure}
    \begin{subfigure}[b]{0.19\textwidth}
        \includegraphics[width=\linewidth]{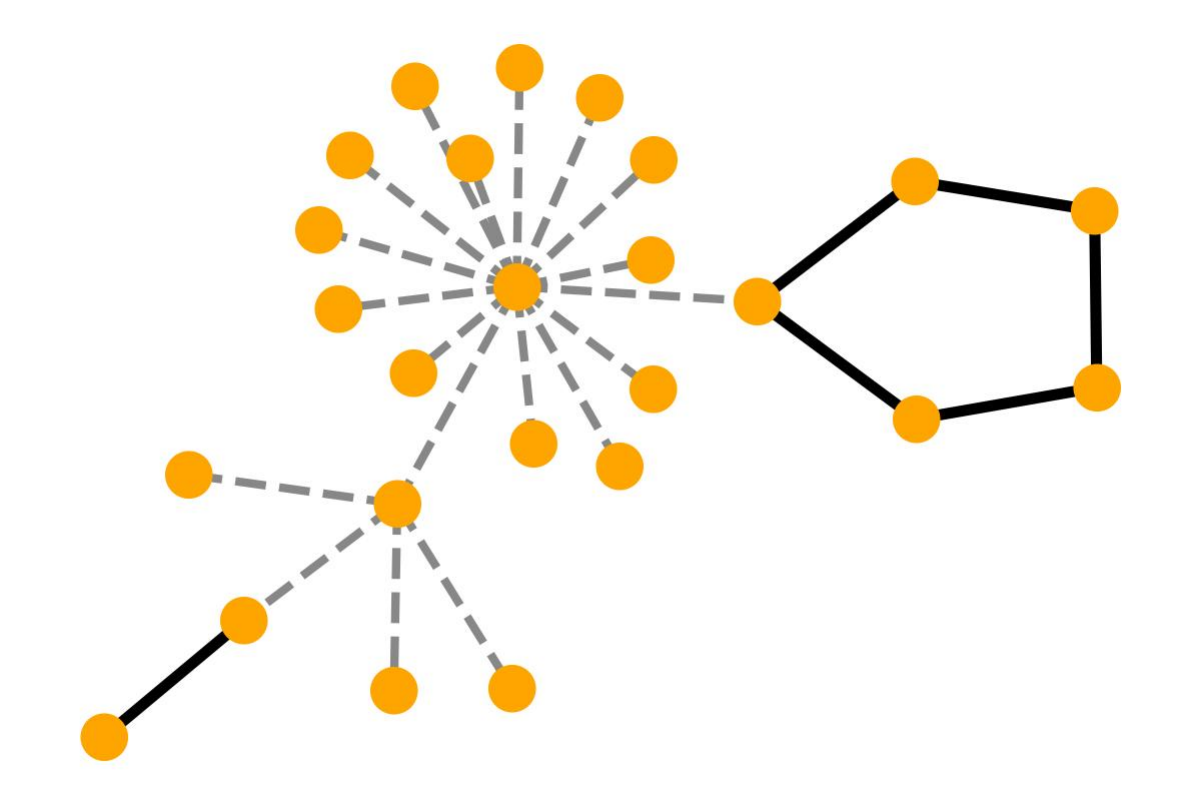}
    \end{subfigure}
    \begin{subfigure}[b]{0.19\textwidth}
        \includegraphics[width=\linewidth]{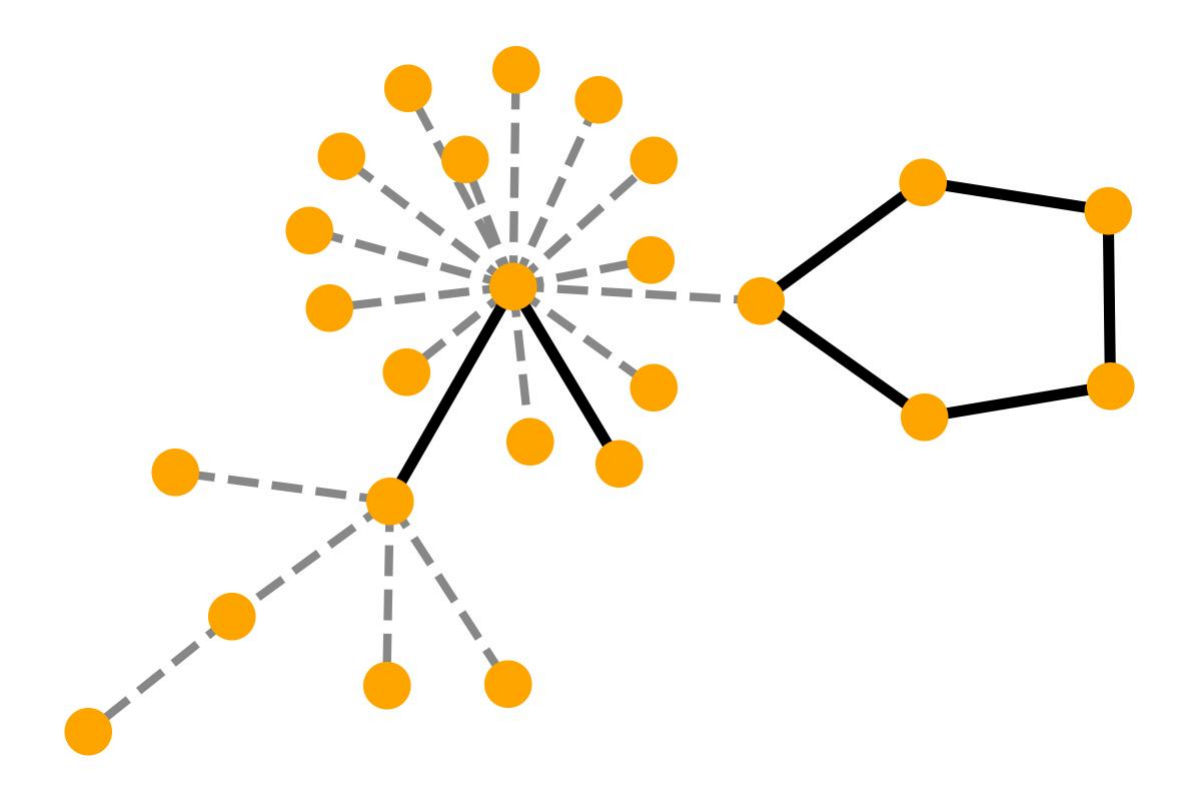}
    \end{subfigure}
    \begin{subfigure}[b]{0.19\textwidth}
        \includegraphics[width=\linewidth]{figures/app/result_ba2/21_mix.pdf}
    \end{subfigure}
    \par\vspace{0.8em}
    \begin{subfigure}[b]{0.19\textwidth}
        \includegraphics[width=\linewidth]{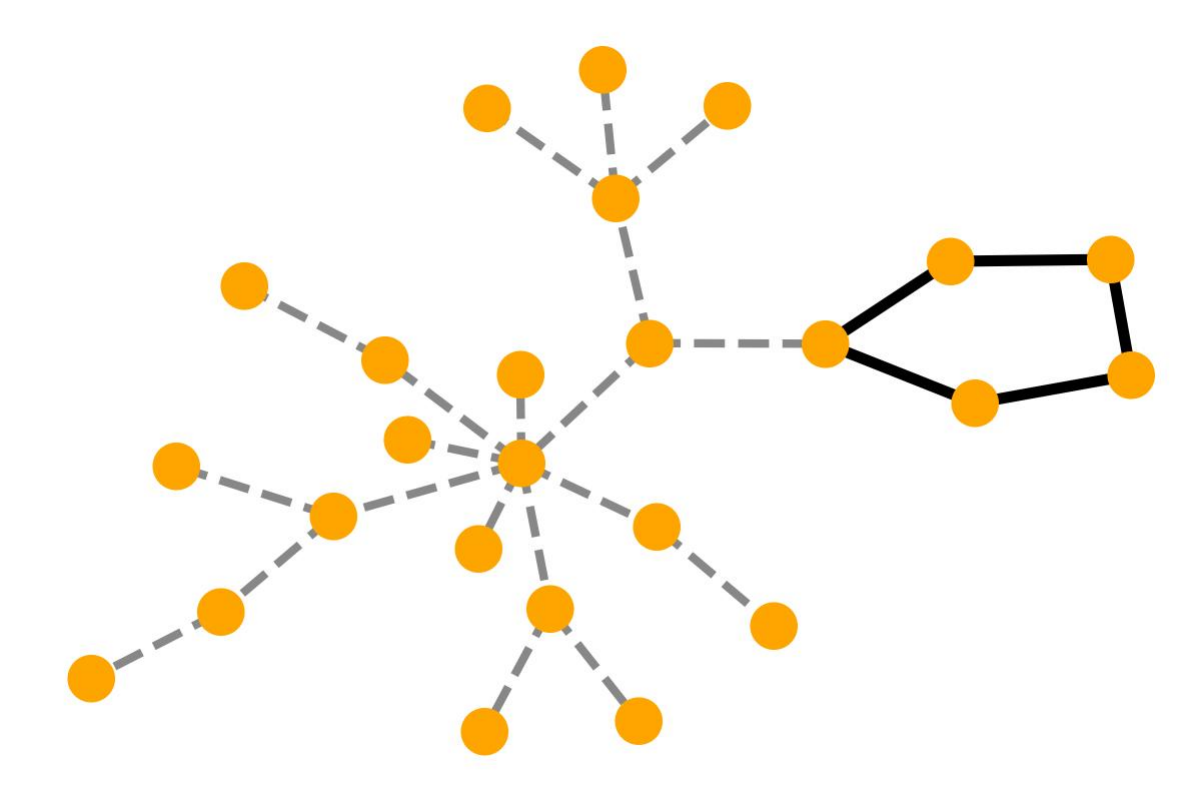}
        \caption{Ground Truth}
    \end{subfigure}
    \begin{subfigure}[b]{0.19\textwidth}
        \includegraphics[width=\linewidth]{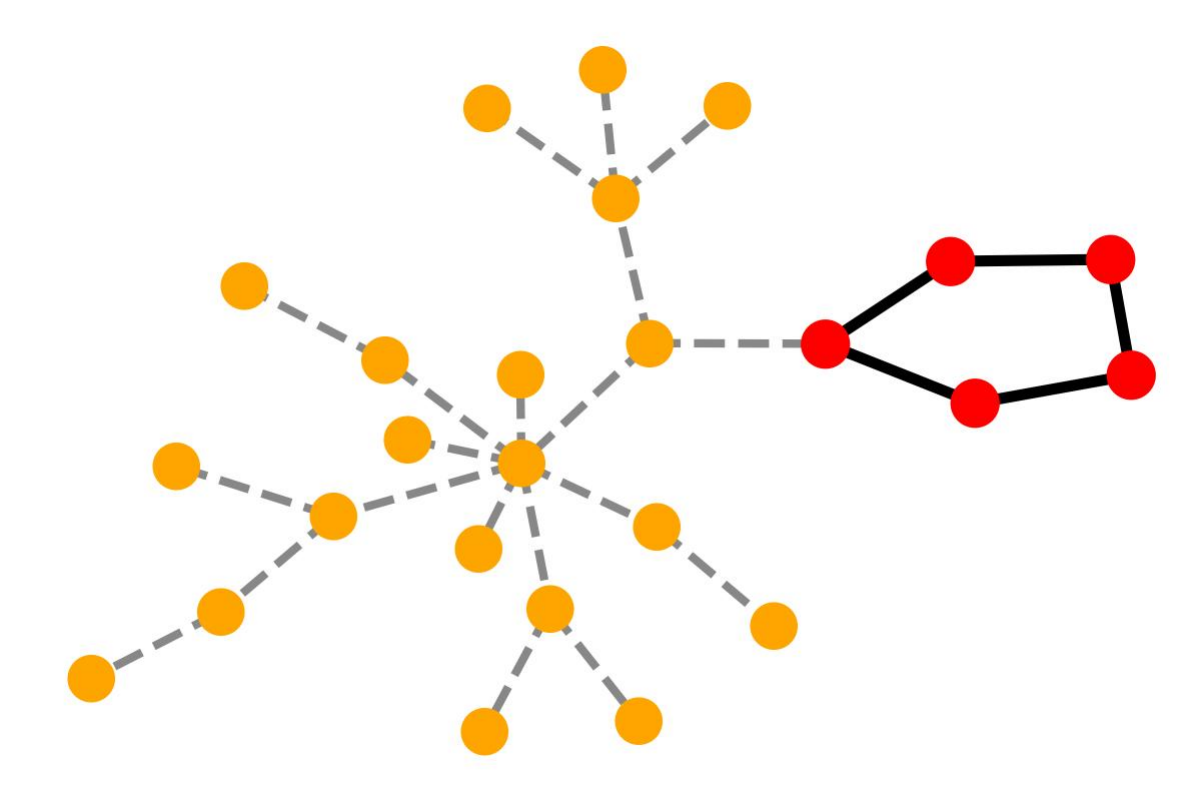}
        \caption{HPME}
    \end{subfigure}
    \begin{subfigure}[b]{0.19\textwidth}
        \includegraphics[width=\linewidth]{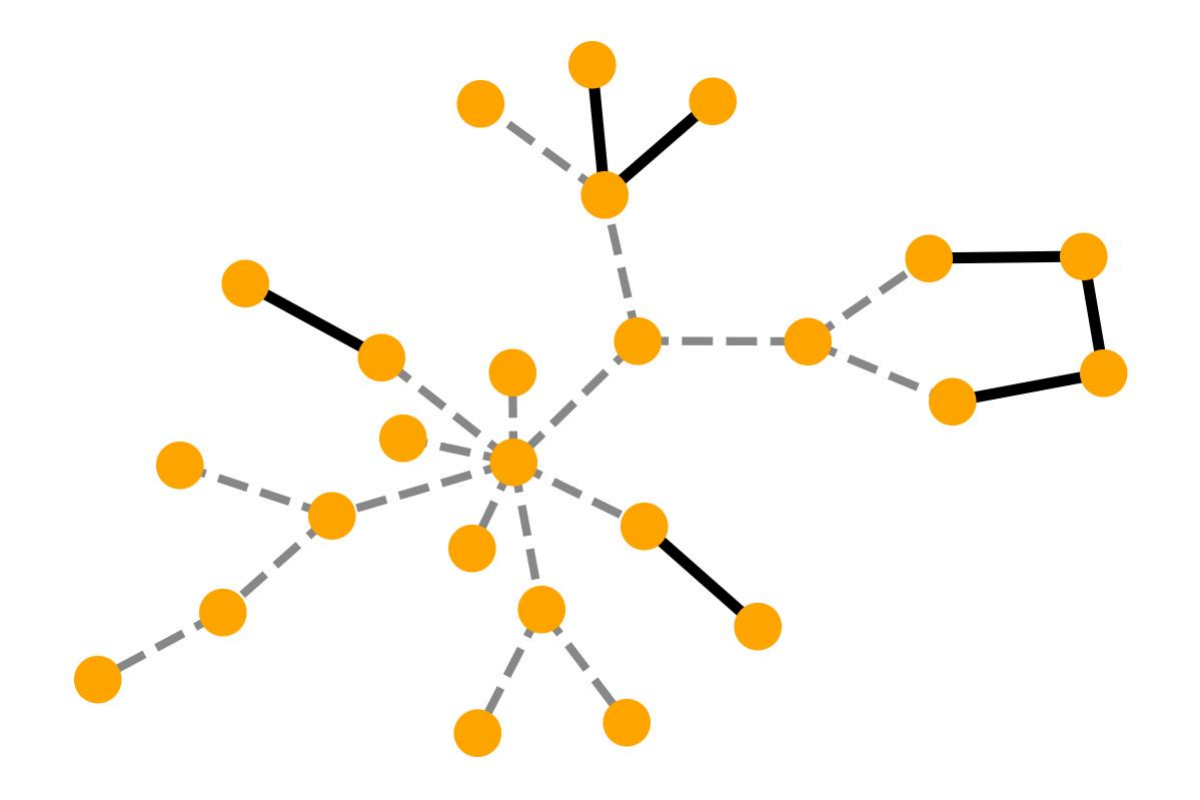}
        \caption{ProxyExplainer}
    \end{subfigure}
    \begin{subfigure}[b]{0.19\textwidth}
        \includegraphics[width=\linewidth]{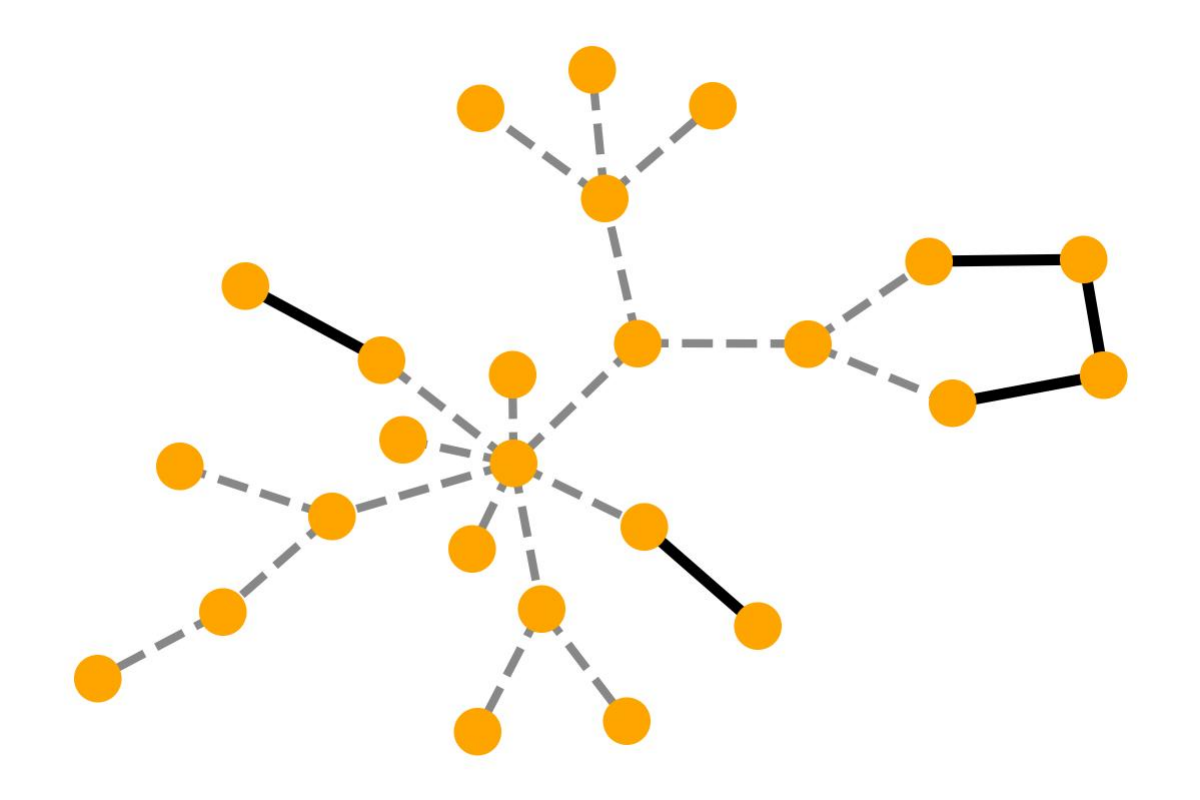}
        \caption{MixupExplainer}
    \end{subfigure}
    \begin{subfigure}[b]{0.19\textwidth}
        \includegraphics[width=\linewidth]{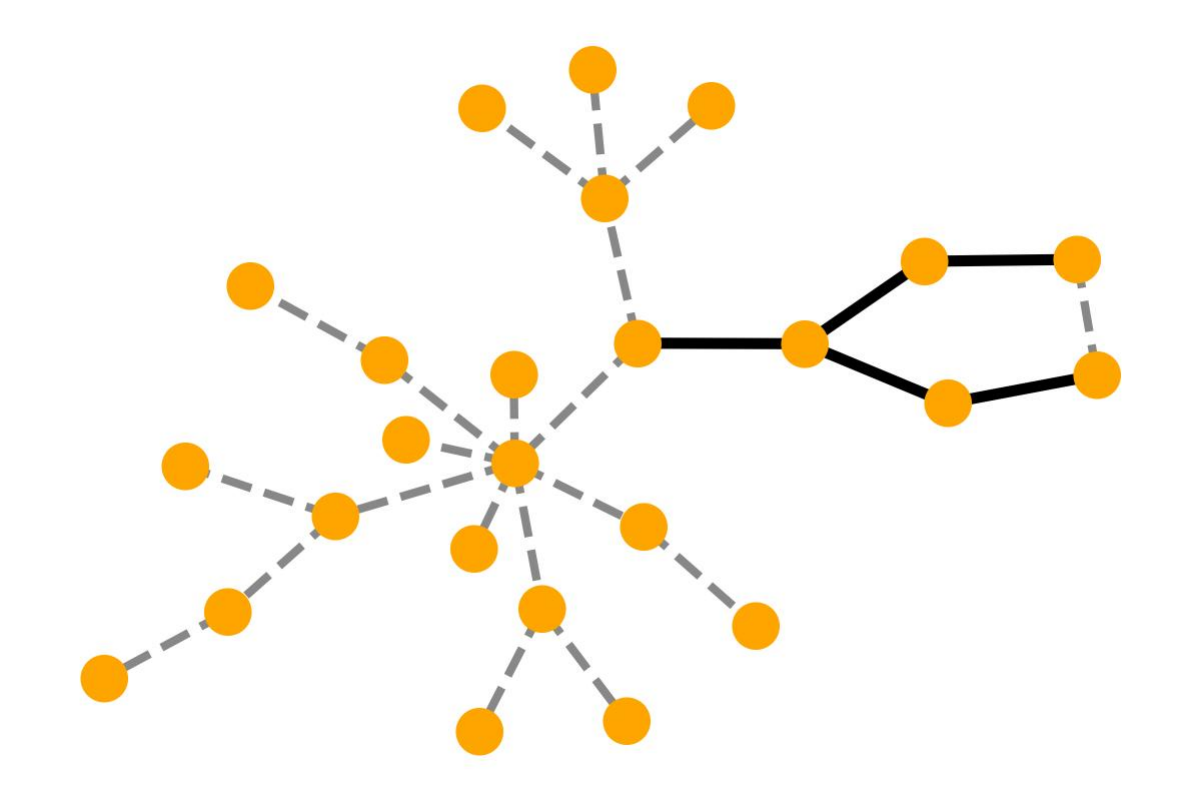}
        \caption{MetaGNN}
    \end{subfigure}
    \caption{Visualization of explanation on BA-2motifs.}
    \label{fig:app:casestudy:BA-2Motifs}
\end{figure*}

\begin{figure*}[h]
    \centering
    \vspace{0.8em}
    \begin{subfigure}[b]{0.19\textwidth}
        \includegraphics[width=\linewidth]{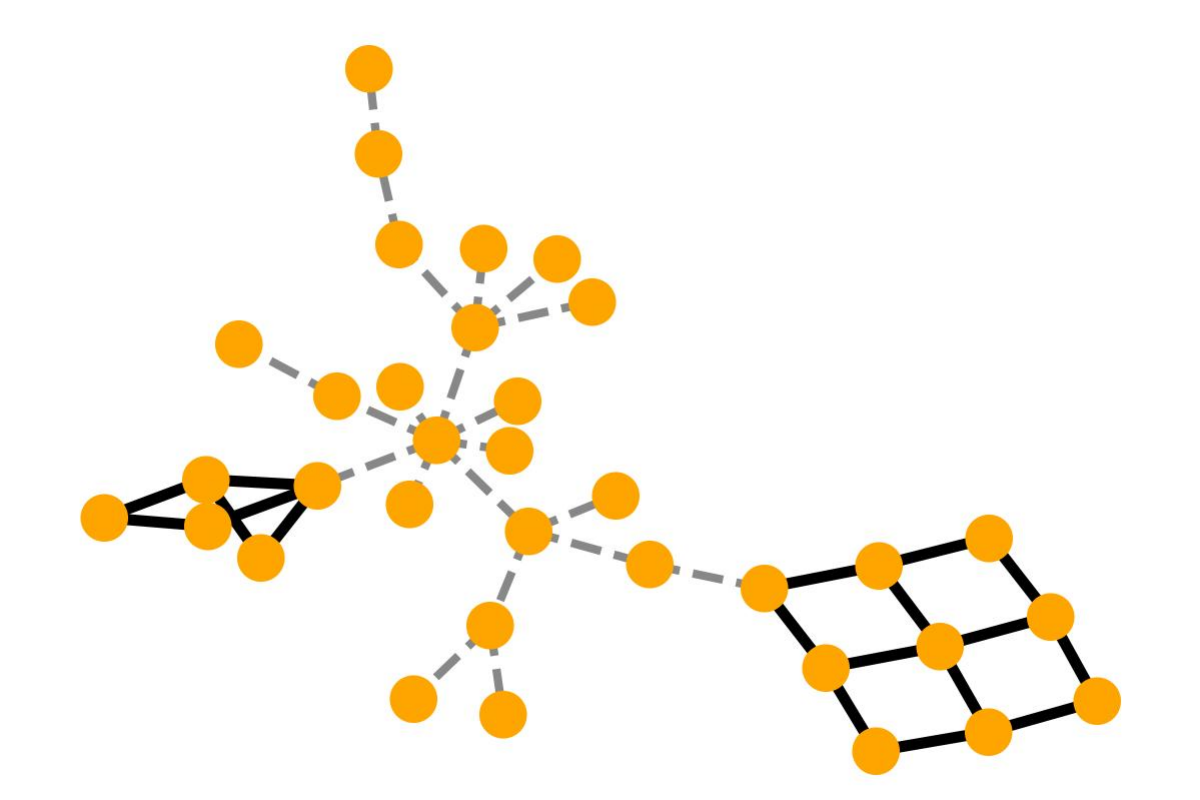}
    \end{subfigure}
    \begin{subfigure}[b]{0.19\textwidth}
        \includegraphics[width=\linewidth]{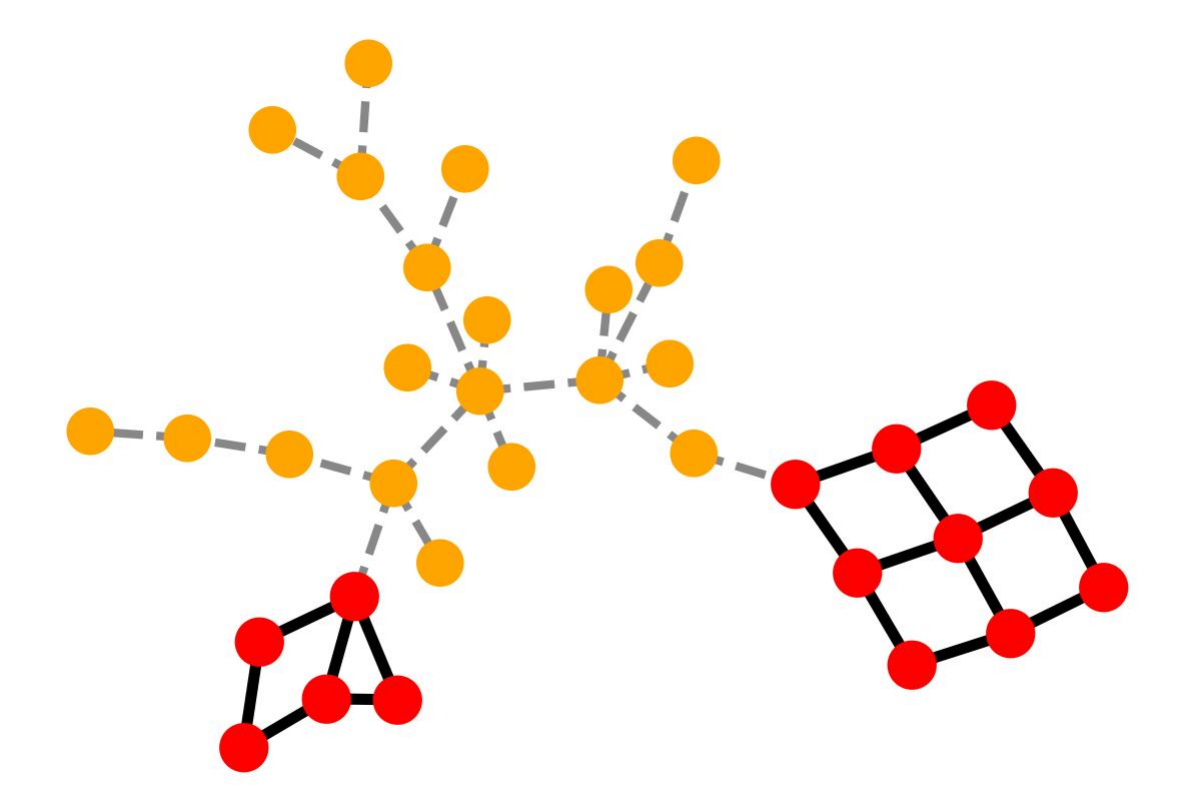}
    \end{subfigure}
    \begin{subfigure}[b]{0.19\textwidth}
        \includegraphics[width=\linewidth]{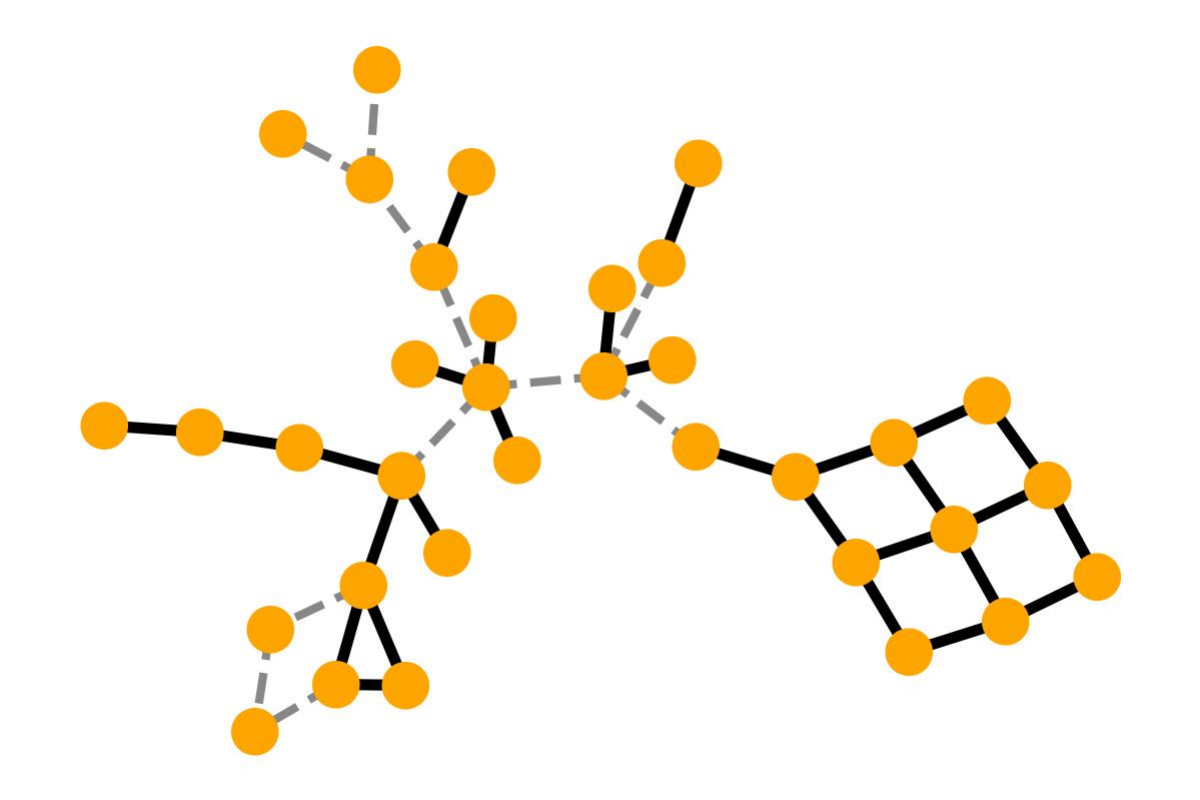}
    \end{subfigure}
    \begin{subfigure}[b]{0.19\textwidth}
        \includegraphics[width=\linewidth]{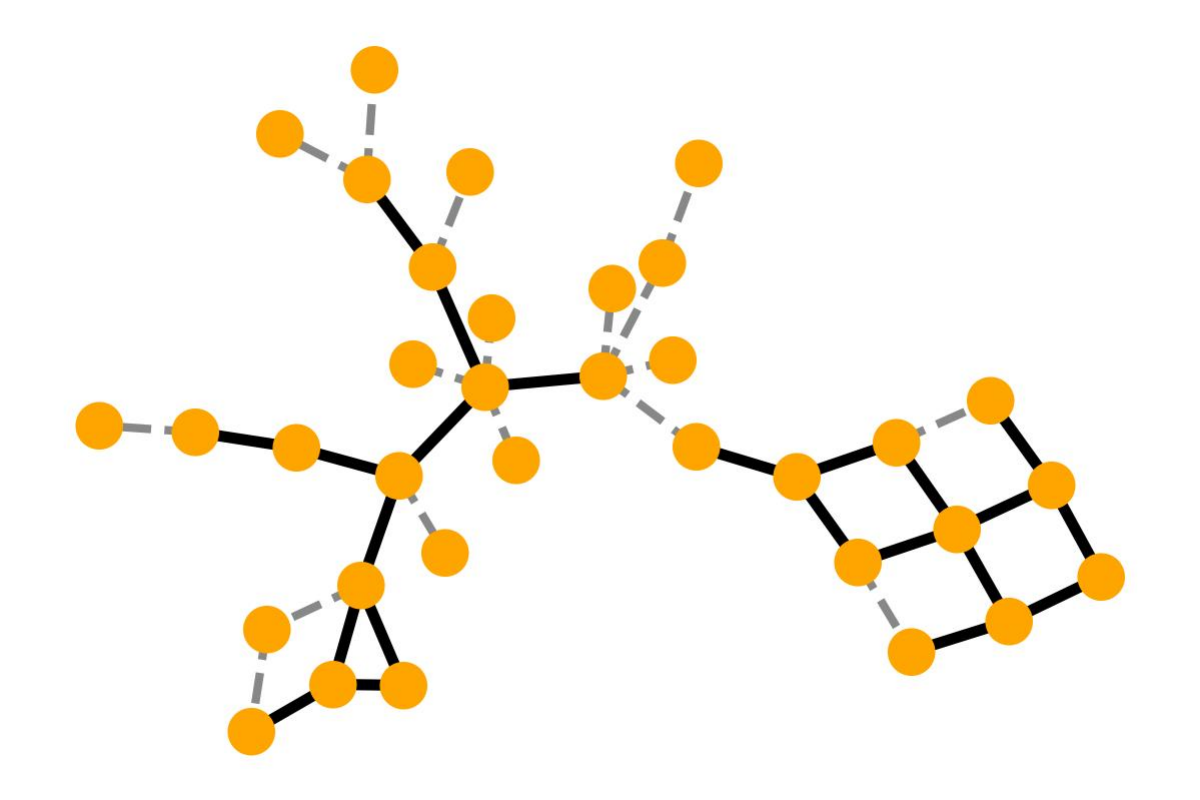}
    \end{subfigure}
    \begin{subfigure}[b]{0.19\textwidth}
        \includegraphics[width=\linewidth]{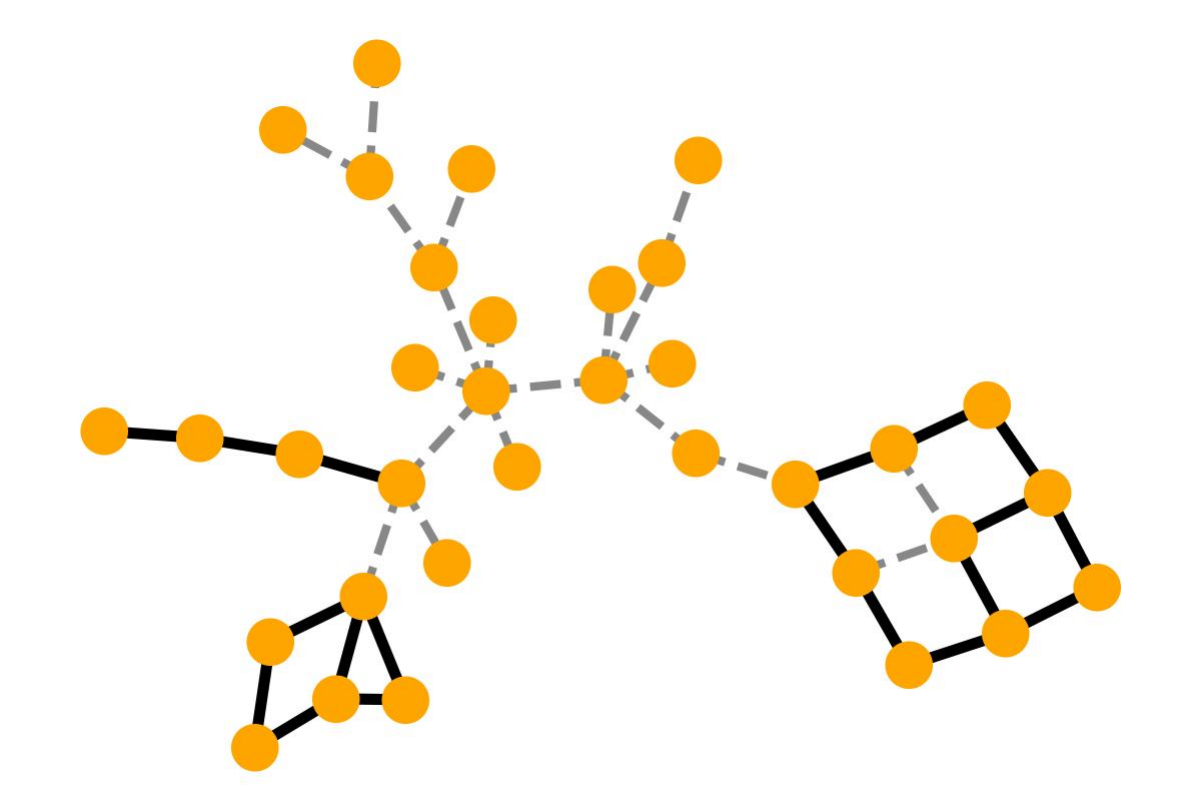}
    \end{subfigure}
    \par\vspace{0.8em}
    \begin{subfigure}[b]{0.19\textwidth}
        \includegraphics[width=\linewidth]{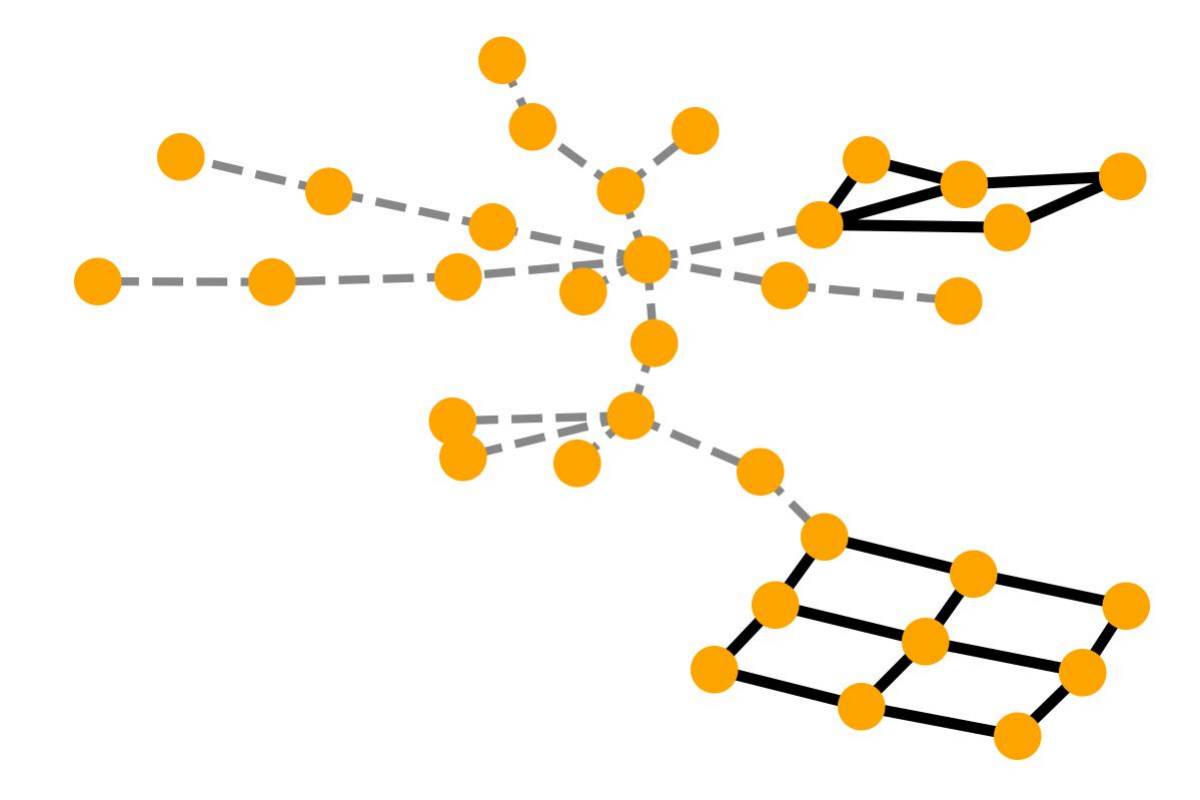}
    \end{subfigure}
    \begin{subfigure}[b]{0.19\textwidth}
        \includegraphics[width=\linewidth]{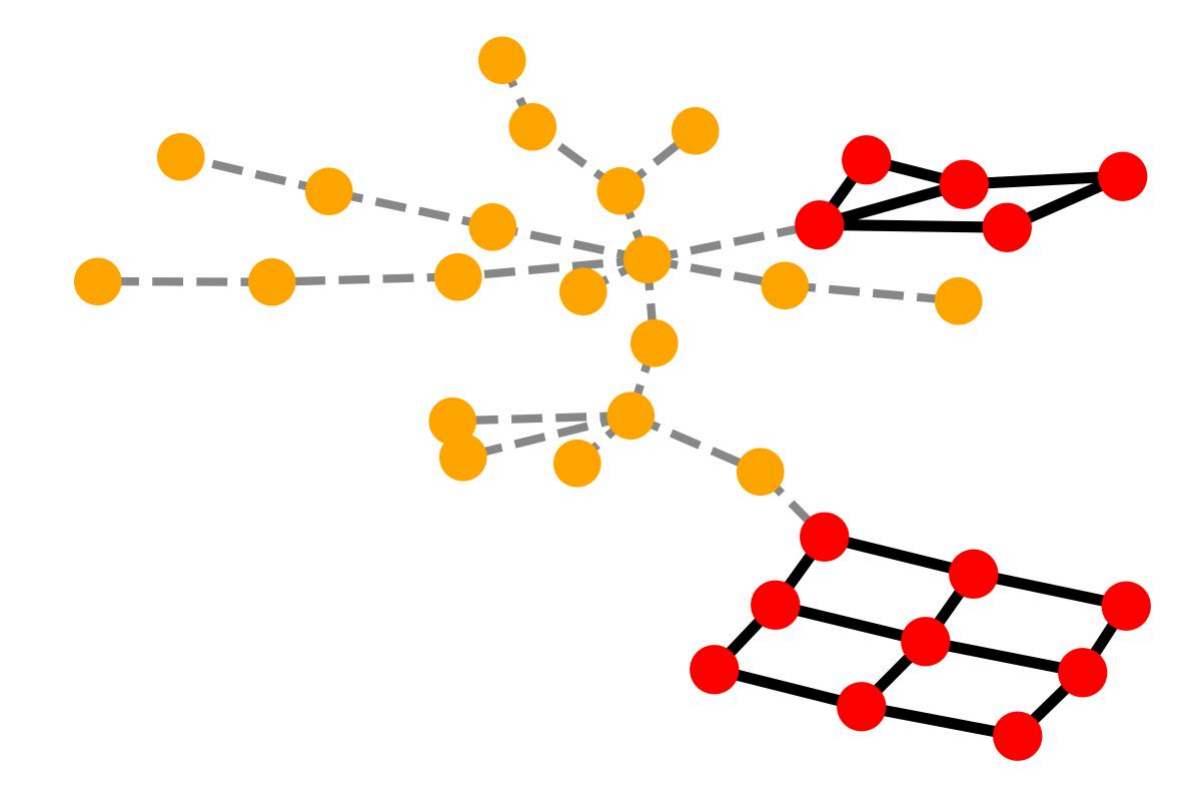}
    \end{subfigure}
    \begin{subfigure}[b]{0.19\textwidth}
        \includegraphics[width=\linewidth]{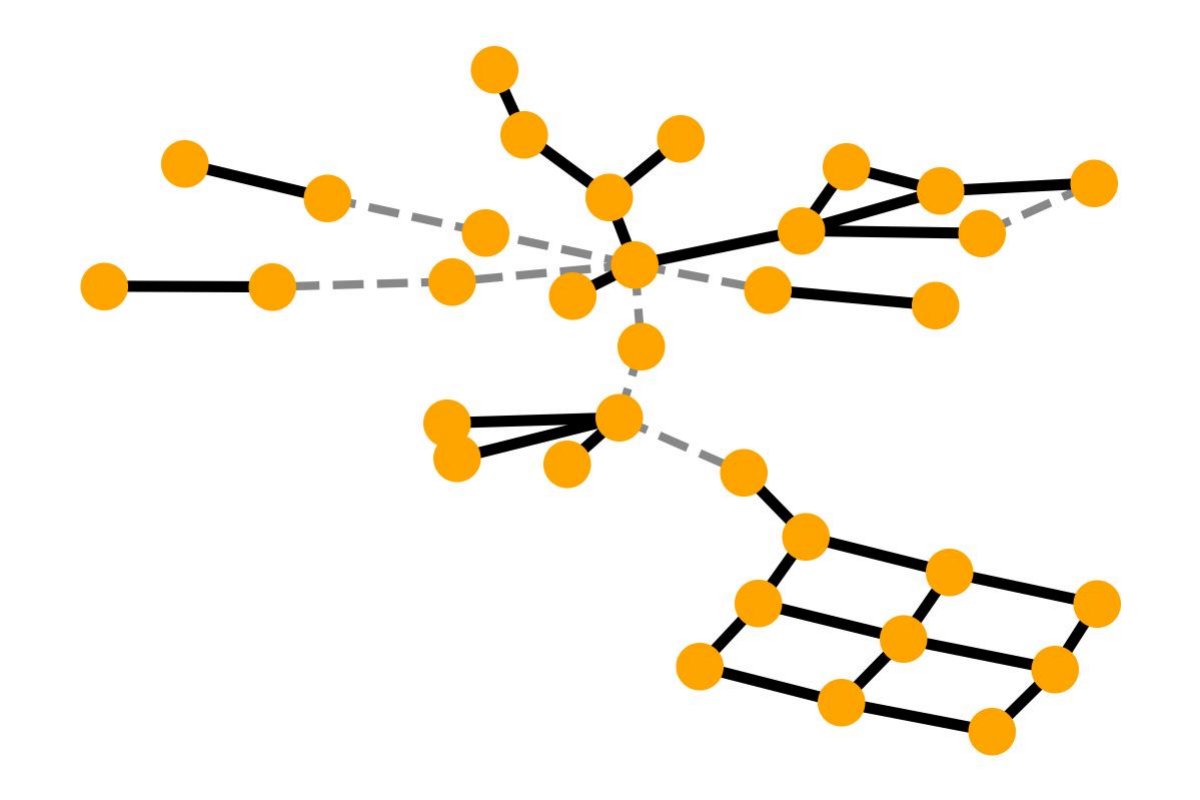}
    \end{subfigure}
    \begin{subfigure}[b]{0.19\textwidth}
        \includegraphics[width=\linewidth]{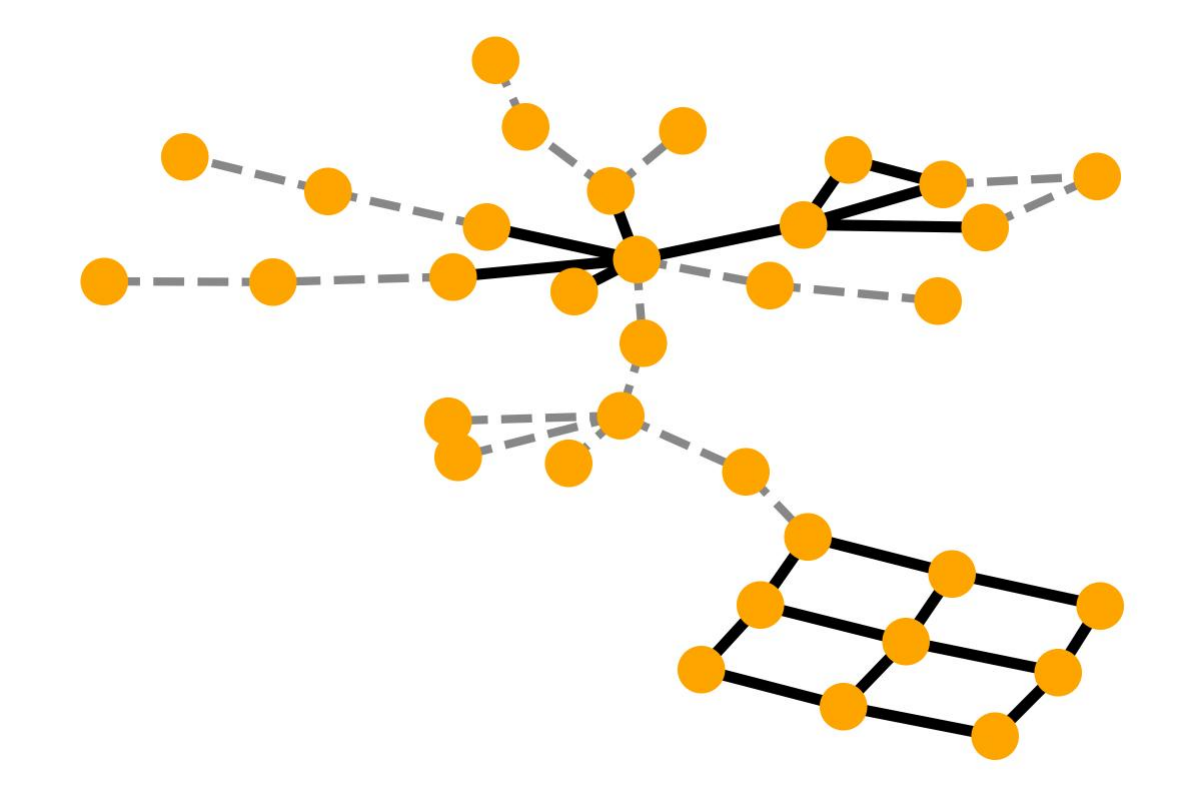}
    \end{subfigure}
    \begin{subfigure}[b]{0.19\textwidth}
        \includegraphics[width=\linewidth]{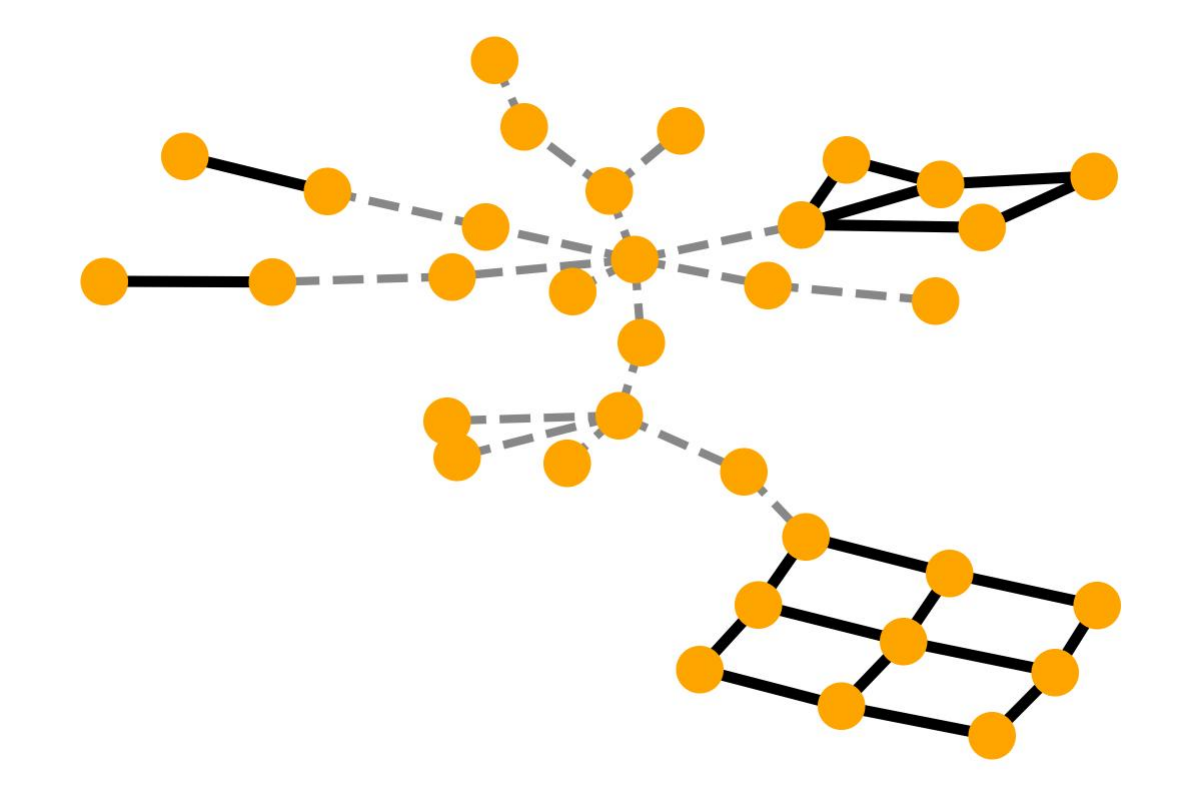}
    \end{subfigure}
    \par\vspace{0.8em}
    \begin{subfigure}[b]{0.19\textwidth}
        \includegraphics[width=\linewidth]{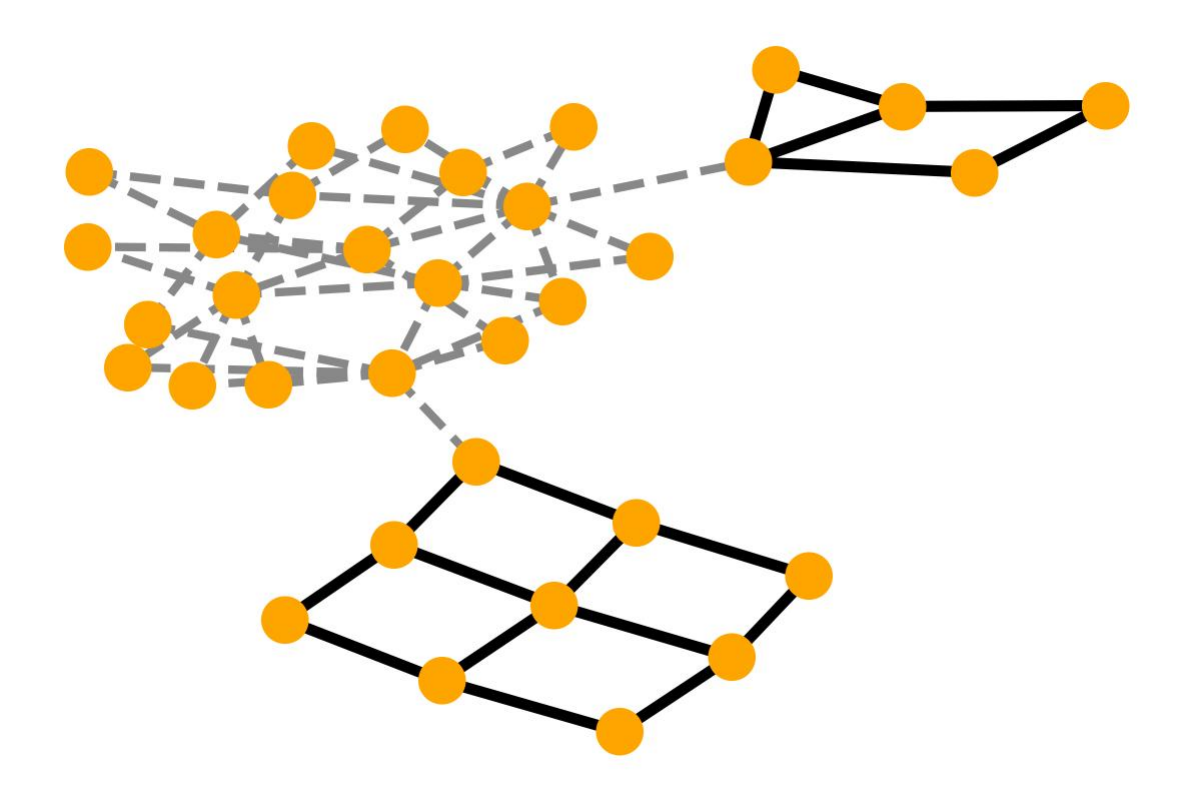}
        \caption{Ground Truth}
    \end{subfigure}
    \begin{subfigure}[b]{0.19\textwidth}
        \includegraphics[width=\linewidth]{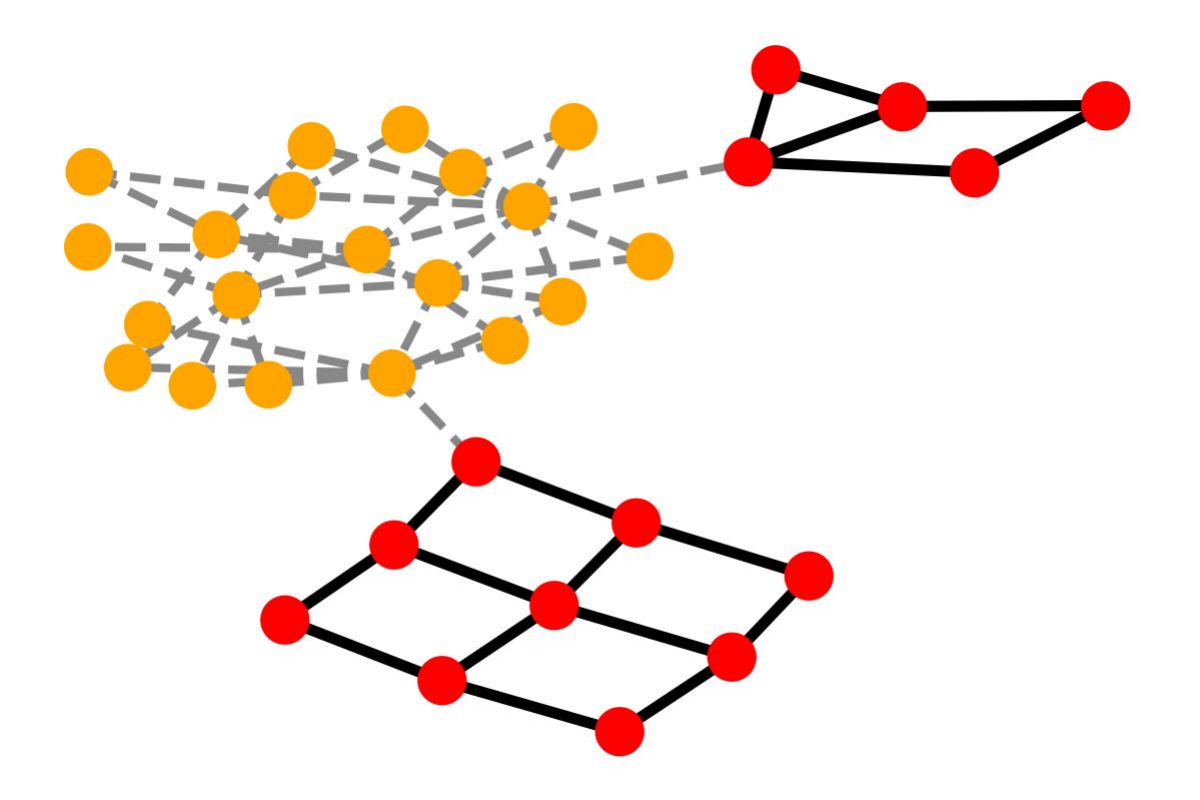}
        \caption{HPME}
    \end{subfigure}
    \begin{subfigure}[b]{0.19\textwidth}
        \includegraphics[width=\linewidth]{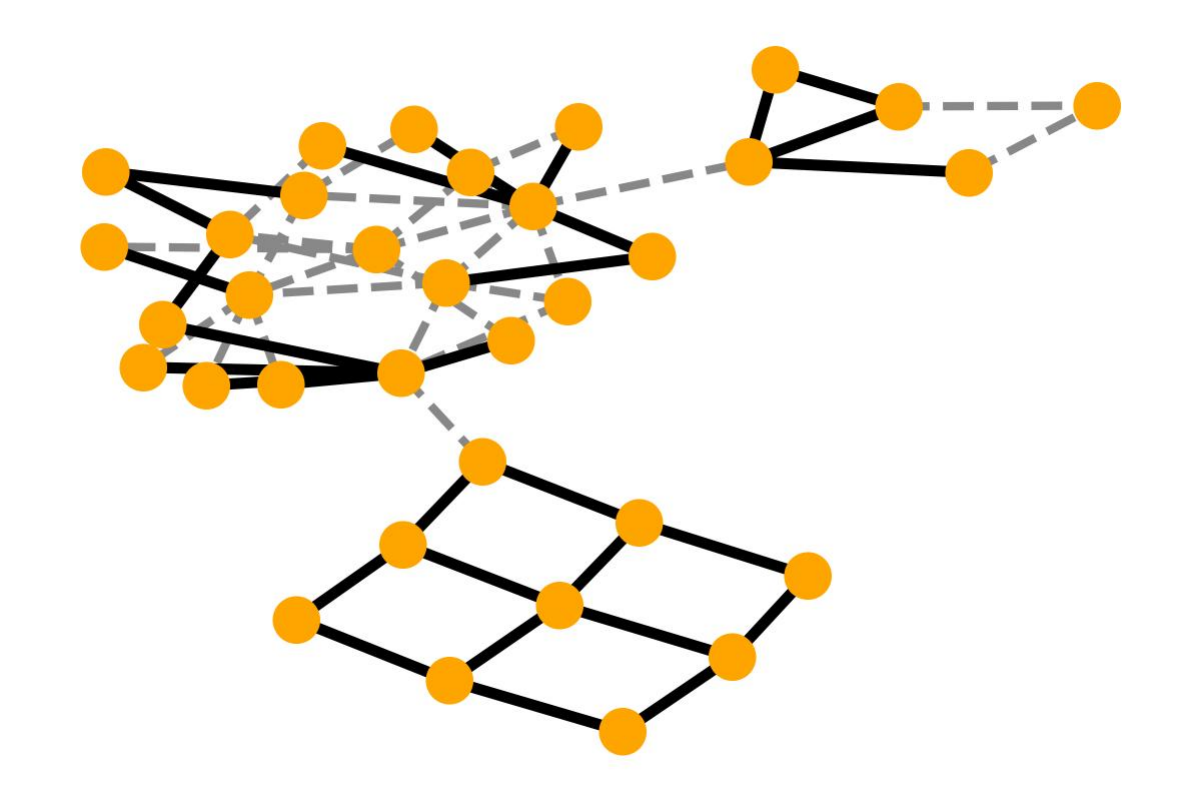}
        \caption{ProxyExplainer}
    \end{subfigure}
    \begin{subfigure}[b]{0.19\textwidth}
        \includegraphics[width=\linewidth]{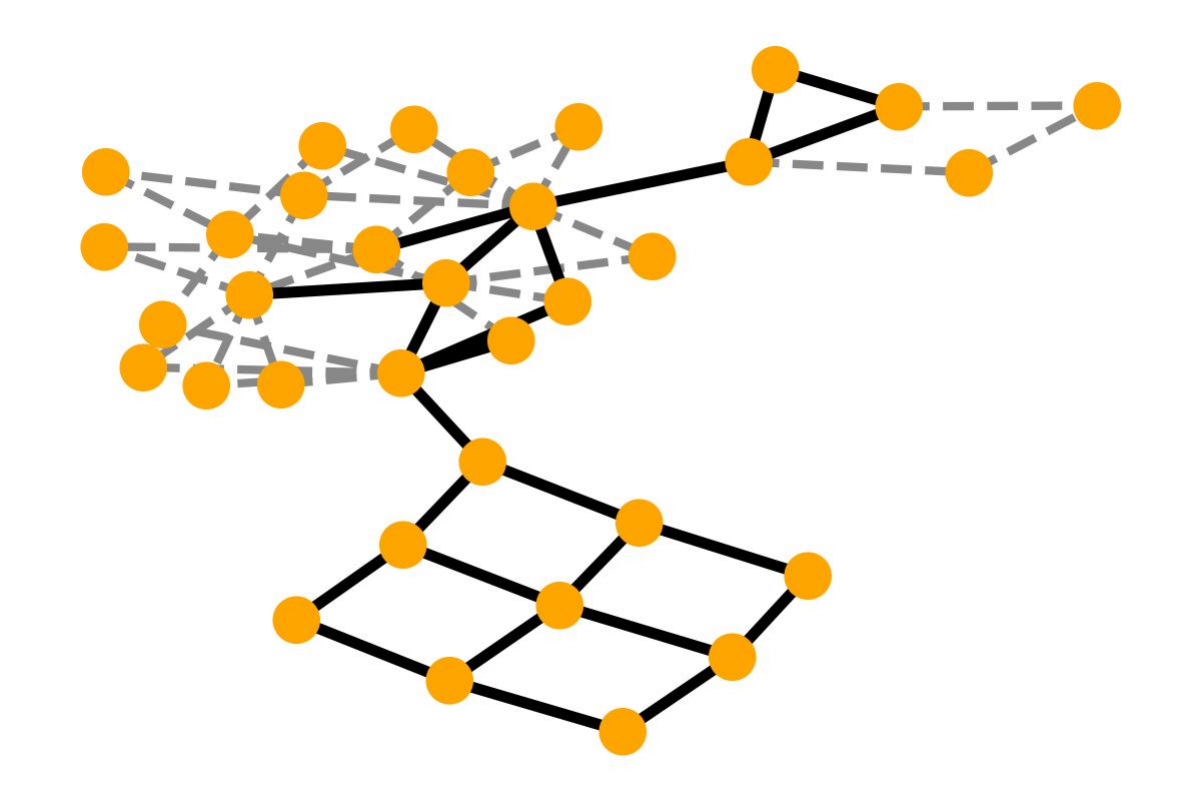}
        \caption{MixupExplainer}
    \end{subfigure}
    \begin{subfigure}[b]{0.19\textwidth}
        \includegraphics[width=\linewidth]{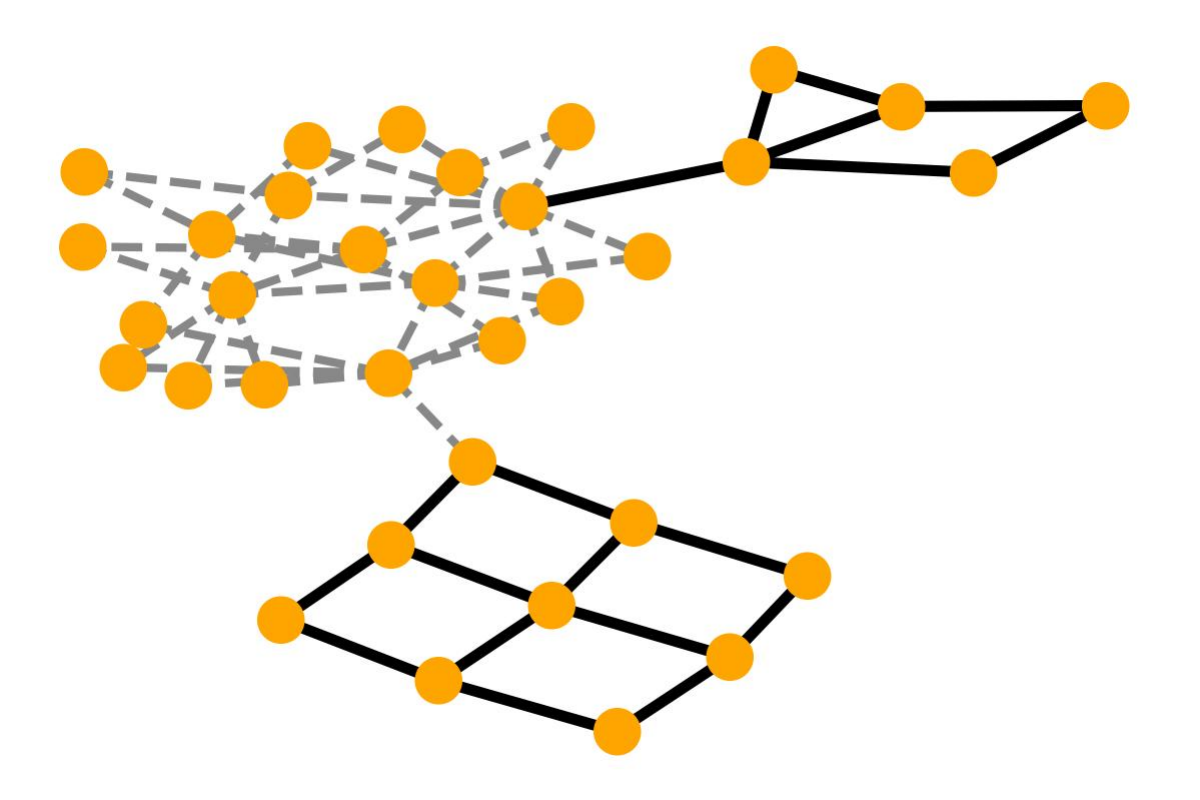}
        \caption{MetaGNN}
    \end{subfigure}
    \caption{Visualization of explanation on BA-HouseAndGrid.}
    \label{fig:app:casestudy:BA-HouseAndGrid}
\end{figure*}

\begin{figure*}[h]
    \centering
    \begin{subfigure}[b]{0.19\textwidth}
        \includegraphics[width=\linewidth]{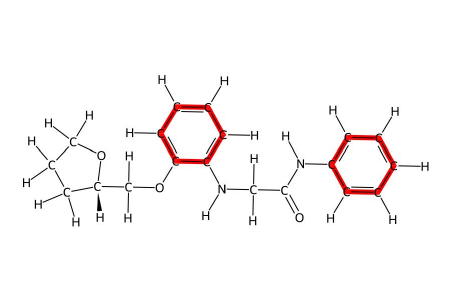}
    \end{subfigure}
    \begin{subfigure}[b]{0.19\textwidth}
        \includegraphics[width=\linewidth]{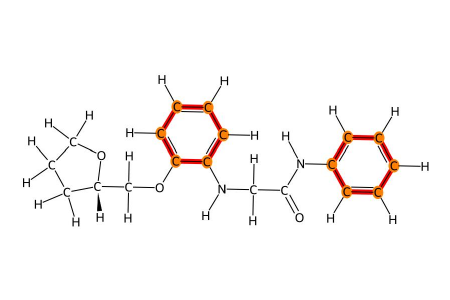}
    \end{subfigure}
    \begin{subfigure}[b]{0.19\textwidth}
        \includegraphics[width=\linewidth]{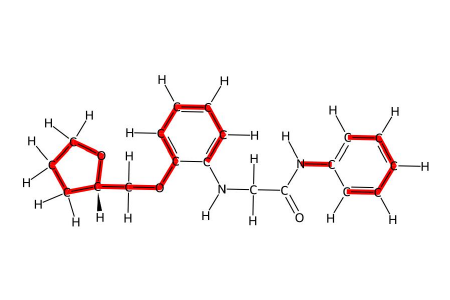}
    \end{subfigure}
    \begin{subfigure}[b]{0.19\textwidth}
        \includegraphics[width=\linewidth]{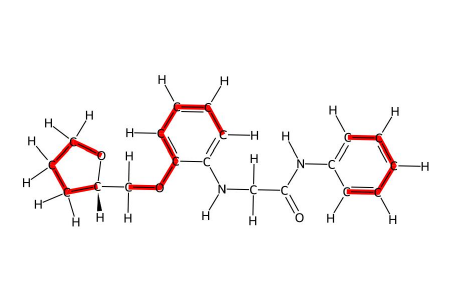}
    \end{subfigure}
    \begin{subfigure}[b]{0.19\textwidth}
        \includegraphics[width=\linewidth]{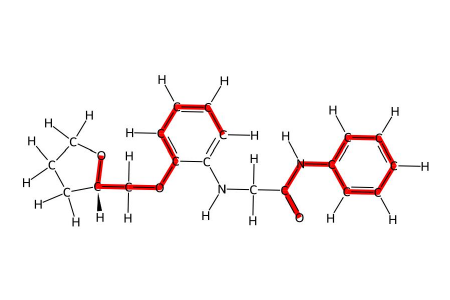}
    \end{subfigure}
    \begin{subfigure}[b]{0.19\textwidth}
        \includegraphics[width=\linewidth]{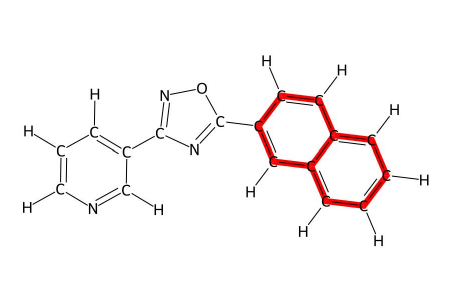}
    \end{subfigure}
    \begin{subfigure}[b]{0.19\textwidth}
        \includegraphics[width=\linewidth]{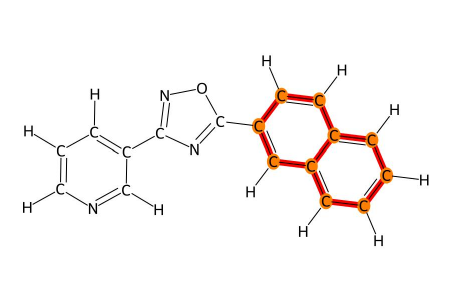}
    \end{subfigure}
    \begin{subfigure}[b]{0.19\textwidth}
        \includegraphics[width=\linewidth]{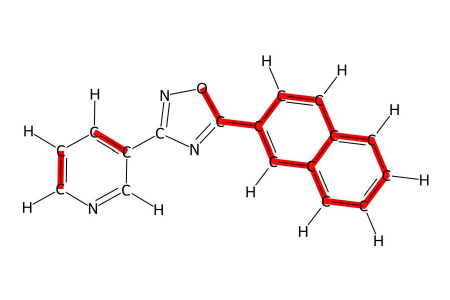}
    \end{subfigure}
    \begin{subfigure}[b]{0.19\textwidth}
        \includegraphics[width=\linewidth]{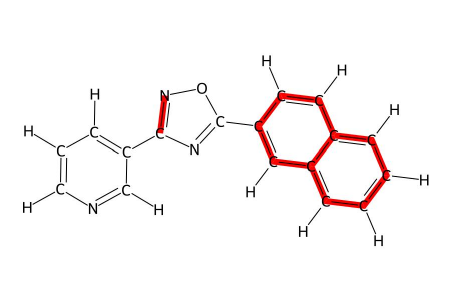}
    \end{subfigure}
    \begin{subfigure}[b]{0.19\textwidth}
        \includegraphics[width=\linewidth]{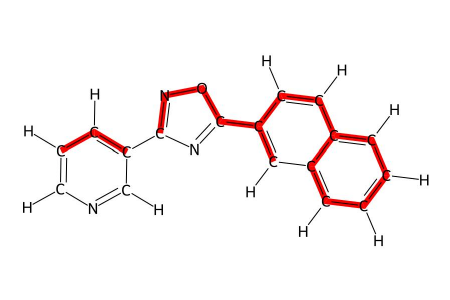}
    \end{subfigure}
    \begin{subfigure}[b]{0.19\textwidth}
        \includegraphics[width=\linewidth]{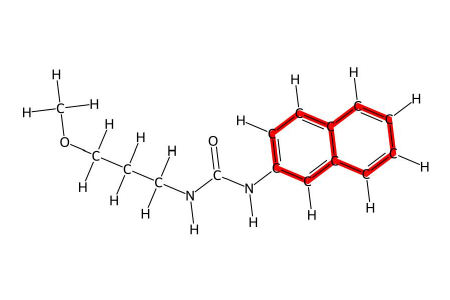}
        \caption{Ground Truth}
    \end{subfigure}
    \begin{subfigure}[b]{0.19\textwidth}
        \includegraphics[width=\linewidth]{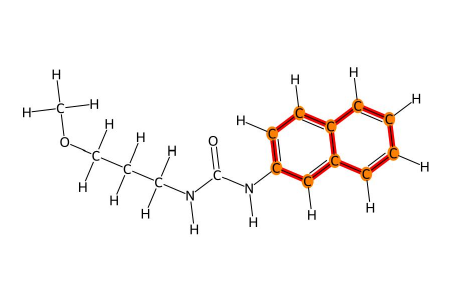}
        \caption{HPME}
    \end{subfigure}
    \begin{subfigure}[b]{0.19\textwidth}
        \includegraphics[width=\linewidth]{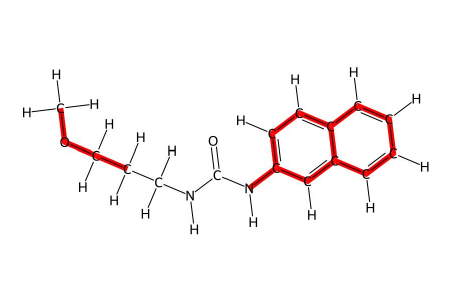}
        \caption{ProxyExplainer}
    \end{subfigure}
    \begin{subfigure}[b]{0.19\textwidth}
        \includegraphics[width=\linewidth]{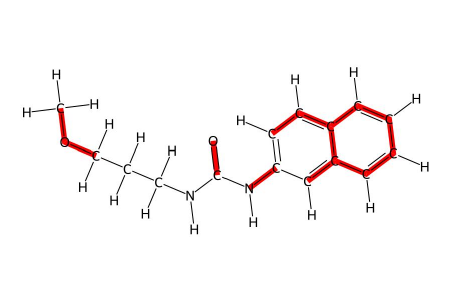}
        \caption{MixupExplainer}
    \end{subfigure}
    \begin{subfigure}[b]{0.19\textwidth}
        \includegraphics[width=\linewidth]{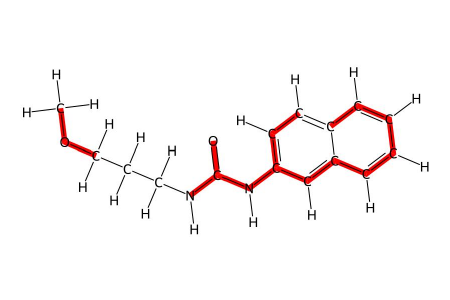}
        \caption{MetaGNN}
    \end{subfigure}
    \caption{Visualization of explanation on Benzene.}
    \label{fig:app:casestudy:Benzene}
\end{figure*}

\begin{figure*}[h]
    \centering
    \vspace{0.8em}
    \begin{subfigure}[b]{0.19\textwidth}
        \includegraphics[width=\linewidth]{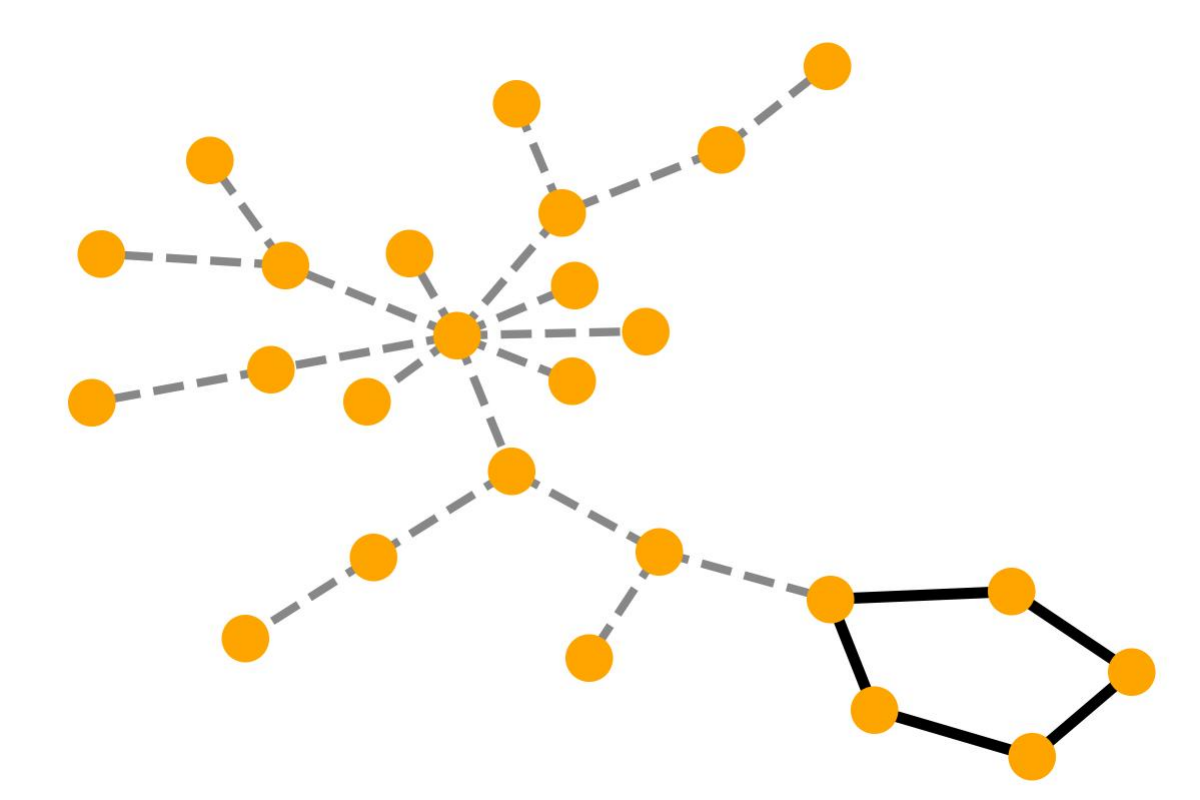}
    \end{subfigure}
    \begin{subfigure}[b]{0.19\textwidth}
        \includegraphics[width=\linewidth]{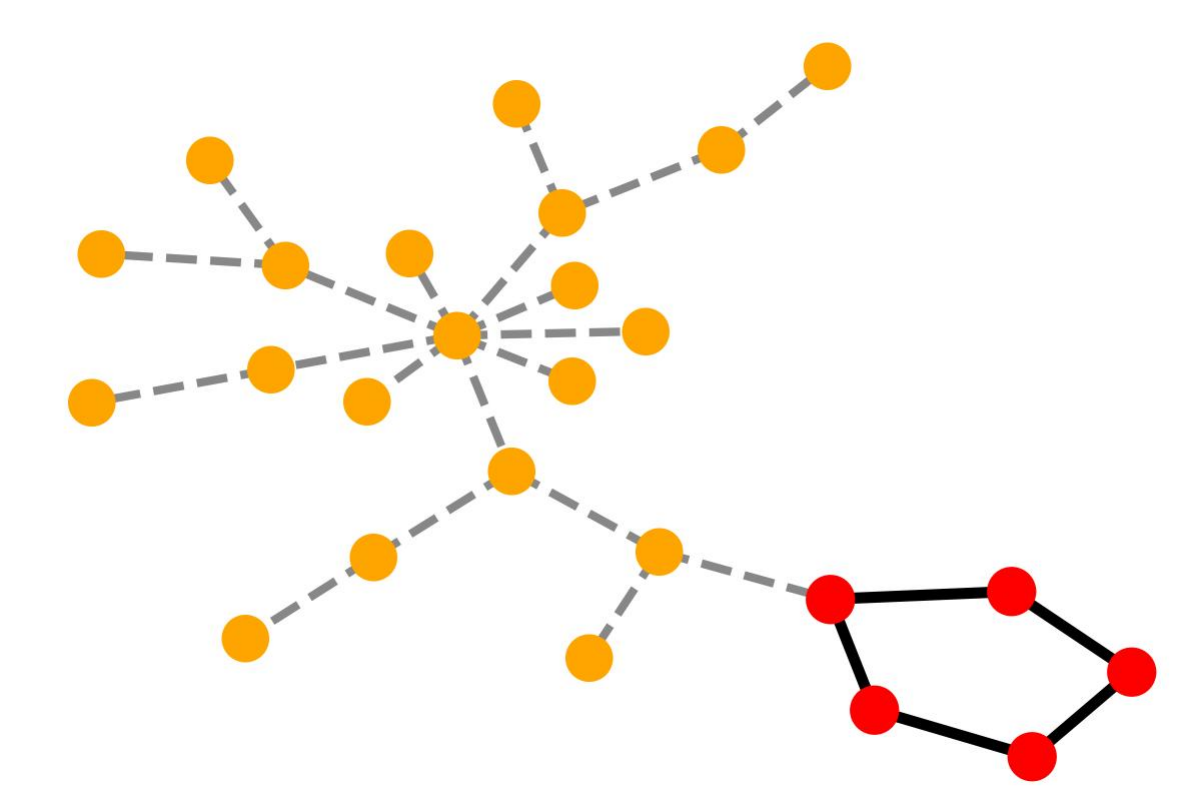}
    \end{subfigure}
    \begin{subfigure}[b]{0.19\textwidth}
        \includegraphics[width=\linewidth]{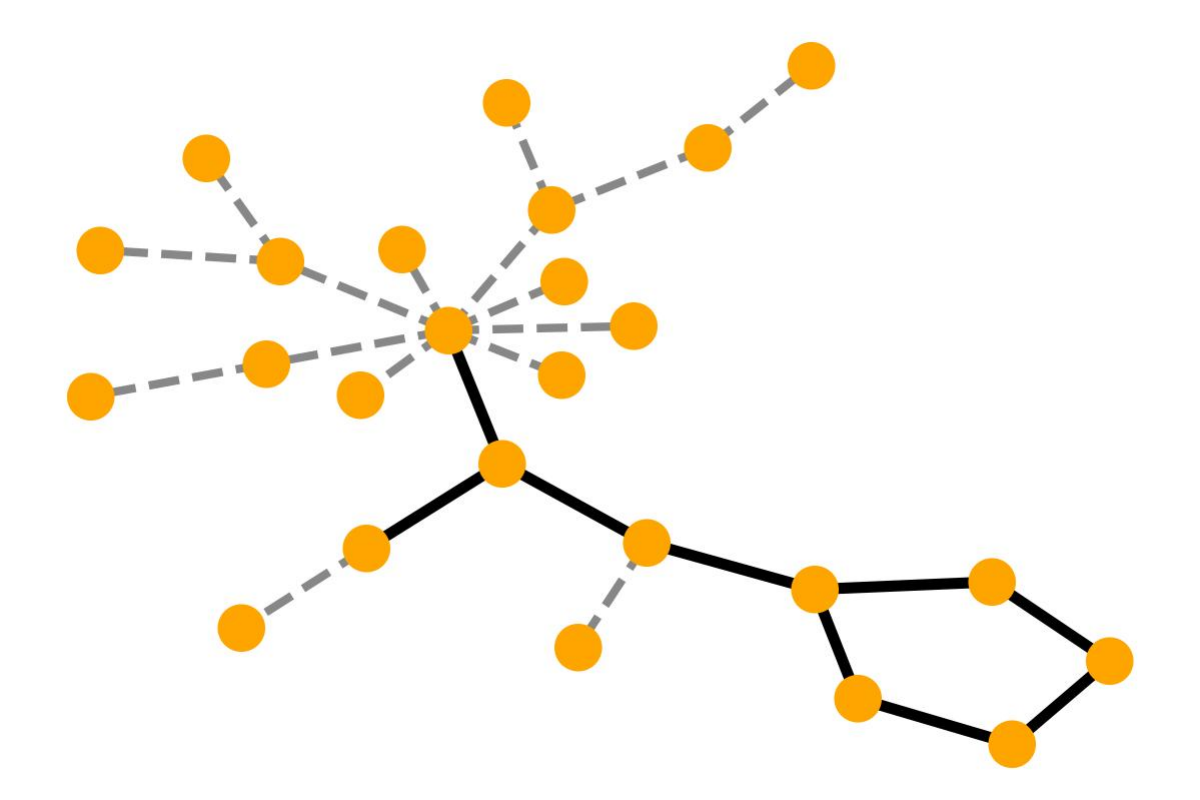}
    \end{subfigure}
    \begin{subfigure}[b]{0.19\textwidth}
        \includegraphics[width=\linewidth]{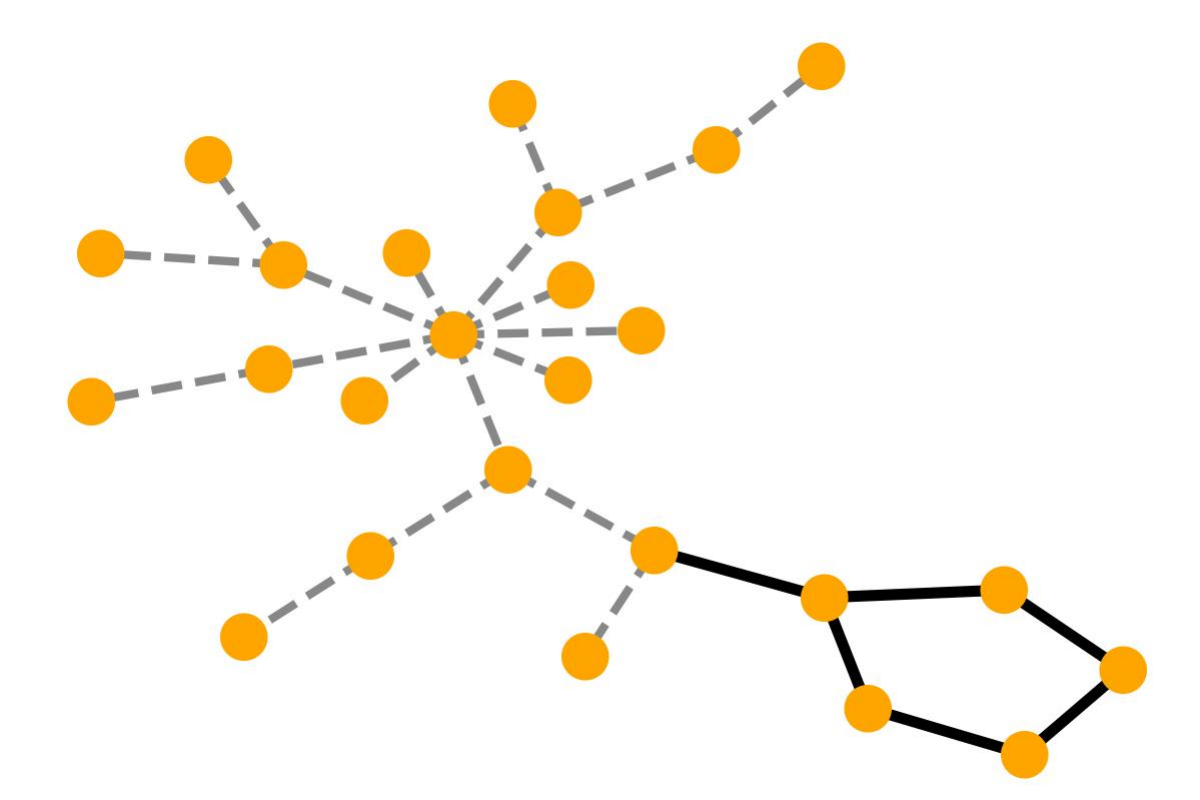}
    \end{subfigure}
    \begin{subfigure}[b]{0.19\textwidth}
        \includegraphics[width=\linewidth]{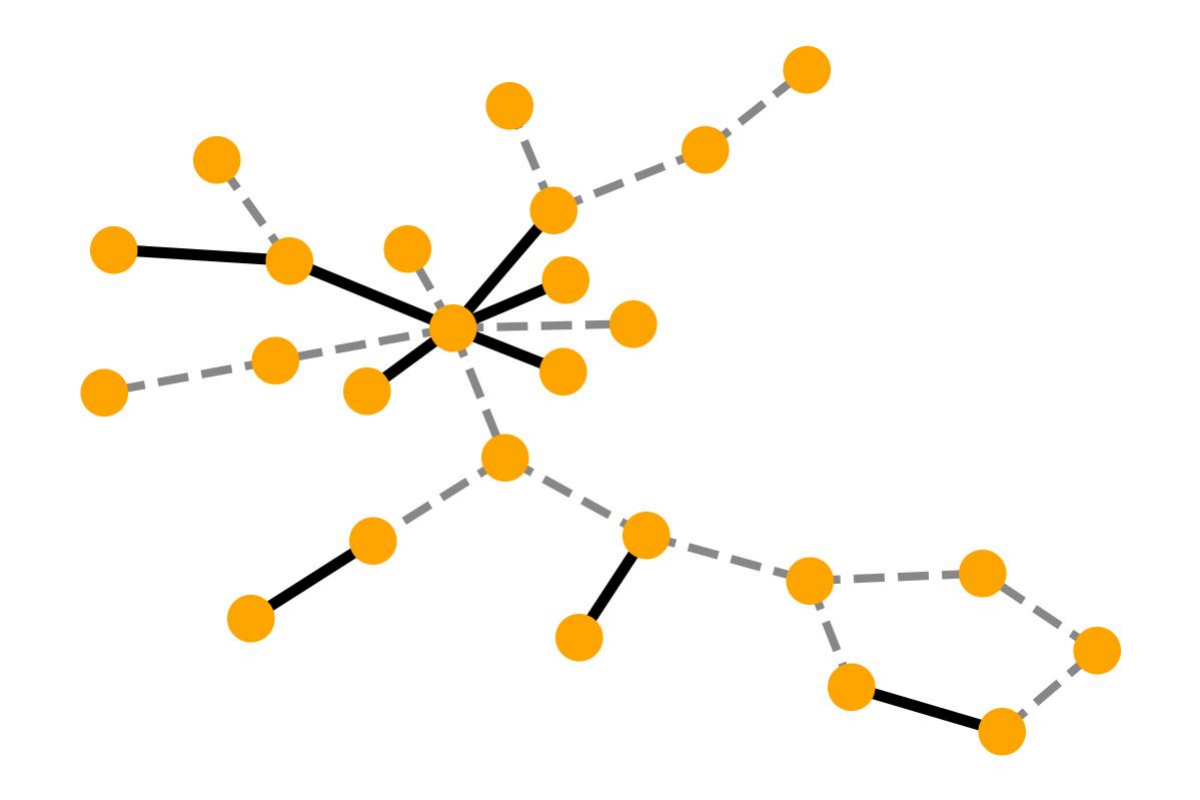}
    \end{subfigure}
    \par\vspace{0.8em}
    \begin{subfigure}[b]{0.19\textwidth}
        \includegraphics[width=\linewidth]{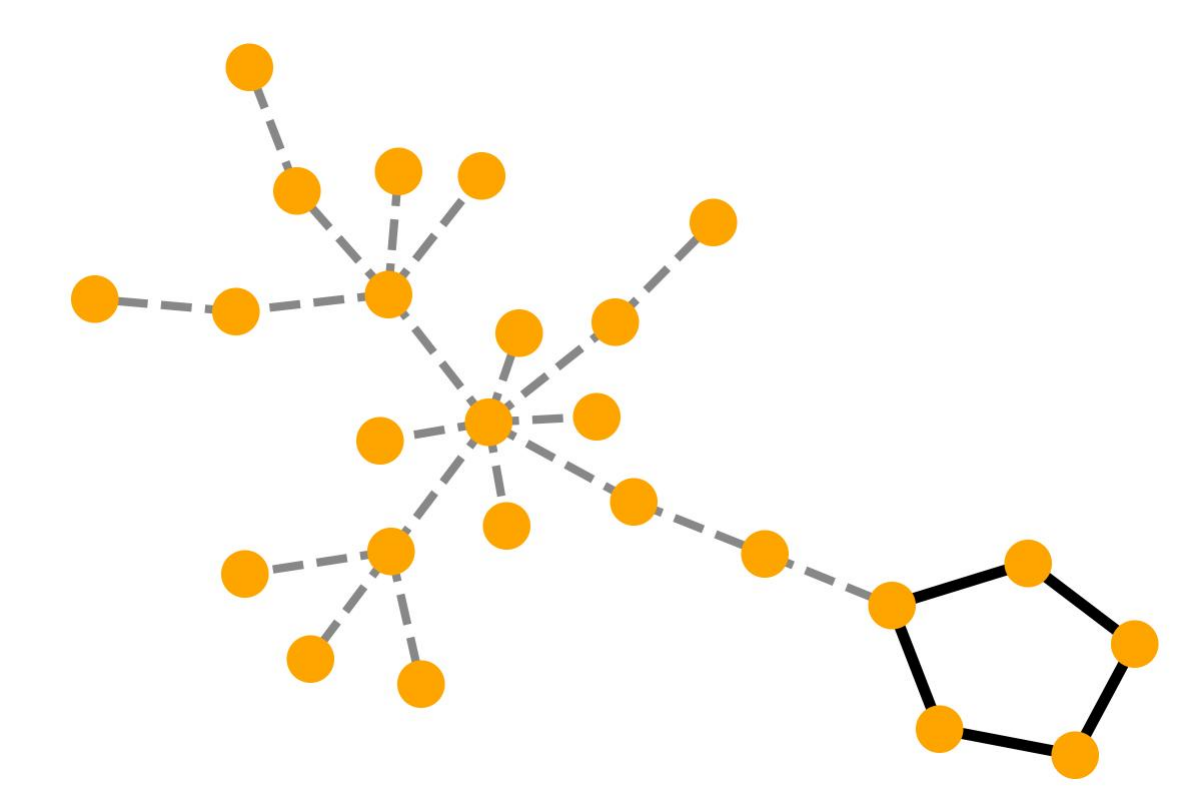}
    \end{subfigure}
    \begin{subfigure}[b]{0.19\textwidth}
        \includegraphics[width=\linewidth]{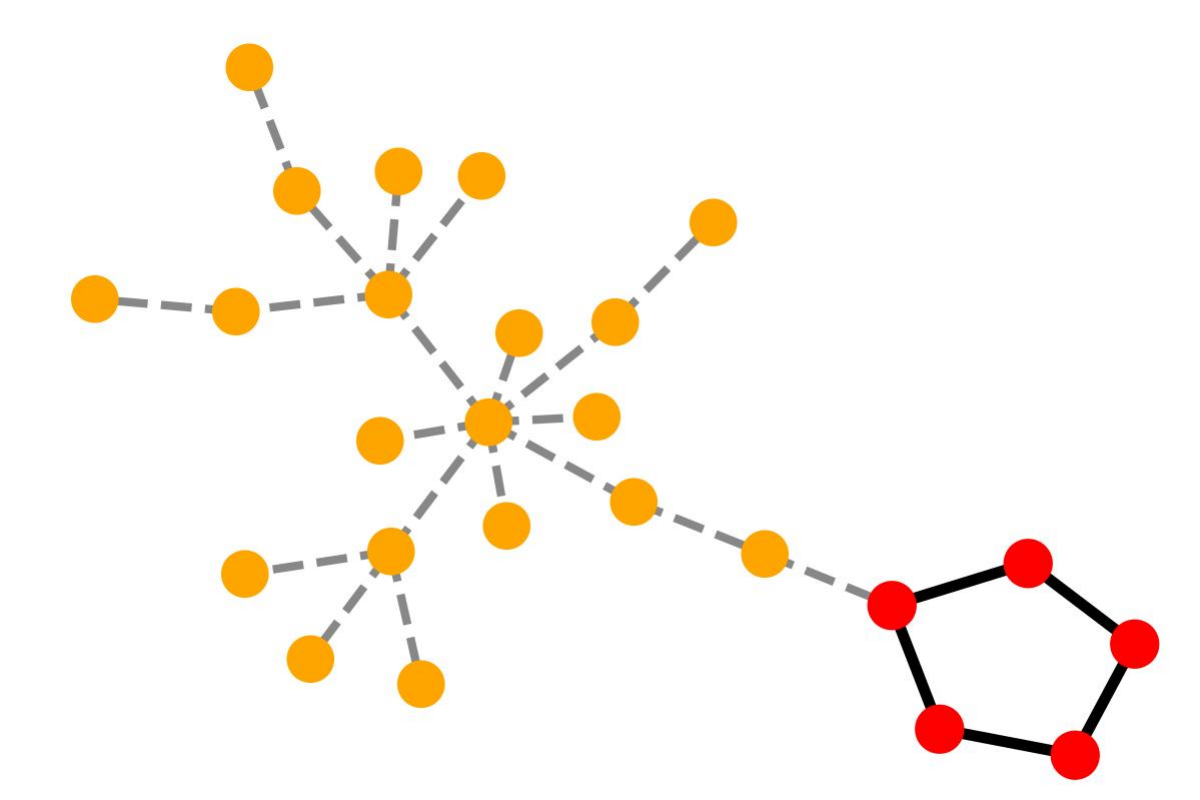}
    \end{subfigure}
    \begin{subfigure}[b]{0.19\textwidth}
        \includegraphics[width=\linewidth]{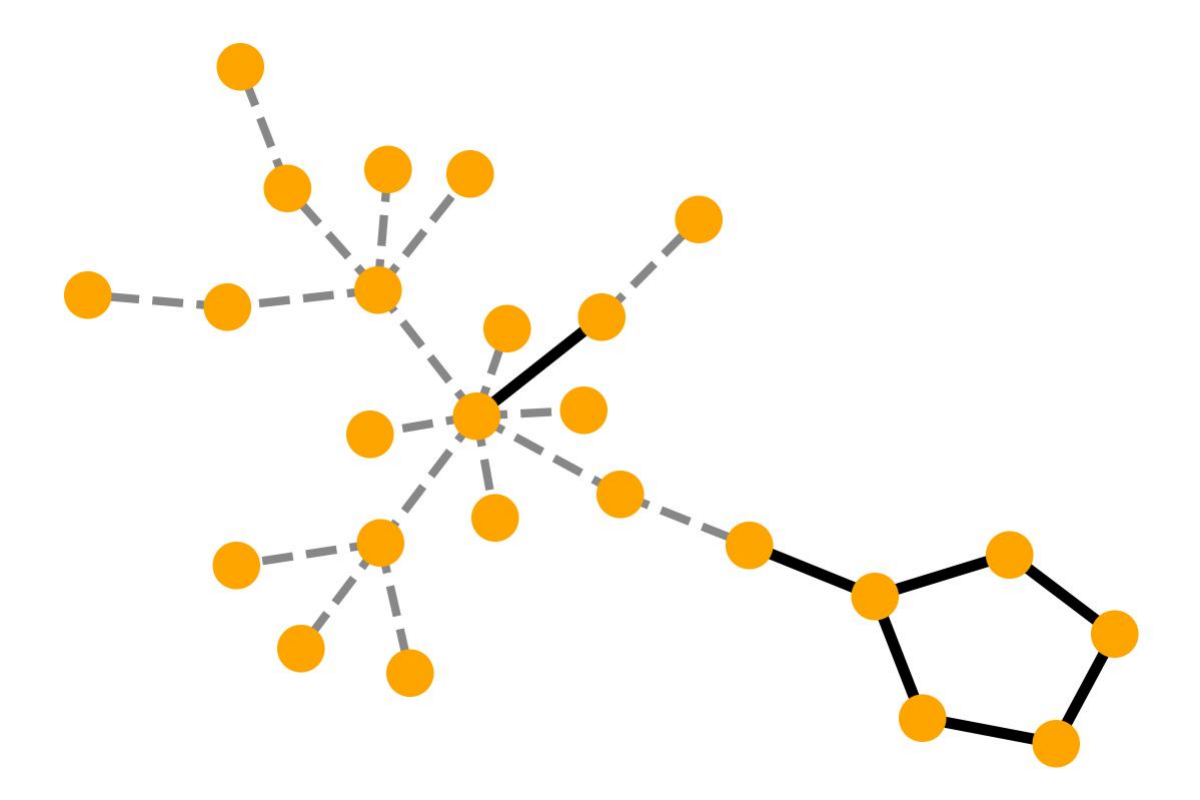}
    \end{subfigure}
    \begin{subfigure}[b]{0.19\textwidth}
        \includegraphics[width=\linewidth]{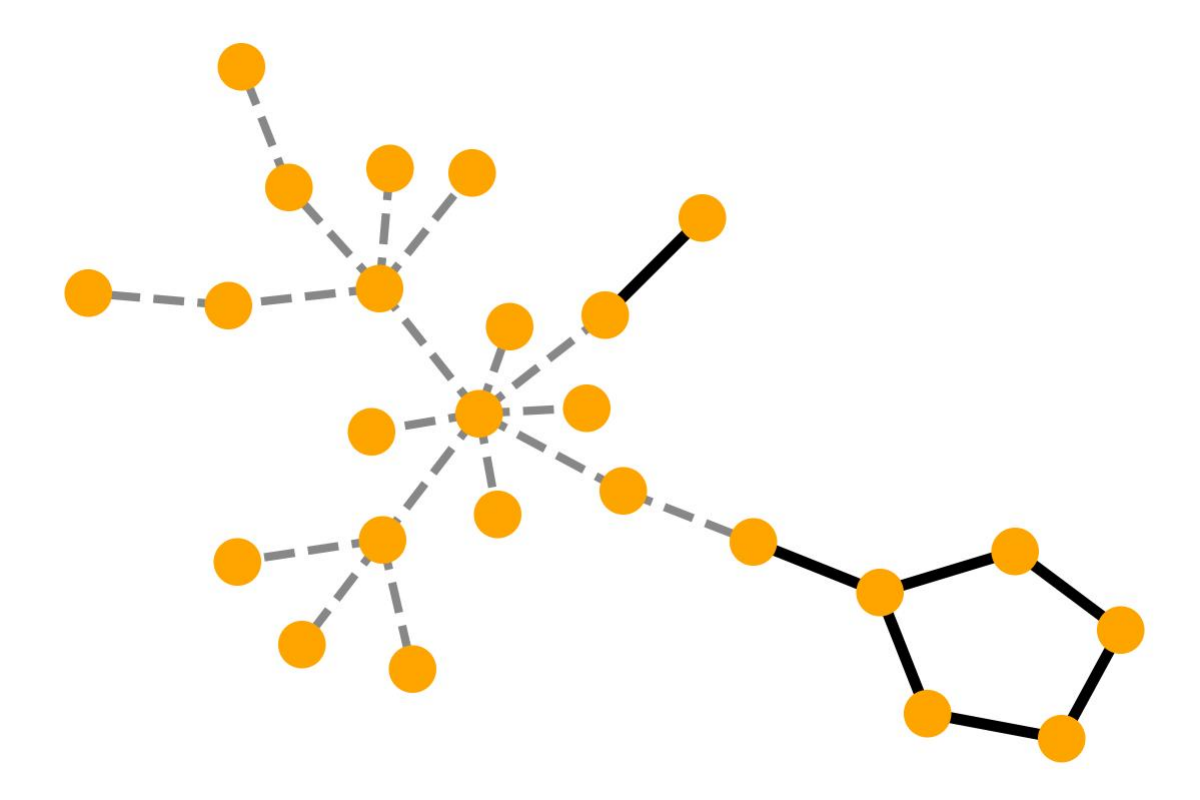}
    \end{subfigure}
    \begin{subfigure}[b]{0.19\textwidth}
        \includegraphics[width=\linewidth]{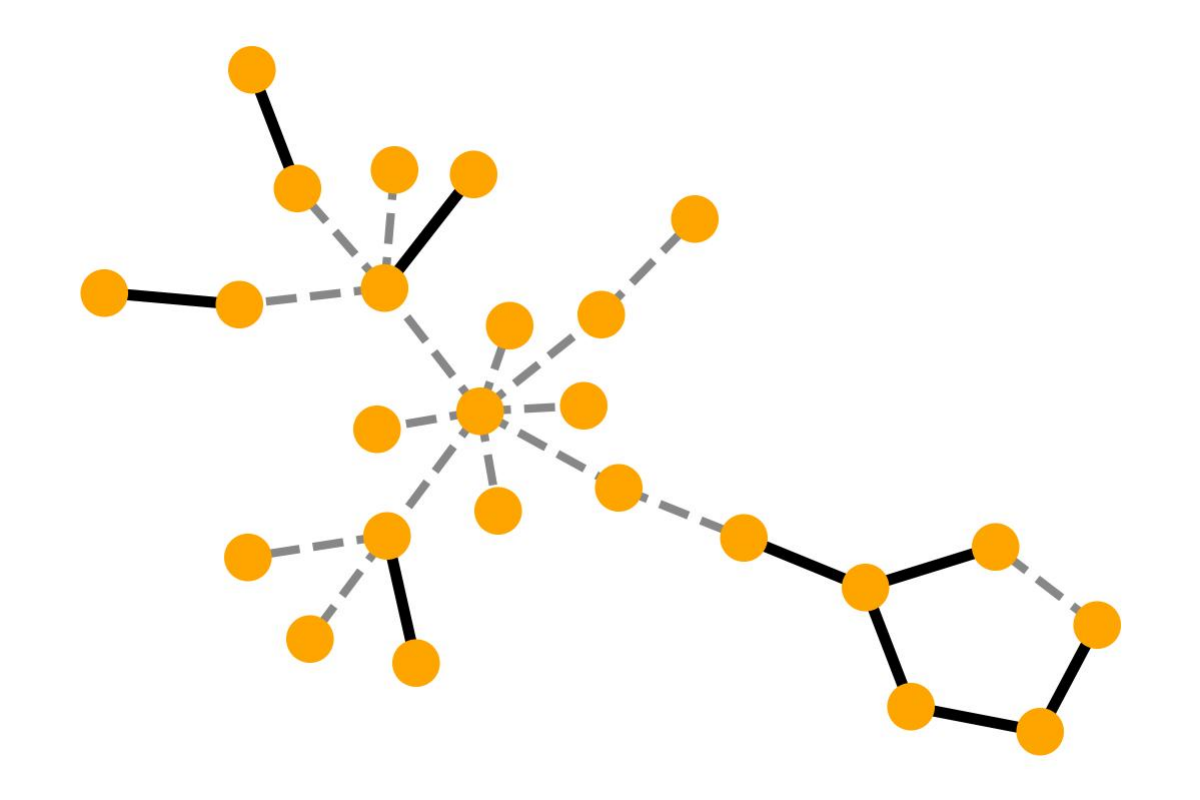}
    \end{subfigure}
    \par\vspace{0.8em}
    \begin{subfigure}[b]{0.19\textwidth}
        \includegraphics[width=\linewidth]{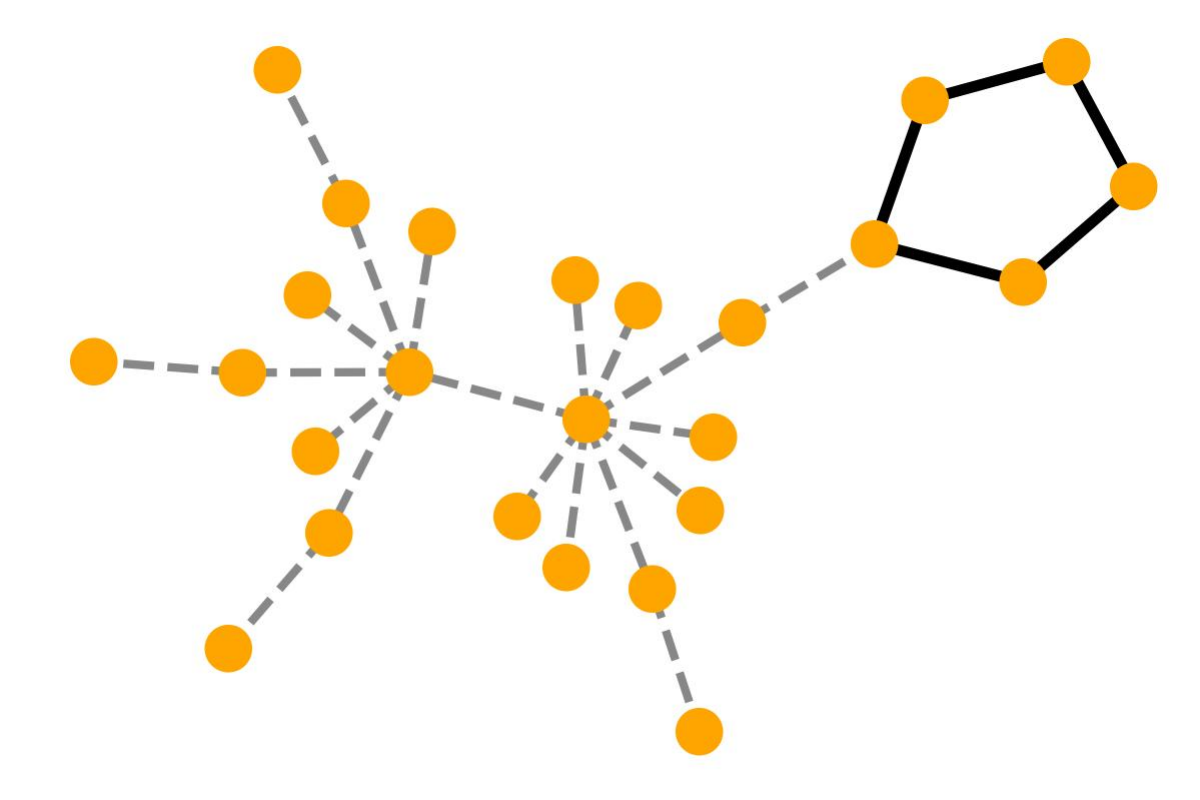}
        \caption{Ground Truth}
    \end{subfigure}
    \begin{subfigure}[b]{0.19\textwidth}
        \includegraphics[width=\linewidth]{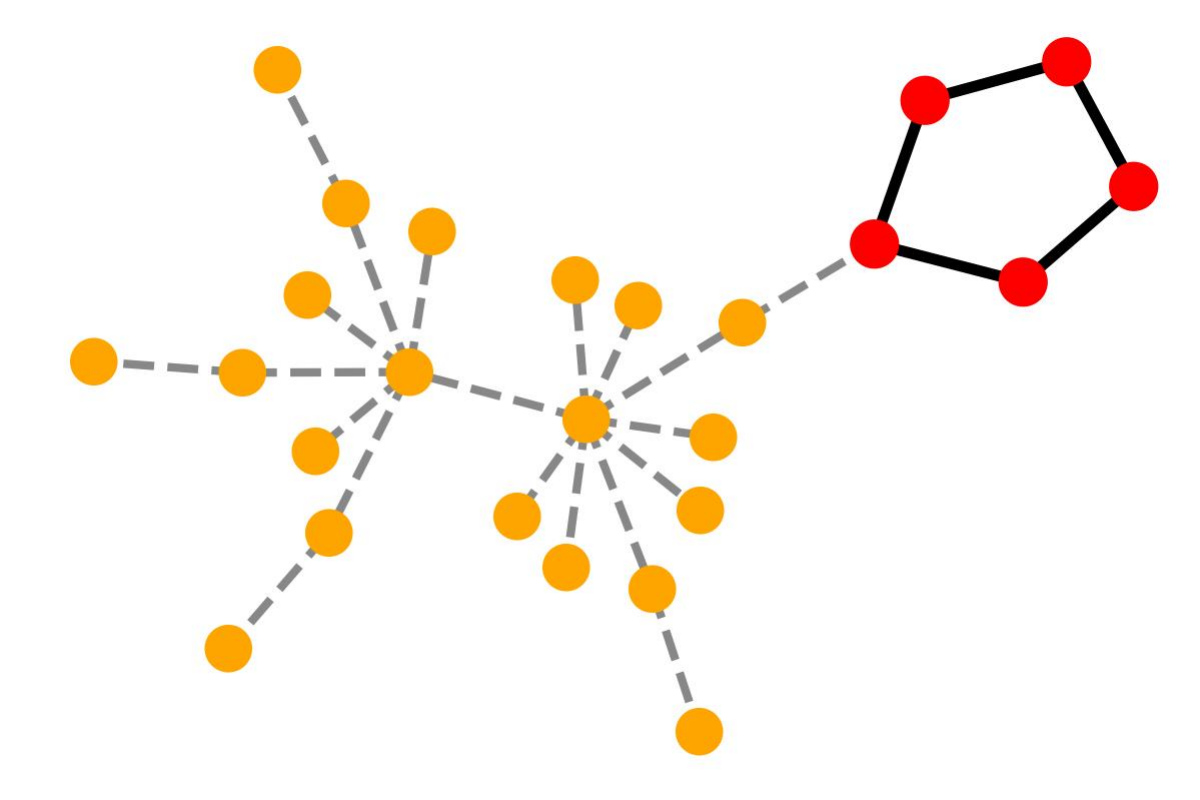}
        \caption{HPME}
    \end{subfigure}
    \begin{subfigure}[b]{0.19\textwidth}
        \includegraphics[width=\linewidth]{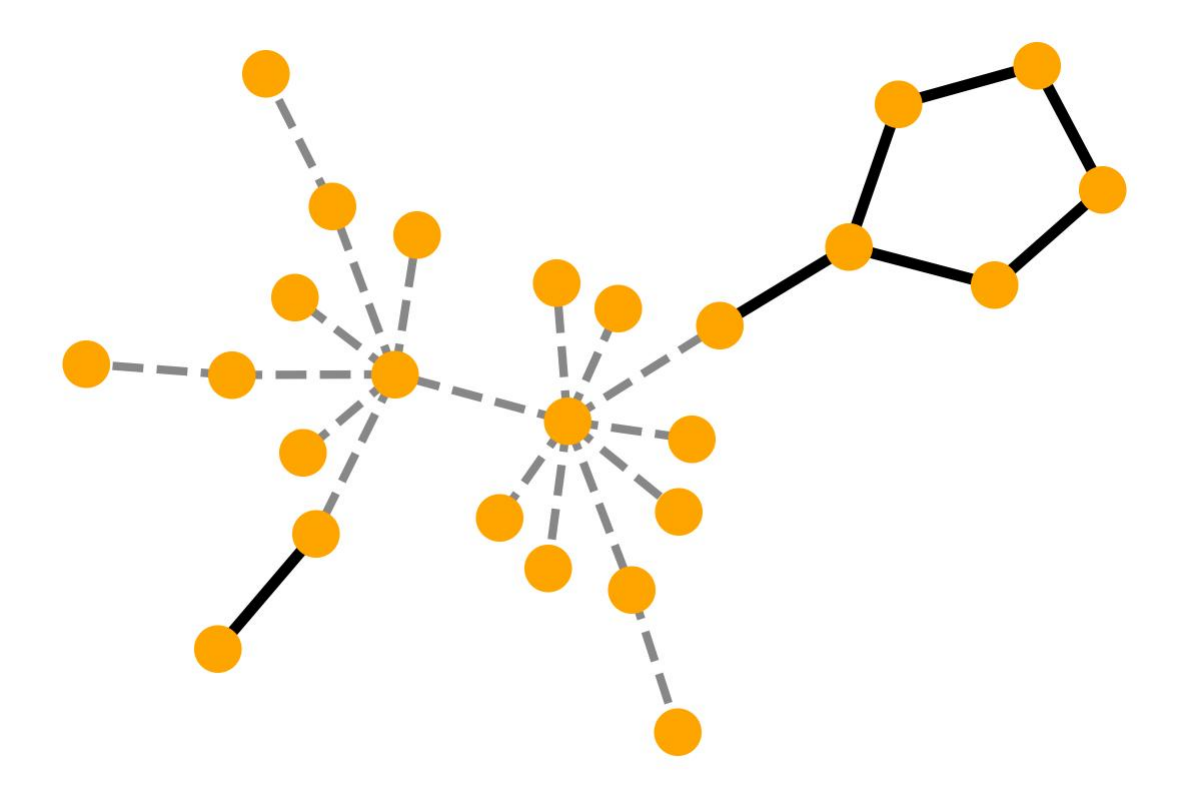}
        \caption{ProxyExplainer}
    \end{subfigure}
    \begin{subfigure}[b]{0.19\textwidth}
        \includegraphics[width=\linewidth]{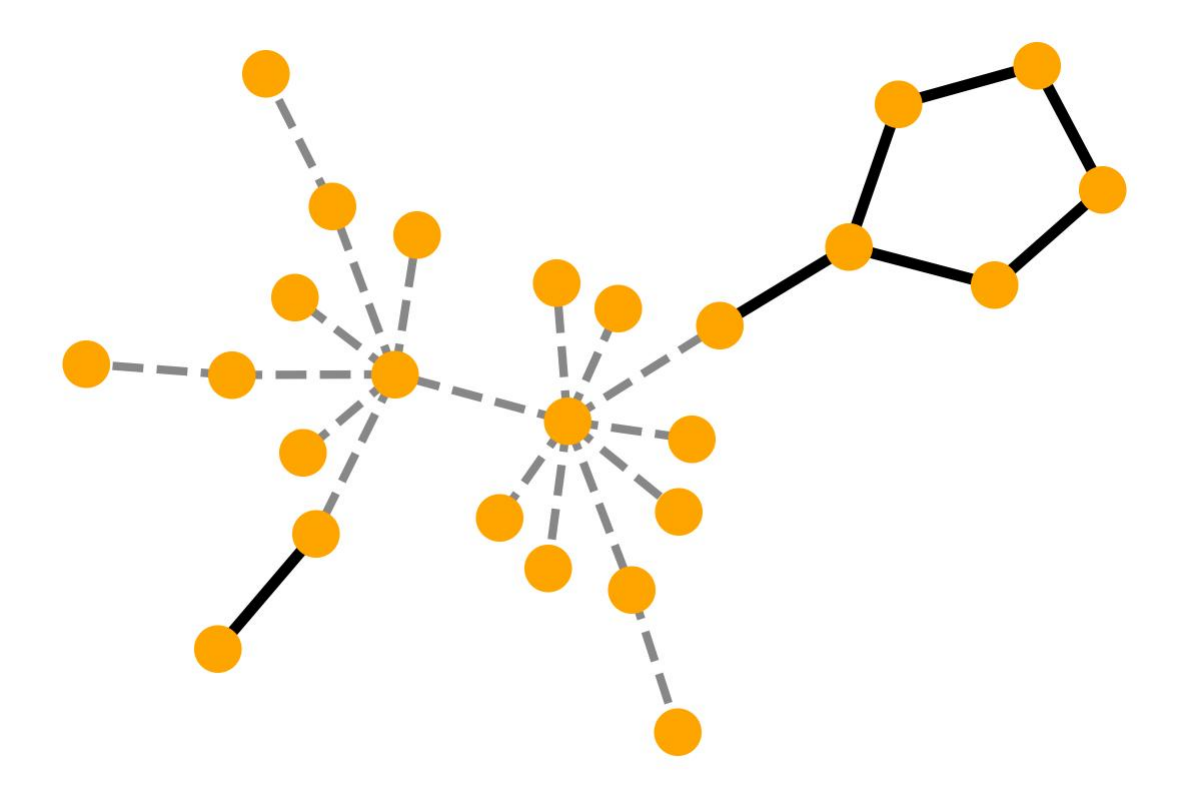}
        \caption{MixupExplainer}
    \end{subfigure}
    \begin{subfigure}[b]{0.19\textwidth}
        \includegraphics[width=\linewidth]{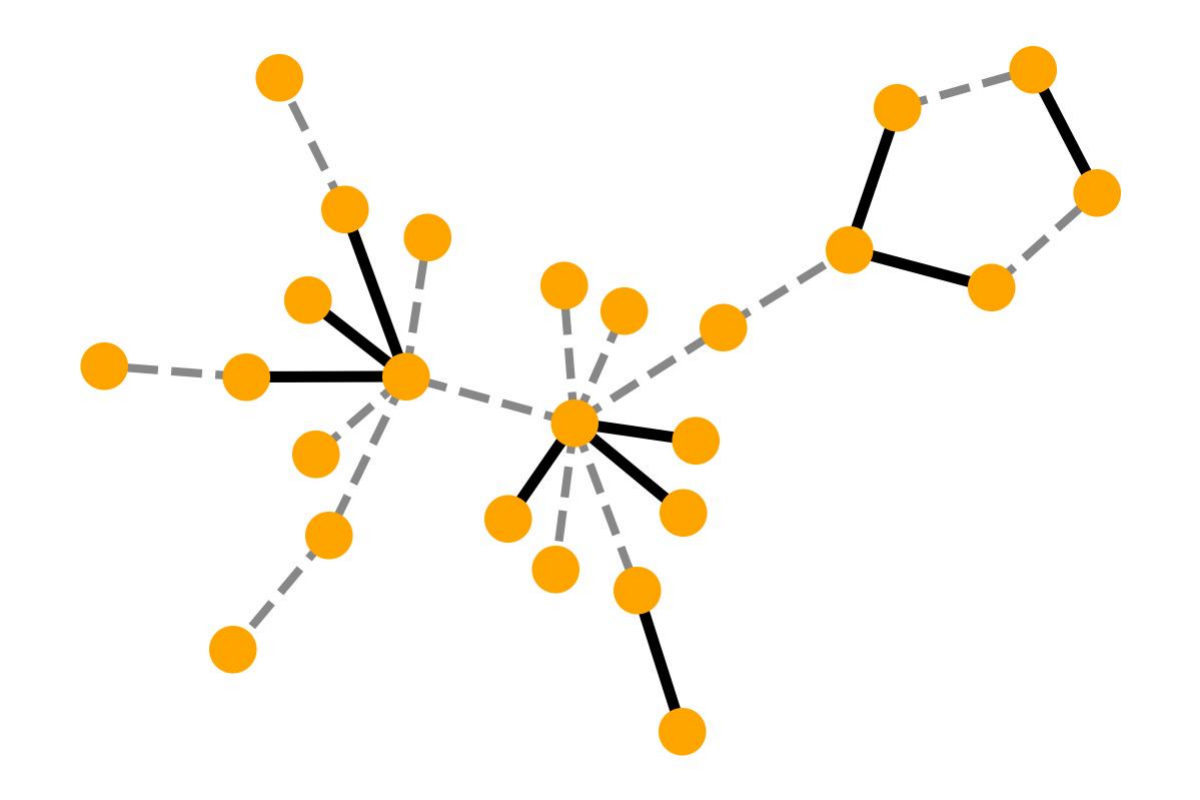}
        \caption{MetaGNN}
    \end{subfigure}
    \caption{Visualization of explanation on BA-Motif-Volume.}
    \label{fig:app:casestudy:BA-Motif-Volume}
\end{figure*}

\begin{figure*}[h]
    \centering
    \vspace{0.8em}
    \begin{subfigure}[b]{0.19\textwidth}
        \includegraphics[width=\linewidth]{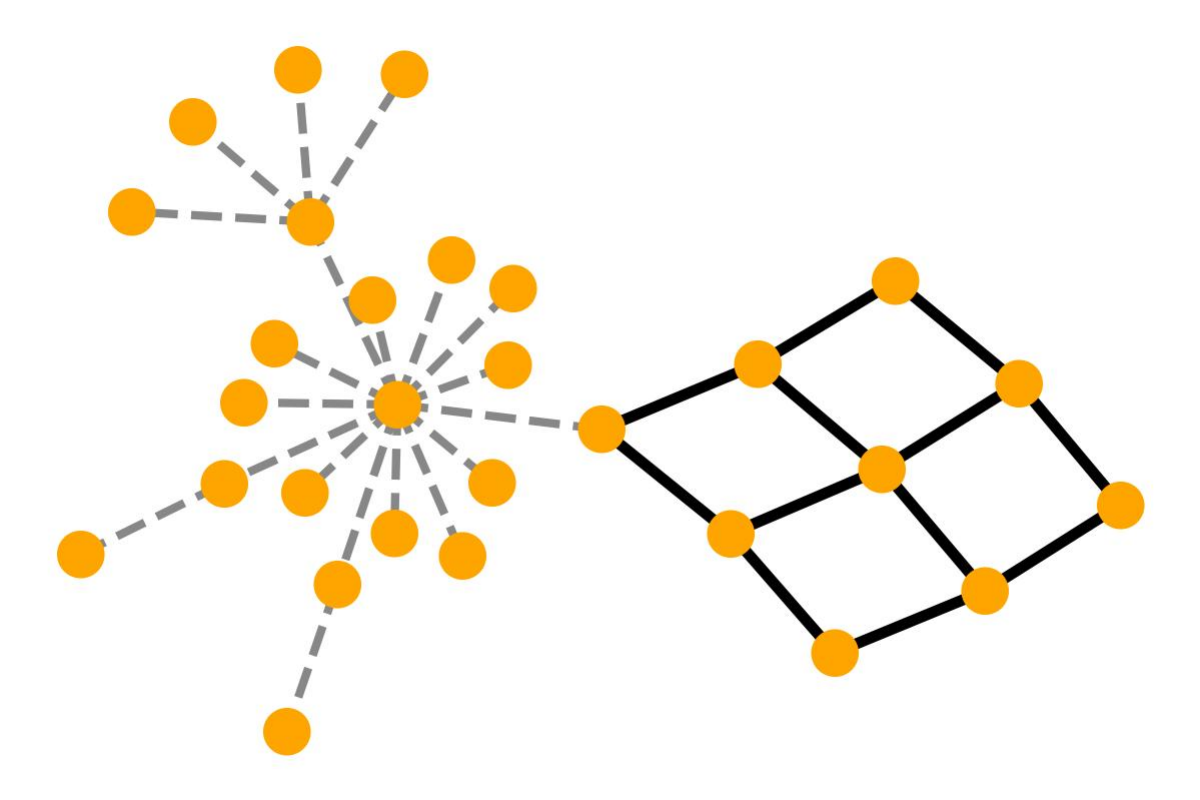}
    \end{subfigure}
    \begin{subfigure}[b]{0.19\textwidth}
        \includegraphics[width=\linewidth]{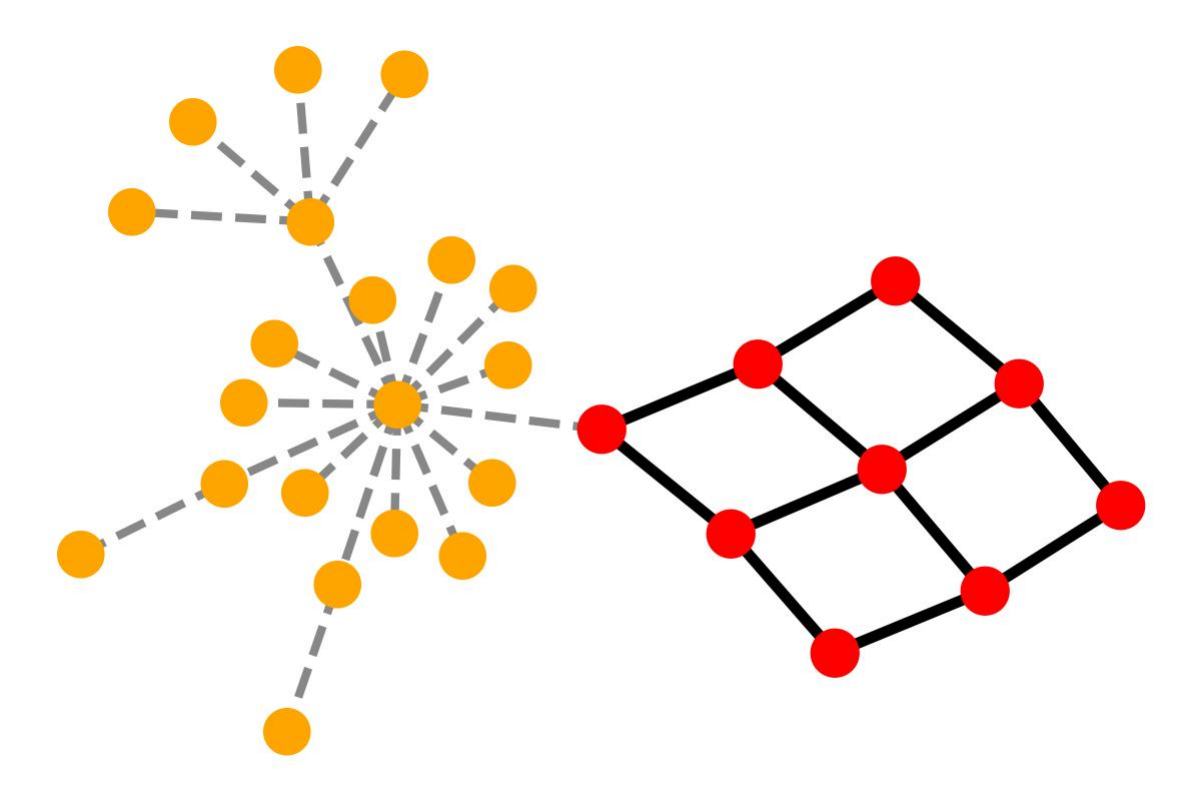}
    \end{subfigure}
    \begin{subfigure}[b]{0.19\textwidth}
        \includegraphics[width=\linewidth]{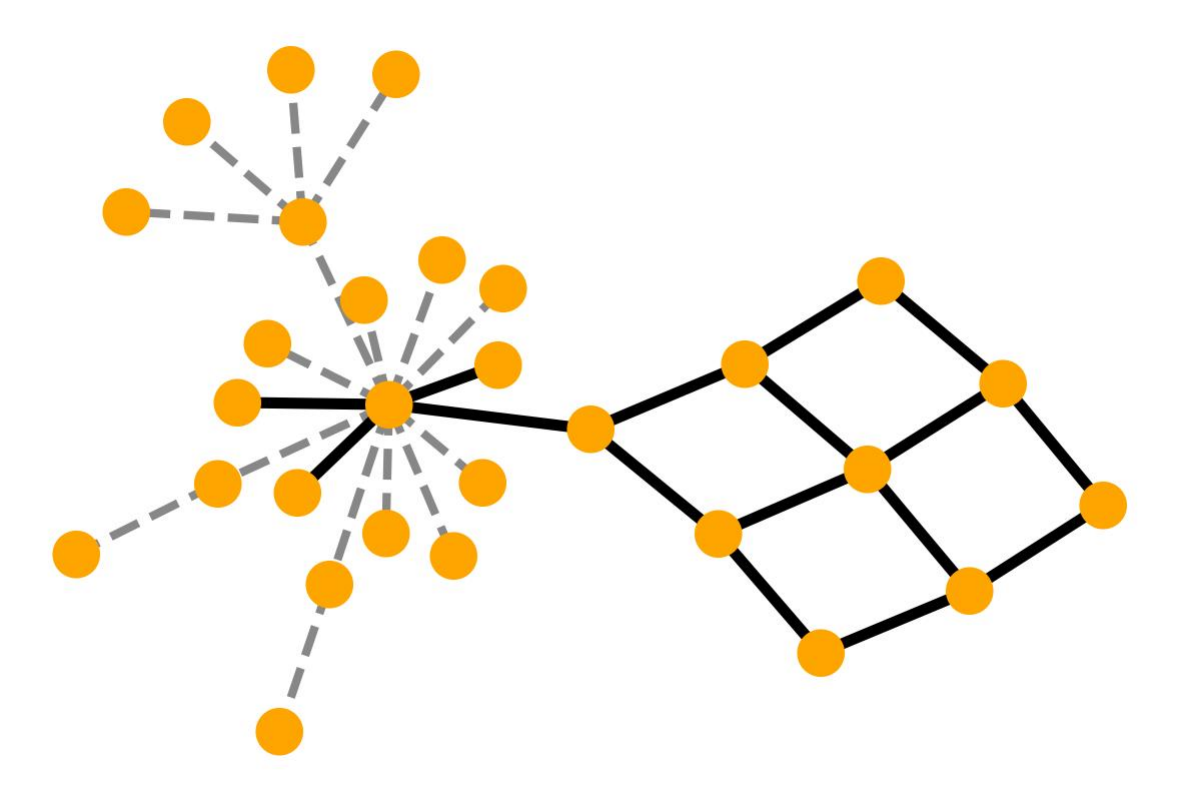}
    \end{subfigure}
    \begin{subfigure}[b]{0.19\textwidth}
        \includegraphics[width=\linewidth]{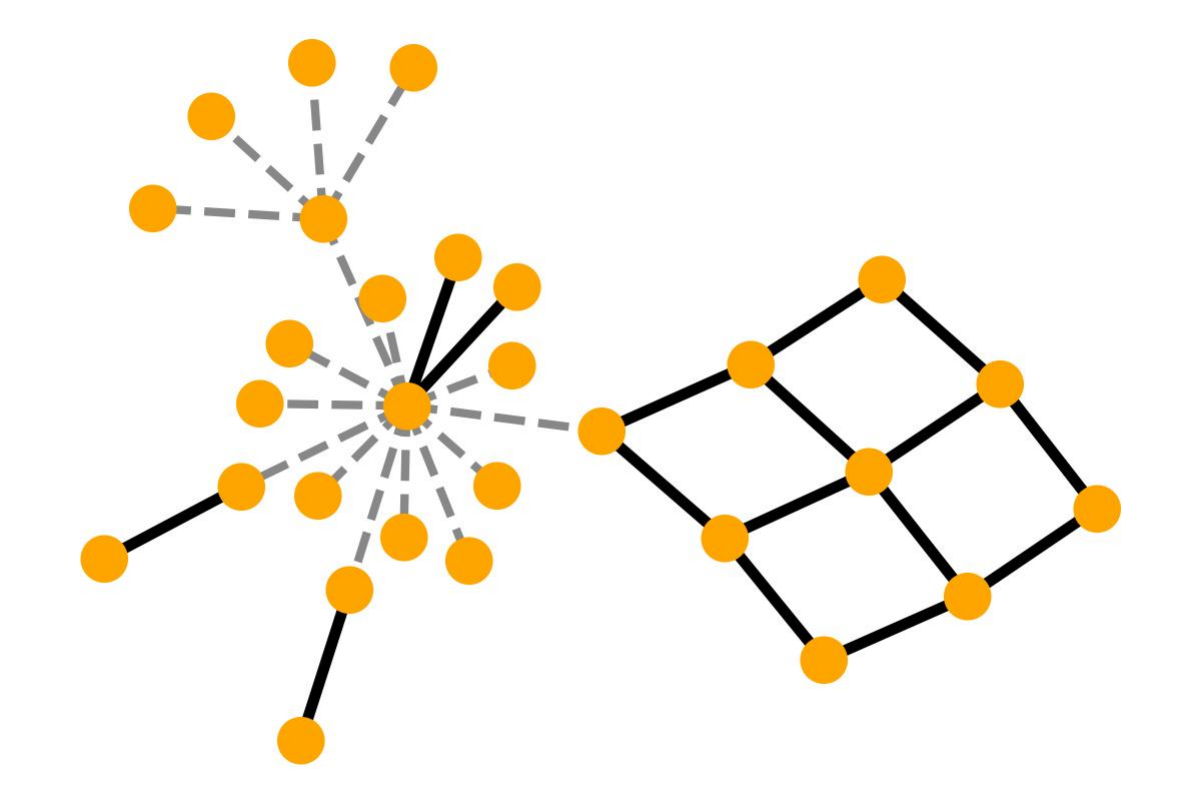}
    \end{subfigure}
    \begin{subfigure}[b]{0.19\textwidth}
        \includegraphics[width=\linewidth]{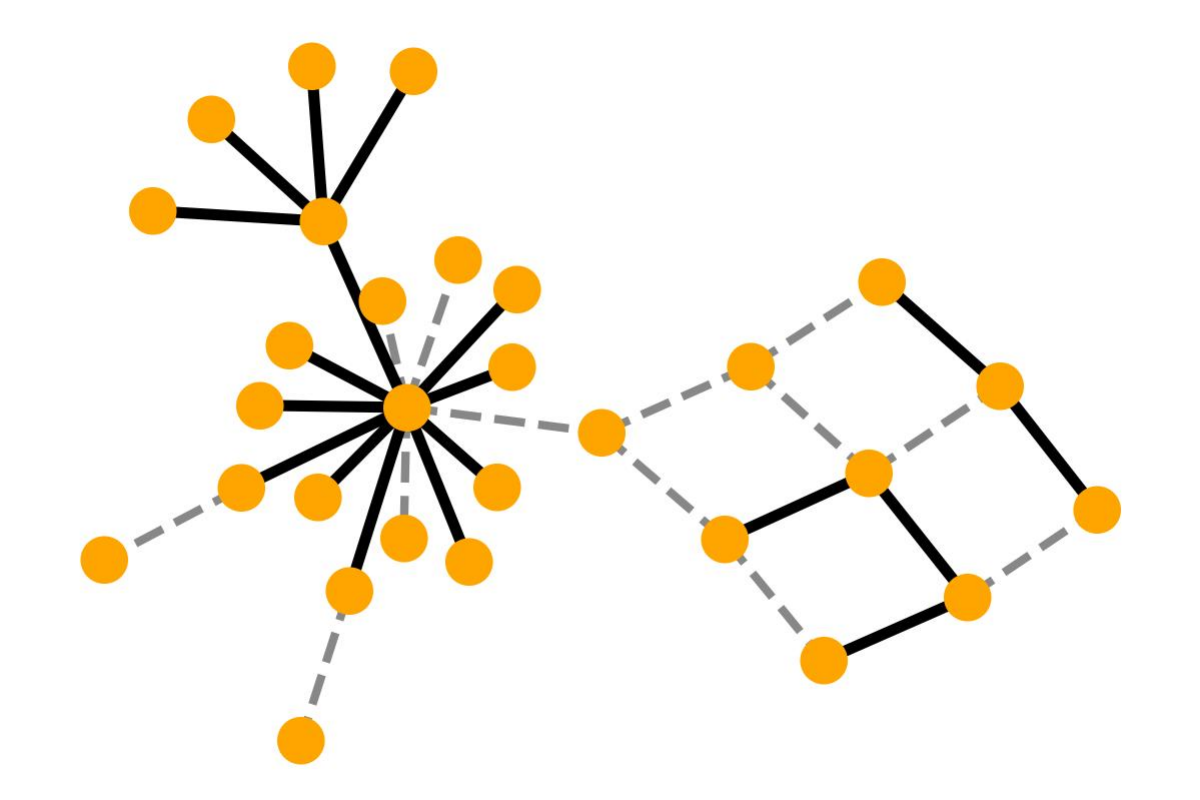}
    \end{subfigure}
    \par\vspace{0.8em}
    \begin{subfigure}[b]{0.19\textwidth}
        \includegraphics[width=\linewidth]{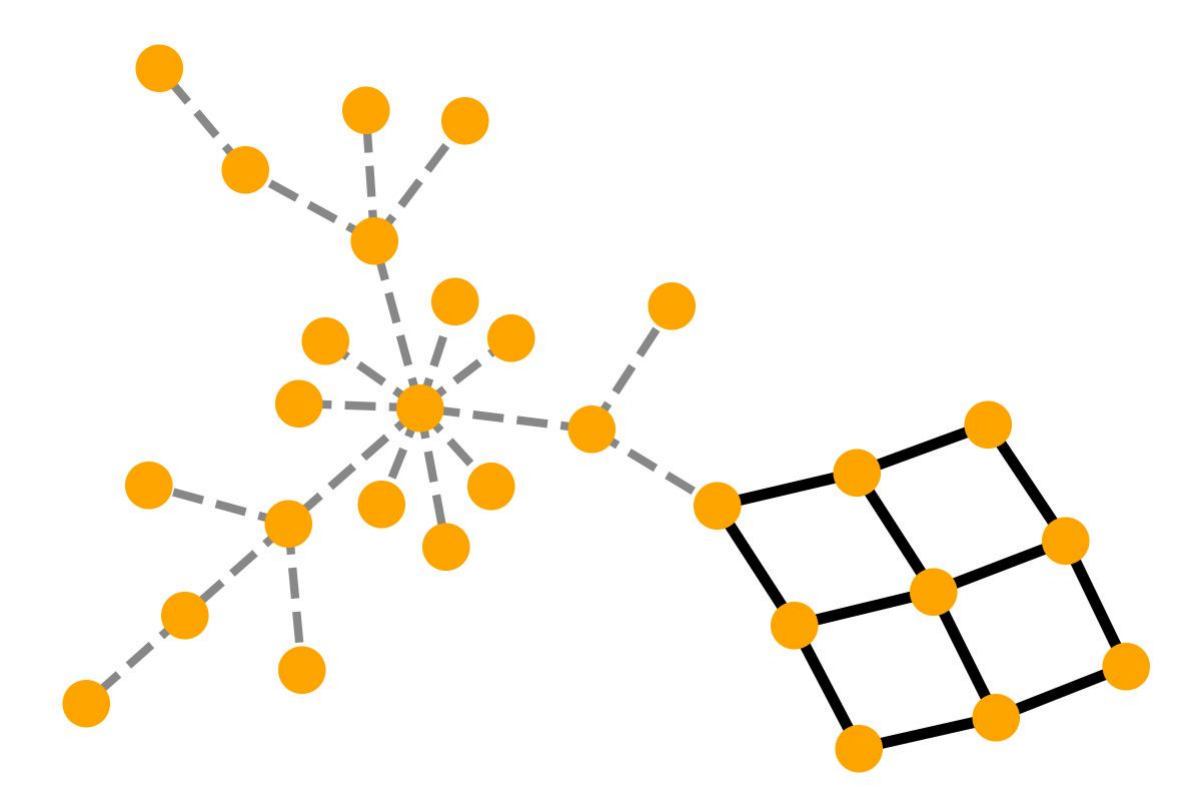}
    \end{subfigure}
    \begin{subfigure}[b]{0.19\textwidth}
        \includegraphics[width=\linewidth]{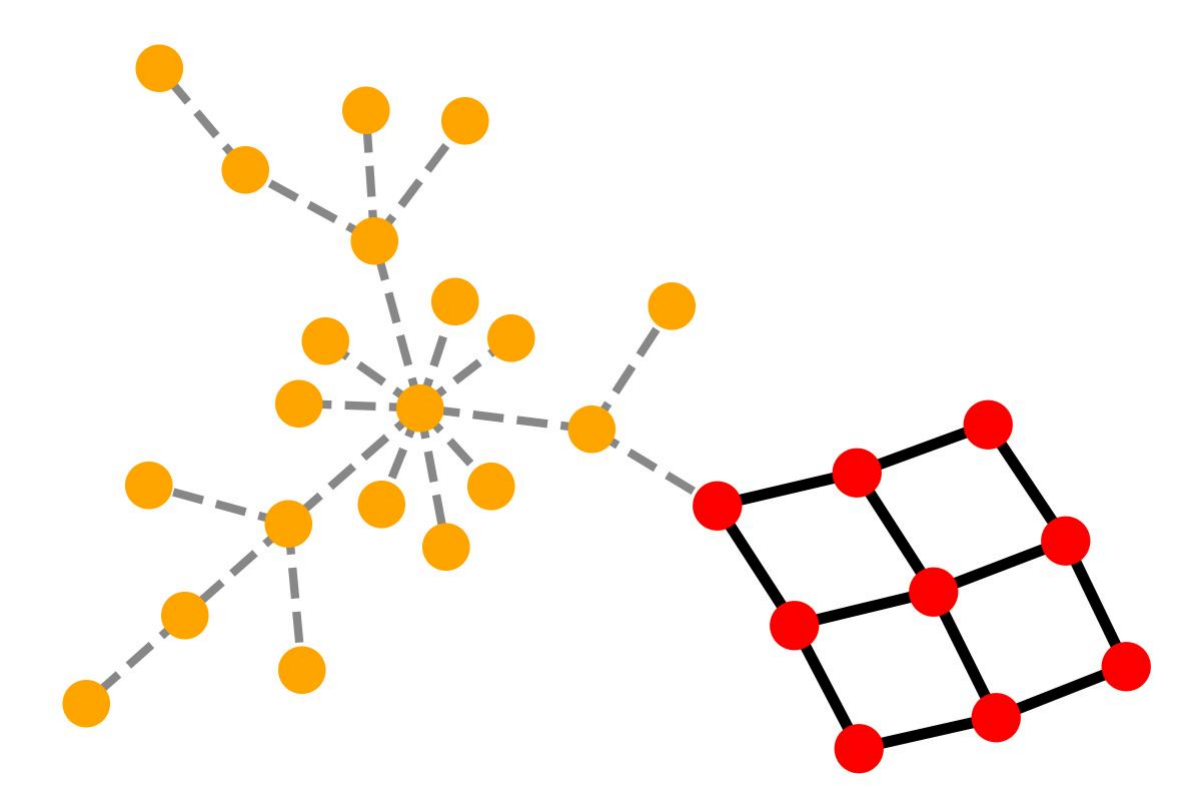}
    \end{subfigure}
    \begin{subfigure}[b]{0.19\textwidth}
        \includegraphics[width=\linewidth]{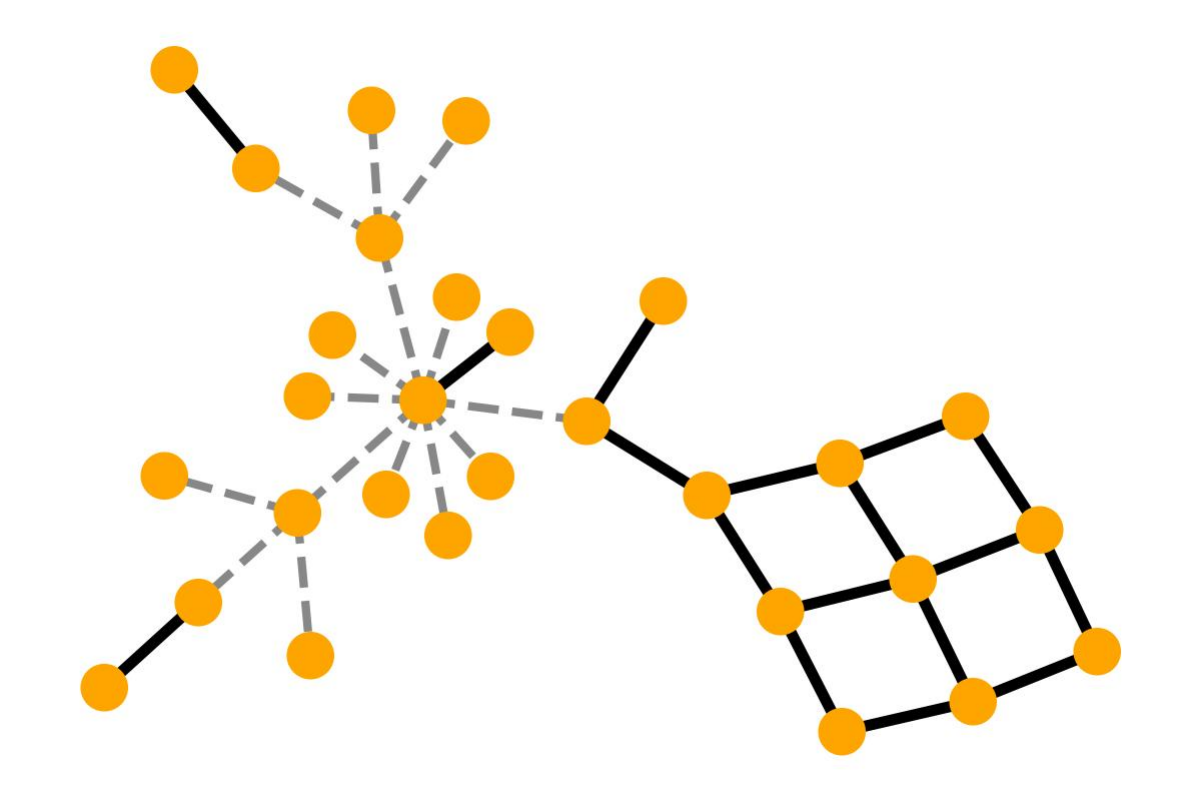}
    \end{subfigure}
    \begin{subfigure}[b]{0.19\textwidth}
        \includegraphics[width=\linewidth]{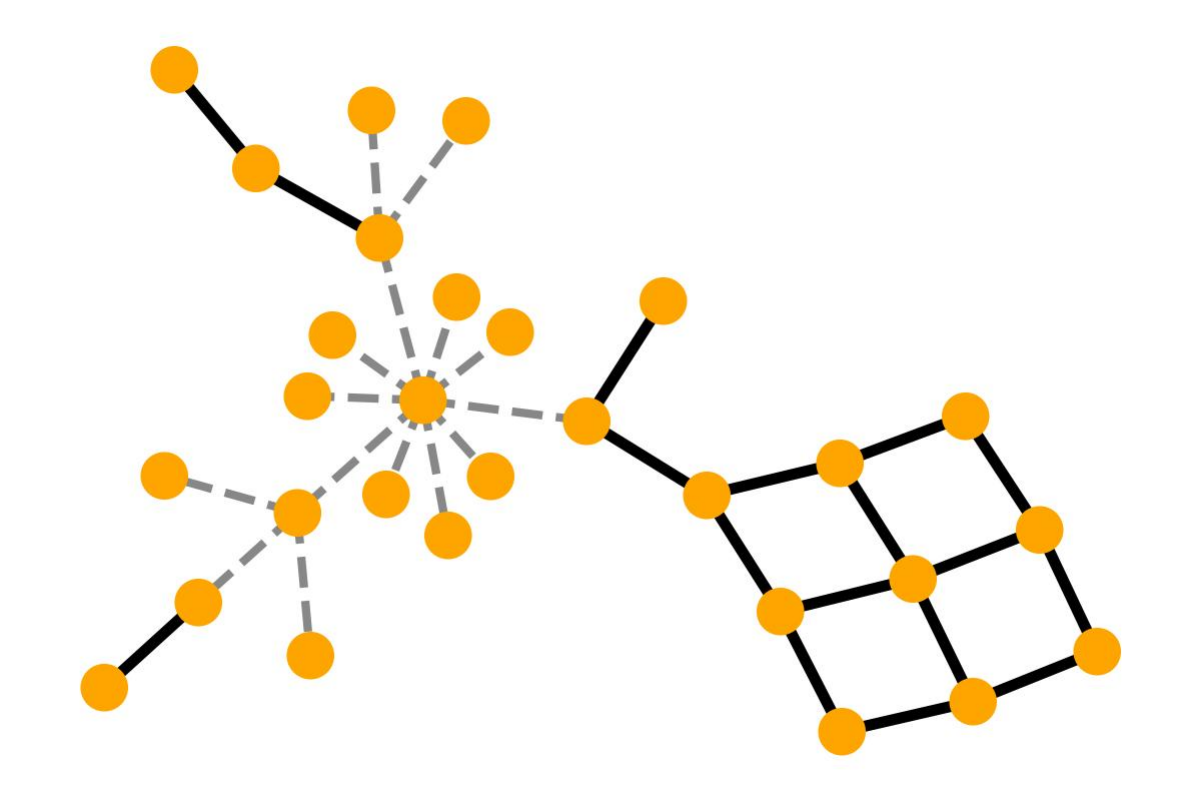}
    \end{subfigure}
    \begin{subfigure}[b]{0.19\textwidth}
        \includegraphics[width=\linewidth]{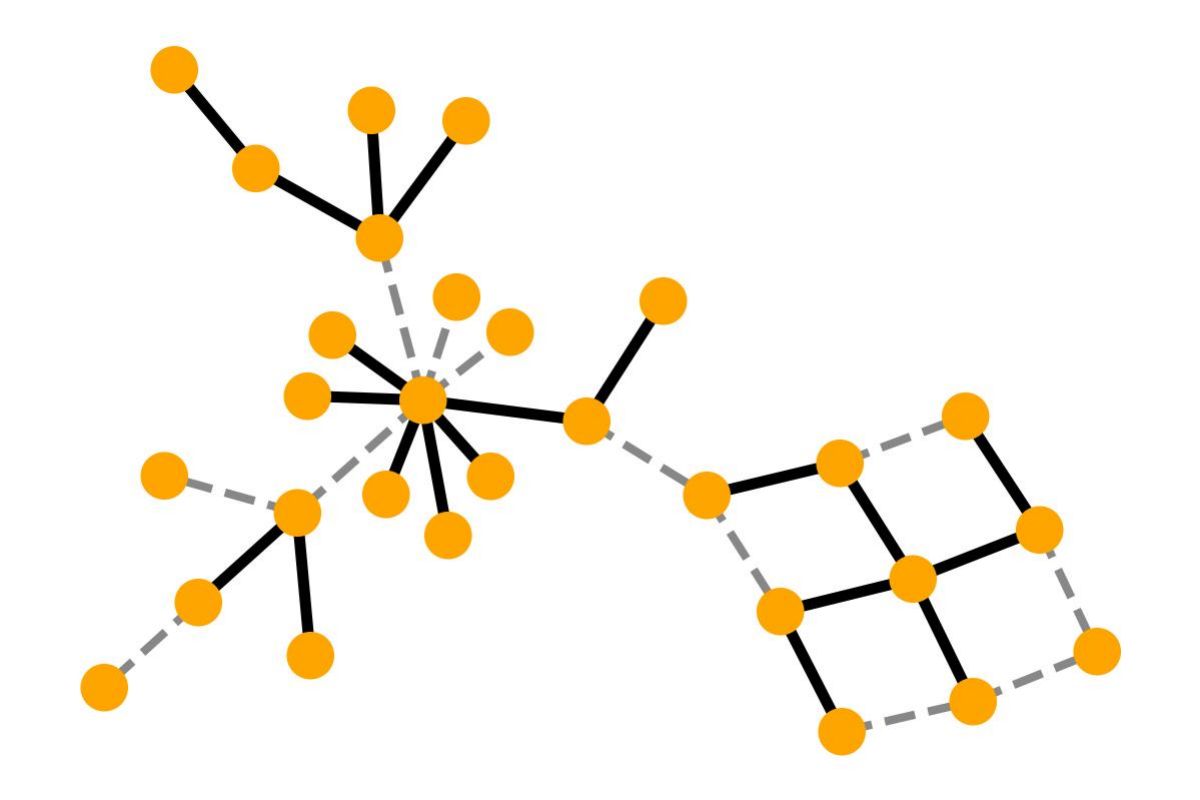}
    \end{subfigure}
    \par\vspace{0.8em}
    \begin{subfigure}[b]{0.19\textwidth}
        \includegraphics[width=\linewidth]{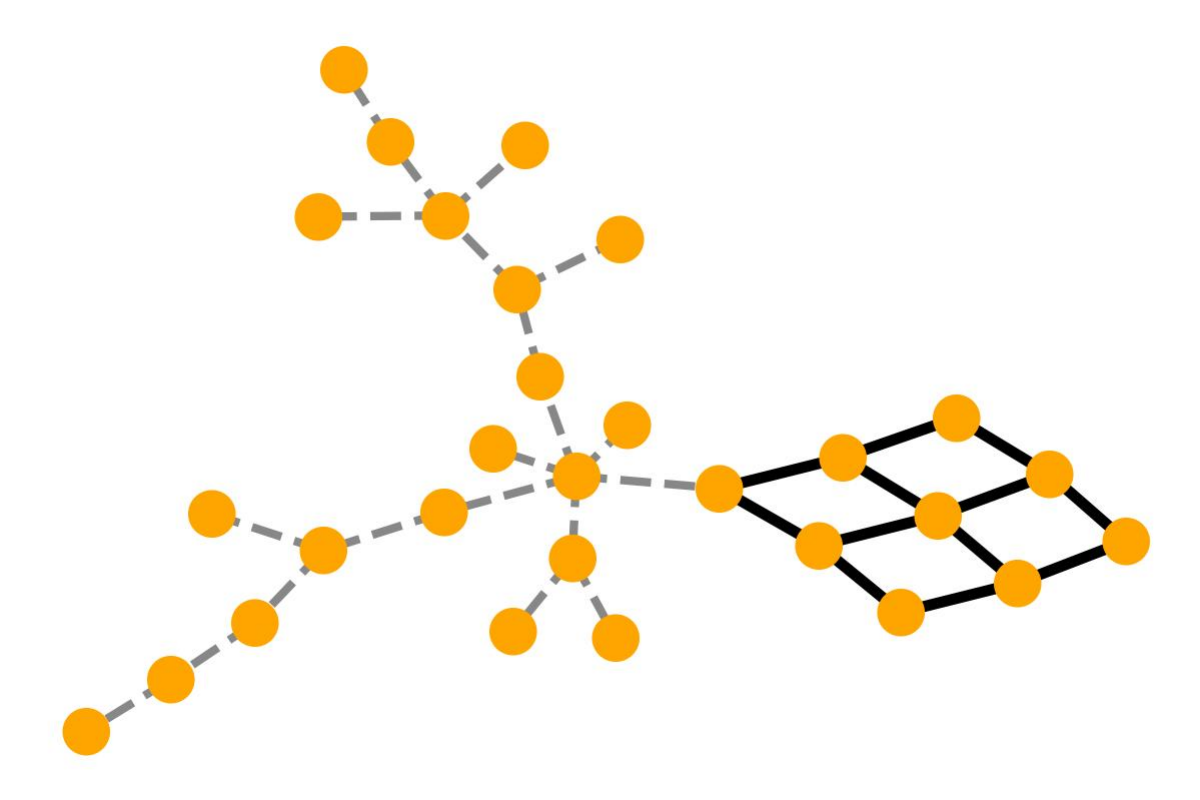}
        \caption{Ground Truth}
    \end{subfigure}
    \begin{subfigure}[b]{0.19\textwidth}
        \includegraphics[width=\linewidth]{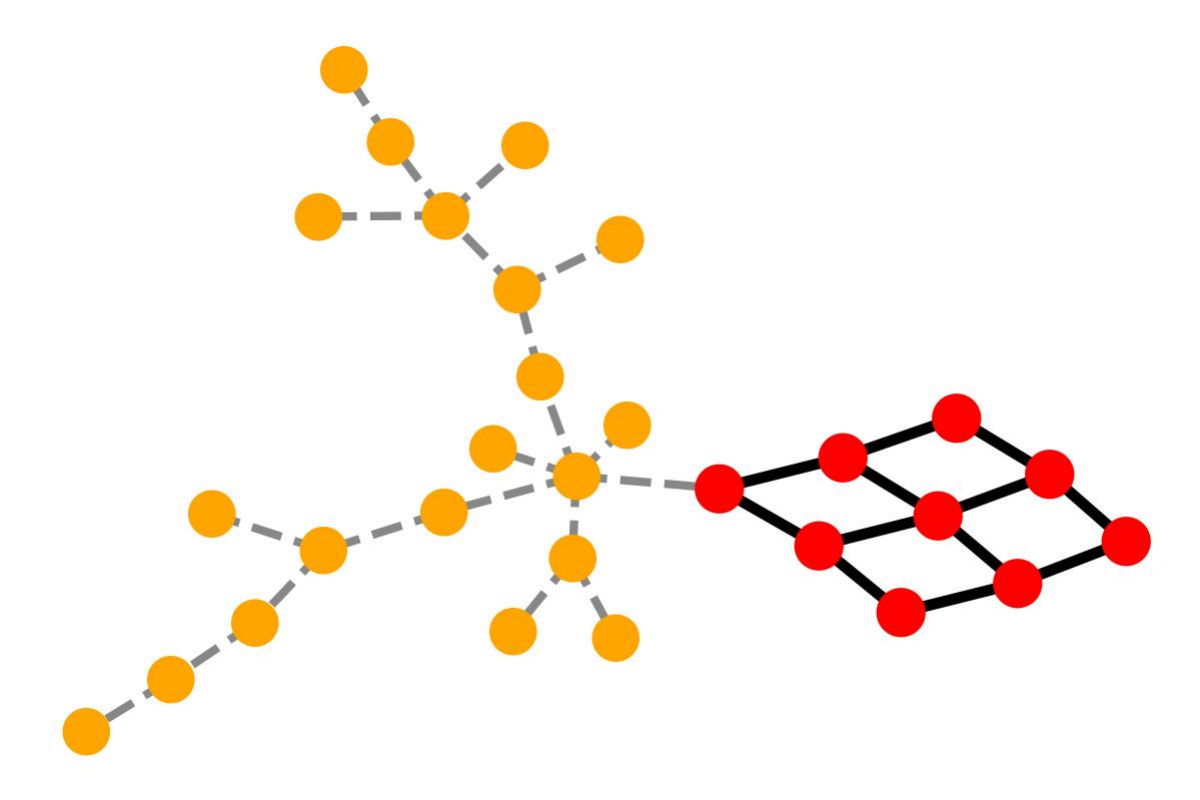}
        \caption{HPME}
    \end{subfigure}
    \begin{subfigure}[b]{0.19\textwidth}
        \includegraphics[width=\linewidth]{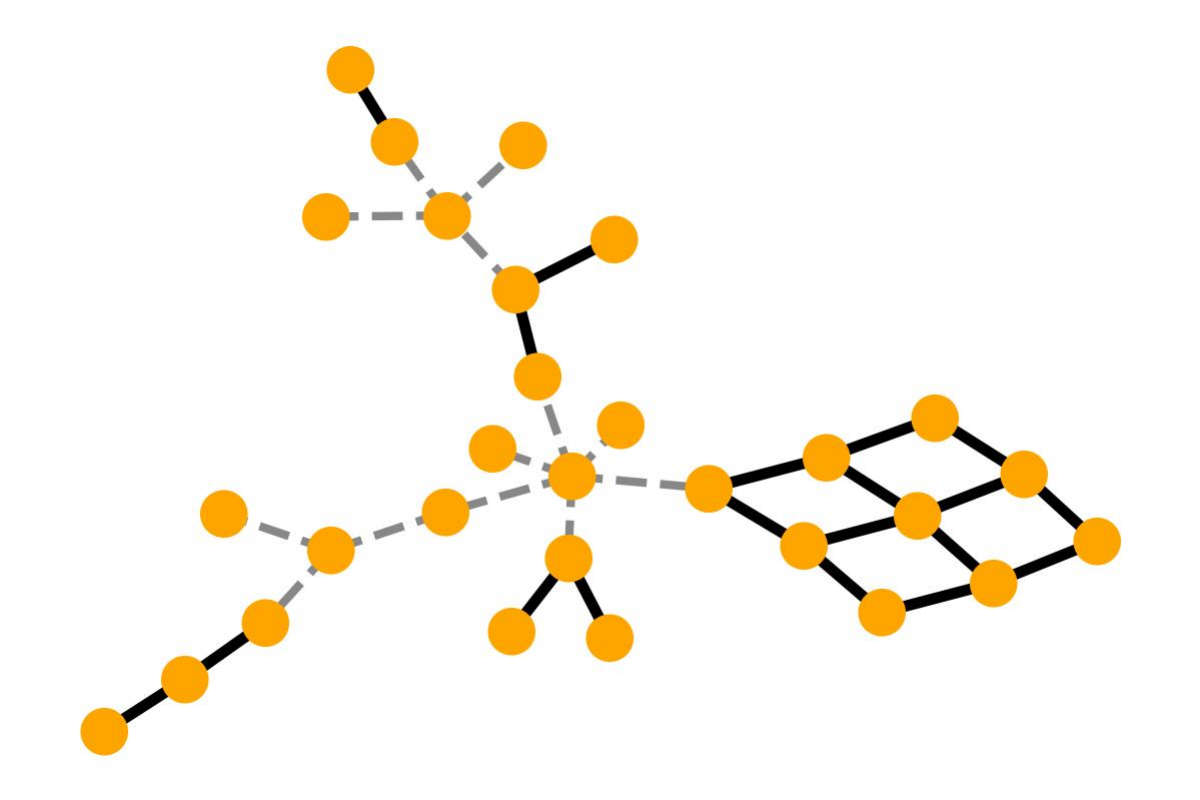}
        \caption{ProxyExplainer}
    \end{subfigure}
    \begin{subfigure}[b]{0.19\textwidth}
        \includegraphics[width=\linewidth]{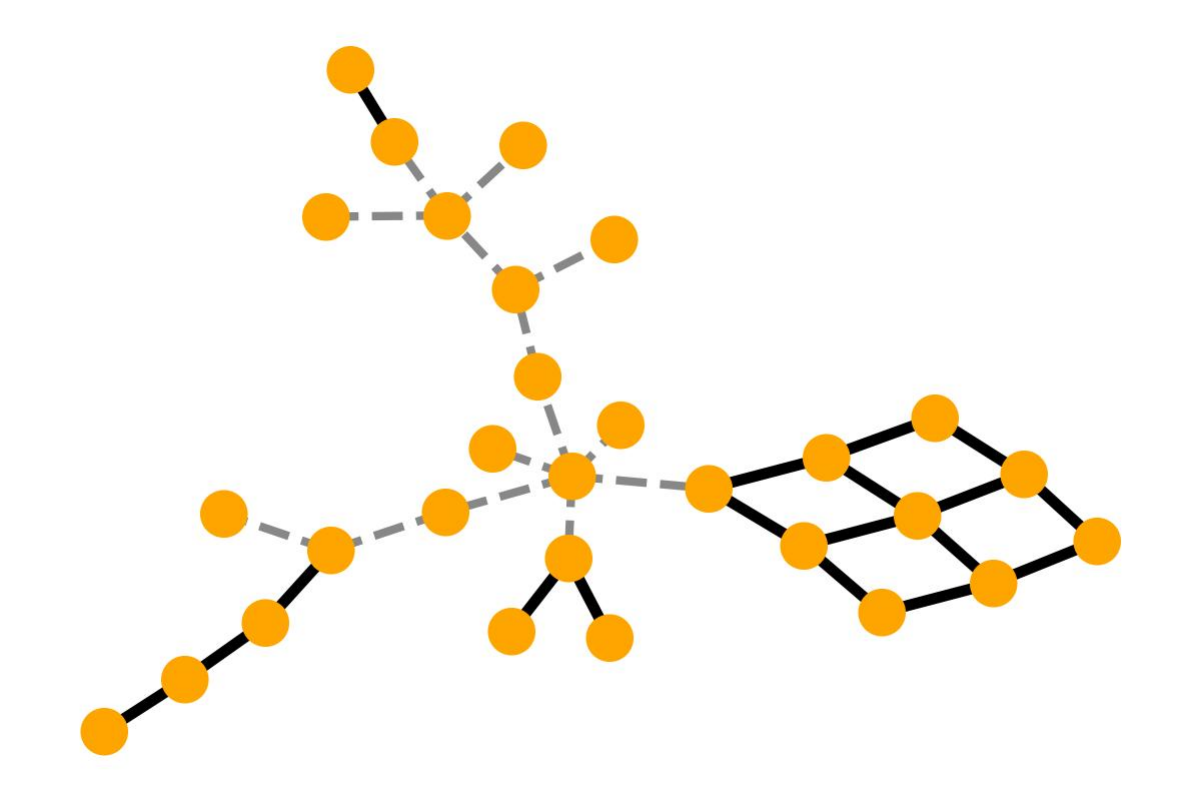}
        \caption{MixupExplainer}
    \end{subfigure}
    \begin{subfigure}[b]{0.19\textwidth}
        \includegraphics[width=\linewidth]{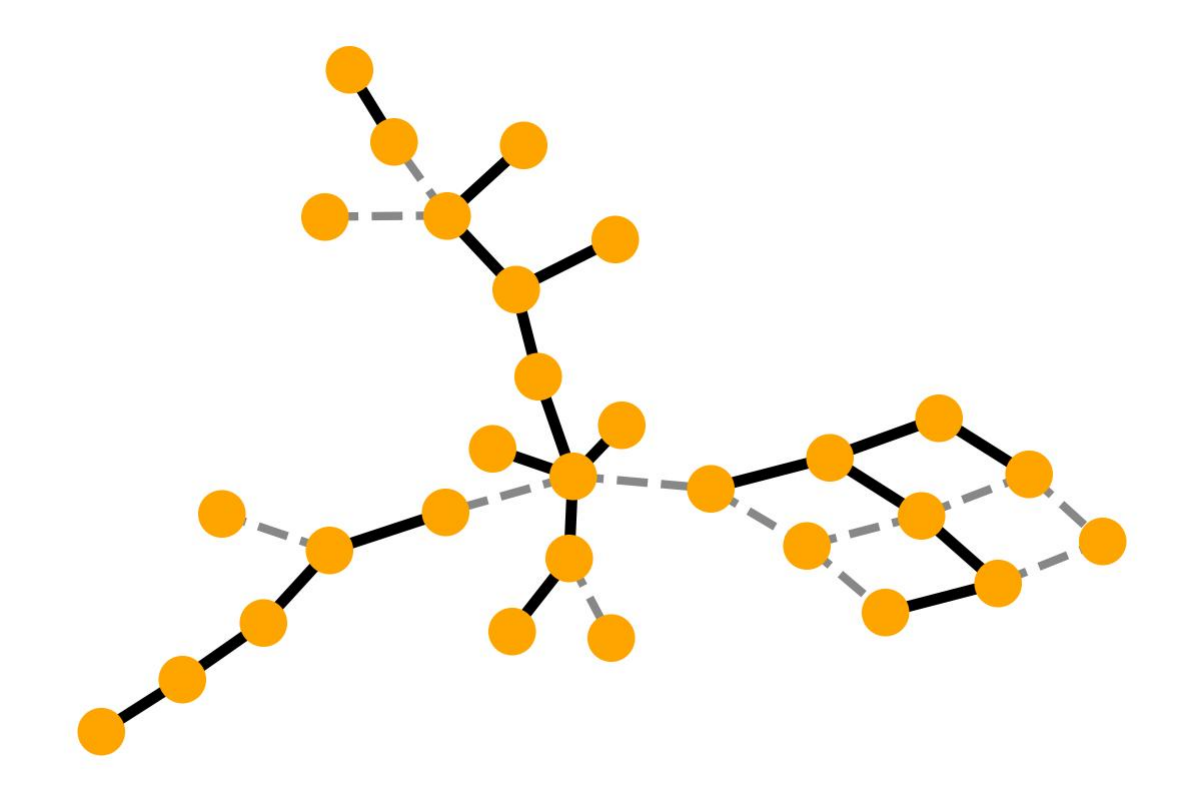}
        \caption{MetaGNN}
    \end{subfigure}
    \caption{Visualization of explanation on House-Grid-Volume.}
    \label{fig:app:casestudy:House-Grid-Volume}
\end{figure*}

We further present extensive visualization experiments on both classification and regression datasets. 
The classification datasets include BA-2Motifs, BA-HouseAndGrid, and Benzene, as shown in \Figref{fig:app:casestudy:BA-2Motifs}, \Figref{fig:app:casestudy:BA-HouseAndGrid}, and \Figref{fig:app:casestudy:Benzene}, respectively. 
The regression datasets include BA-Motif-Volume, and House-OrGrid-Volume, as presented in \Figref{fig:app:casestudy:BA-Motif-Volume}, and \Figref{fig:app:casestudy:House-Grid-Volume}, respectively.
For each dataset, we randomly select three target graphs to evaluate different explainers.

\subsubsection{Visualization of Additional mixup Results} 
\label{sec:more_mixup}
\begin{figure*}[h]
    \centering
    \vspace{0.8em}
    \begin{subfigure}[b]{0.23\textwidth}
        \includegraphics[width=\linewidth]{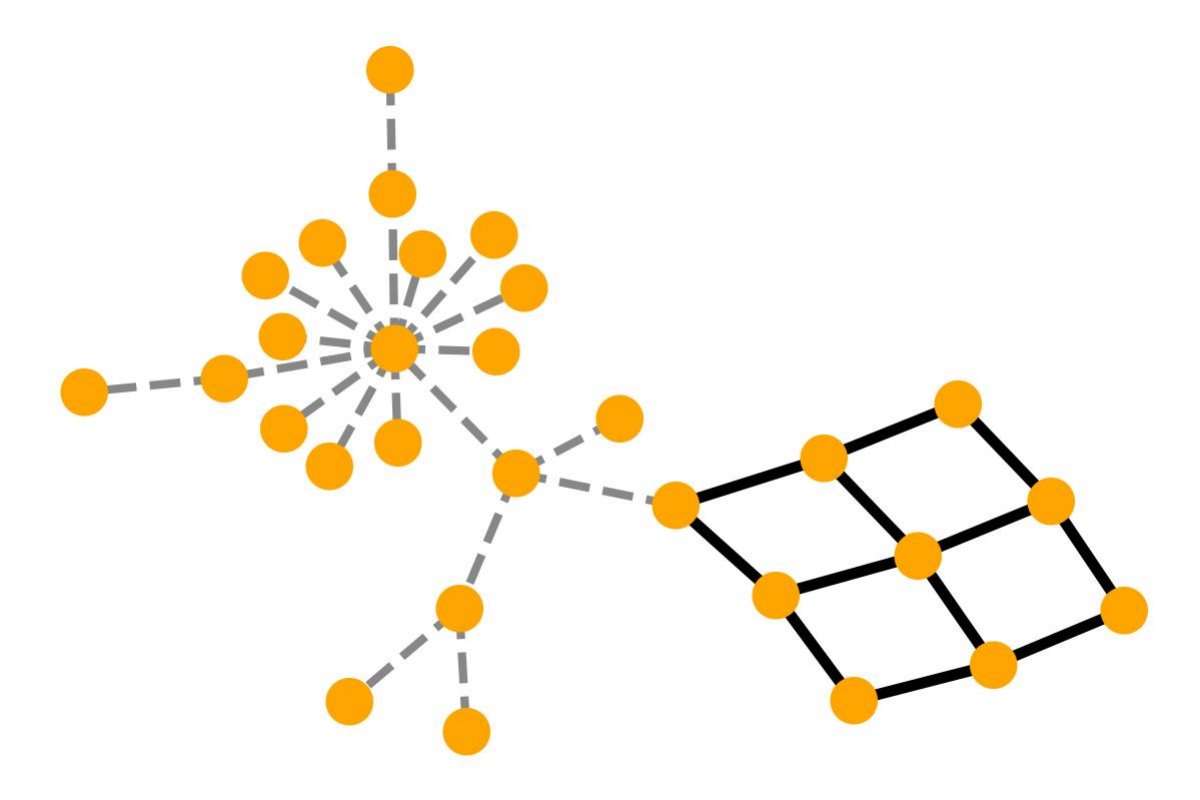}
    \end{subfigure}
    \begin{subfigure}[b]{0.23\textwidth}
        \includegraphics[width=\linewidth]{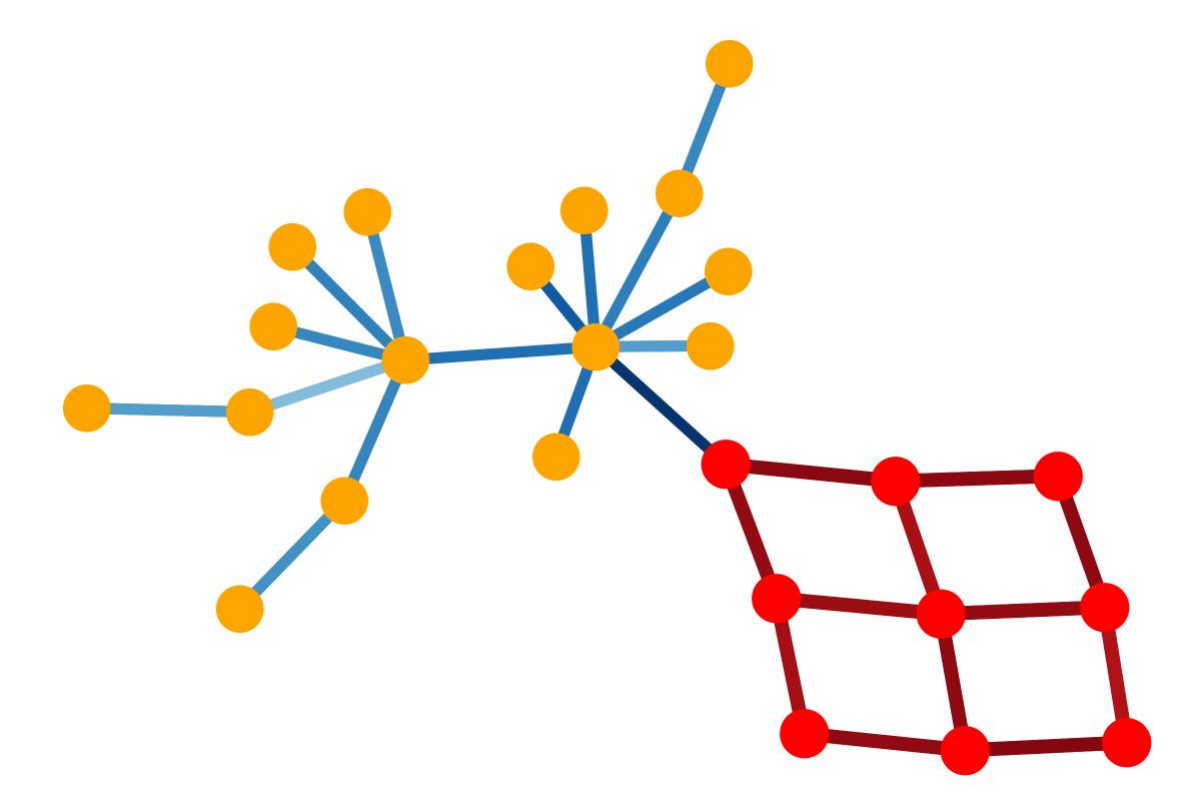}
    \end{subfigure}
    \begin{subfigure}[b]{0.23\textwidth}
        \includegraphics[width=\linewidth]{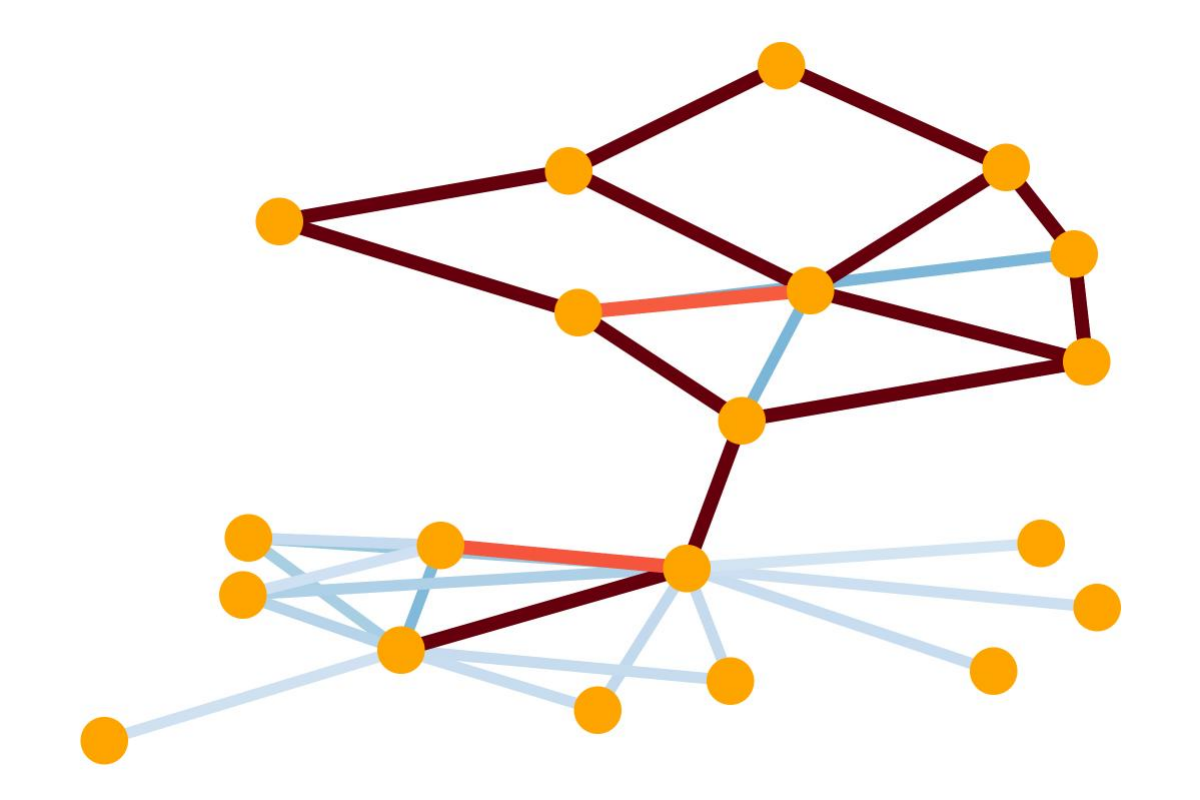}
    \end{subfigure}
    \begin{subfigure}[b]{0.23\textwidth}
        \includegraphics[width=\linewidth]{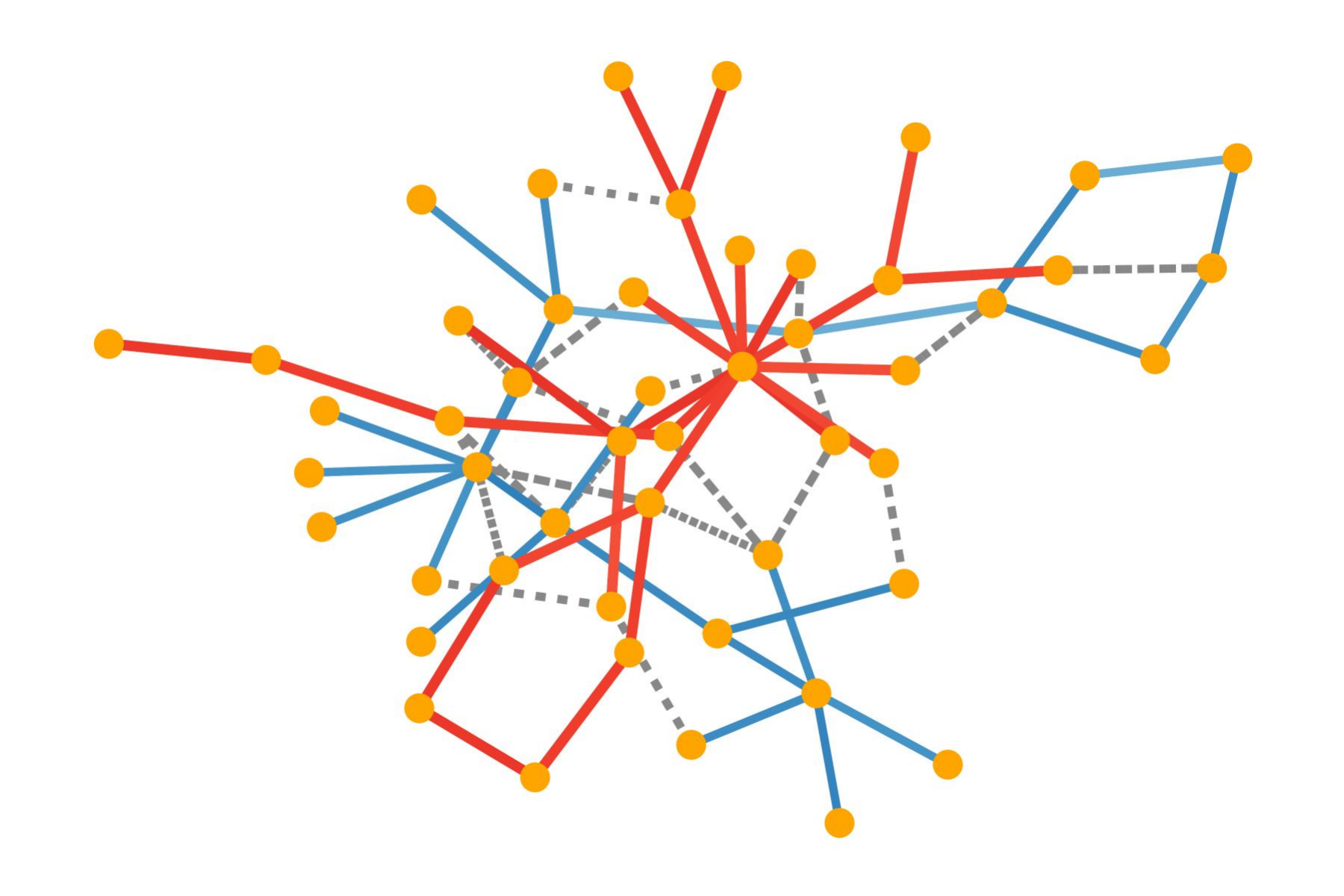}
    \end{subfigure}
    \par\vspace{0.8em}
    
    \begin{subfigure}[b]{0.23\textwidth}
        \includegraphics[width=\linewidth]{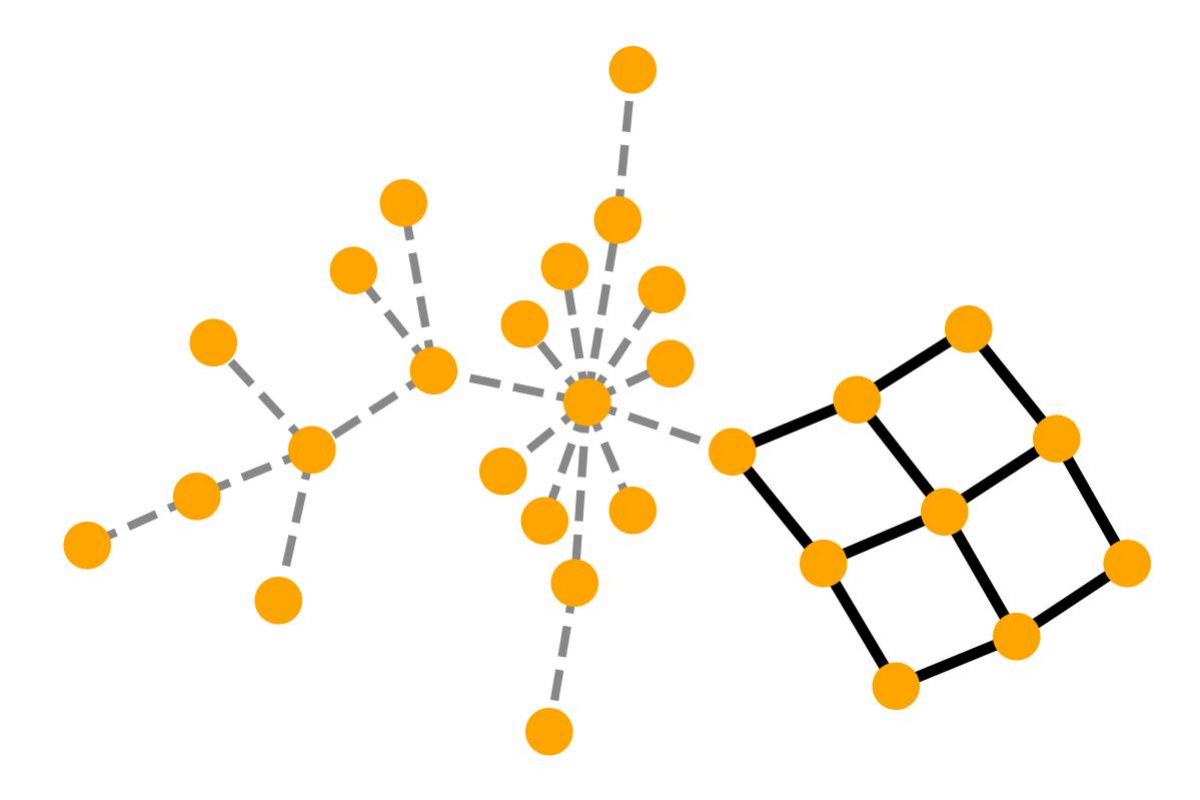}
    \end{subfigure}
    \begin{subfigure}[b]{0.23\textwidth}
        \includegraphics[width=\linewidth]{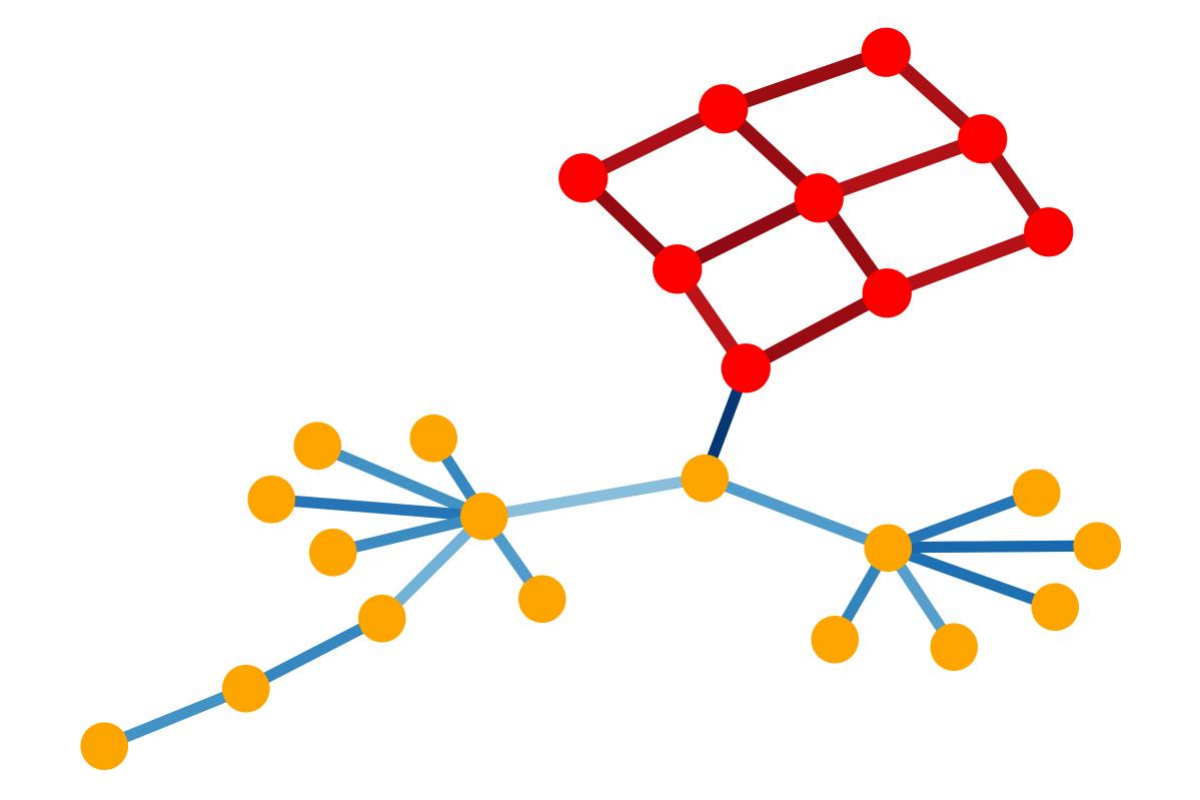}
    \end{subfigure}
    \begin{subfigure}[b]{0.23\textwidth}
        \includegraphics[width=\linewidth]{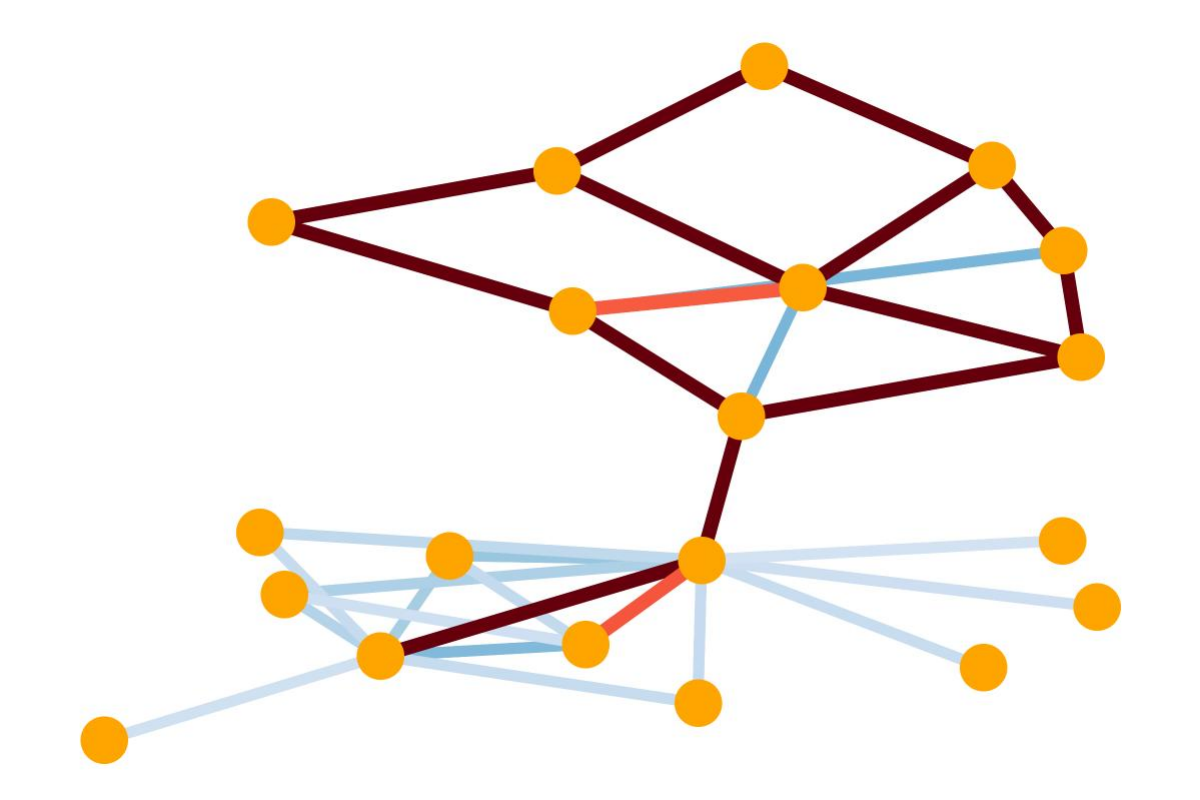}
    \end{subfigure}
    \begin{subfigure}[b]{0.23\textwidth}
        \includegraphics[width=\linewidth]{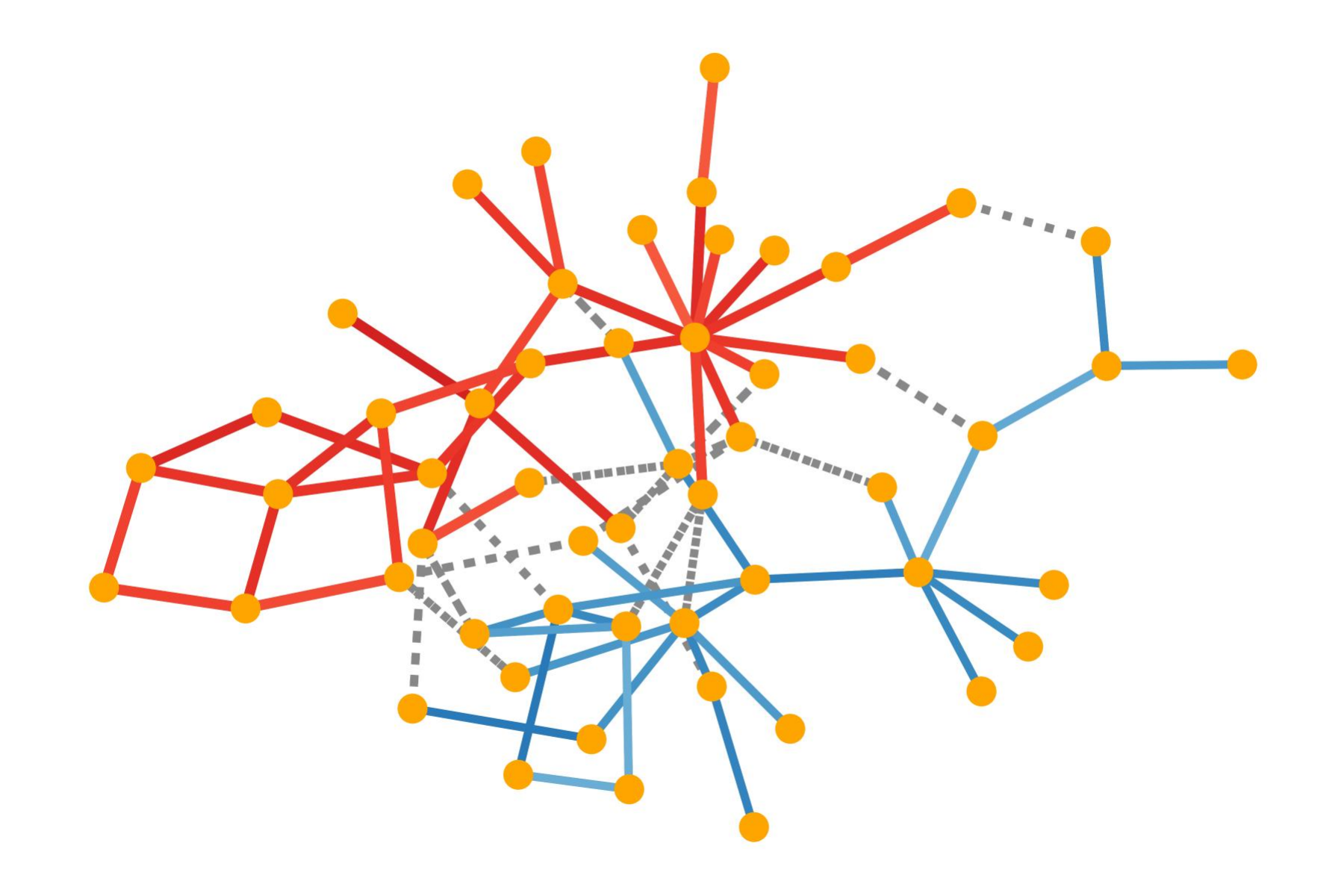}
    \end{subfigure}
    \par\vspace{0.8em}
    
    \begin{subfigure}[b]{0.23\textwidth}
        \includegraphics[width=\linewidth]{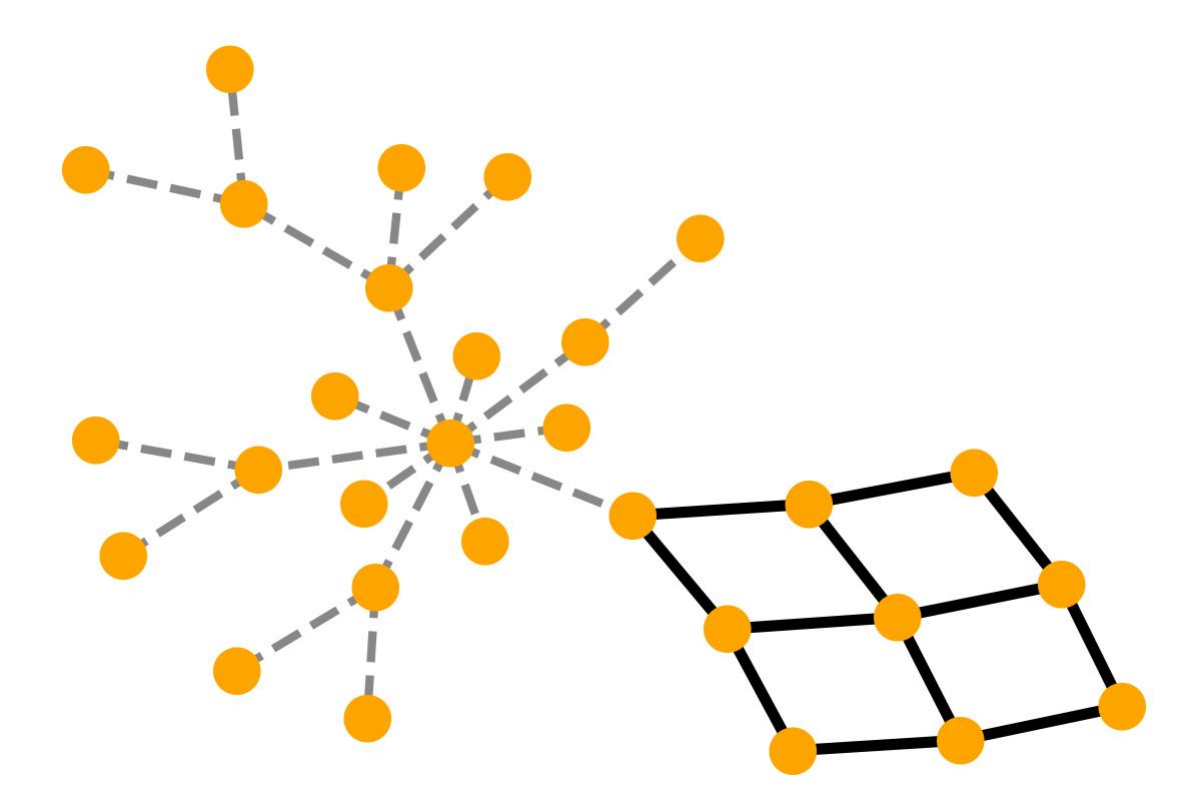}
        \caption{Ground Truth}
    \end{subfigure}
    \begin{subfigure}[b]{0.23\textwidth}
        \includegraphics[width=\linewidth]{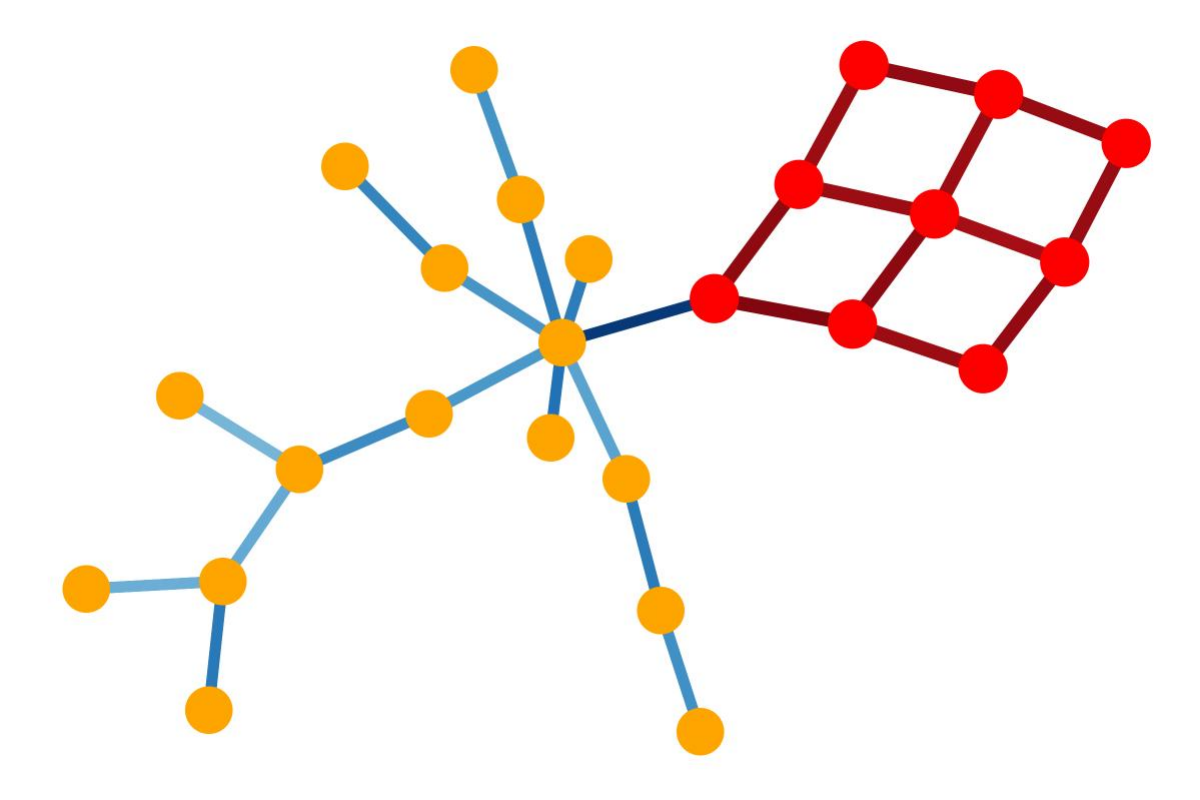}
        \caption{HPME}
    \end{subfigure}
    \begin{subfigure}[b]{0.23\textwidth}
        \includegraphics[width=\linewidth]{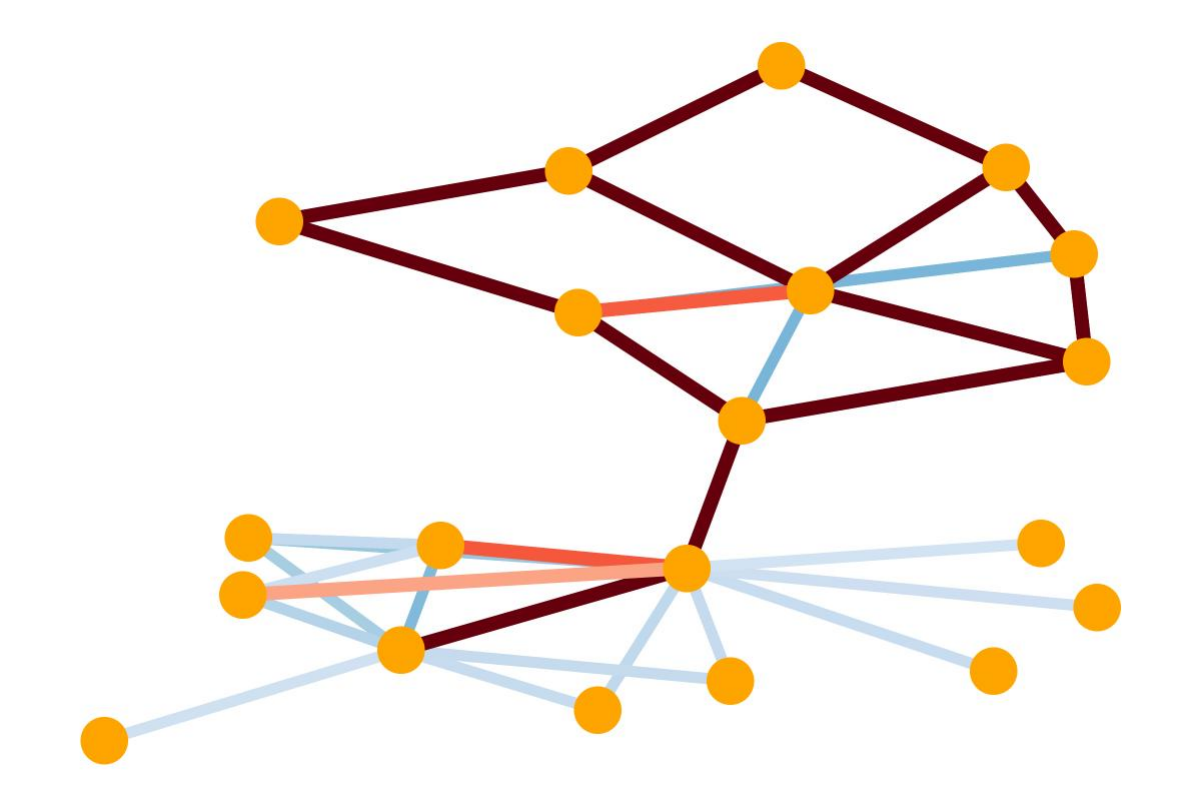}
        \caption{ProxyExplainer}
    \end{subfigure}
    \begin{subfigure}[b]{0.23\textwidth}
        \includegraphics[width=\linewidth]{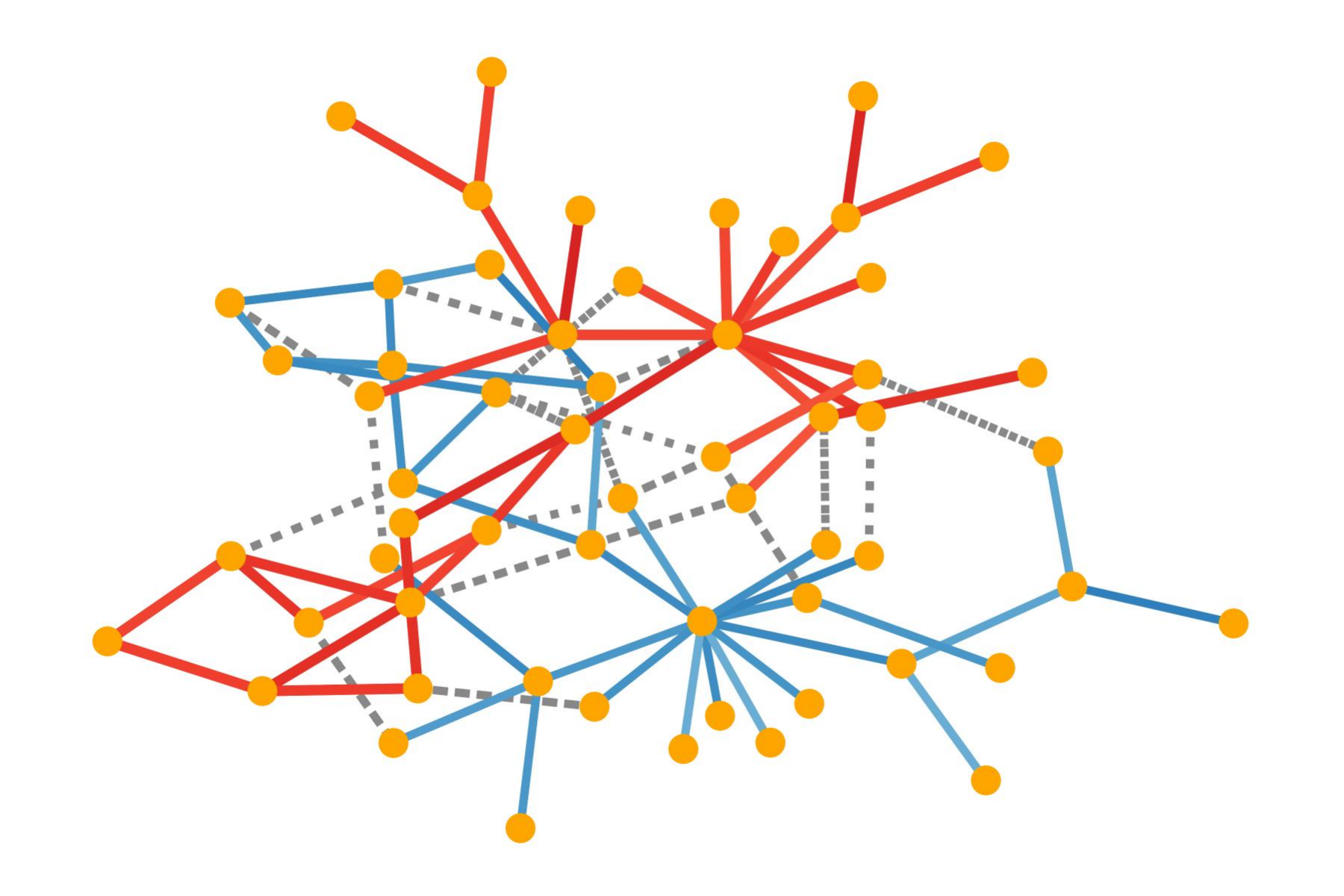}
        \caption{MixupExplainer}
    \end{subfigure}

    \caption{Visualization of mixup graphs on BA-HouseGrid.}
    \label{fig:app:mix:BA-HouseGrid}
\end{figure*}

\begin{figure*}[h]
    \centering
    \vspace{0.8em}
    \begin{subfigure}[b]{0.23\textwidth}
        \includegraphics[width=\linewidth]{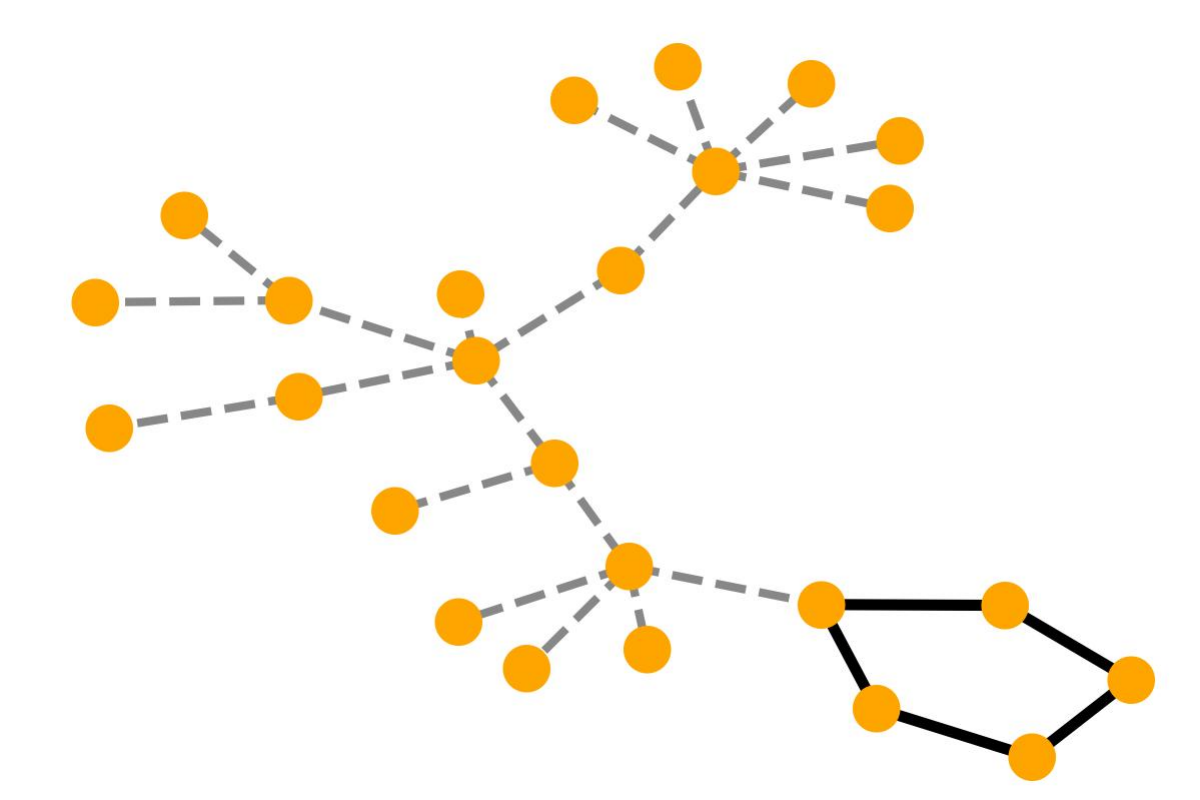}
    \end{subfigure}
    \begin{subfigure}[b]{0.23\textwidth}
        \includegraphics[width=\linewidth]{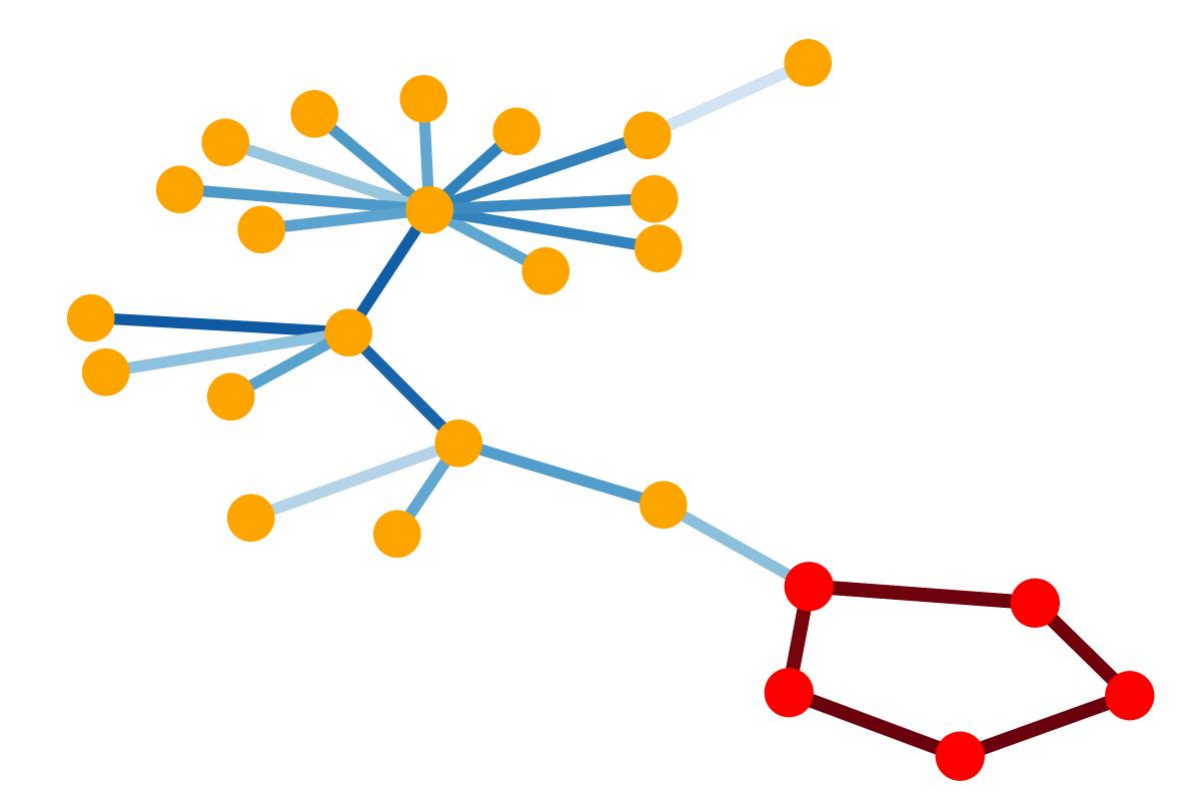}
    \end{subfigure}
    \begin{subfigure}[b]{0.23\textwidth}
        \includegraphics[width=\linewidth]{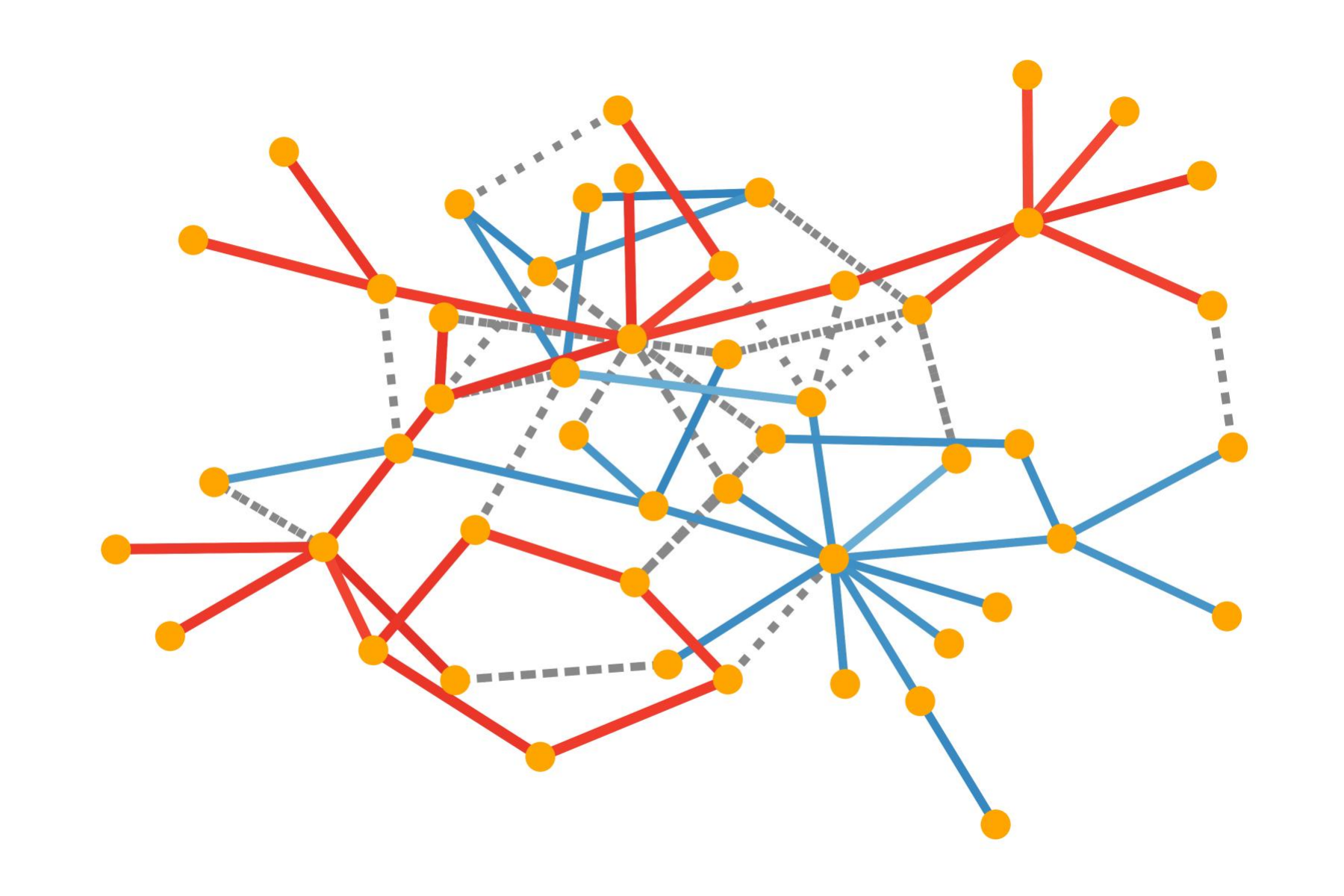}
    \end{subfigure}
    \begin{subfigure}[b]{0.23\textwidth}
        \includegraphics[width=\linewidth]{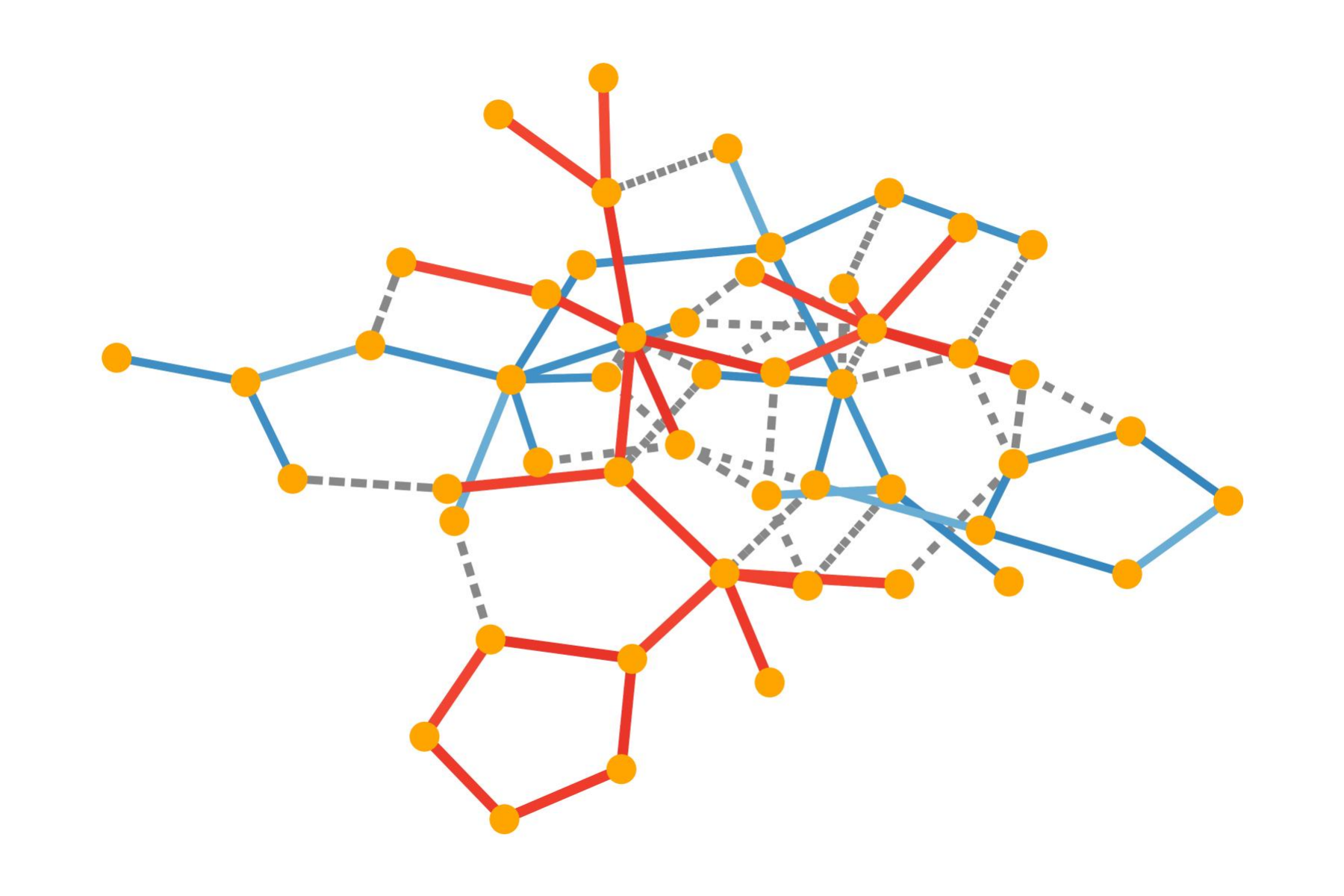}
    \end{subfigure}
    \par\vspace{0.8em}
    
    \begin{subfigure}[b]{0.23\textwidth}
        \includegraphics[width=\linewidth]{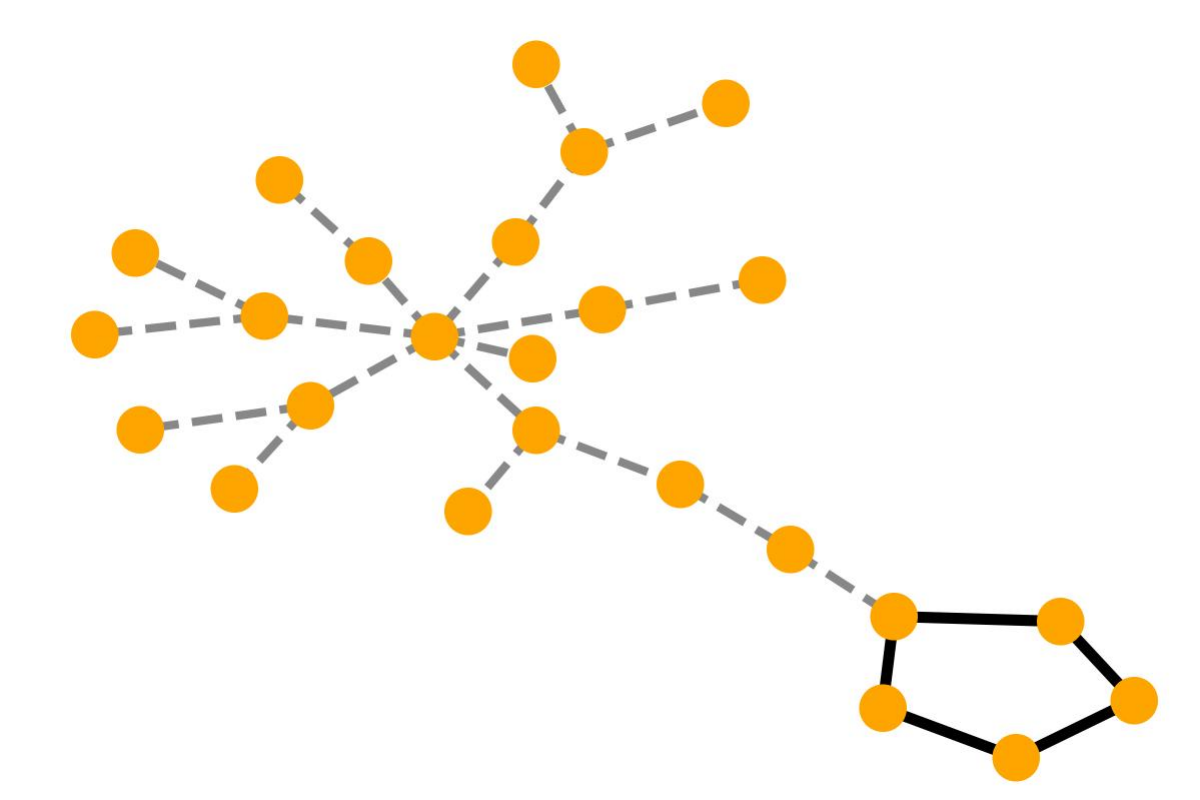}
    \end{subfigure}
    \begin{subfigure}[b]{0.23\textwidth}
        \includegraphics[width=\linewidth]{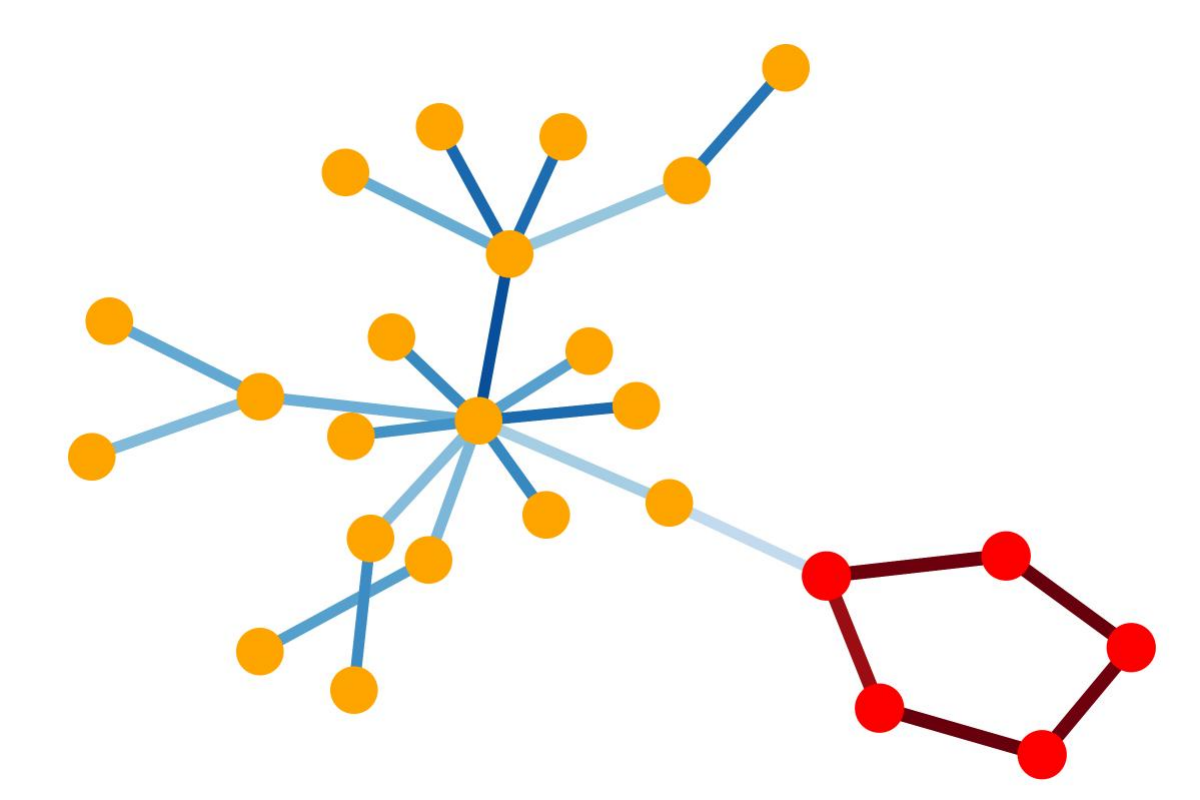}
    \end{subfigure}
    \begin{subfigure}[b]{0.23\textwidth}
        \includegraphics[width=\linewidth]{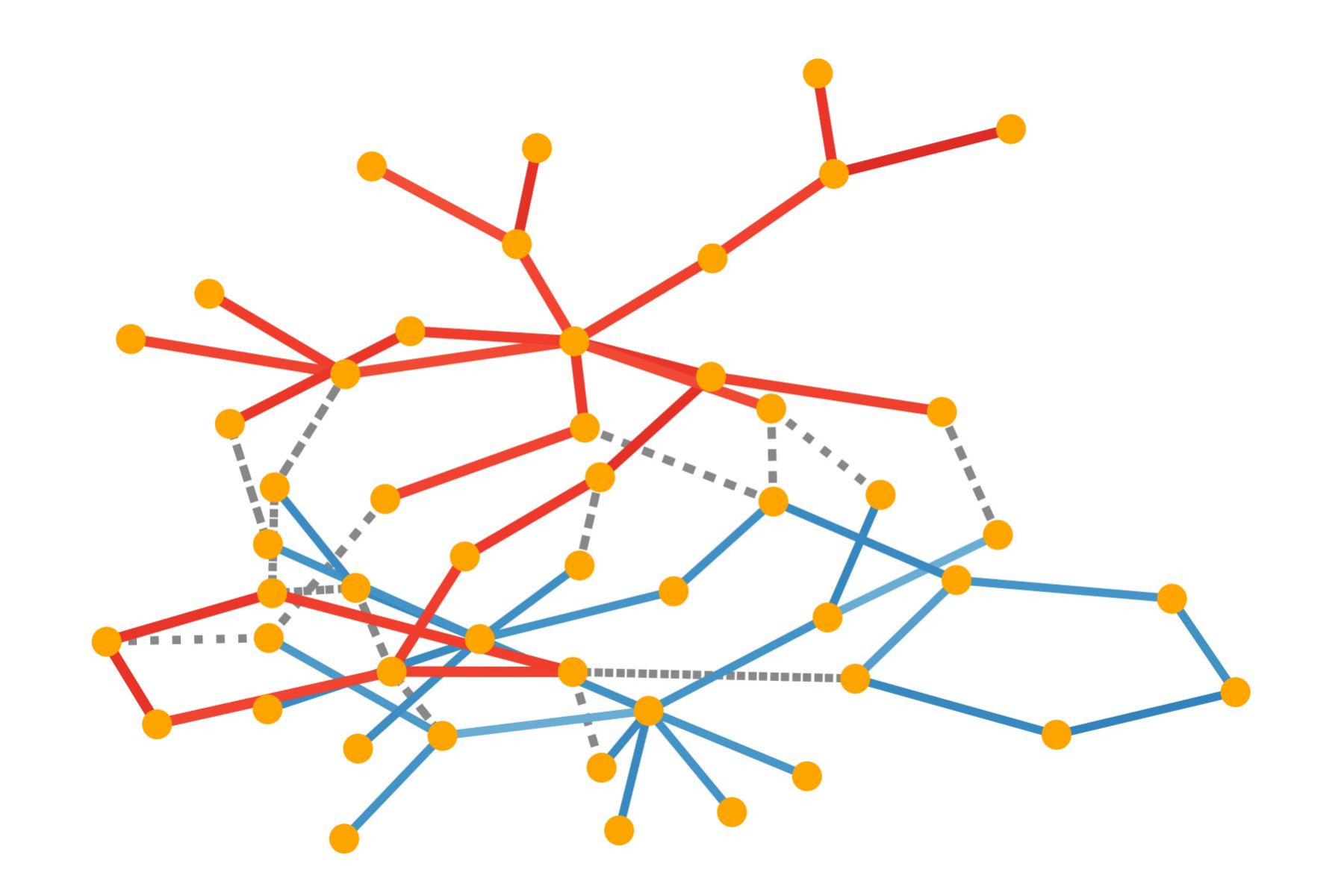}
    \end{subfigure}
    \begin{subfigure}[b]{0.23\textwidth}
        \includegraphics[width=\linewidth]{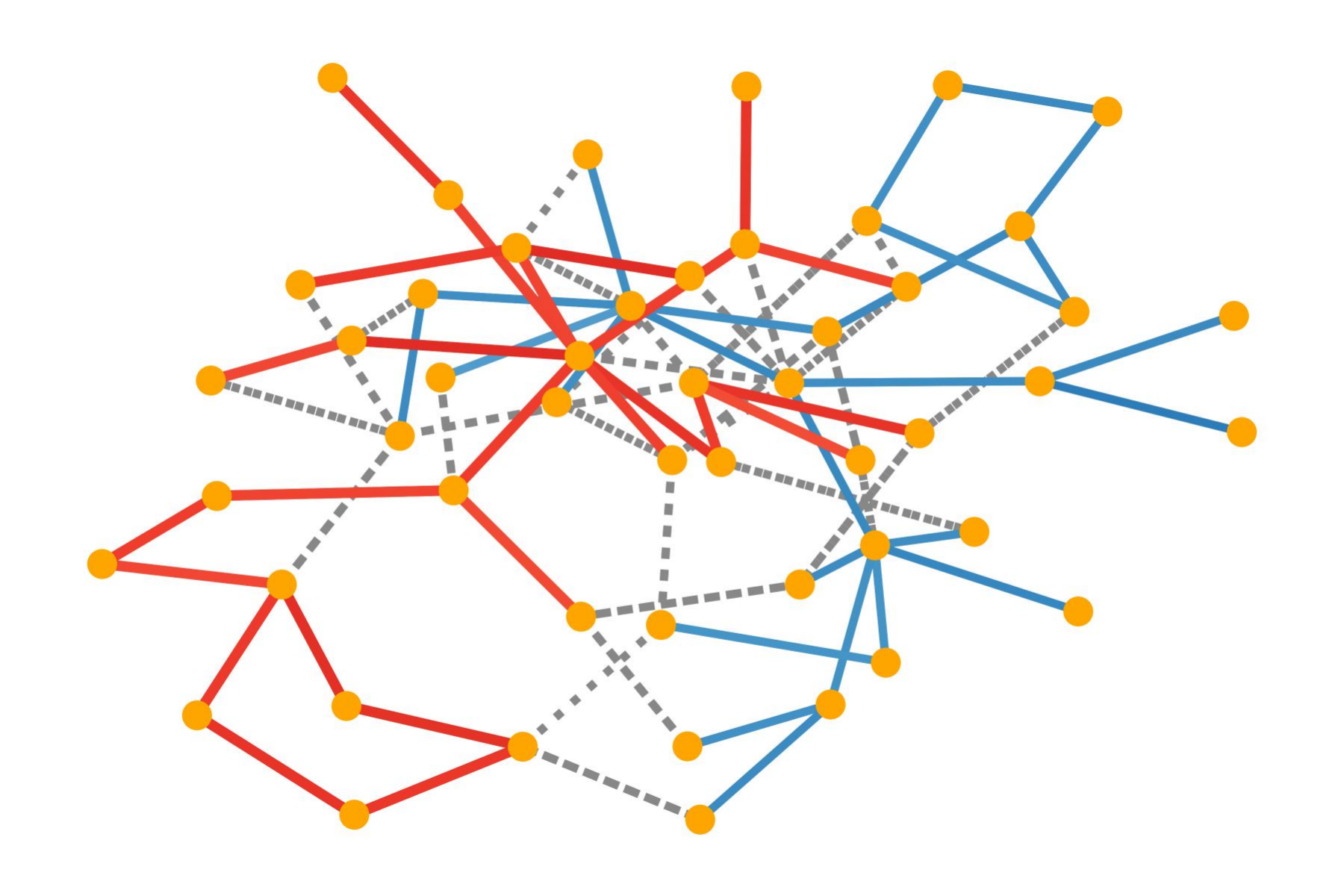}
    \end{subfigure}
    \par\vspace{0.8em}
    
    \begin{subfigure}[b]{0.23\textwidth}
        \includegraphics[width=\linewidth]{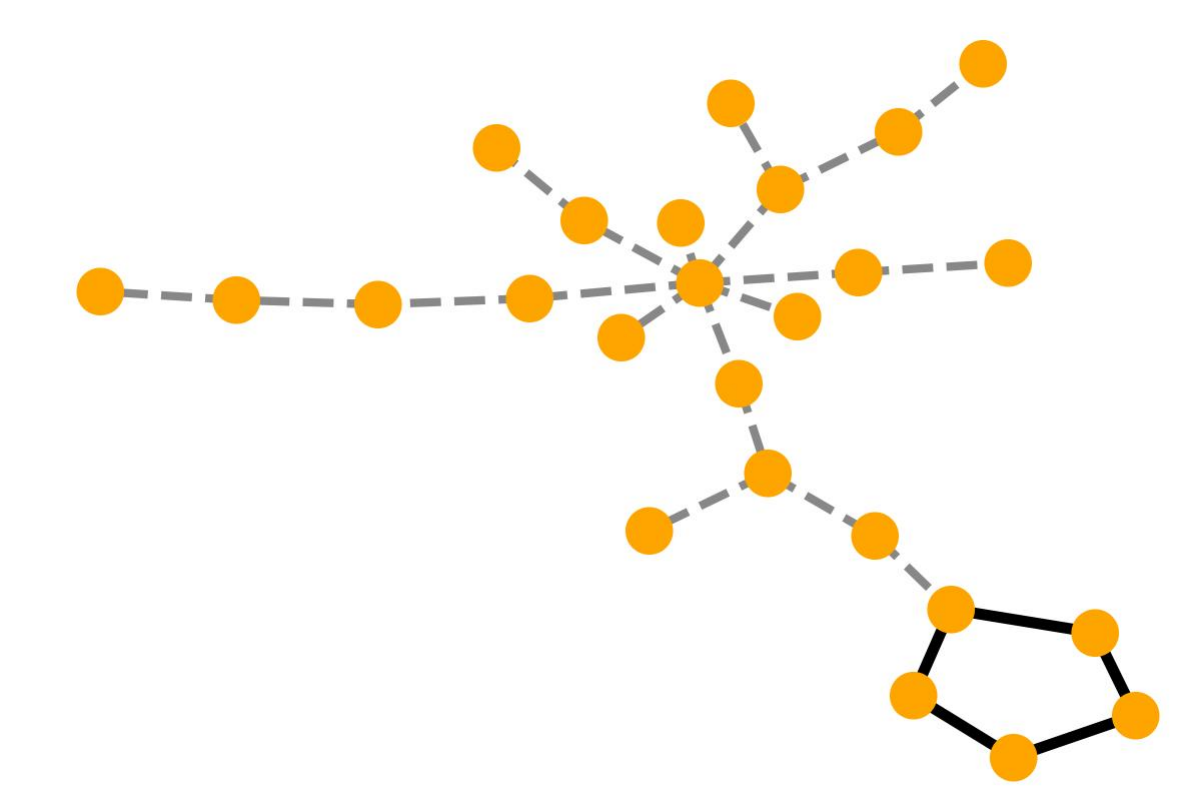}
        \caption{Ground Truth}
    \end{subfigure}
    \begin{subfigure}[b]{0.23\textwidth}
        \includegraphics[width=\linewidth]{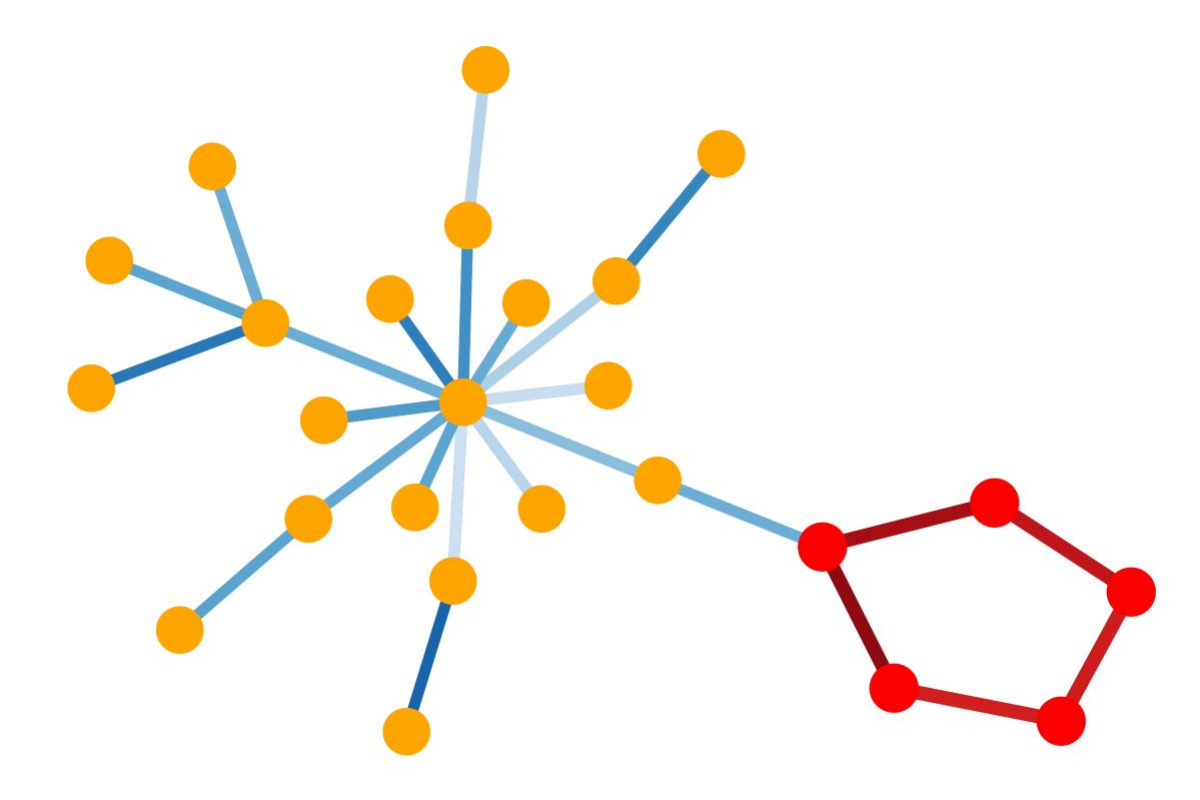}
        \caption{HPME}
    \end{subfigure}
    \begin{subfigure}[b]{0.23\textwidth}
        \includegraphics[width=\linewidth]{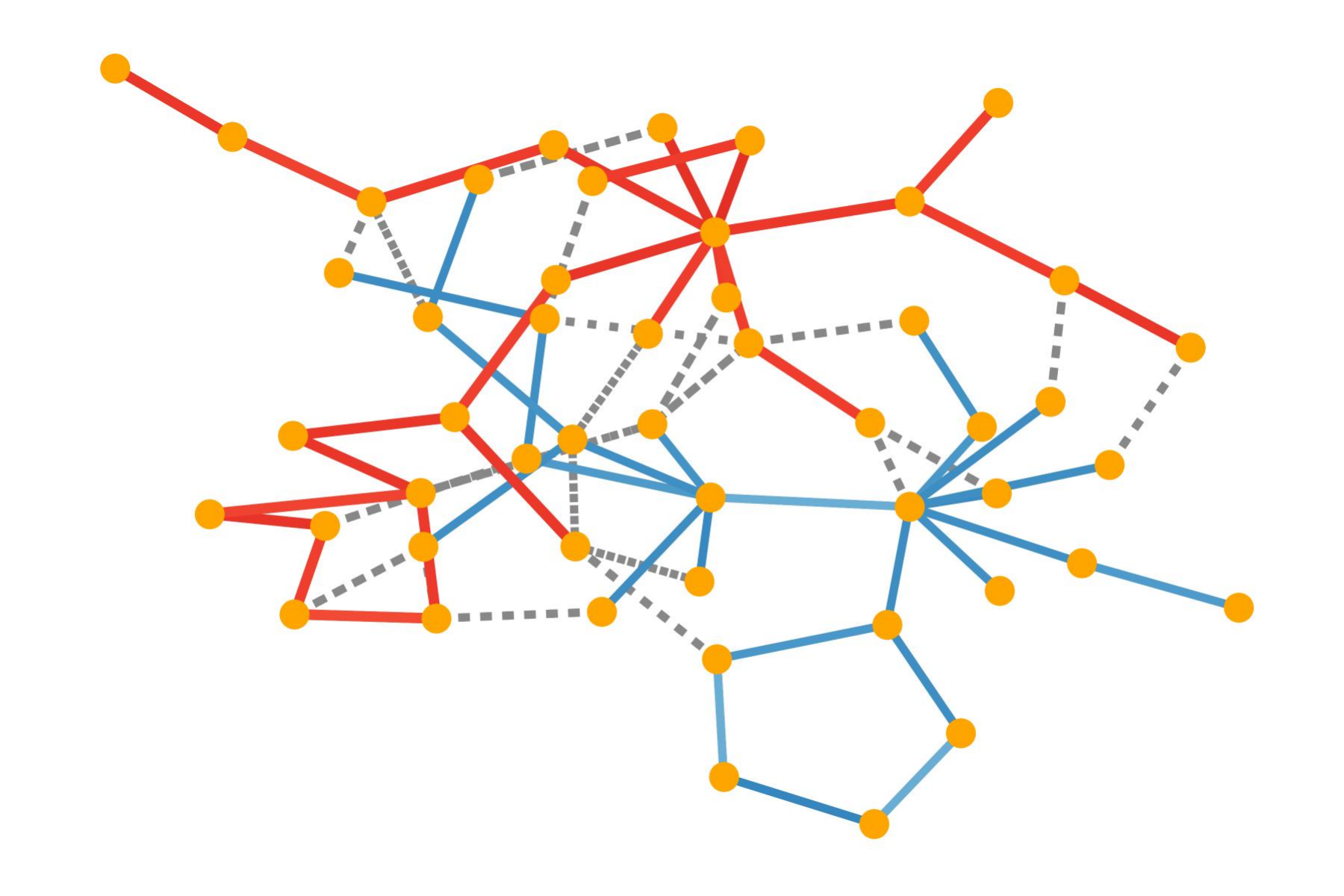}
        \caption{RegExplainer}
    \end{subfigure}
    \begin{subfigure}[b]{0.23\textwidth}
        \includegraphics[width=\linewidth]{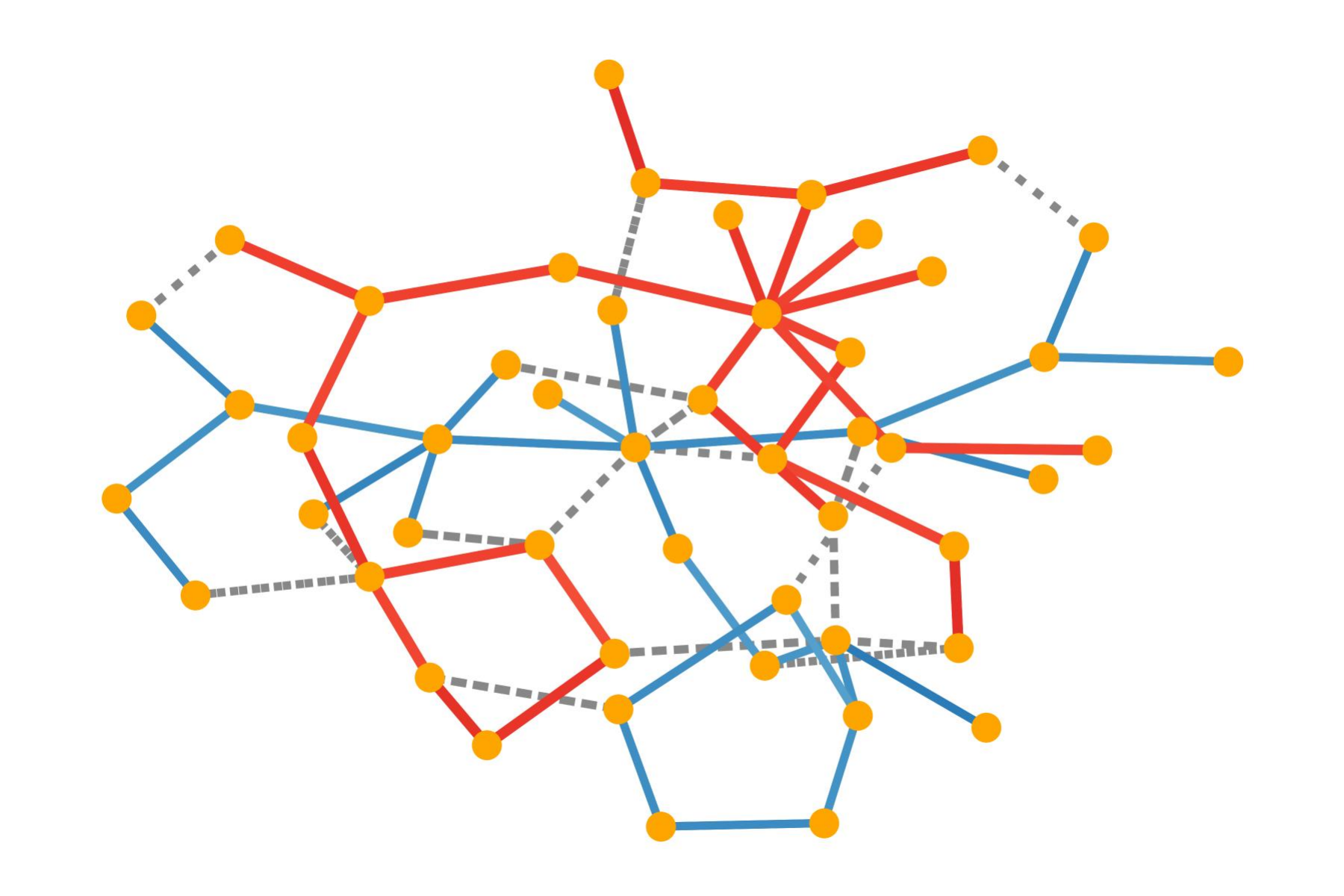}
        \caption{MixupExplainer}
    \end{subfigure}

    \caption{Visualization of mixup graphs on BA-Motif-Volume.}
    \label{fig:app:mix:BA-Motif-Volume}
\end{figure*}

We present the mixup results on the BA-HouseGrid classification dataset and the BA-Motif-Volume regression dataset in \Figref{fig:app:mix:BA-HouseGrid} and \Figref{fig:app:mix:BA-Motif-Volume}, respectively. For each dataset, three graph samples are randomly selected to generate the mixup results, where the edge color intensity indicates the corresponding weight magnitude. It should be noted that ProxyExplainer employs a graph generation strategy to produce mixup results, which leads to fully connected edges. To facilitate visualization, we display only the top 64 edges with the highest weights, corresponding to the maximum number of edges considered in this experiment.

\subsubsection{Analysis of Pooling Size in Structural Mixup} 
\label{sec:size_Analysis}
HPME’s structural mixup first applies top-k graph pooling to extract explanatory subgraphs from both the target graph and its sampled graph. These extracted subgraphs are then exchanged, which requires them to have matching sizes. In this section, we further examine this size constraint.

When mixing with the near graph, the use of embedding-similarity sampling typically selects graphs with similar labels that often share the same functional motifs or structural patterns. As a result, the target graph and its near graph generally produce explanatory subgraphs of similar size, allowing structural mixup to operate reliably.
When mixing with the far graph, the far graph usually differs in label and motif type, leading to potential mismatch in the intrinsic size of the explanatory subgraphs. In such cases, we rely on the assumption that explanatory subgraphs occupy only a small fraction of the full graph. By using a slightly larger top-k ratio, we implicitly pool a superset that includes the far graph’s explanatory region, while any redundant nodes are later refined through the MLP-generated mask. As shown in Table \ref{classification_regression}, HPME maintains strong performance on datasets where explanation sizes vary considerably (e.g., BA-HouseGrid, BAHouse-AndGrid, Fluorid-Carbony, BA-Motif-Counting), demonstrating good robustness and generalization ability.

\begin{figure*}[h]
    \centering
    \vspace{0.8em}
    \begin{subfigure}[b]{0.23\textwidth}
        \includegraphics[width=\linewidth]{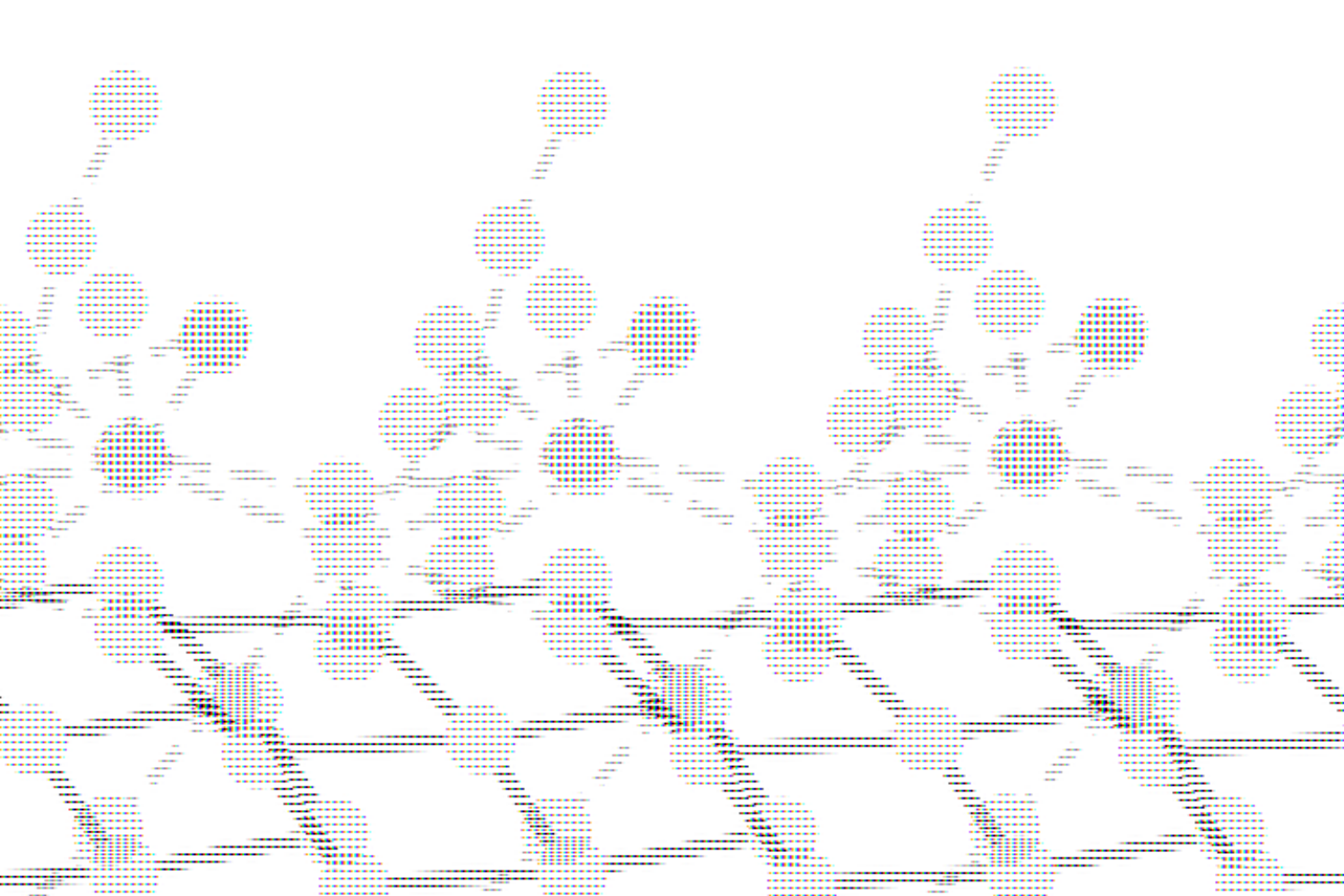}
        \caption{GT}
    \end{subfigure}
    \begin{subfigure}[b]{0.23\textwidth}
        \includegraphics[width=\linewidth]{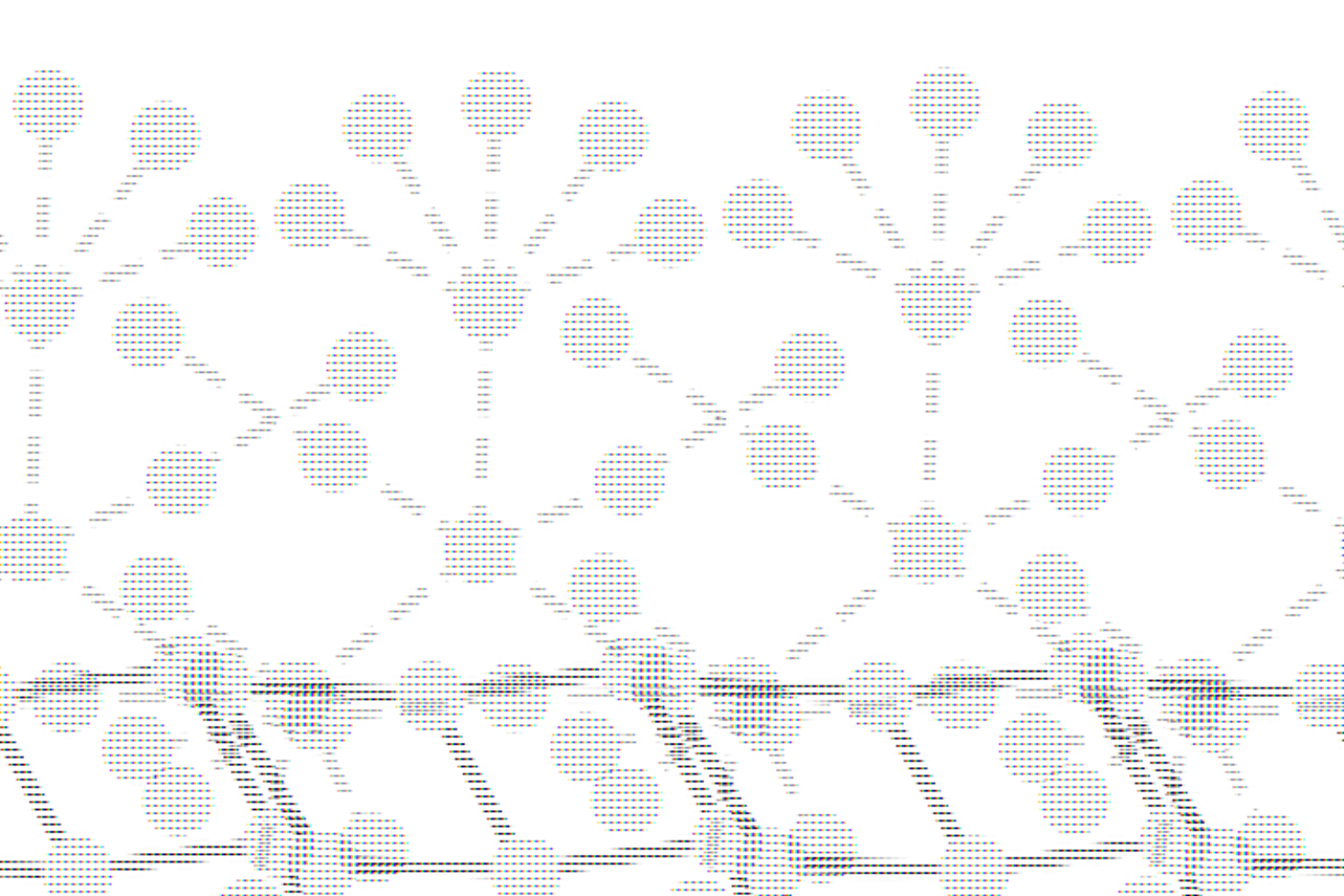}
        \caption{Near GT}
    \end{subfigure}
    \begin{subfigure}[b]{0.23\textwidth}
        \includegraphics[width=\linewidth]{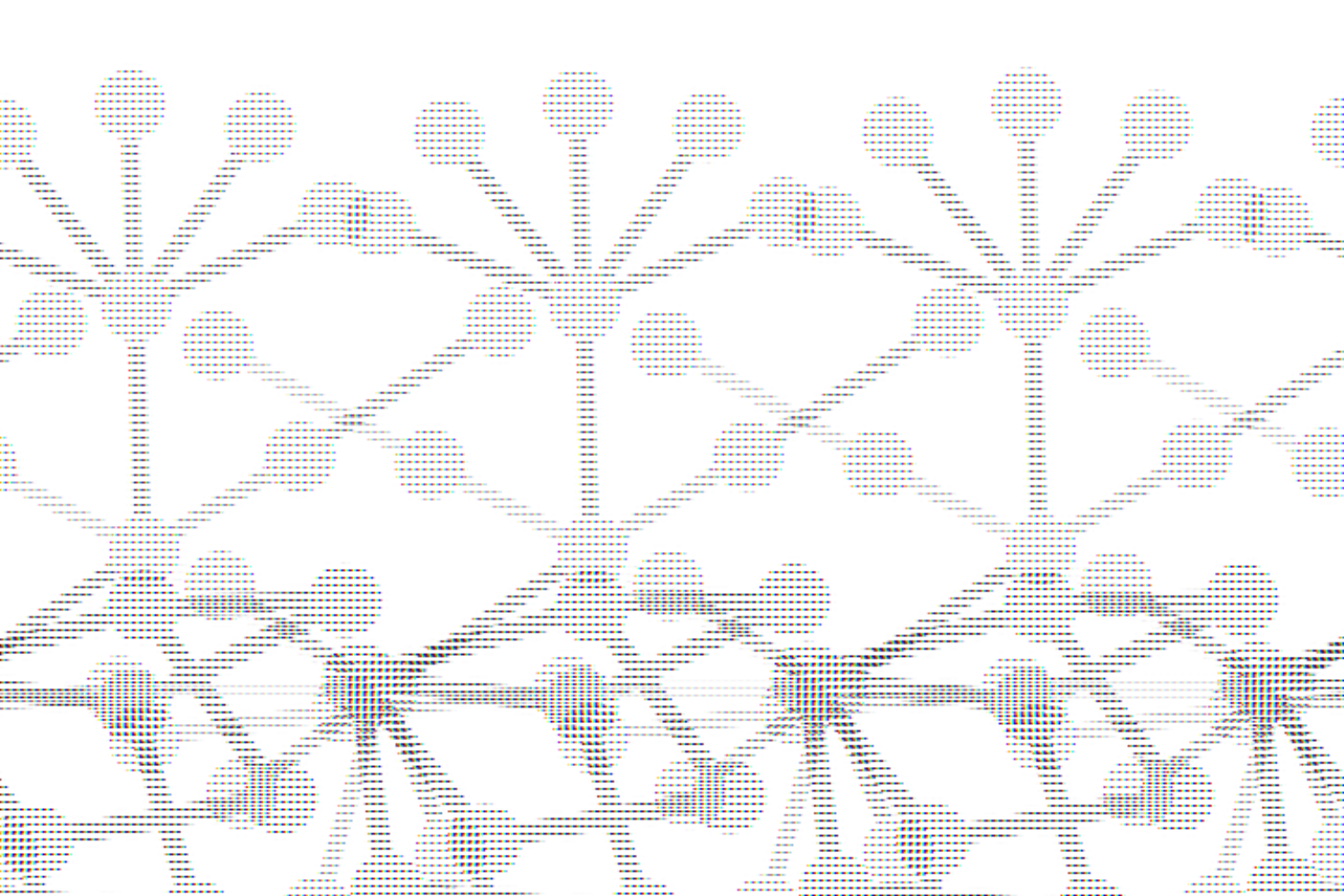}
        \caption{Mixup Graph}
    \end{subfigure}
    \begin{subfigure}[b]{0.23\textwidth}
        \includegraphics[width=\linewidth]{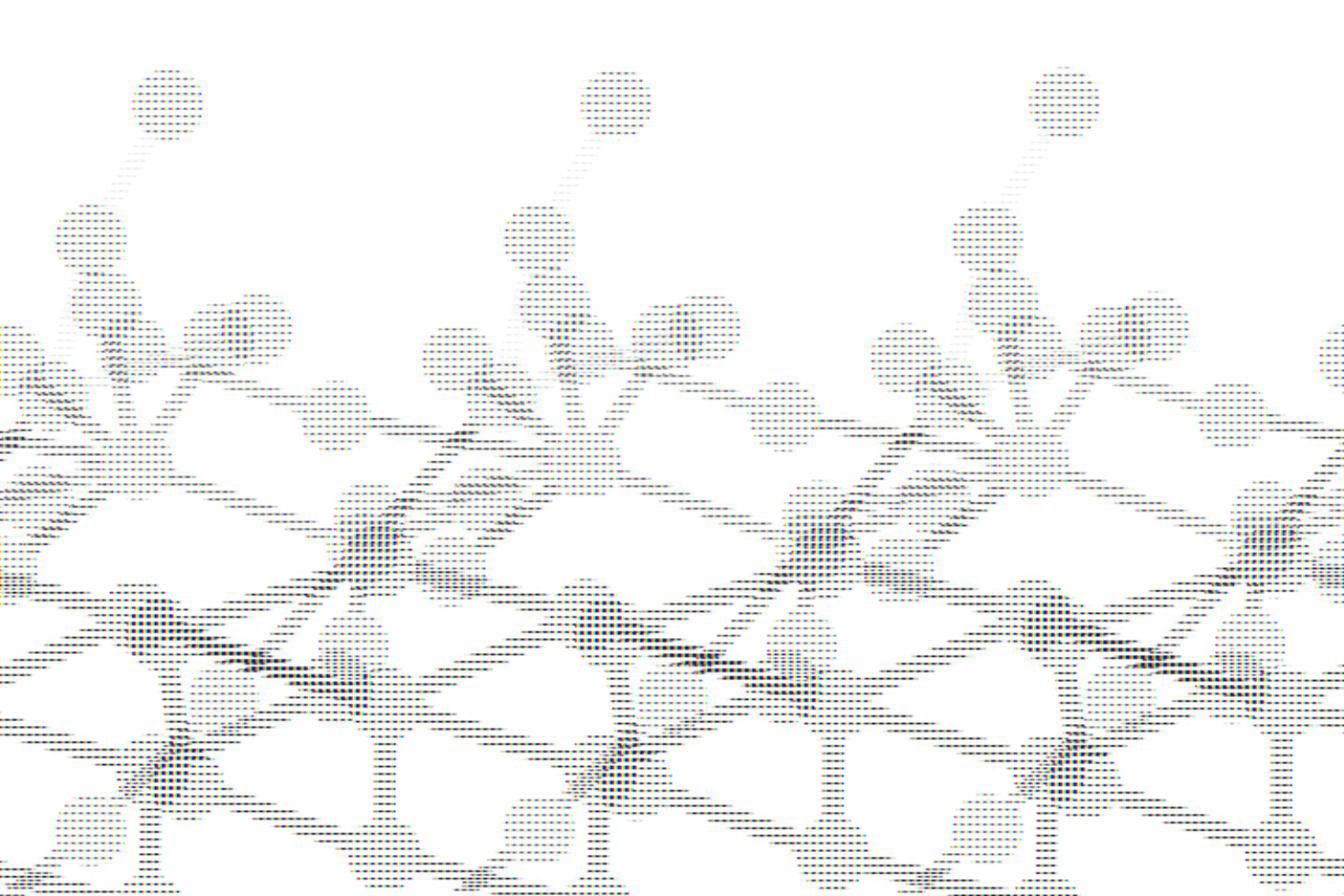}
        \caption{Near Mixup Graph}
    \end{subfigure}
    \vspace{5pt}
    \begin{subfigure}[b]{0.23\textwidth}
        \includegraphics[width=\linewidth]{figures/app/result_size/4374_gt.pdf}
        \caption{GT}
    \end{subfigure}
    \begin{subfigure}[b]{0.23\textwidth}
        \includegraphics[width=\linewidth]{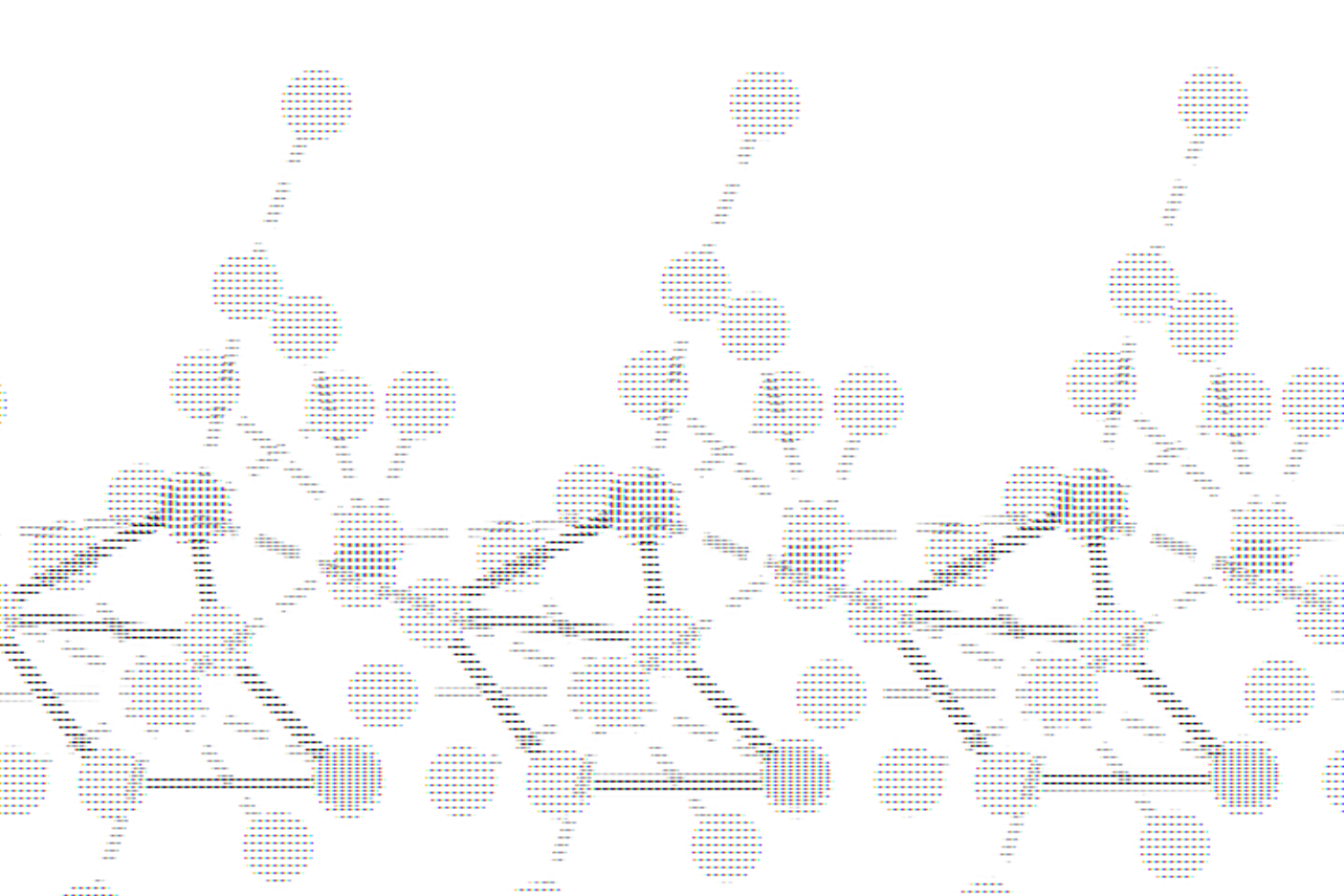}
        \caption{Far GT}
    \end{subfigure}
    \begin{subfigure}[b]{0.23\textwidth}
        \includegraphics[width=\linewidth]{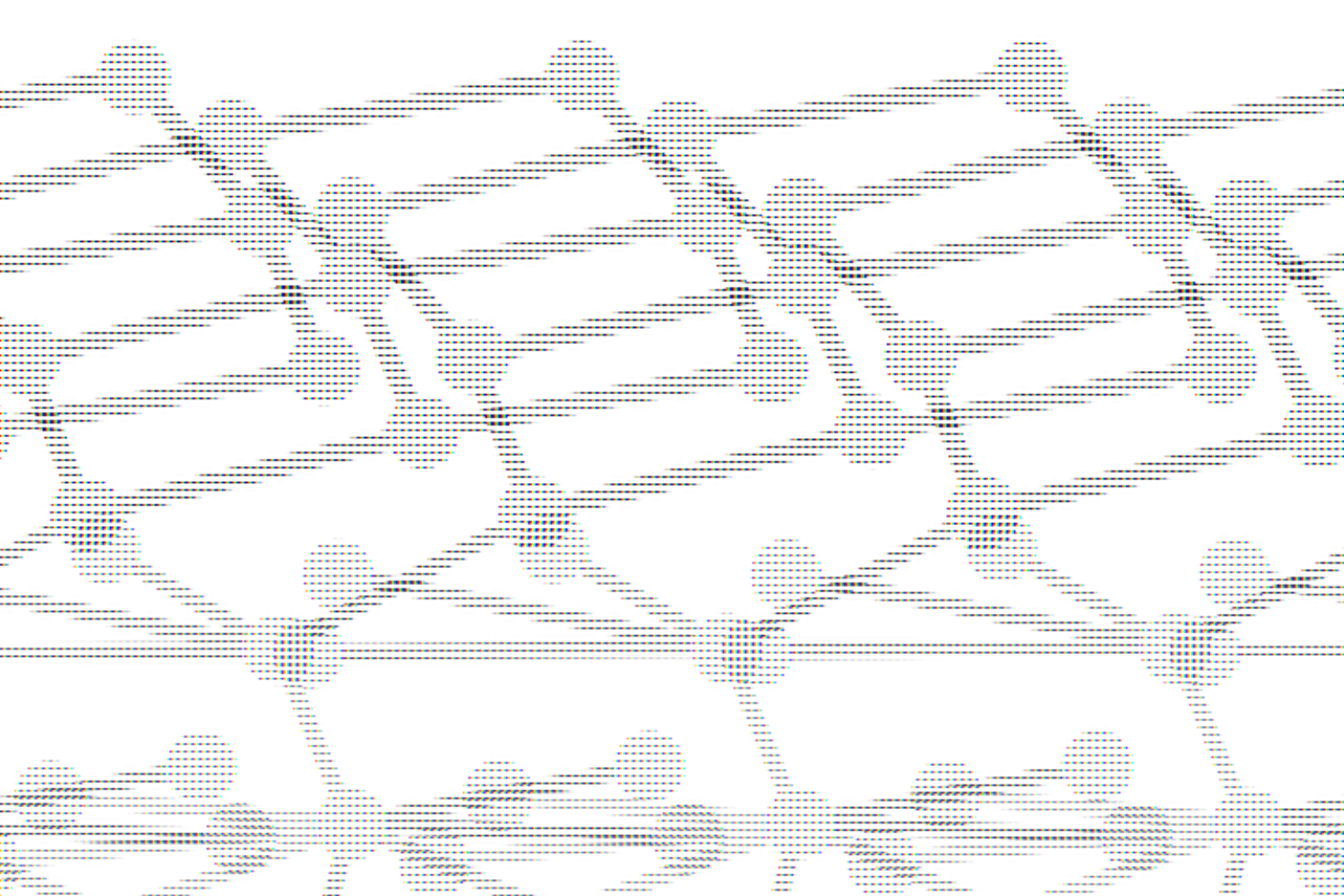}
        \caption{Mixup Graph}
    \end{subfigure}
    \begin{subfigure}[b]{0.23\textwidth}
        \includegraphics[width=\linewidth]{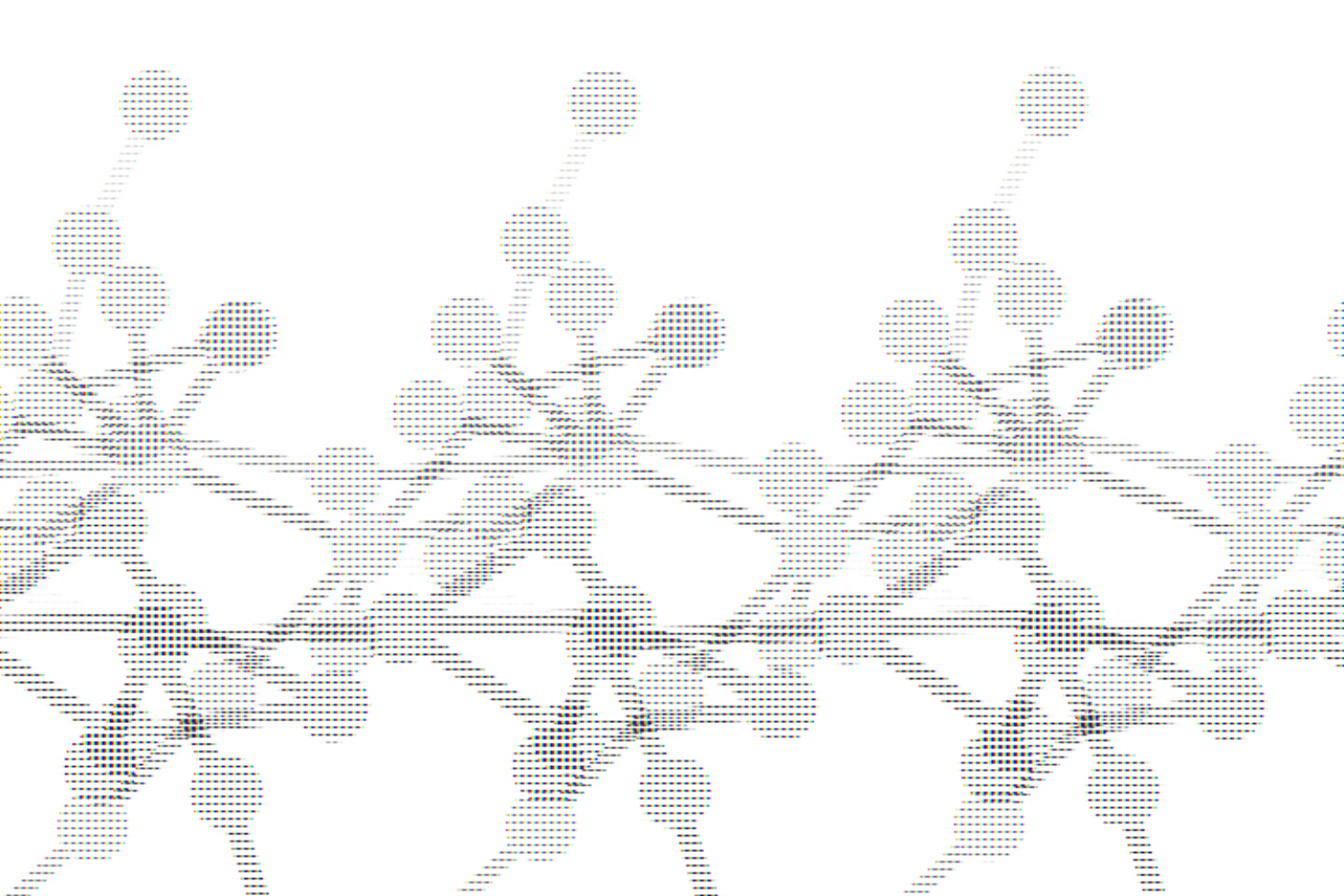}
        \caption{Far Mixup Graph}
    \end{subfigure}

    \caption{Pooling Size Study on BA-HouseGrid.}
    \label{fig:size:BA-HouseGrid}
\end{figure*}

We further conduct a case study on BA-HouseGrid. For a randomly selected target graph, we visualize the structural mixup process with both the near graph and the far graph. The results shown in \Figref{fig:size:BA-HouseGrid} demonstrate that HPME consistently identifies and exchanges the critical motif structures in both cases, enabling HPME to extract accurate and faithful explanations.

\end{document}